\documentclass{article}

\PassOptionsToPackage{numbers, compress}{natbib}

\usepackage[final]{neurips_2022}

\usepackage[utf8]{inputenc} 
\usepackage[T1]{fontenc}    
\usepackage{hyperref}       
\usepackage{url}            
\usepackage{booktabs}       
\usepackage{amsfonts}       
\usepackage{nicefrac}       
\usepackage{microtype}      
\usepackage{xcolor}         

\usepackage{color}
\usepackage{algorithm}
\usepackage{algorithmic}
\usepackage{tabularx}
\usepackage{amsmath}
\usepackage{amssymb}
\usepackage{bm}
\usepackage{graphicx}
\usepackage{booktabs}  
\usepackage{threeparttable} \usepackage{multirow}
\newtheorem{theorem}{\textbf{Theorem}}
\newtheorem{definition}{\textbf{Definition}}
\newtheorem{assumption}{\textbf{Assumption}}
\newtheorem{fact}{\textbf{Fact}}
\newtheorem{setting}{\textbf{Setting}}
\newtheorem{lemma}{\textbf{Lemma}}
\usepackage{cite}
\usepackage{subfigure}
\usepackage{tablefootnote}
\usepackage{footnote}

\title{Distributed Distributionally Robust Optimization with Non-Convex Objectives}

\author{%
    Yang Jiao \\
	Tongji University \\
	\texttt{yangjiao@tongji.edu.cn} \\
    \And
	Kai Yang\thanks{Corresponding author.} \\
	Tongji University\\
	\texttt{kaiyang@tongji.edu.cn} \\
	\And
	Dongjin Song \\
	University of Connecticut \\
	\texttt{dongjin.song@uconn.edu} \\
}

\begin{document}

	\maketitle

	\begin{abstract}
Distributionally Robust Optimization (DRO), which aims to find an optimal decision that minimizes the worst case cost over the ambiguity set of probability distribution, has been widely applied in diverse applications, \textit{e.g.}, network behavior analysis, risk management, \textit{etc.} However, existing DRO techniques face three key challenges: 1) how to deal with the asynchronous updating in a distributed environment;  2) how to leverage the prior distribution effectively; 3) how to properly adjust the degree of robustness according to different scenarios. To this end, we propose an asynchronous distributed algorithm, named \textbf{A}synchronous \textbf{S}ingle-loo\textbf{P} alternat\textbf{I}ve g\textbf{R}adient proj\textbf{E}ction (ASPIRE) algorithm with the it\textbf{E}rative \textbf{A}ctive \textbf{S}\textbf{E}t method (EASE) to tackle the distributed distributionally robust
optimization (DDRO) problem. Furthermore, a new uncertainty set, \textit{i.e.}, constrained $D$-norm uncertainty set, is developed to effectively leverage the prior distribution and flexibly control the degree of robustness. Finally, our theoretical analysis elucidates that the proposed algorithm is guaranteed to converge and the iteration complexity is also analyzed. Extensive empirical studies on real-world datasets demonstrate that the proposed method can not only achieve fast convergence, and remain robust against data heterogeneity as well as malicious attacks, but also tradeoff robustness with performance.
\end{abstract}

\section{Introduction} \label{introduction}
The past decade has witnessed the proliferation of smartphones and Internet of Things (IoT) devices, which generate a plethora of data everyday. Centralized machine learning requires gathering the data to a particular server to train models which incurs high communication overhead \citep{sun2019communication} and suffers privacy risks \citep{sicari2015security}. As a remedy, distributed machine learning methods have been proposed. Considering a distributed system composed of $N$ workers (devices), we denote the dataset of these workers as $\{ {D_1}, \cdots,{D_N}\} $. For the $j^{\rm{th}}$ ($1 \! \le \! j \!\le \! N$) worker, the labeled dataset is given as ${D_j} = \{ {\bf{x}}^i_j,y^i_j\} $, where ${\bf{x}}^i_j \! \in \! \mathbb{R}^{d}$ and $y^i_j \in  \{1,\cdots,c\}$ denote the $i^{\rm{th}}$ data sample and the corresponding label, respectively. The distributed learning tasks can be formulated as the following optimization problem,
\vspace{-1mm}
\begin{align}
\label{eq:0}
\mathop {\min }\limits_{\boldsymbol{w}\in {\boldsymbol{\mathcal{W}}}} \; F(\boldsymbol{w}) \quad {\rm{with}} \quad F(\boldsymbol{w}): = \sum\nolimits_{j} {{f_j}} (\boldsymbol{w}),
\end{align}
where $\boldsymbol{w}\in {\mathbb{R}^p}$ is the model parameter to be learned and ${\boldsymbol{\mathcal{W}}}\!\subseteq\! \mathbb{R}^p$ is a nonempty closed convex set, ${f_j}( \cdot )$ is the empirical risk over the $j^{\rm{th}}$ worker involving only the local data:
\vspace{-1mm}
\begin{equation}
\label{eq:2}
\begin{aligned}
{f_j}({\boldsymbol{w}}) = \sum\nolimits_{i:{\bf{x}}_j^i \in {D_j}}^{} {\frac{1}{{|{D_j}|}}} \mathcal{L}_j({\bf{x}}_j^i,y_j^i;{\boldsymbol{w}}),
\end{aligned}
\end{equation}
where $\mathcal{L}_j$ is the local objective function over the $j^{\rm{th}}$ worker. Problem in Eq. (\ref{eq:0}) arises in numerous areas, such as distributed signal processing \citep{geraci2015energy}, multi-agent optimization \citep{nedic2009distributed}, \textit{etc}. However, such problem does not consider the data heterogeneity \citep{zawad2021curse,qian2020robustness,qian2020towards,liao2021investigation}  among different workers (\textit{i.e.}, data distribution of workers could be substantially different from each other \citep{singhal2021federated}). Indeed, it has been shown that traditional federated approaches, such as FedAvg \citep{mcmahan2017communication}, built for independent and identically distributed (IID) data may perform poorly when applied to Non-IID data \citep{karimireddy2019scaffold}. This issue can be mitigated via learning a robust model that aims to achieve uniformly good performance over all workers by solving the following distributionally robust optimization (DRO) problem in a distributed manner:
\vspace{-1mm}
\begin{align}
\label{eq:4_new}
\mathop {{\rm{min}}}\limits_{\boldsymbol{w}\in {\boldsymbol{\mathcal{W}}}} {\rm{ }}\mathop {{\rm{max}}}\limits_{{\bf{p}} \in {\bf{\Omega} } \subseteq {\Delta _N}} F(\boldsymbol{w},{\bf{p}}): = \sum\nolimits_{j} p_j {{f_j}} (\boldsymbol{w}),
\end{align}
where ${\bf{p}}  =  [{p_1}, \cdots ,{p_N}] \! \in \! {\mathbb{R}^N}$ is the adversarial distribution in $N$ workers, the $j^{\rm{th}}$ entry in this vector, \textit{i.e.}, ${p_j}$ represents the adversarial distribution value for the $j^{\rm{th}}$ worker. ${\Delta _N}=\{ {\bf{p}} \in {\mathbb{R}^{N}_{+}}: {\bf{1}^ \top }{\bf{p}} = 1 \} $  and ${\bf{\Omega} }$ is a subset of ${\Delta _N}$.  Agnostic federated learning (AFL)~\citep{mohri2019agnostic} firstly introduces the distributionally robust (agnostic) loss in federated learning and provides the convergence rate for (strongly) convex functions. However, AFL does not discuss the setting of ${\bf{\Omega} }$. DRFA-Prox~\citep{deng2021distributionally} considers ${\bf{\Omega} }={\Delta _N}$ and imposes a regularizer on adversarial distribution to leverage the prior distribution. Nevertheless, three key challenges have not yet been addressed by prior works. First, whether it is possible to construct an uncertainty framework that can not only flexibly maintain the trade-off between the model robustness and performance but also effectively leverage the prior distribution?  Second, how to design asynchronous algorithms with guaranteed convergence?  Compared to synchronous algorithms, the master in asynchronous algorithms can update its parameters after receiving updates from only a small subset of workers~\citep{zhang2014asynchronous,chang2016asynchronous}.  Asynchronous algorithms are particularly desirable in practice since they can relax strict data dependencies and ensure convergence even in the presence of device failures \citep{zhang2014asynchronous}. Finally, whether it is possible to flexibly adjust the degree of robustness?  Moreover, it is necessary to provide convergence guarantee when the objectives (\textit{i.e.}, ${f_j}({\boldsymbol{w}_j}),\forall j$) are non-convex. 

To this end,  we propose ASPIRE-EASE to effectively address the aforementioned challenges. Firstly, different from existing works, the prior distribution is incorporated within the constraint in our formulation, which can not only leverage the prior distribution more effectively but also achieve guaranteed feasibility for any adversarial distribution within the uncertainty set. The prior distribution can be obtained from side information or uniform distribution \citep{qian2019robust}, which is necessary to construct the uncertainty (ambiguity) set and obtain a more robust model~\citep{deng2021distributionally}.  Specifically, we formulate the prior distribution informed distributionally robust optimization (PD-DRO) problem as:
\vspace{-1mm}
\begin{align}
\label{eq:3}
\mathop {{\rm{min}}}\limits_{\boldsymbol{z}\in{{\boldsymbol{\mathcal{Z}}}},\{{\boldsymbol{w}_j}\in{{\boldsymbol{\mathcal{W}}}}\}} & \mathop {{\rm{max}}} \limits_{{\bf{p}}  \in \boldsymbol{\mathcal{P}}} \sum\nolimits_{j} {{p_j}{f_j}({\boldsymbol{w}_j})} \\ 
{\rm{s.t.}} \; \quad    \boldsymbol{z}&  =    {\boldsymbol{w}_j}, \; j\!=\!1,\!\cdots\!, N , \nonumber\\
{\rm{var.}} \; \;  \; \;  \boldsymbol{z}&,{\boldsymbol{w}_1},{\boldsymbol{w}_2}, \cdots ,{\boldsymbol{w}_N}, \nonumber
\end{align}
where $\boldsymbol{z}\! \in \! {\mathbb{R}^p}$ is the global consensus variable, $\boldsymbol{w}_j\! \in \! {\mathbb{R}^p}$ is the local variable (local model parameter) of $j^{\rm{th}}$ worker and ${\boldsymbol{\mathcal{Z}}}\!\subseteq\! \mathbb{R}^p$ is a nonempty closed convex set.  $\boldsymbol{\mathcal{P}}\!\subseteq\! \mathbb{R}_ + ^N$  is the uncertainty (ambiguity) set of adversarial distribution ${\bf{p}}$, which is set based on the prior distribution. To solve the PD-DRO problem in an asynchronous distributed manner, we first propose \textbf{A}synchronous \textbf{S}ingle-loo\textbf{P} alternat\textbf{I}ve g\textbf{R}adient proj\textbf{E}ction (ASPIRE), which employs \emph{simple} gradient projection steps for the update of primal and dual variables at every iteration, thus is computationally \emph{efficient}. Next, the it\textbf{E}rative \textbf{A}ctive \textbf{S}\textbf{E}t method (EASE) is employed to replace the traditional cutting plane method to improve the computational efficiency and speed up the convergence. We further provide the convergence guarantee for the proposed algorithm. Furthermore, a new uncertainty set, \textit{i.e.}, constrained $D$-norm ($CD$-norm), is proposed in this paper and its advantages include: 1) it can flexibly control the degree of robustness; 2) the resulting subproblem is computationally simple; 3) it can effectively leverage the prior distribution and flexibly set the bounds for every $p_j$.

{\bf{Contributions.}} Our contributions can be summarized as follows:

{\bf{1.}} We formulate a PD-DRO problem with $CD$-norm uncertainty set. PD-DRO incorporates the prior distribution as constraints which can leverage prior distribution more effectively and guarantee robustness. In addition, $CD$-norm is developed to model the ambiguity set around the prior distribution and it provides a flexible way to control the trade-off between model robustness and performance.
    
{\bf{2.}} We develop a \emph{single-loop} \emph{asynchronous} algorithm, namely ASPIRE-EASE, to optimize PD-DRO in an asynchronous distributed manner. ASPIRE employs simple gradient projection steps to update the variables at every iteration, which is computationally efficient. And EASE is proposed to replace cutting plane method to enhance the computational efficiency and speed up the convergence. We demonstrate that even if the objectives ${f_j}({\boldsymbol{w}_j}),\forall j$ are non-convex, the proposed algorithm is guaranteed to converge. We also theoretically derive the iteration complexity of ASPIRE-EASE. 
    
{\bf{3.}} Extensive empirical studies on four different real world datasets demonstrate the superior performance of the proposed algorithm. It is seen that ASPIRE-EASE can not only ensure the model's robustness against data heterogeneity but also mitigate malicious attacks. 

\vspace{-1mm}

\section{Preliminaries}\label{Preliminaries}
\vspace{-1mm}
\subsection{Distributionally Robust Optimization}
\vspace{-0.75mm}
Optimization problems often contain uncertain parameters. 
A small perturbation of the parameters could render the optimal solution of the original optimization problem infeasible or completely meaningless \citep{bertsimas2004price}. Distributionally robust optimization (DRO) \citep{kuhn2019wasserstein,duchi2021learning,blanchet2019quantifying} assumes that the probability distributions of uncertain parameters are unknown but remain in an ambiguity (uncertainty) set and aims to find a decision that minimizes the worst case expected cost over the ambiguity set, whose general form can be expressed as,
\vspace{-1mm}
\begin{align}
\mathop {\min }\limits_{\boldsymbol{x} \in \boldsymbol{\mathcal{X}}} \mathop {\max }\limits_{P \in {{\bf{P}}}} {\mathbb{E}_P}[r(\boldsymbol{x},\boldsymbol{\xi})],
\end{align}
where $\boldsymbol{x} \! \in \! \boldsymbol{\mathcal{X}} $ represents the decision variable,  ${\bf{P}}$ is the ambiguity set of probability distributions $P$ of uncertain parameters $\boldsymbol{\xi}$. Existing methods for solving DRO can be broadly grouped into two widely-used categories \citep{rahimian2019distributionally}: 1) Dual methods \citep{delage2010distributionally,wiesemann2013robust,gao2016distributionally} reformulate the primal DRO problems as deterministic optimization problems through duality theory. Ben-Tal et al. \citep{ben1999robust} reformulate the robust linear optimization (RLO) problem with an ellipsoidal uncertainty set as a second-order cone optimization problem (SOCP).  2) Cutting plane methods \citep{mehrotra2014cutting,bertsimas2016reformulation} (also called adversarial approaches \citep{gorissen2015practical}) continuously solve an approximate problem with a finite number of constraints of the primal DRO problem, and subsequently check whether new constraints are needed to refine the feasible set.  Recently, several new methods~\citep{qian2019robust,levy2020large,hu2021bias} have been developed to solve DRO, which need to solve the inner maximization problem at every iteration.

\vspace{-0.5mm}
\subsection{Cutting Plane Method for PD-DRO}
\vspace{-0.75mm}

In this section, we introduce the cutting plane method for PD-DRO in Eq. (\ref{eq:3}). We first reformulate PD-DRO by introducing an additional variable  $h \!\in\! {{\boldsymbol{\mathcal{H}}}} $ (${{\boldsymbol{\mathcal{H}}}}\! \subseteq\! {\mathbb{R}^1}$ is a nonempty closed convex set) and protection function $g(\{ {\boldsymbol{w}_j}\} )$ \citep{yang2014distributed}. Introducing additional variable $h$  is an epigraph reformulation \citep{ben2009robust,yanikouglu2019survey}. In this case, Eq. (\ref{eq:3}) can be reformulated as the form with uncertainty in the constraints:
\begin{align}
 \label{eq:4-1}
\mathop {{\rm{min}}}\limits_{\boldsymbol{z}\in{{\boldsymbol{\mathcal{Z}}}},\{{\boldsymbol{w}_j}\in{{\boldsymbol{\mathcal{W}}}}\}, h \in {{\boldsymbol{\mathcal{H}}}}} &\quad  h  \nonumber\\
{\rm{s.t.}} \; \sum\nolimits_{j}  {\overline{p} {f_j}({\boldsymbol{w}_j})} \!  + \!   g&(\{ {\boldsymbol{w}_j}\} ) - h \! \le \! 0, \\
\boldsymbol{z}  =    {\boldsymbol{w}_j} , \; j\!=&1,\!\cdots\!, N , \nonumber\\
{\rm{var.}}\quad \boldsymbol{z},{\boldsymbol{w}_1},{\boldsymbol{w}_2}, &\cdots ,{\boldsymbol{w}_N}, h, \nonumber
\end{align}
where $\overline{p} $ is the nominal value of the adversarial distribution for every worker and $g(\{ {\boldsymbol{w}_j}\} ) = \mathop {\max }\limits_{{\bf{p}} \in \boldsymbol{\mathcal{P}}} \sum\nolimits_{j} {({p_j} - \overline{p} ){f_j}({\boldsymbol{w}_j})} $ is the protection function. Eq. (\ref{eq:4-1}) is a semi-infinite program (SIP) which contains infinite constraints and cannot be solved directly \citep{rahimian2019distributionally}{}. Denoting the set of cutting plane parameters in $(t\!+\!1)^{\rm{th}}$ iteration as ${{\bf{A}}^t} \!  \subseteq \!  {\mathbb{R}^N}$, the following function is used to approximate $g(\{ {\boldsymbol{w}_j}\} ) $:
\vspace{-1mm}
\begin{equation}
\begin{aligned}
\overline{g} (\{ {\boldsymbol{w}_j}\} ) = \mathop {\max }\limits_{{\boldsymbol{a}_l} \in {{\bf{A}}^t}} {\boldsymbol{a}^\top_l} {\bf{f}} (\boldsymbol{w}) = \mathop {\max }\limits_{{\boldsymbol{a}_l} \in {{\bf{A}}^t}} \sum\nolimits_{j}  {{a_{l,j}}{{f}_j}({\boldsymbol{w}_j})}, 
\end{aligned}
\end{equation}
where ${\boldsymbol{a}_l = [{a_{l,1}}, \cdots ,{a_{l,N}}]} \! \in \! {\mathbb{R}^N}$ denotes the parameters of $l^{\rm{th}}$ cutting plane in ${{\bf{A}}^t}$ and ${\bf{f}}(\boldsymbol{w}) \! = \! [{{f}_1}(\boldsymbol{w}_1),\cdots, {{f}_N}(\boldsymbol{w}_N)] \! \in \! {\mathbb{R}^N}$. Substituting the protection function $g(\{ {\boldsymbol{w}_j}\} ) $ with $\overline{g} (\{ {\boldsymbol{w}_j}\} )$, we can obtain the following approximate problem:
\vspace{-0.5mm}
\begin{align}
\label{eq:9}
\mathop {{\rm{min}}}\limits_{\boldsymbol{z}\in{{\boldsymbol{\mathcal{Z}}}},\{{\boldsymbol{w}_j}\in{{\boldsymbol{\mathcal{W}}}}\}, h \in {{\boldsymbol{\mathcal{H}}}}} & \quad h \nonumber\\
{\rm{s.t.}} \; \sum\nolimits_{j}\! {(\overline{p}  + {a_{l,j}}){{f}_j}({{\boldsymbol{w}}_j})}&  - h \!\le\! 0,{\rm{     }}\forall {\boldsymbol{a}_l} \! \in \! {{\bf{A}}^t}, \\
\boldsymbol{z}  =    {\boldsymbol{w}_j} , \; j\!=&1,\!\cdots\!, N , \nonumber\\
{\rm{var.}}\quad \boldsymbol{z},{\boldsymbol{w}_1},{\boldsymbol{w}_2}, & \cdots ,{\boldsymbol{w}_N}, h. \nonumber
\end{align}

\section{ASPIRE} 
Distributed optimization is an attractive approach for large-scale learning tasks \citep{yang2008distributed,bottou2018optimization} since it does not require data aggregation, which protects data privacy while also reducing bandwidth requirements \citep{subramanya2021centralized}.  When the neural network models (\textit{i.e.}, ${f_j}({\boldsymbol{w}_j}),\forall j$ are non-convex functions) are used, solving problem in Eq. (\ref{eq:9}) in a distributed manner facing two challenges: 1) Computing the optimal solution to a non-convex subproblem requires a large number of iterations and therefore is highly computationally intensive if not impossible. Thus, the traditional Alternating Direction Method of Multipliers (ADMM) is ineffective. 2) The communication delays of workers may differ significantly \citep{chen2020asynchronous}, thus, asynchronous algorithms are strongly preferred.

To this end, we propose the \textbf{A}synchronous \textbf{S}ingle-loo\textbf{P} alternat\textbf{I}ve g\textbf{R}adient proj\textbf{E}ction (ASPIRE). The advantages of the proposed algorithm include: 1) ASPIRE uses simple gradient projection steps to update variables in each iteration and therefore it is computationally more efficient than the traditional ADMM method, which seeks to find the optimal solution in non-convex (for ${\boldsymbol{w}_j}, \forall j$) and convex (for ${\boldsymbol{z}}$ and $h$) optimization subproblems every iteration,  2) the proposed asynchronous algorithm does not need strict synchronization among different workers. Therefore, ASPIRE remains resilient against communication delays and potential hardware failures from workers. Details of the algorithm are given below. Firstly, we define the node as master which is responsible for updating the global variable $\boldsymbol{z}$, and we define the node which is responsible for updating the local variable ${\boldsymbol{w}_j}$ as worker $j$. In each iteration, the master updates its variables once it receives updates from at least $S$ workers, \textit{i.e.}, active workers, satisfying $1 \le S \le N$. ${{\bf{Q}}^{t + 1}}$  denotes the index subset of workers from which the master  receives updates during $(t+1)^{\rm{th}}$ iteration. We also assume the master will receive updated variables from every worker at least once for each $\tau$ iterations. The augmented Lagrangian function of Eq. (\ref{eq:9}) can be written as:
\begin{align}
\label{eq:11}
\! {L_p}  =  h \! + \! \sum\nolimits_{l}\! {{\lambda _l}(\sum\nolimits_{j}\! {(\overline{p}  + {a_{l,j}}){f_j}({\boldsymbol{w}_j})}  \!-\! h)} \! + \! \sum\nolimits_{j}\! {{\boldsymbol{\phi}^\top_j}\!(\boldsymbol{z}\! -\! {\boldsymbol{w}_j})} \! + \! \sum\nolimits_{j} \! {\frac{{{\kappa _1}}}{2}||\boldsymbol{z} \! - \! {\boldsymbol{w}_j}|{|^2}}\!,  
\end{align}
where ${L_p}\!=\!{L_p}{\rm{(\{ }}{\boldsymbol{w}_j}{\rm{\} ,}}\boldsymbol{z},h,\{ {\lambda _l}\} ,\{ {\boldsymbol{\phi}_j}\})$, ${\lambda _l}\!\in\! {{\boldsymbol{\Lambda}}},\forall l$  and ${\boldsymbol{\phi}_j}\! \in \! {{\boldsymbol{\Phi}}},\forall j$ represent the dual variables of inequality and equality constraints in Eq. (\ref{eq:9}), respectively. ${{\boldsymbol{\Lambda}}}\! \subseteq \! \mathbb{R}^1$ and ${{\boldsymbol{\Phi}}}\! \subseteq \! \mathbb{R}^p$ are nonempty closed convex sets, constant ${\kappa _1} > 0$  is a penalty parameter. Note that Eq. (\ref{eq:11}) does not consider the second-order penalty term for inequality constraint since it will invalidate the distributed optimization. Following \citep{xu2020unified}, the regularized version of Eq. (\ref{eq:11}) is employed to update all variables as follows,
\vspace{-1mm}
\begin{align}
 {\widetilde{L}_p}{\rm{(\{ }}{\boldsymbol{w}_j}{\rm{\} ,}}\boldsymbol{z},h,\{ {\lambda _l}\} ,\{ {\boldsymbol{\phi}_j}\}) 
 ={L_p}- \sum\nolimits_{l} {\frac{{c_1^{t}}}{2}||{\lambda _l}|{|^2}}  - \sum\nolimits_{j} {\frac{{c_2^{t}}}{2}||{\boldsymbol{\phi}_j}|{|^2}},
\end{align}
where $c_1^{t}$ and $c_2^{t}$ denote the regularization terms in $(t+1)^{\rm{th}}$ iteration. To avoid enumerating the whole dataset, the mini-batch loss could be used. A batch of instances with size $m$ can be randomly sampled from each worker during each iteration. The loss function of these instances from $j^{\rm{th}}$ worker is given by ${\hat{f}_j}({\boldsymbol{w}}_j) \! = \! \sum\limits_{i = 1}^m {\frac{1}{m}{{\cal L}_j}({\bf{x}}_j^i,y_j^i;{\boldsymbol{w}}_j)}.$  It is evident that $\mathbb{E}[{\hat{f}_j}({\boldsymbol{w}}_j)]\!=\!{f_j}({\boldsymbol{w}}_j)$ and $\mathbb{E}[\nabla {\hat{f}_j}({\boldsymbol{w}}_j)]\!=\!\nabla{f_j}({\boldsymbol{w}}_j)$. In $(t+1)^{\rm{th}}$ master iteration, the proposed algorithm proceeds as follows.

\textbf{1)} \emph{Active} \emph{workers} update the local variables ${\boldsymbol{w}_j}$ as follows,
\begin{equation}
\label{eq:15}
{\boldsymbol{w}_j^{t+1}} \!=\! \left\{ \begin{array}{l}
{\mathcal{P}_{{\boldsymbol{\mathcal{W}}}}}({\boldsymbol{w}_j^{t}}  -  {\alpha _{\boldsymbol{w}}^{\widetilde{t}_j}}{\nabla _{{\boldsymbol{w}_j}}}{ \widetilde{L}_p}{\rm{(\{ }}{\boldsymbol{w}_j^{\widetilde{t}_j}}{\rm{\} ,}}\boldsymbol{z}^{\widetilde{t}_j},h^{\widetilde{t}_j},\!\{ {\lambda _l^{\widetilde{t}_j}}\} ,\!\{ {\boldsymbol{\phi}_j^{\widetilde{t}_j}}\} {\rm{)}}),\forall j \! \in \! {{\bf{Q}}^{t + 1}},\\
{\boldsymbol{w}_j^t},\forall j \notin {{\bf{Q}}^{t + 1}},
\end{array} \right.
\end{equation}
where $\widetilde{t}_j$ is the last iteration during which worker $j$ was active. It is seen that $\forall j \! \in \! {{\bf{Q}}^{t + 1}}, {\boldsymbol{w}_j^{t}} \!=\! {\boldsymbol{w}_j^{\widetilde{t}_j}} $ and ${\boldsymbol{\phi}_j^{t}} \!=\! {\boldsymbol{\phi}_j^{\widetilde{t}_j}}$. ${\alpha _{\boldsymbol{w}}^{\widetilde{t}_j}}$ represents the step-size and let ${\alpha _{\boldsymbol{w}}^t} \!=\! {\eta _{\boldsymbol{w}}^t}$ when $t\!<\!T_1$ and ${\alpha _{\boldsymbol{w}}^t} \!=\! {\underline{\eta _{\boldsymbol{w}}}}$ when $t \!\ge\! T_1$, where  ${\eta _{\boldsymbol{w}}^t}$ and constant ${\underline{\eta _{\boldsymbol{w}}}}$ will be introduced below. $\mathcal{P}_{{\boldsymbol{\mathcal{W}}}}$ represents the projection onto the closed convex set ${{\boldsymbol{\mathcal{W}}}}$ and we set ${{\boldsymbol{\mathcal{W}}}} = \{ {{{\boldsymbol{w}}_j}}|\;|| {{\boldsymbol{w}}_j}||_{\infty} \! \le {\alpha _1}\} $, ${\alpha _1}$ is a positive constant. And  
then, the active workers ($ j \! \in \! {{\bf{Q}}^{t + 1}}$) transmit their local model parameters ${{\boldsymbol{w}}_j^{t+1}}$ and loss ${f}_j({\boldsymbol{w}_j})$ to the master.

\textbf{2)} After receiving the updates from active workers, the \emph{master} updates the global consensus variable $\boldsymbol{z}$, additional variable $h$ and dual variables ${\lambda _l}$ as follows,
\begin{equation}
\label{eq:16}
\boldsymbol{z}^{t+1}\! =\! {\mathcal{P}_{{\boldsymbol{\mathcal{Z}}}}}({\boldsymbol{z}}^{t} - {\eta _{\boldsymbol{z}}^t}{\nabla _{\boldsymbol{z}}}{\widetilde{L}_p}{\rm{(\{ }}{{\boldsymbol{w}}_j^{t+1}}{\rm{\} ,}}{\boldsymbol{z}}^{t},h^{t},\! \{ {\lambda _l^{t}}\} ,\! \{ {{\boldsymbol{\phi}}_j^{t}}\} {\rm{)}}),
\end{equation}
\begin{equation}
\label{eq:17}
h^{t+1} \!=\! {\mathcal{P}_{{\boldsymbol{\mathcal{H}}}}}(h^{t} - {\eta _h^t}{\nabla _h}{\widetilde{L} _p}{\rm{(\{ }}{{\boldsymbol{w}}_j^{t+1}}{\rm{\} ,}}{\boldsymbol{z}}^{t+1},h^{t},\! \{ {\lambda _l^{t}}\} ,\!\{ {{\boldsymbol{\phi}}_j^{t}}\} {\rm{)}}),
\end{equation}
\begin{equation}
\label{eq:lambda_update}
{\lambda _l^{t+1}} \!=\! {\mathcal{P}_{{\boldsymbol{\Lambda}}} }({\lambda _l^{t}} \! + \! {\rho _1}{\nabla _{{\lambda _l}}}{\widetilde{L}_p}{\rm{(\{ }}{{\boldsymbol{w}}_j^{t+1}}{\rm{\} ,}}{\boldsymbol{z}}^{t+1},h^{t+1},\! \{ {\lambda _l^{t}}\} , \! \{ {{\boldsymbol{\phi}}_j^{t}}\} )), \; l\!=\!1,\!\cdots\!, |{{\bf{A}}^t}|,
\end{equation}

where ${\eta _{\boldsymbol{z}}^t}$, ${\eta _h^t}$ and $\rho_1$ represent the step-sizes. ${\mathcal{P}_{{\boldsymbol{\mathcal{Z}}}}}$, ${\mathcal{P}_{{\boldsymbol{\mathcal{H}}}}}$ and ${\mathcal{P}_{{\boldsymbol{\Lambda}}}}$ respectively represent the projection onto the closed convex sets ${{\boldsymbol{\mathcal{Z}}}}$, ${{\boldsymbol{\mathcal{H}}}}$ and ${{\boldsymbol{\Lambda}}}$. We set ${{\boldsymbol{\mathcal{Z}}}} = \{ {{{\boldsymbol{z}}}}|\;|| {{\boldsymbol{z}}}||_{\infty} \! \le {\alpha _1}\} $, ${{\boldsymbol{\mathcal{H}}}} = \{ h|\; 0 \le \! h \! \le {\alpha _2}\} $ and ${{\boldsymbol{\Lambda}}} = \{ {\lambda _l}|\; 0 \le \! {\lambda _l} \! \le {\alpha _3}\} $, where $\alpha _2$ and $\alpha _3$ are positive constants. $|{{\bf{A}}^t}|$ denotes the number of cutting planes. Then, master broadcasts ${\boldsymbol{z}}^{t+1}$, $h^{t+1}$, $\{{\lambda _l^{t+1}}\}$ to the active workers.

\textbf{3)} \emph{Active} \emph{workers} update the local dual variables ${\boldsymbol{\phi}_j}$ as follows,
\begin{equation}
\label{eq:y_update}
{\boldsymbol{\phi}_j^{t+1}}\! =\! \left\{\! \begin{array}{l}
{\mathcal{P}_{{\boldsymbol{\Phi}}}}({{\boldsymbol{\phi}}_j^{t}}\! +\! {\rho _2}{\nabla _{{{\boldsymbol{\phi}}_j}}}{\widetilde{L}_p}{\rm{(\{ }}{{\boldsymbol{w}}_j^{t+1}}{\rm{\} ,}}{\boldsymbol{z}}^{t+1},h^{t+1},\! \{ {\lambda _l^{t+1}}\} ,\! \{ {{\boldsymbol{\phi}}_j^{t}}\} {\rm{)}}{\rm{)}},\forall j \!\in\! {{\bf{Q}}^{t + 1}},\\
{{\boldsymbol{\phi}}_j^{t}},\forall j \notin {{\bf{Q}}^{t + 1}},
\end{array} \right.
\end{equation}
where $\rho_2$ represents the step-size and  ${\mathcal{P}_{{\boldsymbol{\Phi}}}}$ represents the projection onto the closed convex set  ${{\boldsymbol{\Phi}}}$ and we set ${{\boldsymbol{\Phi}}} = \{ {{{\boldsymbol{\phi}}_j}}|\;|| {{\boldsymbol{\phi}}_j}||_{\infty} \! \le {\alpha _4}\} $, $\alpha _4$ is a positive constant. And master can also obtain $\{{\boldsymbol{\phi}_j^{t+1}}\}$ according to Eq. (\ref{eq:y_update}). It is seen that the projection operation in each step is computationally simple since the closed convex sets have simple structures \citep{bertsekas1997nonlinear}.

\vspace{-0.5mm}
\section{Iterative Active Set Method}\label{sec:ease}
\vspace{-0.5mm}
Cutting plane methods may give rise to numerous linear constraints and lots of extra message passing \citep{yang2014distributed}. Moreover, more iterations are required to obtain the $\varepsilon$-stationary point when the size of a set containing cutting planes increases (which corresponds to a larger $M$), which can be seen in Theorem \ref{theorem 1}. To improve the computational efficiency and speed up the convergence, we consider removing the inactive cutting planes. The proposed it\textbf{E}rative \textbf{A}ctive \textbf{S}\textbf{E}t method (EASE) can be divided into the two steps: during $T_1$ iterations, 1) solving the cutting plane generation subproblem to generate cutting plane, and 2) removing the inactive cutting plane every $k$ iterations, where $k\!>\!0$ is a pre-set constant and can be controlled flexibly.

The cutting planes are generated according to the uncertainty set. For example, if we employ ellipsoid uncertainty set, the cutting plane is generated via solving a SOCP. In this paper, we propose  $CD$-norm uncertainty set, which can be expressed as follows,
\vspace{-1mm}
\begin{align}
\label{eq:D norm}
\boldsymbol{\mathcal{P}} \!= \! \{ {\bf{p}} \!:  - \widetilde{p}_j \! \le \!  p_j -q_j \! \le \! \widetilde{p}_j,   \sum\nolimits_{j} \!|{\frac{{  p_j -q_j }}{ \widetilde{p}_j  }}| \! \le \! \Gamma   ,  {\bf{1}^ \top }{\bf{p}}\! =\! 1  \},
\end{align}

\vspace{-2.5mm}

where  $\Gamma \! \in \! {\mathbb{R}^1}$ can flexibly control the level of robustness, ${\bf{q}}=[q_1, \cdots, q_N] \! \in \! {\mathbb{R}^N}$ represents the prior distribution, $-\widetilde{p}_j$ and $\widetilde{p}_j$ ($\widetilde{p}_j \ge 0$) represent the lower and upper bounds for $p_j -q_j$, respectively. The setting of ${\bf{q}}$ and $\widetilde{p}_j, \forall j$ are based on the prior knowledge. $D$-norm is a classical uncertainty set (which is also called as budget uncertainty set) \citep{bertsimas2004price}. We call Eq. (\ref{eq:D norm}) $CD$-norm uncertainty set since ${\bf{p}}$ is a probability vector so all the entries of this vector are non-negative and add up to exactly one, \textit{i.e.}, ${\bf{1}^ \top }{\bf{p}} = 1$. Due to the special structure of $CD$-norm, the cutting plane generation subproblem is easy to solve and the level of robustness in terms of the outage probability, \textit{i.e.}, probabilistic bounds of the violations of constraints can be flexibly adjusted via a single parameter $\Gamma$. We claim that $l_1$-norm (or twice total variation distance) uncertainty set is closely related to $CD$-norm uncertainty set. Nevertheless, there are two differences: 1) $CD$-norm uncertainty set could be regarded as a weighted $l_1$-norm with additional constraints. 2) $CD$-norm uncertainty set can flexibly set the lower and upper bounds for every $p_j$ (\textit{i.e.}, $q_j \! - \! \widetilde{p}_j  \! \le \!  p_j \! \le \! p_j \! + \! \widetilde{p}_j$), while $0 \! \le \! p_j  \! \le \! 1, \forall j$ in $l_1$-norm uncertainty set. Based on the $CD$-norm uncertainty set, the cutting plane can be derived as follows,

1) Solve the following problem,
\vspace{-1mm}
\begin{align}
\label{eq:18}
& {\bf{p}}^{t + 1}  = \mathop {\arg \max }\limits_{  {p_1},\cdots,{p_N} }  \sum\nolimits_{j} {({p_j}  - \overline{p}  ){{ f}_j}({{\boldsymbol{w}_j}})} \nonumber \vspace{-3mm}\\
{\rm{s.t.}}\;  \sum\nolimits_{j} & |{\frac{{  p_j \!- \!q_j }}{ \widetilde{p}_j  }}|  \! \le \!  \Gamma,\;  - \widetilde{p}_j \! \le \!   p_j \!- \! q_j  \! \le \!  \widetilde{p}_j, \forall{j},\; \sum\nolimits_{j}\! {p_j} \! = \! 1  \\
&{\rm{var.}}\quad \quad \quad \quad \; {p_1},\cdots,{p_N},  \nonumber
\end{align}
where ${\bf{p}}^{t + 1} \! = \! [ p_1^{t + 1},\! \cdots, p_N^{t + 1}]\! \in \! {\mathbb{R}^N} $. Let $\widetilde{{\bf{a}}}^{t + 1} \!= \! {\bf{p}} ^{t + 1} - \overline{{\bf{p}}}  $, where $\overline{{\bf{p}}}    = [\overline{p}  , \cdots ,\overline{p} ] \! \in \! {\mathbb{R}^N}$. This first step aims to obtain the distribution $\widetilde{{\bf{a}}}^{t+1}$  by solving problem in Eq. (\ref{eq:18}). This problem can be effectively solved through combining merge sort \citep{cole1988parallel} (for sorting $\widetilde{p}_j{f}_j({{\boldsymbol{w}_j}}), j\!=\!1,\cdots,N$) with few basic arithmetic operations (for obtaining $p_j^{t + 1}, j\!=\!1,\cdots,N$). Since $N$ is relatively large in distributed system, the arithmetic complexity of solving problem in Eq. (\ref{eq:18}) is dominated by merge sort, which can be regarded as $\mathcal{O}(N\log (N))$.
 
2) Let ${\bf{f}}(\boldsymbol{w}) \! =\! [{{f}_1}(\boldsymbol{w}_1),\cdots , {{f}_N}(\boldsymbol{w}_N)]\!\in\!{\mathbb{R}^N}$, check the feasibility of the following constraints:
\begin{equation}
\label{eq:19}
{\widetilde{{\bf{a}}}^{t+1}}{}^\top {\bf{f}}({\boldsymbol{w}}) \! \le \!\mathop {\max }\limits_{{\boldsymbol{a}_l} \in {{\bf{A}}^t}} {\boldsymbol{a}_l}{}^\top {\bf{f}} (\boldsymbol{w}).
\end{equation}

\vspace{-2mm}

3) If Eq. (\ref{eq:19}) is violated,  $\widetilde{{\bf{a}}}^{t+1}$ will be added into ${{\rm \bf{A}}^t}$:
\begin{equation}
\label{eq:cutting set}
{{\rm \bf{A}}^{t + 1}} = \left\{ \begin{array}{l}
{{\rm \bf{A}}^t} \cup \{ \widetilde{{\bf{a}}}^{t+1} \},{\rm{ if \;  Eq. (\ref{eq:19}) \; is \; violated }},\\
{{\rm \bf{A}}^t},{\rm{ otherwise}},
\end{array} \right.
\end{equation}
when a new cutting plane is added, its corresponding dual variable ${\lambda _{|{{\bf{A}}^t}| + 1}^{t+1}}=0$ will be generated. After the cutting plane subproblem is solved, the inactive cutting plane will be removed, that is:
\begin{equation}
\label{eq:active}
{{\rm \bf{A}}^{t + 1}} =\left\{ \begin{array}{l}
\complement_{ {{\rm \bf{A}}^{t+1}}} \{{\boldsymbol{a}_l}\} ,{\rm{if}} \; \lambda_l^{t+1}\!=\!0 \, {\rm{and}}\, \lambda_l^{t} \!=\! 0, 1\!\le\!l \!\le\! |{{\bf{A}}^t}|, \\
{{\rm \bf{A}}^{t+1}},{\rm{ otherwise}},
\end{array} \right.
\end{equation}
where $\complement_{ {{\rm \bf{A}}^{t+1}}}\{{\boldsymbol{a}_l}\}$ is the complement of $\{{\boldsymbol{a}_l}\}$ in ${ {{\rm \bf{A}}^{t+1}}}$, and the dual variable will be removed. Then master broadcasts ${{\rm \bf{A}}^{t + 1}}$, $\{ \lambda_l^{t+1} \}$ to all workers.  Details of algorithm are summarized in Algorithm \ref{algorithm1}.

\begin{algorithm}[tb]
   \caption{ASPIRE-EASE}
   \label{algorithm1}
\begin{algorithmic}
   \STATE {\bfseries Initialization:}  iteration $t = 0$, variables $\{{{\boldsymbol{w}}_j^0}\}$, ${\boldsymbol{z}}^0$, ${h}^0$, $\{{\lambda _l^0}\}$, $\{{{\boldsymbol{\phi}}_j^0}\}$  and set ${{\bf{A}}^0} $.
   \REPEAT
   \FOR{\emph{active worker}}

   \STATE updates local  ${\boldsymbol{w}_j^{t+1}}$ according to Eq. (\ref{eq:15});
   \ENDFOR
   
   \STATE \emph{active workers} transmit local model parameters and loss to \emph{master};

   \STATE \emph{master} receives updates from \emph{active workers}  \textbf{do}

   \STATE  \quad  updates ${\boldsymbol{z}}^{t+1}$, $h^{t+1}$, $\{{\lambda _l^{t + 1}}\}$, $\{{\boldsymbol{\phi}_j^{t+1}}\}$ in master according to Eq. (\ref{eq:16}), (\ref{eq:17}), (\ref{eq:lambda_update}), (\ref{eq:y_update});

    \STATE  \emph{master} broadcasts ${\boldsymbol{z}}^{t+1}$, $h^{t+1}$, $\{{\lambda _l^{t+1}}\}$ to \emph{active workers};
    
    \FOR{\emph{active worker}}
   \STATE updates local  ${\boldsymbol{\phi}_j^{t+1}}$ according to Eq. (\ref{eq:y_update});
   \ENDFOR
   
   \IF{$(t+1)$ mod $k$ $==$ 0 and $t< T_1$}
   \STATE   \emph{master} updates ${{\rm \bf{A}}^{t + 1}}$ according to Eq. (\ref{eq:cutting set}) and (\ref{eq:active}), and broadcast parameters to all workers;
   
   \ENDIF
  
   \STATE $t =t+1$;
   \UNTIL{convergence}
\end{algorithmic}
\end{algorithm}
	
	\vspace{-2mm}
    \section{Convergence Analysis}\label{convergence analysis}

\vspace{-0.2cm}
\begin{definition}
{\rm{(Stationarity gap)}} Following \citep{xu2020unified,lu2020hybrid,xu2021zeroth}, the \textit{stationarity} \textit{gap} of our problem at $t^{{th}}$ iteration is defined as:
\begin{equation}
\nabla   G^t \! = \! \left[ \begin{array}{l}
   \{   \frac{1}{{{\alpha _{\boldsymbol{w}}^t}}}  (   {{\boldsymbol{w}}_j^t} \! -\!  {\mathcal{P}_{{\boldsymbol{\mathcal{W}}}}} ( {{\boldsymbol{w}}_j^t}  \! -\!  {\alpha _{\boldsymbol{w}}^t} {\nabla _{{{\boldsymbol{w}}_j}}}       {L_p}{\rm{(\{ }}{{\boldsymbol{w}}_j^t}{\rm{\} ,}}{\boldsymbol{z}^t},h^t,\{ {\lambda _l^t}\} ,\{ {{\boldsymbol{\phi}}_j^t}\} {\rm{)}} {\rm{   )  )     \}  } } \vspace{0.3ex}\\
\, \frac{1}{{{\eta _{\boldsymbol{z}}^t}}}   (  {\boldsymbol{z}}^t   \! - \!   {\mathcal{P}_{{\boldsymbol{\mathcal{Z}}}}}(  {\boldsymbol{z}}^t   \! -  \!  {\eta _{\boldsymbol{z}}^t} {\nabla _{\boldsymbol{z}}}       {L_p}{\rm{(\{ }}{{\boldsymbol{w}}_j^t}{\rm{\} ,}}{\boldsymbol{z}^t},h^t,\{ {\lambda _l^t}\} ,\{ {{\boldsymbol{\phi}}_j^t}\} {\rm{)}}  ) ) \vspace{0.3ex}\\
\, \frac{1}{{{\eta _h^t}}}   (   h^t  \!- \! {\mathcal{P}_{{\boldsymbol{\mathcal{H}}}}}(  h^t \! - \! {\eta _h^t} {\nabla _h} {L_p}{\rm{(\{ }}{{\boldsymbol{w}}_j^t}{\rm{\} ,}}{\boldsymbol{z}^t},h^t,\{ {\lambda _l^t}\} ,\{ {{\boldsymbol{\phi}}_j^t}\} {\rm{)}}  ) )  \vspace{0.3ex}\\
\{  \frac{1}{{{\rho _1}}}( {\lambda _l^t}  \! -\!   {\mathcal{P}_{{\boldsymbol{\Lambda}}} }( {\lambda _l^t}  \! + \!   {\rho _1}{\nabla _{{\lambda _l}}} {L_p}{\rm{(\{ }}{{\boldsymbol{w}}_j^t}{\rm{\} ,}}{\boldsymbol{z}^t},h^t,\{ {\lambda _l^t}\} ,\{ {{\boldsymbol{\phi}}_j^t}\} {\rm{)}} {\rm{   )  )     \}  } } \vspace{0.3ex}\\
\{   \frac{1}{{{\rho _2}}}( {{\boldsymbol{\phi}}_j^t}  \! - \!  {\mathcal{P}_{{\boldsymbol{\Phi}}}}( {{\boldsymbol{\phi}}_j^t}  \! + \!  {\rho _2}{\nabla _{{{\boldsymbol{\phi}}_j}}} {L_p}{\rm{(\{ }}{{\boldsymbol{w}}_j^t}{\rm{\} ,}}{\boldsymbol{z}^t},h^t,\{ {\lambda _l^t}\} ,\{ {{\boldsymbol{\phi}}_j^t}\} {\rm{)}} {\rm{   )  )     \}   } }  \end{array} \right],
\end{equation}
\noindent where $\nabla G^t$ is the simplified form of $\nabla G{\rm{(\{ }}{{\boldsymbol{w}}_j^t}{\rm{\} ,}}{\boldsymbol{z}^t},h^t,\{ {\lambda _l^t}\} ,\{ {{\boldsymbol{\phi}}_j^t}\} {\rm{)}}$.
 \end{definition}
 
\vspace{-0.5cm}

\begin{definition}
{\rm{($\varepsilon$-stationary point)}}  ${\rm{(\{ }}{{\boldsymbol{w}}_j^t}{\rm{\} ,}}{\boldsymbol{z}^t},h^t,\{ {\lambda _l^t}\} ,\{ {{\boldsymbol{\phi}}_j^t}\} {\rm{)}}$ is an $\varepsilon$-stationary point ($\varepsilon  \ge 0$) of a differentiable function ${L_p}$,  if $\,||\nabla G^t|| \le \varepsilon $.  $T(\varepsilon )$ is the first iteration index such that $||\nabla G^t|| \! \le \! \varepsilon$, \textit{i.e.}, $T(\varepsilon ) \! = \! \min \{ t \ |\; ||\nabla G^t|| \! \le \! \varepsilon \}  $.
\end{definition}

\vspace{-0.4cm}

\begin{assumption}\label{assumption:1}
{\rm{(Smoothness/Gradient Lipschitz)}} $L_p$ has Lipschitz continuous gradients. We assume that there exists $L>0$ satisfying

\centerline{$\begin{array}{l}
||{\nabla  _\theta }{L_p}( \{ {{\boldsymbol{w}}_j}\},{\boldsymbol{z}},h, \! \{ {\lambda _l}\}, \!\{ {{\boldsymbol{\phi}}_j}\} ) \! - \! {\nabla  _\theta }{L_p}( \{ {\hat{\boldsymbol{w}}_j}\},\hat{\boldsymbol{z}},\hat{h}, \! \{ {\hat{\lambda }_l}\}, \!\{ {\hat{\boldsymbol{\phi}}_j}\}  )|| \vspace{0.5ex}\\
 \le L||[ {{\boldsymbol{w}}_{\rm{cat}}} \! - \!  {{\hat{\boldsymbol{w}}_{\rm{cat}}}}   ;{\boldsymbol{z}} \! - \!{\hat{\boldsymbol{z}}}  ;h \! - \! \hat{h} ; {\boldsymbol{\lambda} _{\rm{cat}}} \! - \! {{\hat{\boldsymbol{\lambda}} _{\rm{cat}}}}  ; {{\boldsymbol{\phi}}_{\rm{cat}}} \! - \! {{\hat{\boldsymbol{\phi}}_{\rm{cat}}}}  ]||,
\end{array}$}

where $\theta \!  \in \! \{ \{ {{\boldsymbol{w}}_j}\} ,{\boldsymbol{z}},h,\{ {\lambda _l}\} ,\{ {{\boldsymbol{\phi}}_j}\} \} $ and $[;]$ represents the concatenation. ${{\boldsymbol{w}}_{\rm{cat}}} \!-\! {{\hat{\boldsymbol{w}}_{\rm{cat}}}}   \!=\![{\boldsymbol{w}}_1 \! - \! {\hat{\boldsymbol{w}}_1};\cdots;{\boldsymbol{w}}_N \! - \! {\hat{\boldsymbol{w}}_N}] \! \in \! {\mathbb{R}^{pN}}$, ${\boldsymbol{\lambda} _{\rm{cat}}} \! - \! {{\hat{\boldsymbol{\lambda}} _{\rm{cat}}}} \!=\! [{\lambda_1} \! - \! {\hat{\lambda}_1};\cdots;{\lambda}_{|{{\bf{A}}^t}|} \! - \! {\hat{\lambda}_{|{{\bf{A}}^t}|}}]\! \in \! {\mathbb{R}^{|{{\bf{A}}^t}|}}$, ${{\boldsymbol{\phi}}_{\rm{cat}}}\!-\!{{\hat{\boldsymbol{\phi}}_{\rm{cat}}}} \!=\![{\boldsymbol{\phi}}_1 \! - \! {\hat{\boldsymbol{\phi}}_1};\cdots;{\boldsymbol{\phi}}_N \! - \! {\hat{\boldsymbol{\phi}}_N}]\! \in \! {\mathbb{R}^{pN}}$.
\end{assumption}

\vspace{-3.5mm}

\begin{assumption}\label{assumption:4}
{\rm{(Boundedness)}} Before obtaining the $\varepsilon$-stationary point (i.e., $t\!\le\! T(\varepsilon )\!-\!1$), we assume variables in master satisfy that $||\boldsymbol{z}^{t+1} \!-\! \boldsymbol{z}^t|{|^2}\!+\!||h^{t + 1}\! -\! h^t|{|^2}\!+\!\sum\nolimits_l||{\lambda _l^{t + 1}} \!-\! {\lambda _l^t}|{|^2} \ge \vartheta $, where $\vartheta >0$ is a relative small constant. The change of the variables in master is upper bounded within $\tau$ iterations:

\vspace{0.5mm}

\centerline{ $\begin{array}{*{20}{l}}
{||\boldsymbol{z}^t - \boldsymbol{z}^{t - k}|{|^2} \! \le \! \tau{k_1}\vartheta}, \;\;
{||h^t - h^{t - k}|{|^2} \! \le \! \tau{k_1}\vartheta},\;\;
{\sum\nolimits_l||{\lambda _l^t} - {\lambda _l^{t - k}}|{|^2} \! \le \! \tau{k_1}\vartheta}, {\forall 1 \! \le \! k \! \le \! \tau },
\end{array}$}

\vspace{-0.8mm}

where $k_1 >0$ is a constant.

\end{assumption}

\vspace{-3.5mm}

\begin{setting}\label{assumption:3}
{\rm{(Bounded $|{{\bf{A}}^t}|$)}} $|{{\bf{A}}^t}| \le M,{\rm{    }}\forall t$, \textit{i.e.}, an upper bound is set for the number of cutting planes.
\end{setting}

\vspace{-3.5mm}

\begin{setting}\label{assumption:2}
{\rm{(Setting of ${c_1^t}$, ${c_2^t}$)}}
${c_1^t}\! = \!\frac{1}{{{\rho _1}{(t+1)^{\frac{1}{6}}}}} \!\ge\! \underline{c}_1$ and ${c_2^t} \!=\! \frac{1}{{{\rho _2}{(t+1)^{\frac{1}{6}}}}} \!\ge\! \underline{c}_2$ are nonnegative non-increasing sequences, where $\underline{c}_1$ and $\underline{c}_2$ are positive constants and meet $M{\underline{c}_1}^2+N{\underline{c}_2}^2 \le \frac{{{\varepsilon ^2}}}{4}$.
\end{setting}

\vspace{-3.5mm}

\begin{theorem}
\label{theorem 1}
{\rm{(Iteration complexity)}} Suppose Assumption 1 and 2 hold. We set  
${\eta _{\boldsymbol{w}}^t} = {\eta _{\boldsymbol{z}}^t} = {\eta _h^t} = \frac{2}{{L + {\rho _1}|{{\bf{A}}^t}|{L^2} + {\rho _2}N{L^2} + 8(\frac{{|{{\bf{A}}^t}|\gamma {L^2}}}{{{\rho _1}({c_1^t})^2}} + \frac{{N\gamma {L^2}}}{{{\rho _2}({c_2^t})^2}})}}$ and $\underline{\eta _{\boldsymbol{w}}}=\frac{2}{{L + {\rho _1}M{L^2} + {\rho _2}N{L^2} + 8(\frac{{M\gamma {L^2}}}{{\rho _1}{\underline{c}_1}^2} + \frac{{N\gamma {L^2}}}{{\rho _2}{\underline{c}_2}^2})}}$. And we set constants  ${\rho _1} \!< \!\min \{\frac{ 2}{{L + 2c_1^0}}  ,\frac{1}{{15\tau{k_1}N{L^2}}}\} $ and $ \rho _2 \!\le\! \frac{2}{{L + 2c_2^0}} $, respectively. For a given $\varepsilon $, we have:
\vspace{-2mm}
{\begin{equation}
     T( \varepsilon ) \! \sim  \! \mathcal{O}(\max  \{ {(\frac{{4M\!{\sigma _1}^2}}{{{\rho _1}^2}}\! +\! \frac{{4N\!{\sigma _2}^2}}{{{\rho _2}^2}}\!)^3}\!\frac{1}{{{\varepsilon ^6}}}, 
{(\frac{{4{{{(d_6+ \frac{{{\rho _2}(N  -  S){{L}^2}}}{2}\!)}\!}^2}\! (\mathop d\limits^ -  +  k_d(\tau \! - \! 1))  {d_5}}}{{{\varepsilon ^2}}}\! +\! (T_1\!+\!\tau)^{\frac{1}{3}})^3}\}), 
\end{equation}}

\vspace{-3.2mm}

where ${\sigma _1}$, ${\sigma _2}$, $\gamma$, $\tau$, $k_d$, $\mathop d\limits^ -  $, ${d_5}$, ${d_6}$ and ${T_1}$ are constants. The detailed proof is given in Appendix \ref{appendix:Theorem1}.
\end{theorem}

There exists a wide array of works
regarding the convergence analysis of various algorithms for nonconvex/convex optimization problems involved in machine learning \citep{jin2020local,xu2021zeroth}.  Our analysis, however, differs from existing works in two aspects.  First, we solve the non-convex PD-DRO in an \emph{asynchronous} \emph{distributed} \emph{manner}. To our best knowledge, there are few works focusing on solving the DRO in a distributed manner. Compared to solving the non-convex PD-DRO in a centralized manner, solving it in an \emph{asynchronous} \emph{distributed} \emph{manner} poses significant challenges in algorithm design and convergence analysis. Secondly, we do not assume the inner problem can be solved nearly optimally for each outer iteration, which is numerically difficult to achieve in practice \citep{bertsekas1997nonlinear}. Instead, ASPIRE-EASE is \emph{single} \emph{loop} and involves simple gradient projection operation at each step.

\vspace{-2mm}
\section{Experiment} \label{experiment}
\vspace{-2mm}
In this section, we conduct experiments on four real-world datasets to assess the performance of the proposed method. Specifically, we evaluate the robustness against data heterogeneity, robustness against malicious attacks and efficiency of the proposed method. Ablation study is also carried out to demonstrate the excellent performance of ASPIRE-EASE.
\vspace{-2.3mm}
\subsection{Datasets and Baseline Methods}\label{dataset}
\vspace{-2.1mm}
We compare the proposed ASPIRE-EASE with baseline methods based on SHL~\citep{gjoreski2018university}, Person Activity~\citep{kaluvza2010agent}, Single Chest-Mounted Accelerometer (SM-AC)~\citep{casale2012personalization} and Fashion MNIST~\citep{xiao2017fashion} datasets. The baseline methods include ${\rm{Ind}}_j$ (learning the model from an individual worker $j$), ${\rm{Mix}}{\rm{_{Even}}}$ (learning the model from all workers with even weights using ASPIRE),  FedAvg \citep{mcmahan2017communication}, AFL \citep{mohri2019agnostic} and DRFA-Prox \citep{deng2021distributionally}. The detailed descriptions of datasets and baselines are given in Appendix \ref{appendix:experiment}.

\vspace{-0.4mm}

In our empirical studies, since the downstream tasks are multi-class classification, the cross entropy loss is used on each worker (\textit{i.e.}, ${\mathcal{L}_j}( \cdot ),\forall j$). For SHL, Person Activity, and SM-AC datasets, we adopt the deep multilayer perceptron~\citep{wang2017time} as the base model. And we use the same logistic regression model as in \citep{mohri2019agnostic,deng2021distributionally} for Fashion MNIST dataset. The base models are trained with SGD. More details are given in Appendix \ref{appendix:experiment}. Following related works in this direction~\citep{qian2019robust,mohri2019agnostic,deng2021distributionally}, worst case performance are reported for the comparison of robustness. Specifically, we use {${\bf{Acc}}_{w}$} and {${\bf{Loss}}_{w}$} to represent the worst case test accuracy and training loss (\textit{i.e.}, the test accuracy and training loss on the worker with worst performance), respectively. We also report the standard deviation ${\bf{Std}}$ of $[\rm{Acc}_1,\cdots , \rm{Acc}_N]$ (the test accuracy on every worker). In the experiment, $S$ is set as 1, that means the master will make an update once it receives a message.  Each experiment is repeated 10 times, both mean and standard deviations are reported. We implement our model with PyTorch and conduct all the experiments on a server with two TITAN V GPUs.

\renewcommand\arraystretch{1.3}
\renewcommand\tabcolsep{4pt}
\begin{table}[t]
\caption{Performance comparisons based on {${\bf{Acc}}_{w}$} (\%) $\uparrow$, ${\bf{Loss}}_{w}$ $\downarrow$ and ${\bf{Std}}$ $\downarrow$  ($\uparrow$ and $\downarrow$ respectively denote higher scores represent better performance and lower scores represent better performance). The boldfaced digits represent the best results, ``$-$'' represents not available.}
\centering
\label{tab:III}
\scalebox{0.61}{
\begin{tabular}{l|cccccccccccc}
\toprule
\multirow{2}{*}{Model}&  
    \multicolumn{3}{c}{SHL}&\multicolumn{3}{c}{Person Activity}&\multicolumn{3}{c}{SC-MA}&\multicolumn{3}{c}{Fashion MNIST}\cr  
    \cmidrule(lr){2-4} \cmidrule(lr){5-7} 
    \cmidrule(lr){8-10}
    \cmidrule(lr){11-13}
    & ${\bf{Acc}}_{w}$$\uparrow$  & ${\bf{Loss}}_{w}$$\downarrow$  & {\bf{Std}}$\downarrow$  & ${\bf{Acc}}_{w}$$\uparrow$ & ${\bf{Loss}}_{w}$ $\downarrow$ & {\bf{Std}}$\downarrow$  & ${\bf{Acc}}_{w}$ $\uparrow$  & ${\bf{Loss}}_{w}$ $\downarrow$ & {\bf{Std}}$\downarrow$   & ${\bf{Acc}}_{w}$ $\uparrow$  & ${\bf{Loss}}_{w}$ $\downarrow$ & {\bf{Std}}$\downarrow$ \\ \hline
${\rm{max}}\{{\rm{Ind}}_j\}$  & 19.06±0.65 &  $-$  &  29.1   & 49.38±0.08  & $-$  &  8.32  & 22.56±0.78  & $-$  &   17.5  & $-$   &  $-$ &  $-$   \\
${\rm{Mix}}{\rm{_{Even}}}$   & 69.87±3.10  & 0.806±0.018  &  4.81    & 56.31±0.69  & 1.165±0.017  & 3.00     & 49.81±0.21  &  1.424±0.024 &   6.99   & 66.80±0.18   & 0.784±0.003  &   10.1   \\ 
FedAvg \citep{mcmahan2017communication}  & 69.96±3.07  & 0.802±0.023  &   5.21   & 56.28±0.63  & 1.154±0.019  &   3.13   & 49.53±0.96  & 1.441±0.015  &   7.17  & 66.58±0.39   & 0.781±0.002  &   10.2   \\ 
AFL  \citep{mohri2019agnostic}    & 78.11±1.99   & 0.582±0.021  &   1.87   & 58.39±0.37   &  1.081±0.014 &    0.99   & 54.56±0.79   & 1.172±0.018  &   3.50   & 77.32±0.15   & 0.703±0.001  &   1.86   \\
DRFA-Prox \citep{deng2021distributionally}    & 78.34±1.46   & 0.532±0.034  &   1.85    & 58.62±0.16  & 1.096±0.037  &   1.26   & 54.61±0.76  &  1.151±0.039 &  4.69   & 77.95±0.51 &  0.702±0.007 &   1.34   \\ \hline
ASPIRE-EASE     & \textbf{79.16±1.13}  & \textbf{0.515±0.019}  &  \textbf{1.02}    & 59.43±0.44  &   1.053±0.010 &  0.82    &  56.31±0.29 &  1.127±0.021 &  \textbf{3.16}  &     \textbf{78.82±0.07}  &  \textbf{0.696±0.004} &   \textbf{1.01}   \cr
ASPIRE-EASE$_{\rm{per}}$     &  78.94±1.27    &  0.521±0.023 &  1.36      & \textbf{59.54±0.21}  &   \textbf{1.051±0.016} &  \textbf{0.79}    &  \textbf{56.71±0.16} &  \textbf{1.119±0.028} &  3.48  &     78.73±0.06  &  0.698±0.006 &   1.09 \cr
\bottomrule  
\end{tabular}}
\vspace{-3mm}
\end{table}

\vspace{-2mm}

\subsection{Results}
\vspace{-2mm}
\textbf{Robustness against Data Heterogeneity}. $\;$
We first assess the robustness of the proposed ASPIRE-EASE by comparing it with baseline methods when data are heterogeneously distributed across different workers. Specifically, we compare the ${\bf{Acc}}_{w}$, ${\bf{Loss}}_{w}$ and ${\bf{Std}}$ of different methods on all datasets. The performance comparison results are shown in Table~\ref{tab:III}.  In this table, we can observe that ${\rm{max}}\{{\rm{Ind}}_j\}$, which represents the best performance of individual training over all workers, exhibits the worst robustness on SHL, Person Activity, and SC-MA. This is because individual training (${\rm{max}}\{{\rm{Ind}}_j\}$) only learns from the data in its local worker and cannot generalize to other workers due to different data distributions. Note that ${\rm{max}}\{{\rm{Ind}}_j\}$ is unavailable for Fashion MNIST since each worker only contains one class of data and cross entropy loss cannot be used in this case. ${\rm{max}}\{{\rm{Ind}}_j\}$ also does not have ${\bf{Loss}}_{w}$, since ${\rm{Ind}}_j$ is trained only on individual worker $j$. The FedAvg and ${\rm{Mix}}{\rm{_{Even}}}$ exhibit better performance than ${\rm{max}}\{{\rm{Ind}}_j\}$ since they consider the data from all workers. Nevertheless, FedAvg and  ${\rm{Mix}}{\rm{_{Even}}}$ only assign the fixed weight for each worker. 
AFL is more robust than FedAvg and  ${\rm{Mix}}{\rm{_{Even}}}$ since it not only utilizes the data from all workers but also considers optimizing the weight of each worker. DRFA-Prox outperforms AFL since it also considers the prior distribution and regards it as a regularizer in the objective function. Finally, we can observe that the proposed ASPIRE-EASE shows excellent robustness, which can be attributed to two factors: 1) ASPIRE-EASE considers data from all workers and can optimize the weight of each worker;
2) compared with DRFA-Prox which uses prior distribution as a regularizer, the prior distribution is incorporated within the constraint in our formulation (Eq. \ref{eq:3}), which can be leveraged more effectively. And it is seen that ASPIRE-EASE can perform periodic communication since ASPIRE-EASE$_{\rm{per}}$, which represents ASPIRE-EASE with periodic communication, also has excellent performance.

Within ASPIRE-EASE, the level of robustness can be controlled by adjusting $\Gamma$. Specially, when $\Gamma=0$, we obtain a nominal optimization problem in which no adversarial distribution is considered. The size of the uncertainty set will increase with $\Gamma$ (when $\Gamma \le N$), which enhances the adversarial robustness of the model. As shown in Figure \ref{fig:gamma}, the robustness of ASPIRE-EASE can be gradually enhanced when  $\Gamma$ increases. More results are available in Figure \ref{appendix:fig:gamma} of Appendix \ref{appendix:experiment}.

\textbf{Robustness against Malicious Attacks}. $\;$
To assess the model robustness against malicious attacks, malicious workers with backdoor attacks \citep{bagdasaryan2020backdoor,wang2019neural}, which attempt to mislead the model training process, are added to the distributed system.  Following \citep{dai2019backdoor}, we report the success attack rate of backdoor attacks for comparison. It can be calculated by checking how many instances in the backdoor dataset can be misled and categorized into the target labels. Lower success attack rates indicate more robustness against backdoor attacks.  The comparison results are summarized in Table~\ref{tab:SAR} and more detailed settings of backdoor attacks are available in Appendix \ref{appendix:experiment}.  In Table~\ref{tab:SAR}, we observe that AFL can be attacked easily since it could assign higher weights to malicious workers. Compared to AFL, FedAvg and ${\rm{Mix}}{\rm{_{Even}}}$ achieve relatively lower success attack rates since they assign equal weights to the malicious workers and other workers. DRFA-Prox can achieve even lower success attack rates since it can leverage the prior distribution to assign lower weights for malicious workers.  The proposed ASPIRE-EASE achieves the lowest success attack rates since it can leverage the prior distribution more effectively. Specifically, it will assign lower weights to malicious workers with tight theoretical guarantees.

\makeatletter\def\@captype{figure}\makeatother 
\begin{minipage}{0.48\textwidth}  
\vspace{-3mm}
\subfigure[Person Activity] 
{\begin{minipage}[t]{0.49\linewidth}
	\centering      
	\includegraphics[scale=0.22]{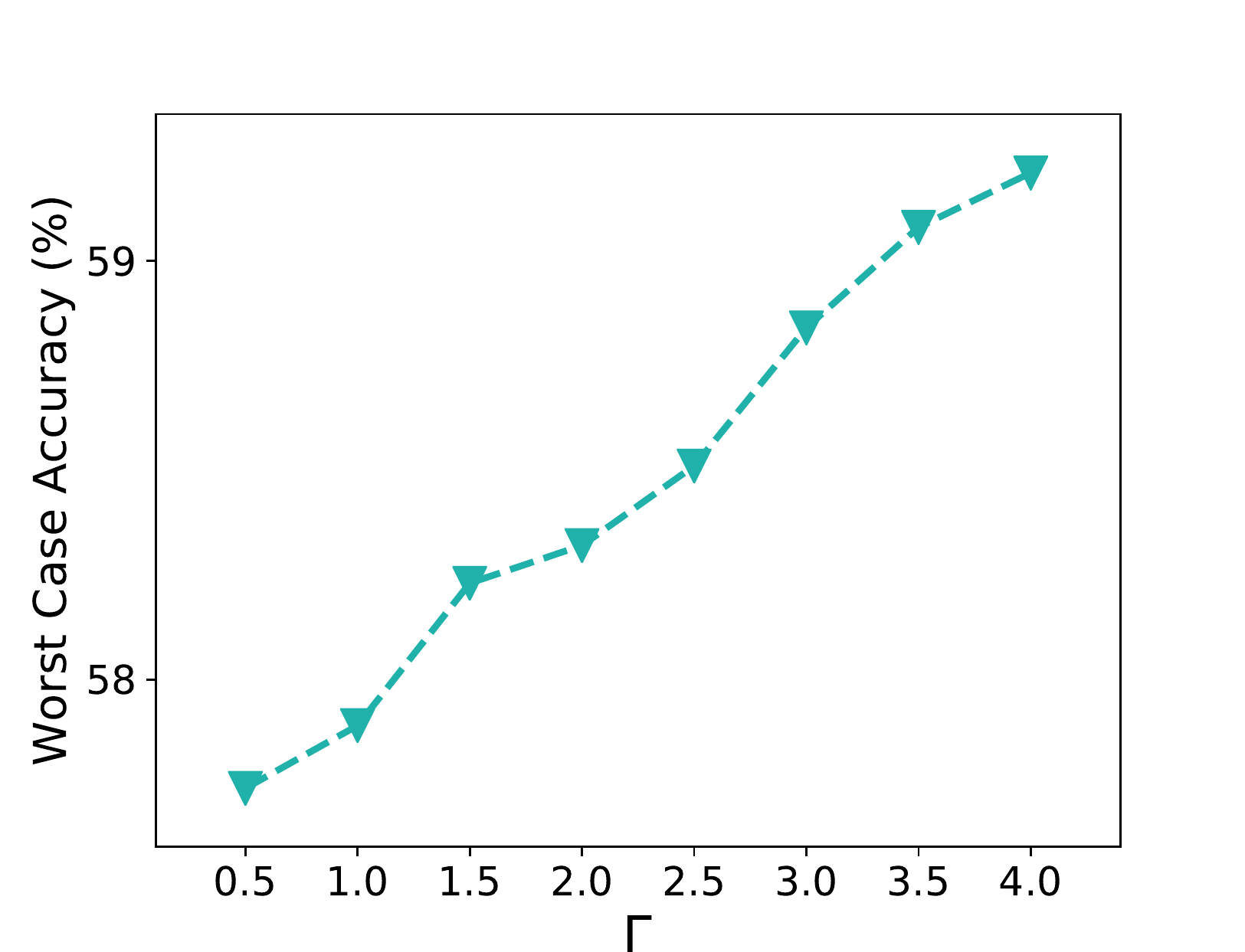}   
	\end{minipage}}
\subfigure[SC-MA] 
{\begin{minipage}[t]{0.49\linewidth}
	\centering      
	\includegraphics[scale=0.22]{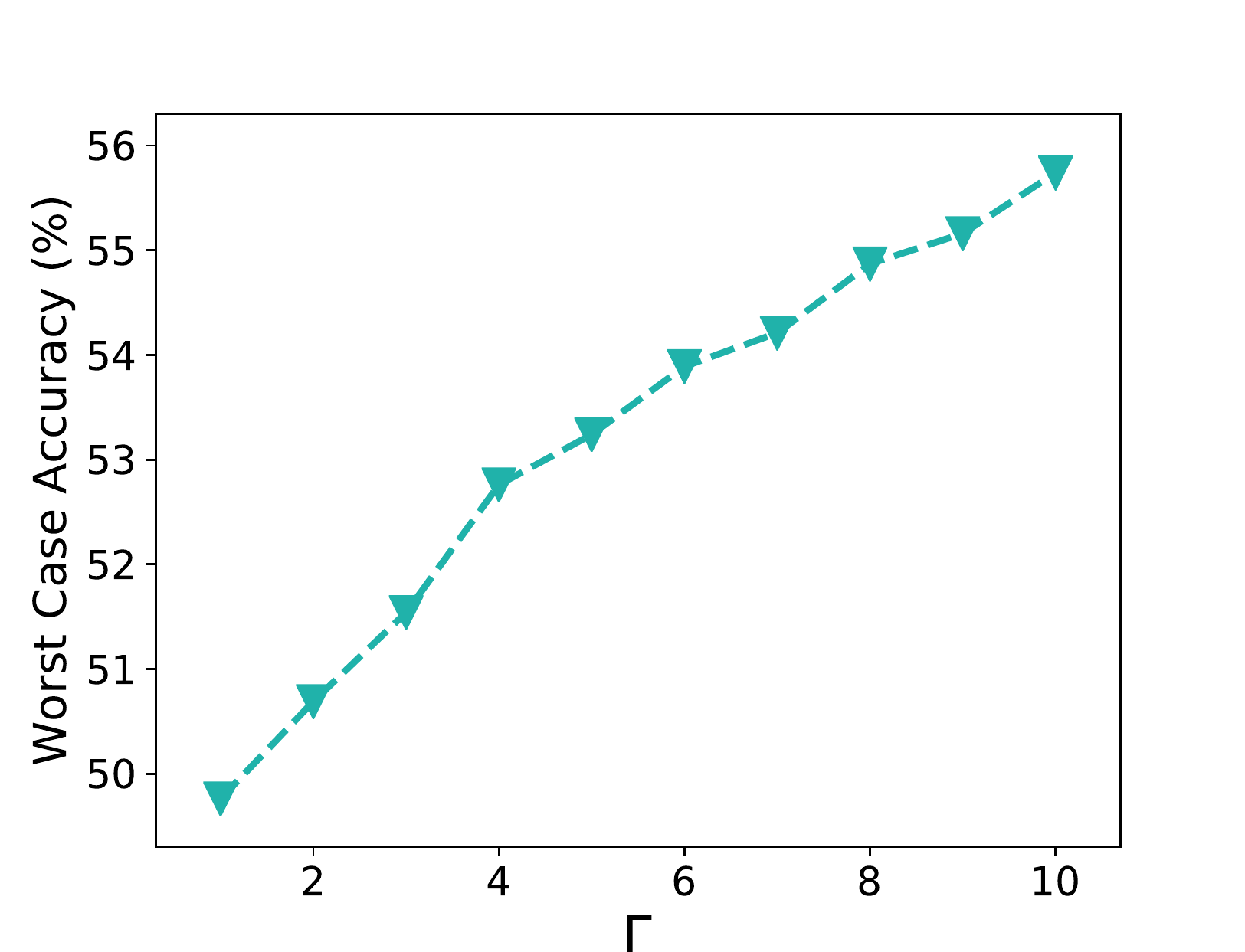}   
	\end{minipage}}
\caption{$\Gamma$ control the degree of robustness (worst case performance in the problem) on  (a) Person Activity, (b) SC-MA  datasets.} 
\label{fig:gamma}  
\vspace{-2mm}
\end{minipage} $\;$
\makeatletter\def\@captype{figure}\makeatother 
\begin{minipage}{0.48\textwidth} 
\vspace{-3.55mm}
\subfigure[Person Activity] 
{\begin{minipage}[t]{0.49\linewidth}
	\centering      
	\includegraphics[scale=0.22]{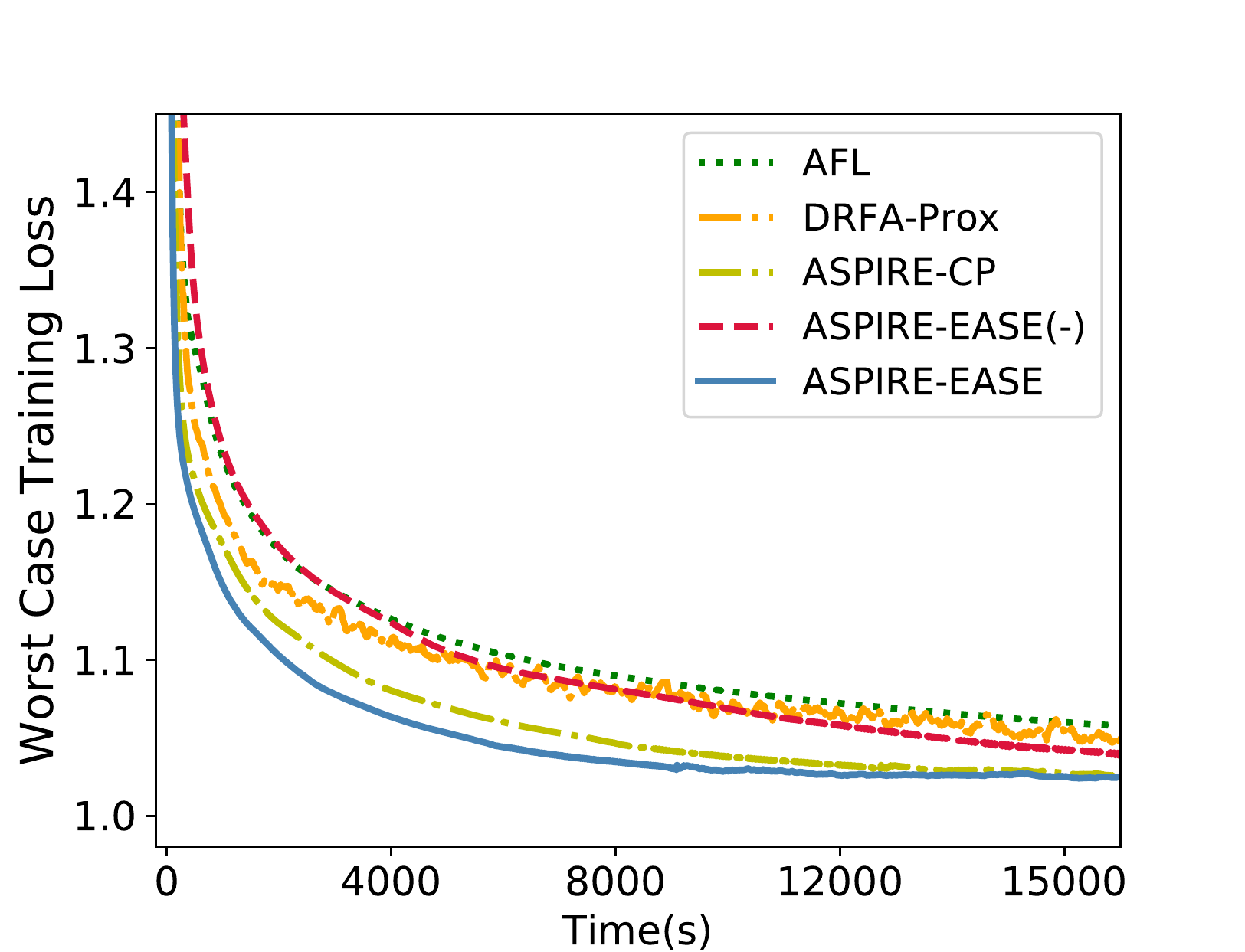}   
	\end{minipage}}
\subfigure[SC-MA] 
{\begin{minipage}[t]{0.49\linewidth}
	\centering      
	\includegraphics[scale=0.22]{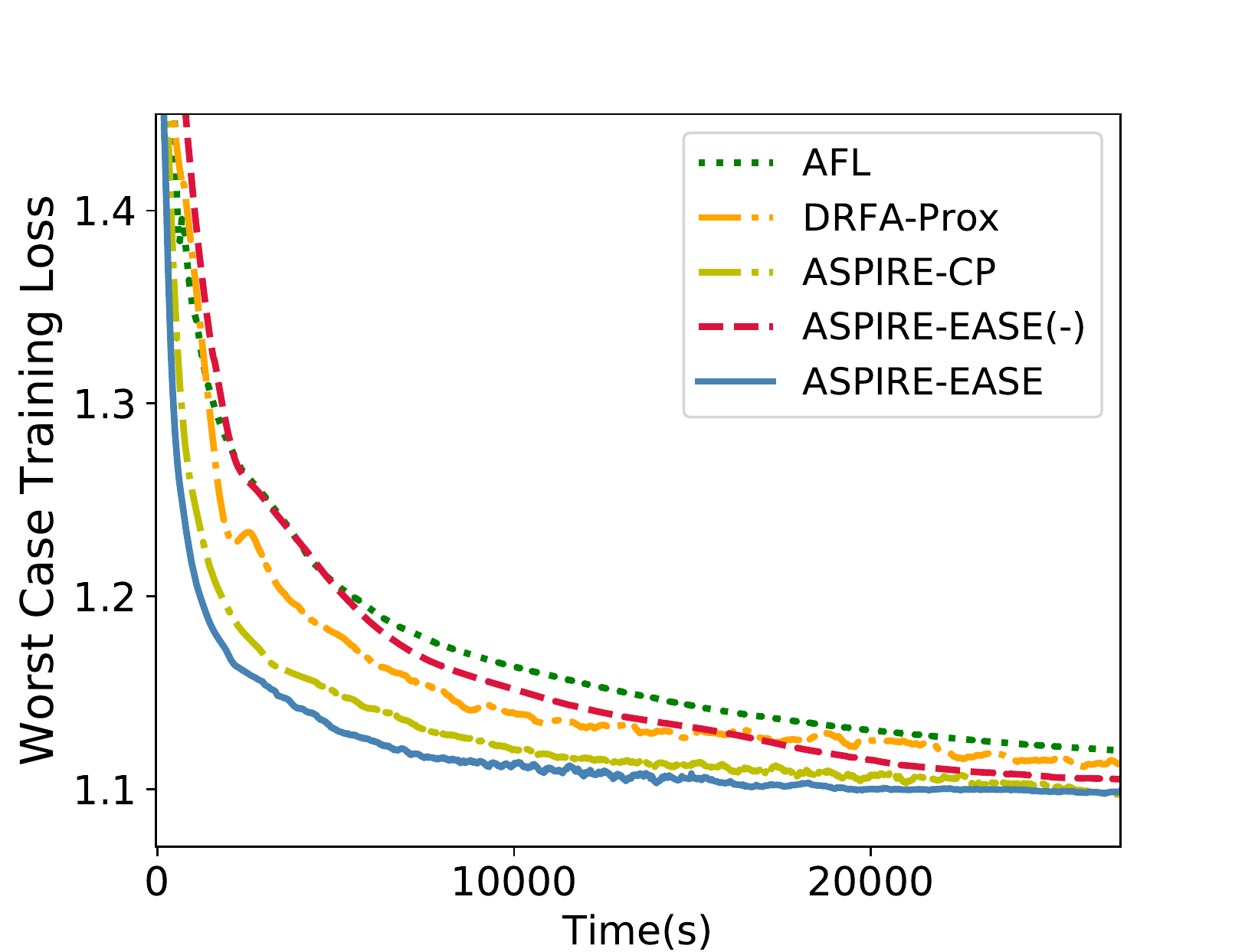}   
	\end{minipage}}
\caption{Comparison of the convergence time on worst case worker on  (a) Person Activity, (b) SC-MA  datasets.} 
\label{fig:time efficientcy}  
\vspace{-1.5mm}
\end{minipage} 
\makeatletter\def\@captype{figure}\makeatother 
\begin{minipage}{0.495\textwidth}  
\subfigure[Person Activity] 
{\begin{minipage}[t]{0.49\linewidth}
	\centering      
	\includegraphics[scale=0.22]{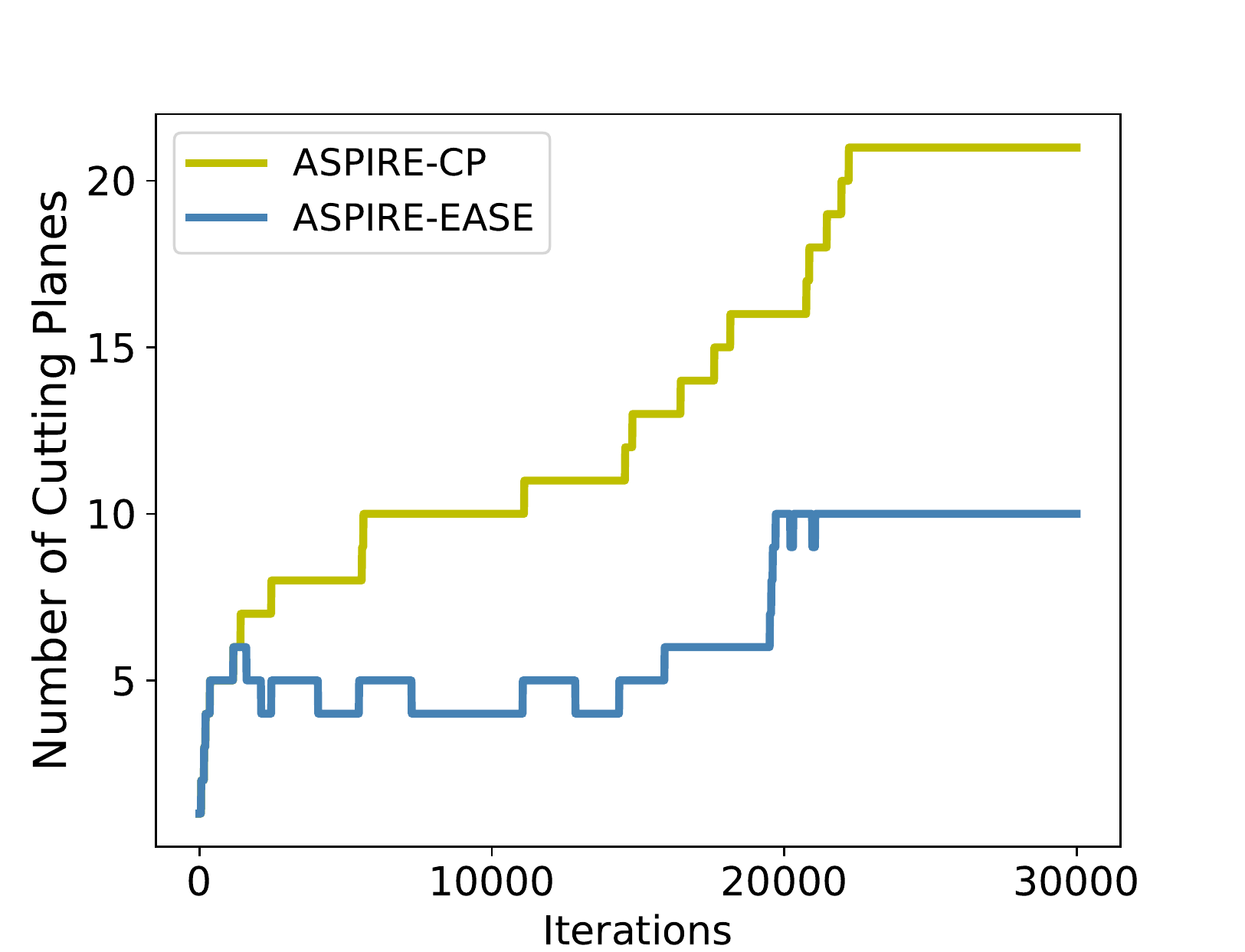}   
	\end{minipage}}
\subfigure[SC-MA] 
{\begin{minipage}[t]{0.49\linewidth}
	\centering      
	\includegraphics[scale=0.22]{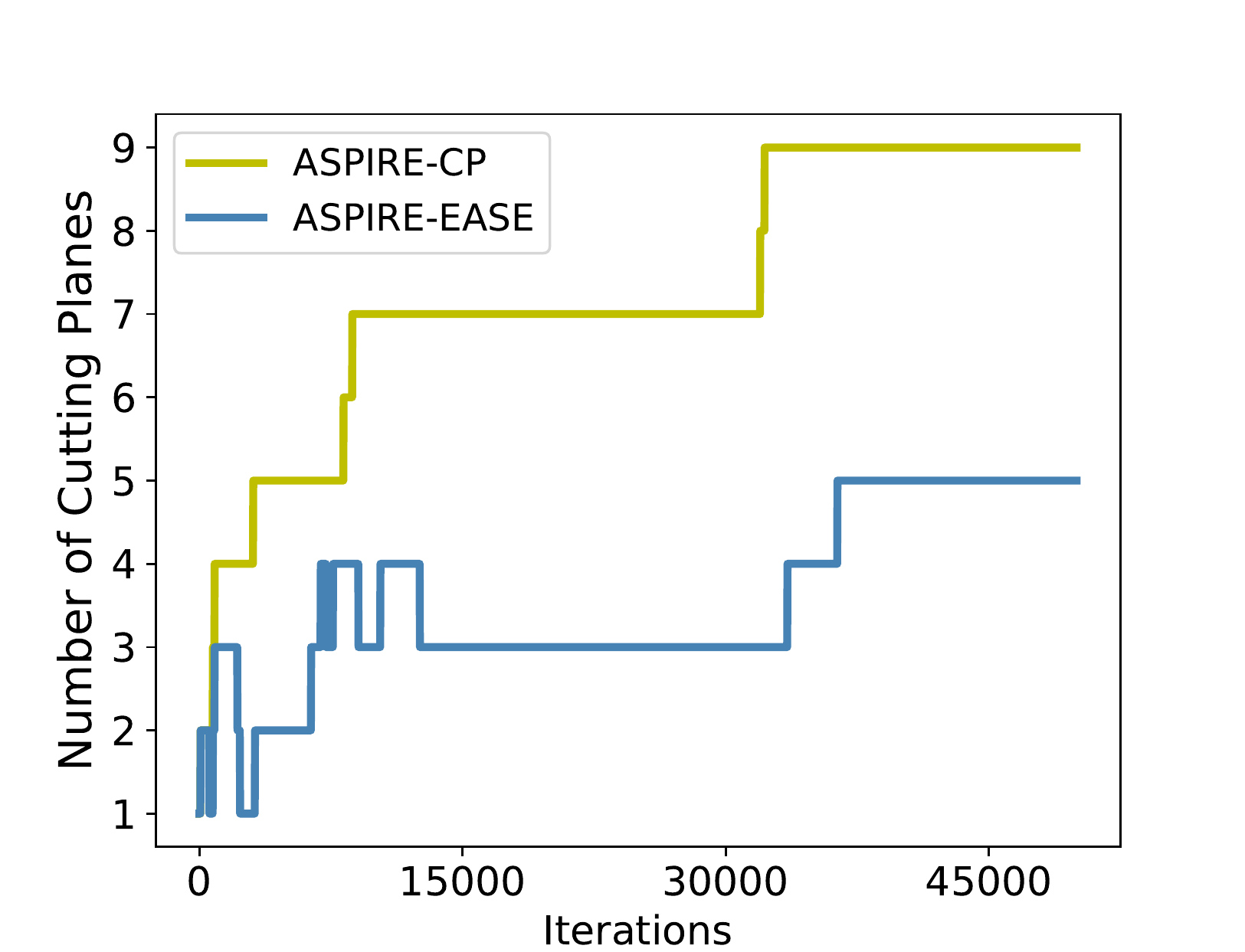}   
	\end{minipage}}
\caption{Comparison of ASPIRE-CP and ASPIRE-EASE regarding the number of cutting planes on  (a) Person Activity, (b) SC-MA datasets.} 
\label{fig:cutting plane}  
\end{minipage} 
\makeatletter\def\@captype{table}\makeatother 
\renewcommand\tabcolsep{1pt}
\renewcommand\arraystretch{1.4}
\begin{minipage}{.49\textwidth} 
\caption{Performance comparisons about the success attack rate ($\%$) $\downarrow$. The boldfaced digits represent the best results.}
\label{tab:SAR}
\scalebox{0.65}{
\begin{tabular}{l|cccc}
\toprule
Model    & SHL     & Person Activity         & SC-MA     & Fashion MNIST        \\ \hline
${\rm{Mix}}{\rm{_{Even}}}$ & 36.21±2.23 &  34.32±2.18  & 52.14±2.89   & 83.18±2.07  \\ 
FedAvg \citep{mcmahan2017communication} & 38.15±3.02  & 33.25±2.49 & 55.39±3.13    &  82.04±1.84\\ 
AFL \citep{mohri2019agnostic} & 68.63±4.24  &  43.66±3.87 & 75.81±4.03  &  90.04±2.52 \\
DRFA-Prox \citep{deng2021distributionally} & 21.23±3.63  &  27.27±3.31 & 30.79±3.65  &  63.24±2.47  \\
\hline
ASPIRE-EASE & \textbf{9.17±1.65}  &  \textbf{22.36±2.33}  & \textbf{14.51±3.21}  & \textbf{45.10±1.64}
\cr
\bottomrule  
\end{tabular}}
\end{minipage}

\vspace{1mm}

\textbf{Efficiency}. $\;$
In Figure \ref{fig:time efficientcy}, we compare the convergence speed of the proposed ASPIRE-EASE with AFL and DRFA-Prox by considering different communication and computation delays for each worker. The proposed ASPIRE-EASE has two variants, ASPIRE-CP (ASPIRE with cutting plane method), ASPIRE-EASE(-)(ASPIRE-EASE without asynchronous setting). More results are available in Figure \ref{appendix:fig:time efficientcy} of Appendix \ref{appendix:experiment}. Based on the comparison, we can observe that the proposed ASPIRE-EASE generally converges faster than baseline methods and its two variants. This is because 1) compared with AFL, DRFA-Prox, and ASPIRE-EASE(-), ASPIRE-EASE is an asynchronous algorithm in which the master updates its parameters only after receiving the updates from active workers instead of all workers; 2) unlike DRFA-Prox, the master in ASPIRE-EASE only needs to communicate with active workers once per iteration; 3) compared with ASPIRE-CP, ASPIRE-EASE utilizes active set method instead of cutting plane method, which is more efficient. It is seen from Figure \ref{fig:time efficientcy} that, the convergence speed of ASPIRE-EASE mainly benefits from the asynchronous setting.

\textbf{Ablation Study}. $\;$
For ASPIRE, compared with cutting plane method, EASE is more efficient since it considers removing the inactive cutting planes. To demonstrate the efficiency of EASE, we firstly compare ASPIRE-EASE with ASPIRE-CP concerning the number of cutting planes used during the training. In Figure \ref{fig:cutting plane}, we can observe that ASPIRE-EASE uses fewer cutting planes than ASPIRE-CP, thus is more efficient. The convergence speed of ASPIRE-EASE and ASPIRE-CP in Figure \ref{fig:time efficientcy} also suggests that ASPIRE-EASE converges much faster than ASPIRE-CP. More results are available in Figure \ref{appendix:fig:time efficientcy} and \ref{appendix:fig:cutting plane}, Appendix \ref{appendix:experiment}.

\vspace{-1.5mm}
\section{Conclusion} \label{conclusion}
\vspace{-1.5mm}
In this paper, we present ASPIRE-EASE method to effectively solve the distributed distributionally robust optimization problem with non-convex objectives. In addition, $CD$-norm uncertainty set has been proposed to effectively incorporate the prior distribution into the problem formulation, which allows for flexible adjustment of the degree of robustness of DRO. Theoretical analysis has also been conducted to analyze the convergence properties and the iteration complexity of ASPIRE-EASE. ASPIRE-EASE exhibits strong empirical performance on multiple real-world datasets and is effective in tackling DRO problems in a fully distributed and asynchronous manner. In the future work, more uncertainty sets could be designed for our framework and more update rule for variables in ASPIRE could be considered.

\bibliographystyle{abbrvnat}
\bibliography{ref}

	
	\newpage
	\appendix
\section*{Appendix}
\section{Proof of Theorem 1}
\label{appendix:Theorem1}

Before proceeding to the detailed proofs, we provide some notations for the clarity in presentation. We use notation $<  \cdot , \cdot  >$ to denote the inner product and we use $||\cdot||$ to denote the $l_2$-norm. $|{{\bf{A}}^t}|$ and $|{{\bf{Q}}^{t + 1}}|$ respectively denote the number of cutting planes and active workers in $(t+1)^{\rm{th}}$ iteration.

Then, we cover some Lemmas which are useful for the deduction of Theorem 1.

\setcounter{equation}{0}
\begin{lemma} \label{lemma 1}
Suppose Assumption 1 and 2 hold, $\forall t \ge T_1+\tau$, we have,
\begin{equation}
\label{eq:A.a111}
\renewcommand{\theequation}{A.\arabic{equation}}
\begin{array}{l}
{L_p}{\rm{(\{ }}{{\boldsymbol{w}}_j^{t+1}}{\rm{\} ,}}\boldsymbol{z}^t,h^t,\{ {\lambda _l^t}\} ,\{ {{\boldsymbol{\phi}}_j^t}\} {\rm{)}} \!-\! {L_p}{\rm{(\{ }}{{\boldsymbol{w}}_j^t}{\rm{\} ,}}\boldsymbol{z}^t,h^t,\{ {\lambda _l^t}\} ,\{ {{\boldsymbol{\phi}}_j^t}\} {\rm{)}}\vspace{1ex}\\
\! \le \! \sum\limits_{j  = 1}^N \! {(\frac{{L  +  1}}{2} \!-\! \frac{1}{{{\eta _{\boldsymbol{w}}^t}}})||{{\boldsymbol{w}}_j^{t+1}} \!-\! {{\boldsymbol{w}}_j^t}|{|^2}}  \! + \! \frac{{3\tau{ k_1 }N{L^2}}}{2}(||\boldsymbol{z}^{t+1} \!-\! \boldsymbol{z}^t|{|^2} \! + \! ||h^{t+1} \!-\! h^t|{|^2} \! + \!  \! \sum\limits_{l = 1}^{|{{\bf{A}}^t}|}\! {||{\lambda _l^{t+1}} \!-\! {\lambda _l^t}|{|^2}}), 
\end{array}
\end{equation}
\begin{equation}
\renewcommand{\theequation}{A.\arabic{equation}}
\label{eq:A.a12}
\begin{array}{l}
{L_p}{\rm{(\{ }}{{\boldsymbol{w}}_j^{t+1}}{\rm{\} ,}}{\boldsymbol{z}}^{t+1},h^t,\{ {\lambda _l^t}\} ,\{ {{\boldsymbol{\phi}}_j^t}\} {\rm{)}} \!-\! {L_p}{\rm{(\{ }}{{\boldsymbol{w}}_j^{t+1}}{\rm{\} ,}}{\boldsymbol{z}^t},h^t,\{ {\lambda _l^t}\} ,\{ {{\boldsymbol{\phi}}_j^t}\} {\rm{)}} 
 \le (\frac{L}{2} \!-\! \frac{1}{{{\eta _{\boldsymbol{z}}^t}}})||{\boldsymbol{z}}^{t+1} \!-\! {\boldsymbol{z}^t}|{|^2},
\end{array}
\end{equation}
\begin{equation}
\renewcommand{\theequation}{A.\arabic{equation}}
\label{eq:A.a13}
\begin{array}{l}
{L_p}{\rm{(\{ }}{{\boldsymbol{w}}_j^{t+1}}{\rm{\} ,}}{\boldsymbol{z}}^{t+1},h^{t+1},\{ {\lambda _l^t}\} ,\{ {{\boldsymbol{\phi}}_j^t}\} {\rm{)}} \!-\! {L_p}{\rm{(\{ }}{{\boldsymbol{w}}_j^{t+1}}{\rm{\} ,}}{\boldsymbol{z}}^{t+1},h^t,\{ {\lambda _l^t}\} ,\{ {{\boldsymbol{\phi}}_j^t}\} {\rm{)}}
 \le (\frac{L}{2} \!-\! \frac{1}{{{\eta _h^t}}})||h^{t+1} \!-\! h^t|{|^2}.
\end{array}
\end{equation}
\end{lemma}

\emph{\textbf{Proof of Lemma \ref{lemma 1}}:} 

According to Assumption 1, we have, 
\begin{equation}
\label{eq:A4-1}
\renewcommand{\theequation}{A.\arabic{equation}}
\begin{array}{l}
{L_p}{\rm{(\{ }}{{\boldsymbol{w}}_1^{t+1}},{{\boldsymbol{w}}_2^t}, \cdots ,{{\boldsymbol{w}}_N^t}{\rm{\} ,}}\boldsymbol{z}^t,h^t,\{ {\lambda _l^t}\} ,\{ {{\boldsymbol{\phi}}_j^t}\} {\rm{)}} \!-\! {L_p}{\rm{(\{ }}{{\boldsymbol{w}}_j^t}{\rm{\} ,}}\boldsymbol{z}^t,h^t,\{ {\lambda _l^t}\} ,\{ {{\boldsymbol{\phi}}_j^t}\} {\rm{)}}\vspace{1.5ex}\\
 \le \left\langle {{\nabla _{{{\boldsymbol{w}}_1}}}{L_p}{\rm{(\{ }}{{\boldsymbol{w}}_j^t}{\rm{\} ,}}\boldsymbol{z}^t,h^t,\{ {\lambda _l^t}\} ,\{ {{\boldsymbol{\phi}}_j^t}\} {\rm{),}}{{\boldsymbol{w}}_1^{t+1}} \!-\! {{\boldsymbol{w}}_1^t}} \right\rangle  \! + \! \frac{L}{2}||{{\boldsymbol{w}}_1^{t+1}} \!-\! {{\boldsymbol{w}}_1^t}|{|^2}, \vspace{3ex}\\
{L_p}{\rm{(\{ }}{{\boldsymbol{w}}_1^{t+1}},{{\boldsymbol{w}}_2^{t+1}}, {{\boldsymbol{w}}_3^t}, \cdots ,{{\boldsymbol{w}}_N^t}{\rm{\} ,}}\boldsymbol{z}^t,h^t,\!\{ {\lambda _l^t}\} ,\!\{ {{\boldsymbol{\phi}}_j^t}\} )
\!-\! {L_p}{\rm{(\{ }}{{\boldsymbol{w}}_1^{t+1}},{{\boldsymbol{w}}_2^t}, \cdots ,{{\boldsymbol{w}}_N^t}{\rm{\} ,}}\boldsymbol{z}^t,h^t,\!\{ {\lambda _l^t}\} ,\!\{ {{\boldsymbol{\phi}}_j^t}\})  \vspace{1.5ex}\\
 \le \left\langle {{\nabla _{{{\boldsymbol{w}}_2}}}{L_p}{\rm{(\{ }}{{\boldsymbol{w}}_j^t}{\rm{\} ,}}\boldsymbol{z}^t,h^t,\{ {\lambda _l^t}\} ,\{ {{\boldsymbol{\phi}}_j^t}\} {\rm{),}}{{\boldsymbol{w}}_2^{t+1}} \!-\! {{\boldsymbol{w}}_2^t}} \right\rangle  \! + \! \frac{L}{2}||{{\boldsymbol{w}}_2^{t+1}} \!-\! {{\boldsymbol{w}}_2^t}|{|^2}, \vspace{2ex}\\
\qquad \qquad \qquad  \qquad \qquad \qquad \qquad \qquad \qquad \vdots  \vspace{2ex}\\
{L_p}{\rm{(\{ }}{{\boldsymbol{w}}_j^{t+1}}{\rm{\} ,}}\boldsymbol{z}^t,h^t,\{ {\lambda _l^t}\} ,\{ {{\boldsymbol{\phi}}_j^t}\} {\rm{)}} \!-\! {L_p}{\rm{(\{ }}{{\boldsymbol{w}}_1^{t+1}}, \cdots ,{{\boldsymbol{w}}_{N \!-\! 1}^{t+1}},{{\boldsymbol{w}}_N^t}{\rm{\} ,}}\boldsymbol{z}^t,h^t,\{ {\lambda _l^t}\} ,\{ {{\boldsymbol{\phi}}_j^t}\} {\rm{)}} \vspace{1.5ex}\\
 \le \left\langle {{\nabla _{{{\boldsymbol{w}}_N}}}{L_p}{\rm{(\{ }}{{\boldsymbol{w}}_j^t}{\rm{\} ,}}\boldsymbol{z}^t,h^t,\{ {\lambda _l^t}\} ,\{ {{\boldsymbol{\phi}}_j^t}\} {\rm{),}}{{\boldsymbol{w}}_N^{t+1}} \!-\! {{\boldsymbol{w}}_N^t}} \right\rangle  \! + \! \frac{L}{2}||{{\boldsymbol{w}}_N^{t+1}} \!-\! {{\boldsymbol{w}}_N^t}|{|^2}.
\end{array}
\end{equation}

\vspace{0.5ex}

Summing up the above inequalities in Eq. (A.\ref{eq:A4-1}), we have,
\begin{equation}
\label{eq:A4}
\renewcommand{\theequation}{A.\arabic{equation}}
\begin{array}{l}
{L_p}{\rm{(\{ }}{{\boldsymbol{w}}_j^{t+1}}{\rm{\} ,}}{\boldsymbol{z}^t},h^t,\{ {\lambda _l^t}\} ,\{ {{\boldsymbol{\phi}}_j^t}\} {\rm{)}} \!-\! {L_p}{\rm{(\{ }}{{\boldsymbol{w}}_j^t}{\rm{\} ,}}{\boldsymbol{z}^t},h^t,\{ {\lambda _l^t}\} ,\{ {{\boldsymbol{\phi}}_j^t}\} {\rm{)}}\\
 \le \sum\limits_{j  = 1}^N {(\left\langle {{\nabla _{{{\boldsymbol{w}}_j}}}{L_p}{\rm{(\{ }}{{\boldsymbol{w}}_j^t}{\rm{\} ,}}{\boldsymbol{z}^t},h^t,\{ {\lambda _l^t}\} ,\{ {{\boldsymbol{\phi}}_j^t}\} {\rm{),}}{{\boldsymbol{w}}_j^{t+1}} \!-\! {{\boldsymbol{w}}_j^t}} \right\rangle  \! + \! \frac{L}{2}||{{\boldsymbol{w}}_j^{t+1}} \!-\! {{\boldsymbol{w}}_j^t}|{|^2})}.
\end{array}
\end{equation}

According to ${\nabla _{{{\boldsymbol{w}}_j}}}{L_p}{\rm{(\{ }}{{\boldsymbol{w}}_j^t}{\rm{\} ,}}{\boldsymbol{z}^t},h^t,\{ {\lambda _l^t}\} ,\{ {{\boldsymbol{\phi}}_j^t}\} {\rm{)}} = {\nabla _{{{\boldsymbol{w}}_j}}}{\widetilde{L}_p}{\rm{(\{ }}{{\boldsymbol{w}}_j^t}{\rm{\} ,}}{\boldsymbol{z}^t},h^t,\{ {\lambda _l^t}\} ,\{ {{\boldsymbol{\phi}}_j^t}\} {\rm{)}}$ and the optimal condition for Eq. (11), for active nodes, \textit{i.e.}, $\forall j \in {{\bf{Q}}^{t + 1}}, \forall t\ge T_1+\tau$, we have,
\begin{equation}
\label{eq:A5}
\renewcommand{\theequation}{A.\arabic{equation}}
\left\langle {{{\boldsymbol{w}}_j^t} - {{\boldsymbol{w}}_j^{t+1}},{{\boldsymbol{w}}_j^{t+1}} - {{\boldsymbol{w}}_j^t}  +  \underline{\eta _{\boldsymbol{w}}}{\nabla _{{{\boldsymbol{w}}_j}}}{L_p}{\rm{(\{ }}{{\boldsymbol{w}}_j^{\widetilde{{t}}_j}}{\rm{\} ,}}{\boldsymbol{z}}^{{\widetilde{{t}}_j}},h^{{\widetilde{{t}}_j}},\{ {\lambda _l^{\widetilde{{t}}_j}}\} ,\{ {{\boldsymbol{\phi}}_j^{\widetilde{{t}}_j}}\} {\rm{)}}} \right\rangle  \ge 0.
\end{equation}

According to Eq. (A.\ref{eq:A5}), $\forall t\ge T_1+\tau$, we have,
\begin{equation}
\label{eq:A5-2}
\renewcommand{\theequation}{A.\arabic{equation}}
\left\langle {{{\boldsymbol{w}}_j^{t+1}} \!-\! {{\boldsymbol{w}}_j^t},{\nabla _{{{\boldsymbol{w}}_j}}}{L_p}{\rm{(\{ }}{{\boldsymbol{w}}_j^{\widetilde{{t}}_j}}{\rm{\} ,}}{\boldsymbol{z}}^{{\widetilde{{t}}_j}},h^{{\widetilde{{t}}_j}},\{ {\lambda _l^{\widetilde{{t}}_j}}\} ,\{ {{\boldsymbol{\phi}}_j^{\widetilde{{t}}_j}}\} {\rm{)}}} \right\rangle \! \le\!  -\frac{1}{\underline{\eta _{\boldsymbol{w}}}}||{{\boldsymbol{w}}_j^{t+1}} - {{\boldsymbol{w}}_j^t}|{|^2} \!\le\!  -\frac{1}{{{\eta _{\boldsymbol{w}}^t}}}||{{\boldsymbol{w}}_j^{t+1}} - {{\boldsymbol{w}}_j^t}|{|^2}.
\end{equation}

And according to the Cauchy-Schwarz inequality, Assumption 1 and 2, we can get,
\begin{equation}
\label{eq:A5-3}
\renewcommand{\theequation}{A.\arabic{equation}}
\begin{array}{l}
\left\langle {{\boldsymbol{w}}_j^{t+1} \!-\! {\boldsymbol{w}}_j^t,{\nabla _{{{\boldsymbol{w}}_j}}}{L_p}{\rm{(\{ }}{{\boldsymbol{w}}_j^t}{\rm{\} ,}}{\boldsymbol{z}^t},h^t,\{ {\lambda _l^t}\} , \{ {{\boldsymbol{\phi}}_j^t}\} {\rm{)}} \!-\! {\nabla _{{{\boldsymbol{w}}_j}}}{L_p}{\rm{(\{ }}{{\boldsymbol{w}}_j^{\widetilde{{t}}_j}}{\rm{\} ,}}{\boldsymbol{z}^{{\widetilde{{t}}_j}}},h^{{\widetilde{{t}}_j}}, \{ {\lambda _l^{\widetilde{{t}}_j}}\} , \!\{ {{\boldsymbol{\phi}}_j^{\widetilde{{t}}_j}}\} {\rm{)}}} \right\rangle \\
\! \le \! \frac{1}{2}||{{\boldsymbol{w}}_j^{t+1}} \!-\! {{\boldsymbol{w}}_j^t}|{|^2} \! + \! \frac{L^2}{2}(||{\boldsymbol{z}^t} \!-\! {\boldsymbol{z}^{{\widetilde{{t}}_j}}}|{|^2} \! + \! ||h^t \!-\! h^{{\widetilde{{t}}_j}}|{|^2} \! + \! \sum\limits_{l = 1}^{|{{\bf{A}}^t}|} {||{\lambda _l^t} \!-\! {\lambda _l^{\widetilde{{t}}_j}}|{|^2}} )\\
\! \le \! \frac{1}{2}||{{\boldsymbol{w}}_j^{t+1}} \!-\! {{\boldsymbol{w}}_j^t}|{|^2} \! + \! \frac{{{3\tau{ k_1 }L^2}}}{2}(||{\boldsymbol{z}}^{t+1} \!-\! {\boldsymbol{z}^t}|{|^2} \! + \! ||h^{t+1} \!-\! h^t|{|^2} \! + \! \sum\limits_{l = 1}^{|{{\bf{A}}^t}|} {||{\lambda _l^{t+1}} \!-\! {\lambda _l^t}|{|^2}} ).
\end{array}
\end{equation}

Combining the above Eq. (A.\ref{eq:A4}), (A.\ref{eq:A5-2}) with Eq.  (A.\ref{eq:A5-3}), we can obtain Eq. (A.\ref{eq:A.a111}), that is,
\begin{equation*}
\begin{array}{l}
{L_p}{\rm{(\{ }}{{\boldsymbol{w}}_j^{t+1}}{\rm{\} ,}}\boldsymbol{z}^t,h^t,\{ {\lambda _l^t}\} ,\{ {{\boldsymbol{\phi}}_j^t}\} {\rm{)}} \!-\! {L_p}{\rm{(\{ }}{{\boldsymbol{w}}_j^t}{\rm{\} ,}}\boldsymbol{z}^t,h^t,\{ {\lambda _l^t}\} ,\{ {{\boldsymbol{\phi}}_j^t}\} {\rm{)}}\vspace{0.5ex}\\
\! \le \! \sum\limits_{j  = 1}^N \! {(\frac{{L  +  1}}{2} \!-\! \frac{1}{{{\eta _{\boldsymbol{w}}^t}}})||{{\boldsymbol{w}}_j^{t+1}} \!-\! {{\boldsymbol{w}}_j^t}|{|^2}}  \! + \! \frac{{3\tau{ k_1 }N{L^2}}}{2}(||\boldsymbol{z}^{t+1} \!-\! \boldsymbol{z}^t|{|^2} \! + \! ||h^{t+1} \!-\! h^t|{|^2} \! + \! \sum\limits_{l = 1}^{|{{\bf{A}}^t}|}\! {||{\lambda _l^{t+1}} \!-\! {\lambda _l^t}|{|^2}}).
\end{array}
\end{equation*}

\vspace{0.5ex}

Following Assumption 1, we have,
\begin{equation}
\renewcommand{\theequation}{A.\arabic{equation}}
\label{eq:A6}
\begin{array}{l}
{L_p}{\rm{(\{ }}{{\boldsymbol{w}}_j^{t+1}}{\rm{\} ,}}{\boldsymbol{z}}^{t+1},h^t,\{ {\lambda _l^t}\} ,\{ {{\boldsymbol{\phi}}_j^t}\} {\rm{)}} \!-\! {L_p}{\rm{(\{ }}{{\boldsymbol{w}}_j^{t+1}}{\rm{\} ,}}{\boldsymbol{z}^t},h^t,\{ {\lambda _l^t}\} ,\{ {{\boldsymbol{\phi}}_j^t}\} {\rm{)}} \vspace{1ex}\\
 \le \left\langle {{\nabla _{\boldsymbol{z}}}{L_p}{\rm{(\{ }}{{\boldsymbol{w}}_j^{t+1}}{\rm{\} ,}}{\boldsymbol{z}^t},h^t,\{ {\lambda _l^t}\} ,\{ {{\boldsymbol{\phi}}_j^t}\} {\rm{)}},{\boldsymbol{z}}^{t+1} \!-\! {\boldsymbol{z}^t}} \right\rangle  \! + \! \frac{L}{2}||{\boldsymbol{z}}^{t+1} \!-\! {\boldsymbol{z}^t}|{|^2}.
\end{array}
\end{equation}

According to ${\nabla _{\boldsymbol{z}}}{L_p}{\rm{(\{ }}{{\boldsymbol{w}}_j^{t+1}}{\rm{\} ,}}{\boldsymbol{z}^t},h^t,\{ {\lambda _l^t}\} ,\{ {{\boldsymbol{\phi}}_j^t}\} {\rm{)}} = {\nabla _{\boldsymbol{z}}}{\widetilde{L}_p}{\rm{(\{ }}{{\boldsymbol{w}}_j^{t+1}}{\rm{\} ,}}{\boldsymbol{z}^t},h^t,\{ {\lambda _l^t}\} ,\{ {{\boldsymbol{\phi}}_j^t}\} {\rm{)}}$ and the optimal condition for Eq. (12), we have,
\begin{equation}
\renewcommand{\theequation}{A.\arabic{equation}}
\label{eq:A7}
\left\langle {{\boldsymbol{z}^t} \!-\! {\boldsymbol{z}}^{t+1},{\boldsymbol{z}}^{t+1} \!-\! {\boldsymbol{z}^t} \! + \! {\eta _{\boldsymbol{z}}^t}{\nabla _{\boldsymbol{z}}}{L_p}{\rm{(\{ }}{{\boldsymbol{w}}_j^{t+1}}{\rm{\} ,}}{\boldsymbol{z}^t},h^t,\{ {\lambda _l^t}\} ,\{ {{\boldsymbol{\phi}}_j^t}\} {\rm{)}}} \right\rangle  \ge 0.
\end{equation}

Combining Eq.  (A.\ref{eq:A6}) with Eq.  (A.\ref{eq:A7}), we can obtain the Eq. (A.\ref{eq:A.a12}), that is,
\begin{equation*}
\begin{array}{l}
{L_p}{\rm{(\{ }}{{\boldsymbol{w}}_j^{t+1}}{\rm{\} ,}}{\boldsymbol{z}}^{t+1},h^t,\{ {\lambda _l^t}\} ,\{ {{\boldsymbol{\phi}}_j^t}\} {\rm{)}} \!-\! {L_p}{\rm{(\{ }}{{\boldsymbol{w}}_j^{t+1}}{\rm{\} ,}}{\boldsymbol{z}^t},h^t,\{ {\lambda _l^t}\} ,\{ {{\boldsymbol{\phi}}_j^t}\} {\rm{)}}
 \le (\frac{L}{2} \!-\! \frac{1}{{{\eta _{\boldsymbol{z}}^t}}})||{\boldsymbol{z}}^{t+1} \!-\! {\boldsymbol{z}^t}|{|^2}.
\end{array}
\end{equation*}

\vspace{0.5ex}

According to Assumption 1, we have:
\begin{equation}
\renewcommand{\theequation}{A.\arabic{equation}}
\label{eq:A8}
\begin{array}{l}
{L_p}{\rm{(\{ }}{{\boldsymbol{w}}_j^{t+1}}{\rm{\} ,}}{\boldsymbol{z}}^{t+1},h^{t+1},\{ {\lambda _l^t}\} ,\{ {{\boldsymbol{\phi}}_j^t}\} {\rm{)}} \!-\! {L_p}{\rm{(\{ }}{{\boldsymbol{w}}_j^{t+1}}{\rm{\} ,}}{\boldsymbol{z}}^{t+1},h^t,\{ {\lambda _l^t}\} ,\{ {{\boldsymbol{\phi}}_j^t}\} {\rm{)}} \vspace{1ex}\\
 \le \left\langle {{\nabla _h}{L_p}{\rm{(\{ }}{{\boldsymbol{w}}_j^{t+1}}{\rm{\} ,}}{\boldsymbol{z}}^{t+1},h^t,\{ {\lambda _l^t}\} ,\{ {{\boldsymbol{\phi}}_j^t}\} {\rm{)}},h^{t+1} \!-\! h^t} \right\rangle  \! + \! \frac{L}{2}||h^{t+1} \!-\! h^t|{|^2}.
\end{array}
\end{equation}

According to ${\nabla _h}{L_p}{\rm{(\{ }}{{\boldsymbol{w}}_j^{t+1}}{\rm{\} ,}}{\boldsymbol{z}}^{t+1},h^t,\{ {\lambda _l^t}\} ,\{ {{\boldsymbol{\phi}}_j^t}\} {\rm{)}} = {\nabla _h}{\widetilde{L}_p}{\rm{(\{ }}{{\boldsymbol{w}}_j^{t+1}}{\rm{\} ,}}{\boldsymbol{z}}^{t+1},h^t,\{ {\lambda _l^t}\} ,\{ {{\boldsymbol{\phi}}_j^t}\} {\rm{)}}$ and the optimal condition for Eq. (13), we have:
\begin{equation}
\renewcommand{\theequation}{A.\arabic{equation}}
\label{eq:A9}
\left\langle {h^t \!-\! h^{t+1},h^{t+1} \!-\! h^t \! + \! {\eta _h^t}{\nabla _h}{L_p}{\rm{(\{ }}{{\boldsymbol{w}}_j^{t+1}}{\rm{\} ,}}{\boldsymbol{z}}^{t+1},h^t,\{ {\lambda _l^t}\} ,\{ {{\boldsymbol{\phi}}_j^t}\} {\rm{)}}} \right\rangle  \ge 0.
\end{equation}

Combining Eq.  (A.\ref{eq:A8}) with Eq.  (A.\ref{eq:A9}), we can show that,
\begin{equation*}
\begin{array}{l}
{L_p}{\rm{(\{ }}{{\boldsymbol{w}}_j^{t+1}}{\rm{\} ,}}{\boldsymbol{z}}^{t+1},h^{t+1},\{ {\lambda _l^t}\} ,\{ {{\boldsymbol{\phi}}_j^t}\} {\rm{)}} \!-\! {L_p}{\rm{(\{ }}{{\boldsymbol{w}}_j^{t+1}}{\rm{\} ,}}{\boldsymbol{z}}^{t+1},h^t,\!\{ {\lambda _l^t}\} ,\!\{ {{\boldsymbol{\phi}}_j^t}\} {\rm{)}}
 \!\le\! (\frac{L}{2} \!-\! \frac{1}{{{\eta _h^t}}})||h^{t+1} \!-\! h^t|{|^2}.
\end{array}
\end{equation*}

\vspace{5ex}

\begin{lemma} \label{lemma 2}
Suppose Assumption 1 and 2 hold, $\forall t \ge T_1 + \tau$, we have:
\begin{equation}
\renewcommand{\theequation}{A.\arabic{equation}}
\label{eq:A10}
\begin{array}{l}
{L_p}{\rm{(\{ }}{{\boldsymbol{w}}_j^{t+1}}{\rm{\} ,}}\boldsymbol{z}^{t+1},h^{t+1},\{ {\lambda _l^{t+1}}\} ,\{ {{\boldsymbol{\phi}}_j^{t+1}}\} {\rm{)}} \!-\! {L_p}{\rm{(\{ }}{{\boldsymbol{w}}_j^t}{\rm{\} ,}}\boldsymbol{z}^t,h^t,\{ {\lambda _l^t}\} ,\{ {{\boldsymbol{\phi}}_j^t}\} {\rm{)}}\vspace{1ex}\\
\! \le \! (\frac{{L \! + \! 1}}{2} \!-\! \frac{1}{{{\eta _{\boldsymbol{w}}^t}}} \! + \! \frac{{|{{\bf{A}}^t}|{L^2}}}{{2{a_1}}} \! + \! \frac{{|{{\bf{Q}}^{t \! + \! 1}}|{L^2}}}{{2{a_3}}})\!\sum\limits_{j = 1}^N \!{||{{\boldsymbol{w}}_j^{t+1}} \!-\! {{\boldsymbol{w}}_j^t}|{|^2}}  \! + \! (\frac{{L \! + \! 3\tau{ k_1 }N{L^2}}}{2} \!-\! \frac{1}{{{\eta _{\boldsymbol{z}}^t}}} \! + \! \frac{{|{{\bf{A}}^t}|{L^2}}}{{2{a_1}}} \! + \! \frac{{|{{\bf{Q}}^{t \! + \! 1}}|{L^2}}}{{2{a_3}}})||\boldsymbol{z}^{t+1} \!-\! \boldsymbol{z}^t|{|^2}\vspace{1ex}\\
 \! +  (\frac{{L \! + \! 3\tau{ k_1 }N{L^2}}}{2} \!-\! \frac{1}{{{\eta _h^t}}} \! + \! \frac{{|{{\bf{A}}^t}|{L^2}}}{{2{a_1}}} \! + \! \frac{{|{{\bf{Q}}^{t \! + \! 1}}|{L^2}}}{{2{a_3}}})||h^{t+1} \!-\! h^t|{|^2} \! + \! (\frac{{{a_1} \! + \! 3\tau{ k_1 }N{L^2}}}{2} \!-\! \frac{{{c_1^{t-1}} - {c_1^t}}}{2} \! + \! \frac{1}{{2{\rho _1}}})\!\sum\limits_{l = 1}^{|{{\bf{A}}^t}|}\! {||{\lambda _l^{t+1}} \!-\! {\lambda _l^t}|{|^2}} \vspace{1ex}\\
 
 \! +  \frac{{{c_1^{t-1}}}}{2} \! \sum\limits_{l = 1}^{|{{\bf{A}}^t}|}\! {(||{\lambda _l^{t+1}}|{|^2} \!-\! ||{\lambda _l^t}|{|^2})}  \! + \! \frac{1}{{2{\rho _1}}} \! \sum\limits_{l = 1}^{|{{\bf{A}}^t}|}\! {||{\lambda _l^t} \!-\! {\lambda _l^{t-1}}|{|^2}} \! + \! (\frac{{{a_3}}}{2} \!-\! \frac{{{c_2^{t-1}} - {c_2^t}}}{2} \! + \! \frac{1}{{2{\rho _2}}}) \! \sum\limits_{j = 1}^N\! {||{{\boldsymbol{\phi}}_j^{t+1}} \!-\! {{\boldsymbol{\phi}}_j^t}|{|^2}} \vspace{1ex}\\
 
 \! +  \frac{{{c_2^{t-1}}}}{2}\sum\limits_{j = 1}^N {(||{{\boldsymbol{\phi}}_j^{t+1}}|{|^2} \!-\! ||{{\boldsymbol{\phi}}_j^t}|{|^2})}  \! + \! \frac{1}{{2{\rho _2}}}\sum\limits_{j = 1}^N {||{{\boldsymbol{\phi}}_j^t} \!-\! {{\boldsymbol{\phi}}_j^{t-1}}|{|^2}}, 
\end{array}
\end{equation}
where $a_1 >0$ and $a_3 >0$ are constants.
\end{lemma}

\emph{\textbf{Proof of Lemma \ref{lemma 2}}:} 

First of all, at $(t+1)^{\rm{th}}$ iteration, the following equations hold and  will be used in the derivation:
\begin{equation*}
\sum\limits_{j = 1}^N \! {||{{\boldsymbol{\phi}}_j^{t+1}} \!-\! {{\boldsymbol{\phi}}_j^t}|{|^2}} \! = \! \sum\limits_{j \in {{\bf{Q}}^{t \! + \! 1}}}\! {||{{\boldsymbol{\phi}}_j^{t+1}} \!-\! {{\boldsymbol{\phi}}_j^t}|{|^2}}, \;
\sum\limits_{j = 1}^N \! {(||{{\boldsymbol{\phi}}_j^{t+1}}|{|^2} \!-\! ||{{\boldsymbol{\phi}}_j^t}|{|^2})} \! = \! \sum\limits_{j \in {{\bf{Q}}^{t + 1}}}\! {(||{{\boldsymbol{\phi}}_j^{t+1}}|{|^2} \!-\! ||{{\boldsymbol{\phi}}_j^t}|{|^2})}. 
\end{equation*}

According to Eq. (14), in $(t+1)^{\rm{th}}$ iteration, $\forall \lambda  \in {\bf{\Lambda}} $, it follows that:
\begin{equation}
\renewcommand{\theequation}{A.\arabic{equation}}
\label{eq:A11}
\left\langle {{\lambda _l^{t+1}} \!-\! {\lambda _l^t} \!-\! {\rho _1}{\nabla _{{\lambda _l}}}{{\widetilde{L}}_p}{\rm{(\{ }}{{\boldsymbol{w}}_j^{t+1}}{\rm{\} ,}}{\boldsymbol{z}}^{t+1},h^{t+1},\{ {\lambda _l^t}\} ,\{ {{\boldsymbol{\phi}}_j^t}\} {\rm{)}},\lambda  \!-\! {\lambda _l^{t+1}}} \right\rangle  \ge 0.
\end{equation}

Let $\lambda  = {\lambda _l^t}$, we can obtain:
\begin{equation}
\renewcommand{\theequation}{A.\arabic{equation}}
\label{eq:A12}
\left\langle {{\nabla _{{\lambda _l}}}{{\widetilde{L}}_p}{\rm{(\{ }}{{\boldsymbol{w}}_j^{t+1}}{\rm{\} ,}}{\boldsymbol{z}}^{t+1},h^{t+1},\{ {\lambda _l^t}\} ,\{ {{\boldsymbol{\phi}}_j^t}\} {\rm{)}} \!-\! \frac{1}{{{\rho _1}}}({\lambda _l^{t+1}} \!-\! {\lambda _l^t}),{\lambda _l^t} \!-\! {\lambda _l^{t+1}}} \right\rangle  \le 0.
\end{equation}

Likewise, in $t^{\rm{th}}$ iteration, we can obtain:
\begin{equation}
\renewcommand{\theequation}{A.\arabic{equation}}
\label{eq:A12}
\left\langle {{\nabla _{{\lambda _l}}}{{\widetilde{L}}_p}{\rm{(\{ }}{{\boldsymbol{w}}_j^t}{\rm{\} ,}}{\boldsymbol{z}^t},h^t,\{ {\lambda _l^{t-1}}\} ,\{ {{\boldsymbol{\phi}}_j^{t-1}}\} {\rm{)}} \!-\! \frac{1}{{{\rho _1}}}({\lambda _l^t} \!-\! {\lambda _l^{t-1}}),{\lambda _l^{t+1}} \!-\! {\lambda _l^t}} \right\rangle  \le 0.
\end{equation}

$\forall t \ge T_1$, since ${\widetilde{L}_p}{\rm{(\{ }}{{\boldsymbol{w}}_j}{\rm{\} ,}}{\boldsymbol{z}},h,\!\{ {\lambda _l}\} ,\!\{ {{\boldsymbol{\phi}}_j}\} {\rm{)}}$ is concave with respect to ${\lambda _l}$, we have,
\begin{equation}
\renewcommand{\theequation}{A.\arabic{equation}}
\label{eq:A14}
\begin{array}{l}
{\widetilde{L}_p}{\rm{(\{ }}{{\boldsymbol{w}}_j^{t+1}}{\rm{\} ,}}{\boldsymbol{z}}^{t+1},h^{t+1},\{ {\lambda _l^{t+1}}\} ,\{ {{\boldsymbol{\phi}}_j^t}\} {\rm{)}} - {\widetilde{L}_p}{\rm{(\{ }}{{\boldsymbol{w}}_j^{t+1}}{\rm{\} ,}}{\boldsymbol{z}}^{t+1},h^{t+1},\{ {\lambda _l^t}\} ,\{ {{\boldsymbol{\phi}}_j^t}\} {\rm{)}}\vspace{1ex}\\
\! \le \! \sum\limits_{l = 1}^{|{{\bf{A}}^t}|} \! {\left\langle \! {{\nabla _{{\lambda _l}}}{{\widetilde{L} }_p}{\rm{(\{ }}{{\boldsymbol{w}}_j^{t+1}}{\rm{\} ,}}{\boldsymbol{z}}^{t+1},h^{t+1},\{ {\lambda _l^t}\} ,\{ {{\boldsymbol{\phi}}_j^t}\} {\rm{)}},{\lambda _l^{t+1}} - {\lambda _l^t}} \! \right\rangle } \\
\! \le \! \sum\limits_{l  =  1}^{|{{\bf{A}}^t}|} \! {( \! \left\langle \! {{\nabla _{{\lambda _l}}}{{\widetilde{L} }_p}{\rm{(\{ }}{{\boldsymbol{w}}_j^{t+1}}{\rm{\} , }}{\boldsymbol{z}}^{t+1},h^{t+1},\!\{ {\lambda _l^t}\} ,\! \{ {{\boldsymbol{\phi}}_j^t}\} {\rm{)}} \! -\! {\nabla _{{\lambda _l}}}{{\widetilde{L}}_p}{\rm{(\{ }}{{\boldsymbol{w}}_j^t}{\rm{\},}}{\boldsymbol{z}^t},h^t,\! \{ {\lambda _l^{t-1}}\}, \!\{ {{\boldsymbol{\phi}}_j^{t-1}}\} {\rm{)}},{\lambda _l^{t+1}}\! -\! {\lambda _l^t}} \! \right\rangle } \vspace{0.5ex}\\
\qquad \quad  + \frac{1}{{{\rho _1}}}\left\langle {{\lambda _l^t} - {\lambda _l^{t-1}},{\lambda _l^{t+1}} - {\lambda _l^t}} \right\rangle).
\end{array}
\end{equation}

Denoting ${{\boldsymbol{v}}_{1,l}^{t + 1}} = {\lambda _l^{t+1}} - {\lambda _l^t} - ({\lambda _l^t} - {\lambda _l^{t-1}})$, we have,
\begin{equation}
\renewcommand{\theequation}{A.\arabic{equation}}
\label{eq:A15}
\begin{array}{l}
\sum\limits_{l = 1}^{|{{\bf{A}}^t}|} \! {\left\langle \! {{\nabla _{{\lambda _l}}}{{\widetilde{L}}_p}{\rm{(\{ }}{{\boldsymbol{w}}_j^{t+1}}{\rm{\},}}{\boldsymbol{z}}^{t+1},h^{t+1}, \! \{ {\lambda _l^t}\} , \{ {{\boldsymbol{\phi}}_j^t}\} {\rm{)}}  \! -\! {\nabla _{{\lambda _l}}}{{\widetilde{L} }_p}{\rm{(\{ }}{{\boldsymbol{w}}_j^t}{\rm{\} ,}}{\boldsymbol{z}^t},h^t, \! \{ {\lambda _l^{t-1}}\} , \! \{ {{\boldsymbol{\phi}}_j^{t-1}}\} {\rm{)}},{\lambda _l^{t+1}}  \! -\! {\lambda _l^t}} \! \right\rangle } \\
 \!= \! \sum\limits_{l = 1}^{|{{\bf{A}}^t}|} \! {\left\langle \! {{\nabla _{{\lambda _l}}}{{\widetilde{L}}_p}{\rm{(\{ }}{{\boldsymbol{w}}_j^{t+1}}{\rm{\},}}{\boldsymbol{z}}^{t+1},h^{t+1},\!\{ {\lambda _l^t}\},\!\{ {{\boldsymbol{\phi}}_j^t}\} {\rm{)}}  \! -\! {\nabla _{{\lambda _l}}}{{\widetilde{L}}_p}{\rm{(\{ }}{{\boldsymbol{w}}_j^t}{\rm{\} ,}}{\boldsymbol{z}^t},h^t, \! \{ {\lambda _l^t}\} , \! \{ {{\boldsymbol{\phi}}_j^t}\} {\rm{)}},{\lambda _l^{t+1}}  \! -\! {\lambda _l^t}} \! \right\rangle } (1a)\\
  \! +\! \sum\limits_{l = 1}^{|{{\bf{A}}^t}|} \! {\left\langle \! {{\nabla _{{\lambda _l}}}{{\widetilde{L}}_p}{\rm{(\{ }}{{\boldsymbol{w}}_j^t}{\rm{\} ,}}{\boldsymbol{z}^t},h^t, \! \{ {\lambda _l^t}\}, \! \{ {{\boldsymbol{\phi}}_j^t}\} {\rm{)}}  \! -\! {\nabla _{{\lambda _l}}}{{\widetilde{L}}_p}{\rm{(\{ }}{{\boldsymbol{w}}_j^t}{\rm{\} ,}}{\boldsymbol{z}^t},h^t, \! \{ {\lambda _l^{t-1}}\} , \! \{ {{\boldsymbol{\phi}}_j^{t-1}}\} {\rm{)}}, {{\boldsymbol{v}}_{1,l}^{t + 1}}} \right\rangle } (1b)\\
  \! +\! \sum\limits_{l = 1}^{|{{\bf{A}}^t}|} \! {\left\langle {{\nabla _{{\lambda _l}}}{{\widetilde{L}}_p}{\rm{(\{ }}{{\boldsymbol{w}}_j^t}{\rm{\} ,}}{\boldsymbol{z}^t},h^t,\!\{ {\lambda _l^t}\} ,\!\{ {{\boldsymbol{\phi}}_j^t}\} {\rm{)}}  \! -\! {\nabla _{{\lambda _l}}}{{\widetilde{L}}_p}{\rm{(\{ }}{{\boldsymbol{w}}_j^t}{\rm{\} ,}}{\boldsymbol{z}^t},h^t,\!\{ {\lambda _l^{t-1}}\} ,\!\{ {{\boldsymbol{\phi}}_j^{t-1}}\} {\rm{)}},{\lambda _l^t}  \! -\! {\lambda _l^{t-1}}} \! \right\rangle } (1c).
\end{array}
\end{equation}

\vspace{0.5ex}

Firstly, we focus on the ($1a$) in Eq.  (A.\ref{eq:A15}), we can write ($1a$) as:
\begin{equation}
\renewcommand{\theequation}{A.\arabic{equation}}
\label{eq:A16}
\begin{array}{l}
\left\langle \!{{\nabla _{{\lambda _l}}}{{\widetilde{L}}_p}{\rm{(\{ }}{{\boldsymbol{w}}_j^{t+1}}{\rm{\} ,}}{\boldsymbol{z}}^{t+1},h^{t+1},\{ {\lambda _l^t}\} ,\{ {{\boldsymbol{\phi}}_j^t}\} {\rm{)}} \! - \! {\nabla _{{\lambda _l}}}{{\widetilde{L}}_p}{\rm{(\{ }}{{\boldsymbol{w}}_j^t}{\rm{\} ,}}{\boldsymbol{z}^t},h^t,\{ {\lambda _l^t}\} ,\{ {{\boldsymbol{\phi}}_j^t}\} {\rm{)}},{\lambda _l^{t+1}} \! - \! {\lambda _l^t}}\! \right\rangle  \vspace{1ex}\\
\! =\! \left\langle {{\nabla _{{\lambda _l}}}{L_p}{\rm{(\{ }}{{\boldsymbol{w}}_j^{t+1}}{\rm{\} ,}}{\boldsymbol{z}}^{t+1},h^{t+1},\{ {\lambda _l^t}\} ,\{ {{\boldsymbol{\phi}}_j^t}\} {\rm{)}} \! - \! {\nabla _{{\lambda _l}}}{L_p}{\rm{(\{ }}{{\boldsymbol{w}}_j^t}{\rm{\} ,}}{\boldsymbol{z}^t},h^t,\{ {\lambda _l^t}\} ,\{ {{\boldsymbol{\phi}}_j^t}\} {\rm{)}},{\lambda _l^{t+1}} \! - \! {\lambda _l^t}} \right\rangle  \vspace{1ex}\\
 \! +  ({c_1^{t-1}} \! - \! {c_1^t})\left\langle {{\lambda _l^t},{\lambda _l^{t+1}} \! - \! {\lambda _l^t}} \right\rangle  \vspace{1ex}\\
\! = \!\left\langle {{\nabla _{{\lambda _l}}}{L_p}{\rm{(\{ }}{{\boldsymbol{w}}_j^{t+1}}{\rm{\} ,}}{\boldsymbol{z}}^{t+1},h^{t+1},\{ {\lambda _l^t}\} ,\{ {{\boldsymbol{\phi}}_j^t}\} {\rm{)}} \! - \! {\nabla _{{\lambda _l}}}{L_p}{\rm{(\{ }}{{\boldsymbol{w}}_j^t}{\rm{\} ,}}{\boldsymbol{z}^t},h^t,\{ {\lambda _l^t}\} ,\{ {{\boldsymbol{\phi}}_j^t}\} {\rm{)}},{\lambda _l^{t+1}} \! - \! {\lambda _l^t}} \right\rangle \vspace{1ex}\\
 \! +  \frac{{{c_1^{t-1}} - {c_1^t}}}{2}(||{\lambda _l^{t+1}}|{|^2} \! - \! ||{\lambda _l^t}|{|^2}) \! - \! \frac{{{c_1^{t-1}} -  {c_1^t}}}{2}||{\lambda _l^{t+1}} \! - \! {\lambda _l^t}|{|^2}.
\end{array}
\end{equation}

And according to Cauchy-Schwarz inequality and Assumption 1, we can obtain,
\begin{equation}
\renewcommand{\theequation}{A.\arabic{equation}}
\label{eq:A17}
\begin{array}{l}
\left\langle {{\nabla _{{\lambda _l}}}{L_p}{\rm{(\{ }}{{\boldsymbol{w}}_j^{t+1}}{\rm{\} ,}}{\boldsymbol{z}}^{t+1},h^{t+1},\{ {\lambda _l^t}\} ,\{ {{\boldsymbol{\phi}}_j^t}\} {\rm{)}}\! -\! {\nabla _{{\lambda _l}}}{L_p}{\rm{(\{ }}{{\boldsymbol{w}}_j^t}{\rm{\} ,}}{\boldsymbol{z}^t},h^t,\{ {\lambda _l^t}\} ,\{ {{\boldsymbol{\phi}}_j^t}\} {\rm{)}},{\lambda _l^{t+1}} - {\lambda _l^t}} \right\rangle \vspace{0.5ex} \\
 \!\le \frac{{{L^2}}}{{2{a_1}}}(\sum\limits_{j = 1}^N {||{{\boldsymbol{w}}_j^{t+1}} - {{\boldsymbol{w}}_j^t}|{|^2}}  \! + \! ||{\boldsymbol{z}}^{t+1} - {\boldsymbol{z}^t}|{|^2} \! + \! ||h^{t+1} - h^t|{|^2}) \! + \! \frac{{{a_1}}}{2}||{\lambda _l^{t+1}} - {\lambda _l^t}|{|^2},
\end{array}
\end{equation}

where $a_1 >0$  is a constant. Combining Eq. (A.\ref{eq:A16}) with Eq. (A.\ref{eq:A17}), we can obtain the upper bound of ($1a$), that is,
\begin{equation}
\renewcommand{\theequation}{A.\arabic{equation}}
\label{eq:A18}
\begin{array}{l}
\sum\limits_{l = 1}^{|{{\bf{A}}^t}|}\! {\left\langle\! {{\nabla _{{\lambda _l}}}{{\widetilde{L}}_p}{\rm{(\{ }}{{\boldsymbol{w}}_j^{t+1}}{\rm{\} ,}}{\boldsymbol{z}}^{t+1},h^{t+1},\{ {\lambda _l^t}\} ,\{ {{\boldsymbol{\phi}}_j^t}\} {\rm{)}} \! - \! {\nabla _{{\lambda _l}}}{{\widetilde{L}}_p}{\rm{(\{ }}{{\boldsymbol{w}}_j^t}{\rm{\} ,}}{\boldsymbol{z}^t},h^t,\{ {\lambda _l^t}\} ,\{ {{\boldsymbol{\phi}}_j^t}\} {\rm{)}},{\lambda _l^{t+1}} \!-\! {\lambda _l^t}} \! \right\rangle } \\
 \!\le \sum\limits_{l = 1}^{|{{\bf{A}}^t}|} ( \frac{{{L^2}}}{{2{a_1}}}(\sum\limits_{j = 1}^N {||{{\boldsymbol{w}}_j^{t+1}} \! - \! {{\boldsymbol{w}}_j^t}|{|^2}}  + ||{\boldsymbol{z}}^{t + 1} \! - \! {\boldsymbol{z}^t}|{|^2} + ||h^{t + 1} \! - \! h^t|{|^2}) + \frac{{{a_1}}}{2}||{\lambda _l^{t + 1}} \! - \! {\lambda _l^t}|{|^2}\vspace{1ex}\\
\qquad \quad + \frac{{{c_1^{t-1}}  -  {c_1^t}}}{2}(||{\lambda _l^{t + 1}}|{|^2} \! - \! ||{\lambda _l^t}|{|^2}) \! - \! \frac{{{c_1^{t-1}}  -  {c_1^t}}}{2}||{\lambda _l^{t + 1}} \! - \! {\lambda _l^t}|{|^2}).
\end{array}
\end{equation}

Secondly, we focus on the ($1b$) in Eq. (A.\ref{eq:A15}). According to Cauchy-Schwarz inequality we can write the ($1b$) as,
\begin{equation}
\renewcommand{\theequation}{A.\arabic{equation}}
\label{eq:A19}
\begin{array}{l}
\sum\limits_{l = 1}^{|{{\bf{A}}^t}|} {\left\langle {{\nabla _{{\lambda _l}}}{{\widetilde{L}}_p}{\rm{(\{ }}{{\boldsymbol{w}}_j^t}{\rm{\} ,}}{\boldsymbol{z}^t},h^t,\{ {\lambda _l^t}\} ,\{ {{\boldsymbol{\phi}}_j^t}\} {\rm{)}} \! - \! {\nabla _{{\lambda _l}}}{{\widetilde{L}}_p}{\rm{(\{ }}{{\boldsymbol{w}}_j^t}{\rm{\} ,}}{\boldsymbol{z}^t},h^t,\{ {\lambda _l^{t-1}}\} ,\{ {{\boldsymbol{\phi}}_j^{t-1}}\} {\rm{)}},{{\boldsymbol{v}}_{1,l}^{t + 1}}} \right\rangle } \\

\! \le \! \sum\limits_{l = 1}^{|{{\bf{A}}^t}|}\! {( \frac{{{a_2}}}{2}\!||{\nabla _{{\lambda _l}}}{{\widetilde{L}}_p}{\rm{(\{ }}{{\boldsymbol{w}}_j^t}{\rm{\} ,}}{\boldsymbol{z}^t},h^t, \! \{ {\lambda _l^t}\} , \! \{ {{\boldsymbol{\phi}}_j^t}\} {\rm{)}} \! - \! {\nabla _{{\lambda _l}}}{{\widetilde{L}}_p}{\rm{(\{ }}{{\boldsymbol{w}}_j^t}{\rm{\} ,}}{\boldsymbol{z}^t},h^t, \! \{ {\lambda _l^{t-1}}\} , \! \{ {{\boldsymbol{\phi}}_j^{t-1}}\} {\rm{)}}|{|^2}\! +\! \frac{1}{{2{a_2}}}\!||{{\boldsymbol{v}}_{1,l}^{t + 1}}|{|^2})},
\end{array}
\end{equation}
where $a_2 >0$  is a constant. Then, we focus on the ($1c$) in Eq. (A.\ref{eq:A15}). Firstly, $\forall \lambda _l$, we have,
\begin{equation}
\renewcommand{\theequation}{A.\arabic{equation}}
\label{eq:A20}
\begin{array}{l}
||{\nabla _{{\lambda _l}}}{\widetilde{L} _p}{\rm{(\{ }}{{\boldsymbol{w}}_j^t}{\rm{\} ,}}{\boldsymbol{z}^t},h^t,\{ {\lambda _l^t}\} ,\{ {{\boldsymbol{\phi}}_j^t}\} {\rm{)}} \! - \! {\nabla _{{\lambda _l}}}{\widetilde{L} _p}{\rm{(\{ }}{{\boldsymbol{w}}_j^t}{\rm{\} ,}}{\boldsymbol{z}^t},h^t,\{ {\lambda _l^{t-1}}\} ,\{ {{\boldsymbol{\phi}}_j^{t-1}}\} {\rm{)||}}\vspace{1ex}\\
 \!= \! ||{\nabla _{{\lambda _l}}}{L_p}{\rm{(\{ }}{{\boldsymbol{w}}_j^t}{\rm{\} ,}}{\boldsymbol{z}^t},h^t, \!\{ {\lambda _l^t}\} ,\! \{ {{\boldsymbol{\phi}}_j^t}\} {\rm{)}} \! - \! {\nabla _{{\lambda _l}}}{L_p}{\rm{(\{ }}{{\boldsymbol{w}}_j^t}{\rm{\} ,}}{\boldsymbol{z}^t},h^t, \! \{ {\lambda _l^{t-1}}\} , \{ {{\boldsymbol{\phi}}_j^t}\} {\rm{)}} \! - \! {c_1^{t-1}}({\lambda _l^t} \! - \! {\lambda _l^{t-1}}){\rm{||}}\vspace{1ex}\\
 \! \le\! (L + {c_1^{t-1}})||{\lambda _l^t}  \! - \! {\lambda _l^{t-1}}{\rm{||}},
\end{array}
\end{equation}
where the last inequality comes from Assumption 1 and the trigonometric inequality. Denoting ${L_1}' = L + {c_1^0}$, we can obtain,
{\begin{equation}
\renewcommand{\theequation}{A.\arabic{equation}}
\label{eq:A22}
\begin{array}{l}
 ||{\nabla _{{\lambda _l}}}{\widetilde{L}_p}{\rm{(\{ }}{{\boldsymbol{w}}_j^t}{\rm{\} ,}}{\boldsymbol{z}^t},h^t,\{ {\lambda _l^t}\} ,\{ {{\boldsymbol{\phi}}_j^t}\} {\rm{)}}\!  -\!  {\nabla _{{\lambda _l}}}{\widetilde{L}_p}{\rm{(\{ }}{{\boldsymbol{w}}_j^t}{\rm{\} ,}}{\boldsymbol{z}^t},h^t,\{ {\lambda _l^{t-1}}\} ,\{ {{\boldsymbol{\phi}}_j^{t-1}}\} {\rm{)||}} \le {L_1}'||{\lambda _l^t}  \! - \! {\lambda _l^{t-1}}{\rm{||}}.
\end{array}
\end{equation}}

Following from Eq. (A.\ref{eq:A22}) and the strong  concavity of ${\widetilde{L}}_p{\rm{(\{ }}{{\boldsymbol{w}}_j}{\rm{\} ,}}{\boldsymbol{z}},h,\{ {\lambda _l}\} ,\{ {{\boldsymbol{\phi}}_j}\} {\rm{)}}$ \textit{w.r.t} ${\lambda _l}$ \citep{nesterov2003introductory,xu2020unified}, we can obtain the upper bound of ($1c$):
{\begin{equation}
\renewcommand{\theequation}{A.\arabic{equation}}
\label{eq:A23}
\begin{array}{l}
\sum\limits_{l = 1}^{|{{\bf{A}}^t}|} {\left\langle {{\nabla _{{\lambda _l}}}{{\widetilde{L}}_p}{\rm{(\{ }}{{\boldsymbol{w}}_j^t}{\rm{\} ,}}{\boldsymbol{z}^t},h^t,\{ {\lambda _l^t}\} ,\{ {{\boldsymbol{\phi}}_j^t}\} {\rm{)}} - {\nabla _{{\lambda _l}}}{{\widetilde{L}}_p}{\rm{(\{ }}{{\boldsymbol{w}}_j^t}{\rm{\} ,}}{\boldsymbol{z}^t},h^t,\{ {\lambda _l^{t-1}}\} ,\{ {{\boldsymbol{\phi}}_j^{t-1}}\} {\rm{)}},{\lambda _l^t}\! - \!{\lambda _l^{t-1}}} \right\rangle } \\
\! \le\! \sum\limits_{l = 1}^{|{{\bf{A}}^t}|} (  - \frac{1}{{{L_1}' + {c_1^{t-1}}}}||{\nabla _{{\lambda _l}}}{\widetilde{L}_p}{\rm{(\{ }}{{\boldsymbol{w}}_j^t}{\rm{\} ,}}{\boldsymbol{z}^t},h^t,\{ {\lambda _l^t}\} ,\{ {{\boldsymbol{\phi}}_j^t}\} {\rm{)}} - {\nabla _{{\lambda _l}}}{\widetilde{L}_p}{\rm{(\{ }}{{\boldsymbol{w}}_j^t}{\rm{\} ,}}{\boldsymbol{z}^t},h^t,\{ {\lambda _l^{t-1}}\} ,\{ {{\boldsymbol{\phi}}_j^{t-1}}\} {\rm{)}}|{|^2} \vspace{0.5ex}\\
\qquad \; \; - \frac{{{c_1^{t-1}}{L_1}'}}{{{L_1}' + {c_1^{t-1}}}}||{\lambda _l^t} \! -\! {\lambda _l^{t-1}}|{|^2}).
\end{array}
\end{equation}}

In addition, the following inequality can be obtained,
{\begin{equation}
\renewcommand{\theequation}{A.\arabic{equation}}
\label{eq:A24}
\begin{array}{l}
\frac{1}{{{\rho _1}}}\left\langle {{\lambda _l^t} - {\lambda _l^{t-1}},{\lambda _l^{t + 1}} - {\lambda _l^t}} \right\rangle 
 \le \frac{1}{{2{\rho _1}}}||{\lambda _l^{t + 1}} - {\lambda _l^t}|{|^2} - \frac{1}{{2{\rho _1}}}||{{\boldsymbol{v}}_{1,l}^{t + 1}}|{|^2} + \frac{1}{{2{\rho _1}}}||{\lambda _l^t} - {\lambda _l^{t-1}}|{|^2}.
\end{array}
\end{equation}}

Combining Eq. (A.\ref{eq:A14}), (A.\ref{eq:A15}), (A.\ref{eq:A18}), (A.\ref{eq:A19}), (A.\ref{eq:A23}), (A.\ref{eq:A24}), $ \frac{{{\rho _1}}}{2} \le \frac{1}{{{L_1}' + c_1^0}} $, and setting  ${a_2} = {\rho _1}$, we have:
{\begin{equation}
\renewcommand{\theequation}{A.\arabic{equation}}
\label{eq:A25}
\begin{array}{l}
{L_p}{\rm{(\{ }}{{\boldsymbol{w}}_j^{t+1}}{\rm{\} ,}}{\boldsymbol{z}}^{t + 1},h^{t+1},\{ {\lambda _l^{t+1}}\} ,\{ {{\boldsymbol{\phi}}_j^t}\} {\rm{)}} \! - \! {L_p}{\rm{(\{ }}{{\boldsymbol{w}}_j^{t+1}}{\rm{\} ,}}{\boldsymbol{z}}^{t + 1},h^{t+1},\{ {\lambda _l^t}\} ,\{ {{\boldsymbol{\phi}}_j^t}\} {\rm{)}} \vspace{1ex}\\

\! \le \! \sum\limits_{l = 1}^{|{{\bf{A}}^t}|} \! {(\! \left\langle \! {{\nabla _{{\lambda _l}}}{{\widetilde{L}}_p}{\rm{(\{ }}{{\boldsymbol{w}}_j^{t+1}}{\rm{\},}}{\boldsymbol{z}}^{t + 1},h^{t+1}, \! \{ {\lambda _l^t}\} , \! \{ {{\boldsymbol{\phi}}_j^t}\} {\rm{)}} \! - \! {\nabla _{{\lambda _l}}}{{\widetilde{L}}_p}{\rm{(\{ }}{{\boldsymbol{w}}_j^t}{\rm{\} ,}}{\boldsymbol{z}^t},h^t, \! \{ {\lambda _l^{t-1}}\} , \! \{ {{\boldsymbol{\phi}}_j^{t-1}}\} {\rm{)}},{\lambda _l^{t+1}} \! - \! {\lambda _l^t}} \! \right\rangle } \vspace{1ex}\\

 \!  +  \frac{1}{{{\rho _1}}}\left\langle {{\lambda _l^t} \! - \! {\lambda _l^{t-1}},{\lambda _l^{t+1}} \! - \! {\lambda _l^t}} \right\rangle  \!  + \! \frac{{{c_1^t}}}{2}(||{\lambda _l^{t+1}}|{|^2} \! - \! ||{\lambda _l^t}|{|^2})) \vspace{1ex}\\
\! \le\! \sum\limits_{l = 1}^{|{{\bf{A}}^t}|} {(\frac{{{L^2}}}{{2{a_1}}}(\sum\limits_{j = 1}^N {||{{\boldsymbol{w}}_j^{t+1}} \! - \! {{\boldsymbol{w}}_j^t}|{|^2}}  \!  + \! ||{\boldsymbol{z}}^{t + 1} \! - \! {\boldsymbol{z}^t}|{|^2} \!  + \! ||h^{t+1} \! - \! h^t|{|^2})} \vspace{1ex}\\

 \!  +  (\frac{{{a_1}}}{2} \! - \! \frac{{{c_1^{t-1}} \! - \! {c_1^t}}}{2} \!  + \! \frac{1}{{2{\rho _1}}})||{\lambda _l^{t+1}} \! - \! {\lambda _l^t}|{|^2} \!  + \! \frac{{{c_1^{t-1}} }}{2}(||{\lambda _l^{t+1}}|{|^2} \! - \! ||{\lambda _l^t}|{|^2}) \!  + \! \frac{1}{{2{\rho _1}}}||{\lambda _l^t} \! - \! {\lambda _l^{t-1}}|{|^2})\vspace{1ex}\\
\! =\! \frac{{|{{\bf{A}}^t}|{L^2}}}{{2{a_1}}}(\sum\limits_{j = 1}^N {||{{\boldsymbol{w}}_j^{t+1}} \! - \! {{\boldsymbol{w}}_j^t}|{|^2}}  \!  + \! ||{\boldsymbol{z}}^{t + 1} \! - \! {\boldsymbol{z}^t}|{|^2} \!  + \! ||h^{t+1} \! - \! h^t|{|^2})\\
 \!  +  (\frac{{{a_1}}}{2} \! - \! \frac{{{c_1^{t-1}} \! - \! {c_1^t}}}{2} \!  + \! \frac{1}{{2{\rho _1}}})\sum\limits_{l = 1}^{|{{\bf{A}}^t}|} {||{\lambda _l^{t+1}} \! - \! {\lambda _l^t}|{|^2}}  \!  + \! \frac{{{c_1^{t-1}}}}{2}\sum\limits_{l = 1}^{|{{\bf{A}}^t}|} {(||{\lambda _l^{t+1}}|{|^2} \! - \! ||{\lambda _l^t}|{|^2})}  + \frac{1}{{2{\rho _1}}}\sum\limits_{l = 1}^{|{{\bf{A}}^t}|} {||{\lambda _l^t} \! - \! {\lambda _l^{t-1}}|{|^2}}.
\end{array}
\end{equation}}

According to Eq. (15), $\forall {\boldsymbol{\phi}} \in {{\boldsymbol{\Phi}}}$, it follows that,
{\begin{equation}
\renewcommand{\theequation}{A.\arabic{equation}}
\label{eq:A26}
\left\langle {{{\boldsymbol{\phi}}_j^{t+1}} \! - \! {{\boldsymbol{\phi}}_j^t} - {\rho _2}{\nabla _{{{\boldsymbol{\phi}}_j}}}{{\widetilde{L}}_p}{\rm{(\{ }}{{\boldsymbol{w}}_j^{t+1}}{\rm{\} ,}}{\boldsymbol{z}}^{t+1},h^{t+1},\{ {\lambda _l^{t+1}}\} ,\{ {{\boldsymbol{\phi}}_j^t}\} {\rm{)}},{\boldsymbol{\phi}} \! - \! {{\boldsymbol{\phi}}_j^{t+1}}} \right\rangle  \ge 0.
\end{equation}}

Choosing ${\boldsymbol{\phi}} = {{\boldsymbol{\phi}}_j^t}$, we can obtain,
{\begin{equation}
\renewcommand{\theequation}{A.\arabic{equation}}
\label{eq:A27}
\left\langle {{\nabla _{{{\boldsymbol{\phi}}_j}}}{{\widetilde{L}}_p}{\rm{(\{ }}{{\boldsymbol{w}}_j^{t+1}}{\rm{\} ,}}{\boldsymbol{z}}^{t+1},h^{t+1},\{ {\lambda _l^{t+1}}\} ,\{ {{\boldsymbol{\phi}}_j^t}\} {\rm{)}} \!-\! \frac{1}{{{\rho _2}}}({{\boldsymbol{\phi}}_j^{t+1}}\! -\! {{\boldsymbol{\phi}}_j^t}),{{\boldsymbol{\phi}}_j^t}\! -\! {{\boldsymbol{\phi}}_j^{t+1}}} \right\rangle  \le 0.
\end{equation}}

Likewise, we have,
{\begin{equation}
\renewcommand{\theequation}{A.\arabic{equation}}
\label{eq:A28}
\left\langle {{\nabla _{{{\boldsymbol{\phi}}_j}}}{{\widetilde{L}}_p}{\rm{(\{ }}{{\boldsymbol{w}}_j^t}{\rm{\} ,}}{\boldsymbol{z}^t},h^t,\{ {\lambda _l^t}\} ,\{ {{\boldsymbol{\phi}}_j^{t-1}}\} {\rm{)}} \!-\! \frac{1}{{{\rho _2}}}({{\boldsymbol{\phi}}_j^t}\! -\! {{\boldsymbol{\phi}}_j^{t-1}}),{{\boldsymbol{\phi}}_j^{t+1}} \!-\! {{\boldsymbol{\phi}}_j^t}} \right\rangle  \le 0.
\end{equation}}

Since ${\widetilde{L}_p}{\rm{(\{ }}{{\boldsymbol{w}}_j}{\rm{\} ,}}{\boldsymbol{z}},h,\{ {\lambda _l}\} ,\{ {{\boldsymbol{\phi}}_j}\} {\rm{)}}$ is concave with respect to ${{\boldsymbol{\phi}}_j}$ and follows from Eq. (A.\ref{eq:A28}):
{\begin{equation}
\renewcommand{\theequation}{A.\arabic{equation}}
\label{eq:A29}
\begin{array}{l}
{\widetilde{L}_p}{\rm{(\{ }}{{\boldsymbol{w}}_j^{t+1}}{\rm{\} ,}}{\boldsymbol{z}}^{t+1},h^{t+1},\{ {\lambda _l^{t+1}}\} ,\{ {{\boldsymbol{\phi}}_j^{t+1}}\} {\rm{)}} \! - \! {\widetilde{L}_p}{\rm{(\{ }}{{\boldsymbol{w}}_j^{t+1}}{\rm{\} ,}}{\boldsymbol{z}}^{t+1},h^{t+1},\{ {\lambda _l^{t+1}}\} ,\{ {{\boldsymbol{\phi}}_j^t}\} {\rm{)}} \vspace{0.5ex}\\

\! \le \! \sum\limits_{j = 1}^N \! {\left\langle \! {{\nabla  _{{{\boldsymbol{\phi}}_j}}}{{\widetilde{L}}_p}{\rm{(\{ }}{{\boldsymbol{w}}_j^{t+1}}{\rm{\} ,}}{\boldsymbol{z}}^{t+1},h^{t+1},\{ {\lambda _l^{t+1}}\} ,\{ {{\boldsymbol{\phi}}_j^t}\} {\rm{)}},{{\boldsymbol{\phi}}_j^{t+1}} \! - \! {{\boldsymbol{\phi}}_j^t}}\! \right\rangle } \vspace{0.5ex} \\
 
\! \le \! \sum\limits_{j = 1}^N \! {(\! \left\langle \! {{\nabla \! _{{{\boldsymbol{\phi}}_j}}}{{\widetilde{L}}_p}{\rm{(\{ }}{{\boldsymbol{w}}_j^{t+1}}{\rm{\} ,}}{\boldsymbol{z}}^{t+1},h^{t+1}, \! \{ {\lambda _l^{t+1}}\}, \! \{ {{\boldsymbol{\phi}}_j^t}\} {\rm{)}} \! - \! {\nabla \! _{{{\boldsymbol{\phi}}_j}}}{{\widetilde{L}}_p}{\rm{(\{ }}{{\boldsymbol{w}}_j^t}{\rm{\},}}{\boldsymbol{z}^t},h^t, \!\{ {\lambda _l^t}\} , \! \{ {{\boldsymbol{\phi}}_j^{t-1}}\} {\rm{)}},  {{\boldsymbol{\phi}}_j^{t+1}} \! - \! {{\boldsymbol{\phi}}_j^t}}\! \right\rangle } \vspace{1ex}\\
 \qquad \quad +  \frac{1}{{{\rho _2}}}\left\langle {{{\boldsymbol{\phi}}_j^t} \! - \! {{\boldsymbol{\phi}}_j^{t-1}},{{\boldsymbol{\phi}}_j^{t+1}} \! - \! {{\boldsymbol{\phi}}_j^t}}  \right\rangle ).
\end{array}
\end{equation}}

Denoting ${{\boldsymbol{v}}_{2,l}^{t + 1}} = {{\boldsymbol{\phi}}_j^{t+1}} - {{\boldsymbol{\phi}}_j^t} - ({{\boldsymbol{\phi}}_j^t} - {{\boldsymbol{\phi}}_j^{t-1}})$, we can write the first term in the last inequality of Eq.  (A.\ref{eq:A29}) as
{\begin{equation}
\renewcommand{\theequation}{A.\arabic{equation}}
\label{eq:A30}
\begin{array}{l}
\sum\limits_{j = 1}^N \! {\left\langle \! {{\nabla _{{{\boldsymbol{\phi}}_j}}}{{\widetilde{L}}_p}{\rm{(\{ }}{{\boldsymbol{w}}_j^{t+1}}{\rm{\},}}{\boldsymbol{z}}^{t+1},h^{t+1}, \!\{ {\lambda _l^{t+1}}\},\{ {{\boldsymbol{\phi}}_j^t}\} {\rm{)}} \! - \! {\nabla _{{{\boldsymbol{\phi}}_j}}}{{\widetilde{L}}_p}{\rm{(\{ }}{{\boldsymbol{w}}_j^t}{\rm{\},}}{\boldsymbol{z}^t},h^t, \{ {\lambda _l^t}\},\{ {{\boldsymbol{\phi}}_j^{t-1}}\} {\rm{)}},{{\boldsymbol{\phi}}_j^{t+1}} \! - \! {{\boldsymbol{\phi}}_j^t}} \! \right\rangle } \vspace{0.5ex}\\

 \!= \! \sum\limits_{j = 1}^N \! {\left\langle \! {{\nabla\! _{{{\boldsymbol{\phi}}_j}}}{{\widetilde{L}}_p}{\rm{(\{ }}{{\boldsymbol{w}}_j^{t+1}}{\rm{\} ,}}{\boldsymbol{z}}^{t+1},h^{t+1}, \! \{ {\lambda _l^{t+1}}\}, \! \{ {{\boldsymbol{\phi}}_j^t}\} {\rm{)}} \! - \! {\nabla \! _{{{\boldsymbol{\phi}}_j}}}{{\widetilde{L}}_p}{\rm{(\{ }}{{\boldsymbol{w}}_j^t}{\rm{\},}}{\boldsymbol{z}^t},h^t, \! \{ {\lambda _l^t}\} , \! \{ {{\boldsymbol{\phi}}_j^t}\} {\rm{)}},{{\boldsymbol{\phi}}_j^{t+1}} \! - \! {{\boldsymbol{\phi}}_j^t}} \! \right\rangle } (2a) \vspace{0.5ex}\\
 \! + \! \sum\limits_{j = 1}^N \! {\left\langle\! {{\nabla _{{{\boldsymbol{\phi}}_j}}}{{\widetilde{L}}_p}{\rm{(\{ }}{{\boldsymbol{w}}_j^t}{\rm{\} ,}}{\boldsymbol{z}^t},h^t, \!\{ {\lambda _l^t}\} , \!\{ {{\boldsymbol{\phi}}_j^t}\} {\rm{)}} \! - \! {\nabla _{{{\boldsymbol{\phi}}_j}}}{{\widetilde{L}}_p}{\rm{(\{ }}{{\boldsymbol{w}}_j^t}{\rm{\} ,}}{\boldsymbol{z}^t},h^t,\!\{ {\lambda _l^t}\}, \!\{ {{\boldsymbol{\phi}}_j^{t-1}}\} {\rm{)}},{{\boldsymbol{v}}_{2,l}^{t + 1}}} \!\right\rangle } (2b) \vspace{0.5ex}\\
 
 \! + \! \sum\limits_{j = 1}^N {\left\langle\! {{\nabla _{{{\boldsymbol{\phi}}_j}}}{{\widetilde{L}}_p}{\rm{(\{ }}{{\boldsymbol{w}}_j^t}{\rm{\} ,}}{\boldsymbol{z}^t},h^t,\!\{ {\lambda _l^t}\},\!\{ {{\boldsymbol{\phi}}_j^t}\} {\rm{)}} \! - \! {\nabla _{{{\boldsymbol{\phi}}_j}}}{{\widetilde{L}}_p}{\rm{(\{ }}{{\boldsymbol{w}}_j^t}{\rm{\},}}{\boldsymbol{z}^t},h^t,\!\{ {\lambda _l^t}\} ,\!\{ {{\boldsymbol{\phi}}_j^{t-1}}\} {\rm{)}},{{\boldsymbol{\phi}}_j^t} \! - \! {{\boldsymbol{\phi}}_j^{t-1}}}\! \right\rangle } (2c).
\end{array}
\end{equation}}

We firstly focus on the ($2a$) in Eq. (A.\ref{eq:A30}), we can write the ($2a$) as,
{\begin{equation}
\renewcommand{\theequation}{A.\arabic{equation}}
\label{eq:A31}
\begin{array}{l}
\left\langle\! {{\nabla _{{{\boldsymbol{\phi}}_j}}}{{\widetilde{L} }_p}{\rm{(\{ }}{{\boldsymbol{w}}_j^{t+1}}{\rm{\} ,}}{\boldsymbol{z}}^{t+1},h^{t+1},\{ {\lambda _l^{t+1}}\},\{ {{\boldsymbol{\phi}}_j^t}\} {\rm{)}}\! - \!{\nabla _{{{\boldsymbol{\phi}}_j}}}{{\widetilde{L} }_p}{\rm{(\{ }}{{\boldsymbol{w}}_j^t}{\rm{\} ,}}{\boldsymbol{z}^t},h^t,\{ {\lambda _l^t}\},\{ {{\boldsymbol{\phi}}_j^t}\} {\rm{)}},{{\boldsymbol{\phi}}_j^{t+1}}\! -\! {{\boldsymbol{\phi}}_j^t}}\! \right\rangle \vspace{1ex} \\
\! = \!\left\langle {{\nabla _{{{\boldsymbol{\phi}}_j}}}{L_p}{\rm{(\{ }}{{\boldsymbol{w}}_j^{t+1}}{\rm{\},}}{\boldsymbol{z}}^{t+1},h^{t+1},\{ {\lambda _l^{t+1}}\} ,\{ {{\boldsymbol{\phi}}_j^t}\} {\rm{)}} \!-\! {\nabla _{{{\boldsymbol{\phi}}_j}}}{L_p}{\rm{(\{ }}{{\boldsymbol{w}}_j^t}{\rm{\} ,}}{\boldsymbol{z}^t},h^t,\{ {\lambda _l^t}\} ,\{ {{\boldsymbol{\phi}}_j^t}\} {\rm{)}},{{\boldsymbol{\phi}}_j^{t+1}} \!-\! {{\boldsymbol{\phi}}_j^t}} \right\rangle \vspace{1ex} \\
\! + ({c_2^{t-1}} - {c_2^t})\left\langle {{{\boldsymbol{\phi}}_j^t},{{\boldsymbol{\phi}}_j^{t+1}} - {{\boldsymbol{\phi}}_j^t}} \right\rangle \vspace{1ex}\\

\! = \! \left\langle  {{\nabla _{{{\boldsymbol{\phi}}_j}}}{L_p}{\rm{(\{ }}{{\boldsymbol{w}}_j^{t+1}}{\rm{\},}}{\boldsymbol{z}}^{t+1},h^{t+1},\{ {\lambda _l^{t+1}}\},\{ {{\boldsymbol{\phi}}_j^t}\} {\rm{)}} \!-\! {\nabla _{{{\boldsymbol{\phi}}_j}}}{L_p}{\rm{(\{ }}{{\boldsymbol{w}}_j^t}{\rm{\},}}{\boldsymbol{z}^t},h^t,\{ {\lambda _l^t}\} , \{ {{\boldsymbol{\phi}}_j^t}\} {\rm{)}},{{\boldsymbol{\phi}}_j^{t+1}}\! -\! {{\boldsymbol{\phi}}_j^t}} \right\rangle \vspace{1ex}
 \vspace{1ex}\\
  \! +  \frac{{{c_2^{t-1}}  -  {c_2^t}}}{2}(||{{\boldsymbol{\phi}}_j^{t+1}}|{|^2} \! - \! ||{{\boldsymbol{\phi}}_j^t}|{|^2}) \! - \! \frac{{{c_2^{t-1}}  -  {c_2^t}}}{2}||{{\boldsymbol{\phi}}_j^{t+1}} \! - \! {{\boldsymbol{\phi}}_j^t}|{|^2}).
\end{array}
\end{equation}}

\vspace{1ex}

And according to Cauchy-Schwarz inequality and Assumption 1, we can obtain,
{\begin{equation}
\renewcommand{\theequation}{A.\arabic{equation}}
\label{eq:A32}
\begin{array}{l}
\left\langle {{\nabla _{{{\boldsymbol{\phi}}_j}}}{L_p}{\rm{(\{ }}{{\boldsymbol{w}}_j^{t+1}}{\rm{\},}}{\boldsymbol{z}}^{t + 1},h^{t+1},\{ {\lambda _l^{t+1}}\},\{ {{\boldsymbol{\phi}}_j^t}\} {\rm{)}}\! -\! {\nabla _{{{\boldsymbol{\phi}}_j}}}{L_p}{\rm{(\{ }}{{\boldsymbol{w}}_j^t}{\rm{\} ,}}{\boldsymbol{z}^t},h^t,\{ {\lambda _l^t}\} ,\{ {{\boldsymbol{\phi}}_j^t}\} {\rm{)}},{{\boldsymbol{\phi}}_j^{t+1}}\! -\! {{\boldsymbol{\phi}}_j^t}} \right\rangle \vspace{1.5ex}\\
\! =\! \left\langle {{\nabla _{{{\boldsymbol{\phi}}_j}}}{L_p}{\rm{(\{ }}{{\boldsymbol{w}}_j^{t+1}}{\rm{\} ,}}{\boldsymbol{z}}^{t+1},h^{t+1},\{ {\lambda _l^t}\} ,\{ {{\boldsymbol{\phi}}_j^t}\} {\rm{)}}\! -\! {\nabla _{{{\boldsymbol{\phi}}_j}}}{L_p}{\rm{(\{ }}{{\boldsymbol{w}}_j^t}{\rm{\},}}{\boldsymbol{z}^t},h^t,\{ {\lambda _l^t}\} , \{ {{\boldsymbol{\phi}}_j^t}\} {\rm{)}},{{\boldsymbol{\phi}}_j^{t+1}}\! -\! {{\boldsymbol{\phi}}_j^t}} \right\rangle \vspace{1ex}\\
\! \le \! \frac{{{L^2}}}{{2{a_3}}}(\sum\limits_{j = 1}^N {||{{\boldsymbol{w}}_j^{t+1}}\! -\! {{\boldsymbol{w}}_j^t}|{|^2}}  + ||{\boldsymbol{z}}^{t+1} \!-\! {\boldsymbol{z}^t}|{|^2} + ||h^{t+1}\! -\! h^t|{|^2}) + \frac{{{a_3}}}{2}||{{\boldsymbol{\phi}}_j^{t+1}}\! -\! {{\boldsymbol{\phi}}_j^t}|{|^2},
\end{array}
\end{equation}}

where $a_3 >0$ is a constant. Thus, we can obtain the upper bound of ($2a$) by combining the above Eq. (A.\ref{eq:A31}) and Eq. (A.\ref{eq:A32}),
{\begin{equation}
\renewcommand{\theequation}{A.\arabic{equation}}
\label{eq:A33}
\begin{array}{l}
\sum\limits_{j = 1}^N \! {\left\langle\! {{\nabla _{{{\boldsymbol{\phi}}_j}}}{{\widetilde{L} }_p}{\rm{(\{ }}{{\boldsymbol{w}}_j^{t+1}}{\rm{\},}}{\boldsymbol{z}}^{t+1},h^{t+1},\{ {\lambda _l^{t+1}}\},\{ {{\boldsymbol{\phi}}_j^t}\} {\rm{)}} \! - \! {\nabla _{{{\boldsymbol{\phi}}_j}}}{{\widetilde{L} }_p}{\rm{(\{ }}{{\boldsymbol{w}}_j^t}{\rm{\},}}{\boldsymbol{z}^t},h^t,\{ {\lambda _l^t}\} ,\{ {{\boldsymbol{\phi}}_j^t}\} {\rm{)}},{{\boldsymbol{\phi}}_j^{t+1}} \! - \! {{\boldsymbol{\phi}}_j^t}}\! \right\rangle } \vspace{0.5ex}\\
\! = \! \sum\limits_{j \in {{\bf{Q}}^{t + 1}}}^{} \! {\left\langle\! {{\nabla _{{{\boldsymbol{\phi}}_j}}}{{\widetilde{L} }_p}{\rm{(\{ }}{{\boldsymbol{w}}_j^{t+1}}{\rm{\},}}{\boldsymbol{z}}^{t+1},h^{t+1}, \! \{ {\lambda _l^{t+1}}\}, \! \{ {{\boldsymbol{\phi}}_j^t}\} {\rm{)}} \! - \! {\nabla _{{{\boldsymbol{\phi}}_j}}}{{\widetilde{L} }_p}{\rm{(\{ }}{{\boldsymbol{w}}_j^t}{\rm{\},}}{\boldsymbol{z}^t},h^t, \! \{ {\lambda _l^t}\} , \! \{ {{\boldsymbol{\phi}}_j^t}\} {\rm{)}},{{\boldsymbol{\phi}}_j^{t+1}} \! - \! {{\boldsymbol{\phi}}_j^t}}\! \right\rangle } \vspace{0.5ex}\\
\! \le \! \sum\limits_{j \in {{\bf{Q}}^{t + 1}}}^{} \! {(\frac{{{L^2}}}{{2{a_3}}}(\sum\limits_{j = 1}^N {||{{\boldsymbol{w}}_j^{t+1}} \! - \! {{\boldsymbol{w}}_j^t}|{|^2}}  \! + \! ||{\boldsymbol{z}}^{t+1} \! - \! {\boldsymbol{z}^t}|{|^2} \! + \! ||h^{t+1} \! - \! h^t|{|^2}) \! + \! \frac{{{a_3}}}{2}||{{\boldsymbol{\phi}}_j^{t+1}} \! - \! {{\boldsymbol{\phi}}_j^t}|{|^2}} \vspace{1ex}\\
 \qquad \qquad +  \frac{{{c_2^{t-1}}  -  {c_2^t}}}{2}(||{{\boldsymbol{\phi}}_j^{t+1}}|{|^2} \! - \! ||{{\boldsymbol{\phi}}_j^t}|{|^2}) \! - \! \frac{{{c_2^{t-1}}  -  {c_2^t}}}{2}||{{\boldsymbol{\phi}}_j^{t+1}} \! - \! {{\boldsymbol{\phi}}_j^t}|{|^2}).
\end{array}
\end{equation}}

\vspace{1ex}

Next we focus on the ($2b$) in Eq. (A.\ref{eq:A30}). According to Cauchy-Schwarz inequality we can write the ($2b$) as
\begin{equation}
\renewcommand{\theequation}{A.\arabic{equation}}
\label{eq:A34}
\begin{array}{l}
\sum\limits_{j = 1}^N \!\left\langle \! {{\nabla _{{{\boldsymbol{\phi}}_j}}}{{\widetilde{L}}_p}{\rm{(\{ }}{{\boldsymbol{w}}_j^t}{\rm{\},}}{\boldsymbol{z}^t},h^t, \{ {\lambda _l^t}\}, \!\{ {{\boldsymbol{\phi}}_j^t}\} {\rm{)}} \!-\! {\nabla _{{{\boldsymbol{\phi}}_j}}}{{\widetilde{L}}_p}{\rm{(\{ }}{{\boldsymbol{w}}_j^t}{\rm{\} ,}}{\boldsymbol{z}^t},h^t,\{ {\lambda _l^t}\},\{ {{\boldsymbol{\phi}}_j^{t-1}}\} {\rm{)}},{\boldsymbol{v}}_{2,l}^{t+1}} \! \right\rangle \\
\! \le \! \sum\limits_{j = 1}^N \! {(\frac{{{a_4}}}{2}||{\nabla _{{{\boldsymbol{\phi}}_j}}}{{\widetilde{L}}_p}{\rm{(\{ }}{{\boldsymbol{w}}_j^t}{\rm{\},}}{\boldsymbol{z}^t},h^t, \! \{ {\lambda _l^t}\} , \!\{ {{\boldsymbol{\phi}}_j^t}\} {\rm{)}} \!-\! {\nabla _{{{\boldsymbol{\phi}}_j}}}{{\widetilde{L}}_p}{\rm{(\{ }}{{\boldsymbol{w}}_j^t}{\rm{\},}}{\boldsymbol{z}^t},h^t, \! \{ {\lambda _l^t}\} , \!\{ {{\boldsymbol{\phi}}_j^{t-1}}\} {\rm{)}}|{|^2}\! +\! \frac{1}{{2{a_4}}}\!||{\boldsymbol{v}}_{2,l}^{t+1}|{|^2})},
\end{array}
\end{equation}
where $a_4>0$ is a constant. Then, we focus on the ($2c$) in Eq. (A.\ref{eq:A30}), we have,
{\begin{equation}
\renewcommand{\theequation}{A.\arabic{equation}}
\label{eq:A35}
\begin{array}{l}
||{\nabla _{{{\boldsymbol{\phi}}_j}}}{\widetilde{L}_p}{\rm{(\{ }}{{\boldsymbol{w}}_j^t}{\rm{\} ,}}{\boldsymbol{z}^t},h^t,\{ {\lambda _l^t}\} ,\{ {{\boldsymbol{\phi}}_j^t}\} {\rm{)}} \!-\! {\nabla _{{{\boldsymbol{\phi}}_j}}}{\widetilde{L}_p}{\rm{(\{ }}{{\boldsymbol{w}}_j^t}{\rm{\} ,}}{\boldsymbol{z}^t},h^t,\{ {\lambda _l^t}\} ,\{ {{\boldsymbol{\phi}}_j^{t-1}}\} {\rm{)||}}\vspace{1.5ex}\\

\! \le\! ||{\nabla _{{{\boldsymbol{\phi}}_j}}}{L_p}{\rm{(\{ }}{{\boldsymbol{w}}_j^t}{\rm{\},}}{\boldsymbol{z}^t},h^t, \! \{ {\lambda _l^t}\} , \!\{ {{\boldsymbol{\phi}}_j^t}\} {\rm{)}}\! -\! {\nabla _{{{\boldsymbol{\phi}}_j}}}{L_p}{\rm{(\{ }}{{\boldsymbol{w}}_j^t}{\rm{\} ,}}{\boldsymbol{z}^t},h^t, \! \{ {\lambda _l^t}\}, \! \{ {{\boldsymbol{\phi}}_j^{t-1}}\} {\rm{)}}||\! +\! {c_2^{t-1}}||{{\boldsymbol{\phi}}_j^t}\! -\! {{\boldsymbol{\phi}}_j^{t-1}}{\rm{||}}\vspace{1.5ex}\\

\! \le\! (L + {c_2^{t-1}})||{{\boldsymbol{\phi}}_j^t} \!-\! {{\boldsymbol{\phi}}_j^{t-1}}{\rm{||}},
\end{array}
\end{equation}}

where the last inequality comes from Assumption 1 and the trigonometric inequality. Denoting ${L_2}' = L  +  {c_2^0}$, we can obtain,
{\begin{equation}
\renewcommand{\theequation}{A.\arabic{equation}}
\label{eq:A36}
\begin{array}{l}
||{\nabla _{{{\boldsymbol{\phi}}_j}}}{\widetilde{L}_p}{\rm{(\{ }}{{\boldsymbol{w}}_j^t}{\rm{\} ,}}{\boldsymbol{z}^t},h^t,\{ {\lambda _l^t}\} ,\{ {{\boldsymbol{\phi}}_j^t}\} {\rm{)}}\! -\! {\nabla _{{{\boldsymbol{\phi}}_j}}}{\widetilde{L}_p}{\rm{(\{ }}{{\boldsymbol{w}}_j^t}{\rm{\} ,}}{\boldsymbol{z}^t},h^t,\{ {\lambda _l^t}\} ,\{ {{\boldsymbol{\phi}}_j^{t-1}}\} {\rm{)||}}
 \le {L_2}'||{{\boldsymbol{\phi}}_j^t} \!-\! {{\boldsymbol{\phi}}_j^{t-1}}{\rm{||}}.
\end{array}
\end{equation}}

\vspace{1ex}

Following Eq. (A.\ref{eq:A36}) and the strong  concavity of ${\widetilde{L}}_p{\rm{(\{ }}{{\boldsymbol{w}}_j}{\rm{\} ,}}{\boldsymbol{z}},h,\{ {\lambda _l}\} ,\{ {{\boldsymbol{\phi}}_j}\} {\rm{)}}$ \textit{w.r.t} ${\boldsymbol{\phi}}_j$, we can obtain the upper bound of ($2c$),
{\begin{equation}
\renewcommand{\theequation}{A.\arabic{equation}}
\label{eq:A37}
\begin{array}{l}
\sum\limits_{j = 1}^N \! {\left\langle\! {{\nabla _{{{\boldsymbol{\phi}}_j}}}{{\widetilde{L}}_p}{\rm{(\{ }}{{\boldsymbol{w}}_j^t}{\rm{\},}}{\boldsymbol{z}^t},h^t, \! \{ {\lambda _l^t}\}, \! \{ {{\boldsymbol{\phi}}_j^t}\} {\rm{)}}\! - \!{\nabla _{{{\boldsymbol{\phi}}_j}}}{{\widetilde{L}}_p}{\rm{(\{ }}{{\boldsymbol{w}}_j^t}{\rm{\},}}{\boldsymbol{z}^t},h^t, \!\{ {\lambda _l^t}\} , \!\{ {{\boldsymbol{\phi}}_j^{t-1}}\} {\rm{)}},{{\boldsymbol{\phi}}_j^t}\! -\! {{\boldsymbol{\phi}}_j^{t-1}}}\! \right\rangle } \vspace{0.5ex}\\
\! \le \! \sum\limits_{j = 1}^N \! {( - \frac{1}{{{L_2}' + {c_2^{t-1}}}}||{\nabla _{{{\boldsymbol{\phi}}_j}}}{{\widetilde{L}}_p}{\rm{(\{ }}{{\boldsymbol{w}}_j^t}{\rm{\} ,}}{\boldsymbol{z}^t},h^t, \! \{ {\lambda _l^t}\} , \! \{ {{\boldsymbol{\phi}}_j^t}\} {\rm{)}}\! -\! {\nabla _{{{\boldsymbol{\phi}}_j}}}{{\widetilde{L}}_p}{\rm{(\{ }}{{\boldsymbol{w}}_j^t}{\rm{\} ,}}{\boldsymbol{z}^t},h^t, \! \{ {\lambda _l^t}\} , \!\{ {{\boldsymbol{\phi}}_j^{t-1}}\} )|{|^2}} \vspace{1ex}\\
\qquad \;\; - \frac{{{c_2^{t-1}}{L_2}'}}{{{L_2}' + {c_2^{t-1}}}}||{{\boldsymbol{\phi}}_j^t} \!-\! {{\boldsymbol{\phi}}_j^{t-1}}|{|^2}).
\end{array}
\end{equation}}

In addition, the following inequality can also be obtained,
{\begin{equation}
\renewcommand{\theequation}{A.\arabic{equation}}
\label{eq:A38}
\begin{array}{l}
\sum\limits_{j = 1}^N {\frac{1}{{{\rho _2}}}\left\langle {{{\boldsymbol{\phi}}_j^t} \!-\! {{\boldsymbol{\phi}}_j^{t-1}},{{\boldsymbol{\phi}}_j^{t+1}} \!-\! {{\boldsymbol{\phi}}_j^t}} \right\rangle } 
 \le \sum\limits_{j = 1}^N {(\frac{1}{{2{\rho _2}}}||{{\boldsymbol{\phi}}_j^{t+1}} \!-\! {{\boldsymbol{\phi}}_j^t}|{|^2} - \frac{1}{{2{\rho _2}}}||{{\boldsymbol{v}}_{2,l}^{t+1}}|{|^2} + \frac{1}{{2{\rho _2}}}||{{\boldsymbol{\phi}}_j^t} - {{\boldsymbol{\phi}}_j^{t-1}}|{|^2})}.
\end{array}
\end{equation}}

Combining Eq. (A.\ref{eq:A29}), (A.\ref{eq:A30}), (A.\ref{eq:A33}), (A.\ref{eq:A34}), (A.\ref{eq:A37}), (A.\ref{eq:A38}),  $ \frac{{{\rho _2}}}{2} \le \frac{1}{{{L_2}' + c_2^0}}$, and setting ${a_4} = {\rho _2}$, we have,
{\begin{equation}
\renewcommand{\theequation}{A.\arabic{equation}}
\label{eq:A39}
\begin{array}{l}
{L_p}{\rm{(\{ }}{{\boldsymbol{w}}_j^{t+1}}{\rm{\} ,}}{\boldsymbol{z}}^{t+1},h^{t+1},\{ {\lambda _l^{t+1}}\},\{ {{\boldsymbol{\phi}}_j^{t+1}}\} {\rm{)}} \! - \! {L_p}{\rm{(\{ }}{{\boldsymbol{w}}_j^{t+1}}{\rm{\} ,}}{\boldsymbol{z}}^{t+1},h^{t+1},\{ {\lambda _l^{t+1}}\},\{ {{\boldsymbol{\phi}}_j^t}\} {\rm{)}}\vspace{1ex}\\

\! \le \! \sum\limits_{j = 1}^N \! {(\!\left\langle\! {{\nabla \! _{{{\boldsymbol{\phi}}_j}}}{{\widetilde{L}}_p}{\rm{(\{ }}{{\boldsymbol{w}}_j^{t+1}}{\rm{\} ,}}{\boldsymbol{z}}^{t+1},h^{t+1}, \! \{ {\lambda _l^{t+1}}\} , \! \{ {{\boldsymbol{\phi}}_j^t}\} {\rm{)}} \! - \! {\nabla \! _{{{\boldsymbol{\phi}}_j}}}{{\widetilde{L}}_p}{\rm{(\{ }}{{\boldsymbol{w}}_j^t}{\rm{\} ,}}{\boldsymbol{z}^t},h^t, \! \{ {\lambda _l^t}\}, \{ {{\boldsymbol{\phi}}_j^{t-1}}\} {\rm{)}},{{\boldsymbol{\phi}}_j^{t+1}} \! - \! {{\boldsymbol{\phi}}_j^t}}\! \right\rangle } \vspace{0.5ex}\\
 \qquad \;\; +  \frac{1}{{{\rho _2}}}\left\langle {{{\boldsymbol{\phi}}_j^t} \! - \! {{\boldsymbol{\phi}}_j^{t-1}},{{\boldsymbol{\phi}}_j^{t+1}} \! - \! {{\boldsymbol{\phi}}_j^t}} \right\rangle  \! + \! \frac{{{c_2^t}}}{2}(||{{\boldsymbol{\phi}}_j^{t+1}}|{|^2} \! - \! ||{{\boldsymbol{\phi}}_j^t}|{|^2}))\vspace{1ex}\\
\! \le \! \frac{{|{{\bf{Q}}^{t + 1}}|{L^2}}}{{2{a_3}}}(\sum\limits_{j = 1}^N {||{{\boldsymbol{w}}_j^{t+1}} \! - \! {{\boldsymbol{w}}_j^t}|{|^2}}  \! + \! ||{\boldsymbol{z}}^{t+1} \! - \! {\boldsymbol{z}^t}|{|^2} \! + \! ||h^{t+1} \! - \! h^t|{|^2})\vspace{0.5ex}\\
 \! +  (\frac{{{a_3}}}{2} \! - \! \frac{{{c_2^{t-1}}  -  {c_2^t}}}{2} \! + \! \frac{1}{{2{\rho _2}}})\! \sum\limits_{j = 1}^N\! {||{{\boldsymbol{\phi}}_j^{t+1}} \! - \! {{\boldsymbol{\phi}}_j^t}|{|^2}}  \! + \! \frac{{{c_2^{t-1}}}}{2}\! \sum\limits_{j = 1}^N\! {(||{{\boldsymbol{\phi}}_j^{t+1}}|{|^2} \! - \! ||{{\boldsymbol{\phi}}_j^t}|{|^2})} 
 \! + \! \frac{1}{{2{\rho _2}}}\! \sum\limits_{j = 1}^N \! {||{{\boldsymbol{\phi}}_j^t} \! - \! {{\boldsymbol{\phi}}_j^{t-1}}|{|^2}}. 
\end{array}
\end{equation}}

By combining Lemma \ref{lemma 1} with Eq. (A.\ref{eq:A25}) and Eq. (A.\ref{eq:A39}), we conclude the proof of Lemma \ref{lemma 2}.

\vspace{3ex}

\begin{lemma} \label{lemma3}
Firstly,  we denote ${S_1^{t+1}}$, ${S_2^{t+1}}$ and ${F^{t+1}}$  as,
\begin{equation}
\renewcommand{\theequation}{A.\arabic{equation}}
\label{eq:A40}
{S_1^{t+1}} = \frac{4}{{{\rho _1}^2{c_1^{t+1}}}}\sum\limits_{l = 1}^{|{{\bf{A}}^t}|} {||{\lambda _l^{t+1}} \! - \! {\lambda _l^t}|{|^2}}  \! - \! \frac{4}{{{\rho _1}}}(\frac{{{c_1^{t-1}}}}{{{c_1^t}}} \! - \! 1)\sum\limits_{l = 1}^{|{{\bf{A}}^t}|} {||{\lambda _l^{t+1}}|{|^2}},
\end{equation}
\begin{equation}
\renewcommand{\theequation}{A.\arabic{equation}}
\label{eq:A41}
{S_2^{t+1}} = \frac{4}{{{\rho _2}^2{c_2^{t+1}}}}\sum\limits_{j = 1}^N {||{{\boldsymbol{\phi}}_j^{t+1}} \! - \! {{\boldsymbol{\phi}}_j^t}|{|^2}}  \! - \! \frac{4}{{{\rho _2}}}(\frac{{{c_2^{t-1}}}}{{{c_2^t}}} \! - \! 1)\sum\limits_{j = 1}^N {||{{\boldsymbol{\phi}}_j^{t+1}}|{|^2}}, 
\end{equation}
\begin{equation}
\renewcommand{\theequation}{A.\arabic{equation}}
\label{eq:A42}
\begin{array}{l}
F^{t+1} = {L_p}{\rm{(\{ }}{{\boldsymbol{w}}_j^{t+1}}{\rm{\} ,}}{\boldsymbol{z}}^{t+1},h^{t+1},\{ {\lambda _l^{t+1}}\} ,\{ {{\boldsymbol{\phi}}_j^{t+1}}\} {\rm{)}}  +  {S_1^{t+1}}  +  {S_2^{t+1}}\vspace{1ex}\\
 \quad  \quad   \quad \;    \! -  \frac{7}{{2{\rho _1}}}\!\sum\limits_{l = 1}^{|{{\bf{A}}^t}|}\! {||{\lambda _l^{t+1}} \! - \! {\lambda _l^t}|{|^2}} 
 \! - \! \frac{c_1^t}{{2}}\!\sum\limits_{l = 1}^{|{{\bf{A}}^t}|}\! {||{\lambda _l^{t+1}}|{|^2}} 
 
 \! - \! \frac{7}{{2{\rho _2}}}\! \sum\limits_{j = 1}^N\! {||{{\boldsymbol{\phi}}_j^{t+1}} \! - \! {{\boldsymbol{\phi}}_j^t}|{|^2}}
 
  \! - \! \frac{c_2^t}{{2}} \! \sum\limits_{j = 1}^N\! {||{{\boldsymbol{\phi}}_j^{t+1}}|{|^2}},
\end{array}
\end{equation}
then $\forall t \ge T_1 + \tau$, we have,
{\begin{equation}
\renewcommand{\theequation}{A.\arabic{equation}}
\label{eq:A43}
\begin{array}{l}
F^{t+1} \! - \! F^{t}
\! \le \! (\frac{{L  +  1}}{2} \! - \! \frac{1}{{{\eta _{\boldsymbol{w}}^t}}} \! + \! \frac{{{\rho _1}|{{\bf{A}}^t}|{L^2}}}{2} \! + \! \frac{{{\rho _2}|{{\bf{Q}}^{t + 1}}|{L^2}}}{2} \! + \! \frac{{8|{{\bf{A}}^t}|{L^2}}}{{{\rho _1}({c_1^t}){^2}}} \! + \! \frac{{8N{L^2}}}{{{\rho _2}({c_2^t}){^2}}})\sum\limits_{j = 1}^N {||{{\boldsymbol{w}}_j^{t+1}} \! - \! {{\boldsymbol{w}}_j^t}|{|^2}} \vspace{1ex}\\

\quad \quad \quad \quad \;\;  \! +  (\frac{{L  +  3\tau{ k_1 }N{L^2}}}{2} \! - \! \frac{1}{{{\eta _{\boldsymbol{z}}^t}}} \! + \! \frac{{{\rho _1}|{{\bf{A}}^t}|{L^2}}}{2} \! + \! \frac{{{\rho _2}|{{\bf{Q}}^{t + 1}}|{L^2}}}{2} \! + \! \frac{{8|{{\bf{A}}^t}|{L^2}}}{{{\rho _1}({c_1^t}){^2}}} \! + \! \frac{{8N{L^2}}}{{{\rho _2}({c_2^t}){^2}}})||\boldsymbol{z}^{t+1} \! - \! \boldsymbol{z}^t|{|^2}\vspace{1ex}\\

\quad \quad \quad \quad \;\; \! +  (\frac{{L  +  3\tau{ k_1 }N{L^2}}}{2} \! - \! \frac{1}{{{\eta _h^t}}} \! + \! \frac{{{\rho _1}|{{\bf{A}}^t}|{L^2}}}{2} \! + \! \frac{{{\rho _2}|{{\bf{Q}}^{t + 1}}|{L^2}}}{2} \! + \! \frac{{8|{{\bf{A}}^t}|{L^2}}}{{{\rho _1}({c_1^t}){^2}}} \! + \! \frac{{8N{L^2}}}{{{\rho _2}({c_2^t}){^2}}})||h^{t+1} \! - \! h^t|{|^2}\vspace{1ex}\\
 
\quad \quad \quad \quad \;\; \! -  (\frac{1}{{10{\rho _1}}} \! - \! \frac{{3\tau{ k_1 }N{L^2}}}{2})\!\sum\limits_{l = 1}^{|{{\bf{A}}^t}|}\! {||{\lambda _l^{t+1}} \! - \! {\lambda _l^t}|{|^2}}  \! - \! \frac{1}{{10{\rho _2}}}\!\sum\limits_{j = 1}^N\! {||{{\boldsymbol{\phi}}_j^{t+1}} \! - \! {{\boldsymbol{\phi}}_j^t}|{|^2}}  \! + \! \frac{{{c_1^{t-1}}  -  {c_1^t}}}{2}\!\sum\limits_{l = 1}^{|{{\bf{A}}^t}|}\! {||{\lambda _l^{t+1}}|{|^2}}  \vspace{0.5ex}\\

\quad \quad \quad \quad \;\; \! +  \frac{{{c_2^{t-1}}  -  {c_2^t}}}{2}\sum\limits_{j = 1}^N {||{{\boldsymbol{\phi}}_j^{t+1}}|{|^2}}
 \! + \! \frac{4}{{{\rho _1}}}(\frac{{{c_1^{t-2}}}}{{{c_1^{t-1}}}} \! - \! \frac{{{c_1^{t-1}}}}{{{c_1^t}}})\sum\limits_{l = 1}^{|{{\bf{A}}^t}|} {||{\lambda _l^t}|{|^2}}  \! + \! \frac{4}{{{\rho _2}}}(\frac{{{c_2^{t-2}}}}{{{c_2^{t-1}}}} \! - \! \frac{{{c_2^{t-1}}}}{{{c_2^t}}})\sum\limits_{j = 1}^N {||{{\boldsymbol{\phi}}_j^t}|{|^2}}.
\end{array}
\end{equation}}
\end{lemma}

\vspace{-2mm}

\emph{\textbf{Proof of Lemma \ref{lemma3}}:} 

Let ${a_1} = \frac{1}{{{\rho _1}}}$, ${a_3} = \frac{1}{{{\rho _2}}}$ and substitute them into Lemma \ref{lemma 2}, $\forall t \ge T_1 + \tau$, we have,
{\begin{equation}
\renewcommand{\theequation}{A.\arabic{equation}}
\label{eq:A44}
\begin{array}{l}
{L_p}{\rm{(\{ }}{{\boldsymbol{w}}_j^{t+1}}{\rm{\} ,}}{\boldsymbol{z}}^{t + 1},h^{t+1},\{ {\lambda _l^{t+1}}\} ,\{ {{\boldsymbol{\phi}}_j^{t+1}}\} {\rm{)}} - {L_p}{\rm{(\{ }}{{\boldsymbol{w}}_j^t}{\rm{\} ,}}\boldsymbol{z}^t,h^t,\{ {\lambda _l^t}\} ,\{ {{\boldsymbol{\phi}}_j^t}\} {\rm{)}}\\
\! \le\! (\frac{{L + 1}}{2} \!-\! \frac{1}{{{\eta _{\boldsymbol{w}}^t}}}\! +\! \frac{{{\rho _1}|{{\bf{A}}^t}|{L^2}}}{2} \!+\! \frac{{{\rho _2}|{{\bf{Q}}^{t + 1}}|{L^2}}}{2})\! \sum\limits_{j = 1}^N\! {||{{\boldsymbol{w}}_j^{t+1}}\! -\! {{\boldsymbol{w}}_j^t}|{|^2}} \\

\! + (\frac{{L + 3\tau{ k_1 }N{L^2}}}{2} \!-\! \frac{1}{{{\eta _{\boldsymbol{z}}^t}}}\! + \!\frac{{{\rho _1}|{{\bf{A}}^t}|{L^2}}}{2}\! +\! \frac{{{\rho _2}|{{\bf{Q}}^{t + 1}}|{L^2}}}{2})||{\boldsymbol{z}}^{t + 1}\! -\! \boldsymbol{z}^t|{|^2}\\

\! + (\frac{{L + 3\tau{ k_1 }N{L^2}}}{2}\! -\! \frac{1}{{{\eta _h^t}}}\! +\! \frac{{{\rho _1}|{{\bf{A}}^t}|{L^2}}}{2}\!+\! \frac{{{\rho _2}|{{\bf{Q}}^{t + 1}}|{L^2}}}{2})||h^{t+1} \!-\! h^t|{|^2}\\

\!+ (\frac{{3\tau{ k_1 }N{L^2}}}{2} \!+\! \frac{1}{{{\rho _1}}} \!-\! \frac{{{c_1^{t-1}} - {c_1^t}}}{2})\! \sum\limits_{l = 1}^{|{{\bf{A}}^t}|} \! {||{\lambda _l^{t+1}} \!-\! {\lambda _l^t}|{|^2}} \\

\! + \frac{{{c_1^{t-1}} }}{2}\! \sum\limits_{l = 1}^{|{{\bf{A}}^t}|} \! {(||{\lambda _l^{t+1}}|{|^2}\! -\! ||{\lambda _l^t}|{|^2})} \! +\! \frac{1}{{2{\rho _1}}} \!\sum\limits_{l = 1}^{|{{\bf{A}}^t}|} \! {||{\lambda _l^t} - {\lambda _l^{t-1}}|{|^2}} \! +\! (\frac{1}{{{\rho _2}}} \!-\! \frac{{{c_2^{t-1}} - {c_2^t}}}{2}) \!\sum\limits_{j = 1}^N \! {||{{\boldsymbol{\phi}}_j^{t+1}}\! -\! {{\boldsymbol{\phi}}_j^t}|{|^2}} \\

\! + \frac{{{c_2^{t-1}}}}{2} \! \sum\limits_{j = 1}^N \! {(||{{\boldsymbol{\phi}}_j^{t+1}}|{|^2} \!- \!||{{\boldsymbol{\phi}}_j^t}|{|^2})}\!  +\! \frac{1}{{2{\rho _2}}}\!\sum\limits_{j = 1}^N\! {||{{\boldsymbol{\phi}}_j^t} - {{\boldsymbol{\phi}}_j^{t-1}}|{|^2}}.
\end{array}
\end{equation}}

According to Eq. (14), in $(t+1)^{\rm{th}}$ iteration, it follows that:
{\begin{equation}
\renewcommand{\theequation}{A.\arabic{equation}}
\label{eq:A45}
\left\langle {{\lambda _l^{t+1}}\! -\! {\lambda _l^t}\! -\! {\rho _1}{\nabla _{{\lambda _l}}}{{\widetilde{L}}_p}{\rm{(\{ }}{{\boldsymbol{w}}_j^{t+1}}{\rm{\} ,}}{\boldsymbol{z}}^{t+1},h^{t+1},\{ {\lambda _l^t}\} ,\{ {{\boldsymbol{\phi}}_j^t}\} {\rm{)}},{\lambda _l^t}\! -\! {\lambda _l^{t+1}}} \right\rangle  \ge 0.
\end{equation}}

Similar to Eq. (A.\ref{eq:A45}), in $t^{\rm{th}}$ iteration, we have,
{\begin{equation}
\renewcommand{\theequation}{A.\arabic{equation}}
\label{eq:A46}
\left\langle {{\lambda _l^t}\! -\! {\lambda _l^{t-1}} \!-\! {\rho _1}{\nabla _{{\lambda _l}}}{{\widetilde{L}}_p}{\rm{(\{ }}{{\boldsymbol{w}}_j^t}{\rm{\} ,}}{\boldsymbol{z}^t},h^t,\{ {\lambda _l^{t-1}}\} ,\{ {{\boldsymbol{\phi}}_j^{t-1}}\} {\rm{)}},{\lambda _l^{t+1}} \!-\! {\lambda _l^t}} \right\rangle  \ge 0.
\end{equation}}

$\forall t \ge T_1$, we can obtain the following inequality, 
{\begin{equation}
\renewcommand{\theequation}{A.\arabic{equation}}
\label{eq:A47}
\begin{array}{l}
\! \sum\limits_{l = 1}^{|{{\bf{A}}^t}|}\! \frac{1}{{{\rho _1}}} \left\langle {{{\boldsymbol{v}}_{1,l}^{t + 1}},{\lambda _l^{t+1}} \! - \! {\lambda _l^t}} \right\rangle  \\
\! \le \! \sum\limits_{l = 1}^{|{{\bf{A}}^t}|} ( \left\langle \!  {{\nabla _{{\lambda _l}}}{{\widetilde{L}}_p}{\rm{(\{ }}{{\boldsymbol{w}}_j^{t+1}}{\rm{\},}}{\boldsymbol{z}}^{t+1},h^{t+1},\{ {\lambda _l^t}\},\{ {{\boldsymbol{\phi}}_j^t}\} {\rm{)}} \! - \! {\nabla _{{\lambda _l}}}{{\widetilde{L}}_p}{\rm{(\{ }}{{\boldsymbol{w}}_j^t}{\rm{\},}}{\boldsymbol{z}^t},h^t,\{ {\lambda _l^t}\} ,\{ {{\boldsymbol{\phi}}_j^t}\} {\rm{)}},{\lambda _l^{t+1}} \! - \! {\lambda _l^t}} \! \right\rangle  \vspace{0.5ex}\\

 \qquad \;  + \!  \left\langle \! {{\nabla _{{\lambda _l}}}{{\widetilde{L}}_p}{\rm{(\{ }}{{\boldsymbol{w}}_j^t}{\rm{\},}}{\boldsymbol{z}^t},h^t,\{ {\lambda _l^t}\} ,\{ {{\boldsymbol{\phi}}_j^t}\} {\rm{)}} \! - \! {\nabla _{{\lambda _l}}}{{\widetilde{L}}_p}{\rm{(\{ }}{{\boldsymbol{w}}_j^t}{\rm{\},}}{\boldsymbol{z}^t},h^t,\{ {\lambda _l^{t-1}}\},\{ {{\boldsymbol{\phi}}_j^{t-1}}\} {\rm{)}},{{\boldsymbol{v}}_{1,l}^{t + 1}}} \! \right\rangle \vspace{0.5ex} \\
 
 \qquad \; + \!  \left\langle \! {{\nabla _{{\lambda _l}}}{{\widetilde{L}}_p}{\rm{(\{ }}{{\boldsymbol{w}}_j^t}{\rm{\},}}{\boldsymbol{z}^t},h^t,\{ {\lambda _l^t}\},\{ {{\boldsymbol{\phi}}_j^t}\} {\rm{)}} \! - \! {\nabla _{{\lambda _l}}}{{\widetilde{L}}_p}{\rm{(\{ }}{{\boldsymbol{w}}_j^t}{\rm{\},}}{\boldsymbol{z}^t},h^t,\{ {\lambda _l^{t-1}}\},\{ {{\boldsymbol{\phi}}_j^{t-1}}\} {\rm{)}},{\lambda _l^t} \! - \! {\lambda _l^{t-1}}} \! \right\rangle ). 
\end{array}
\end{equation}}

Since we have the following equality, 
{\begin{equation}
\renewcommand{\theequation}{A.\arabic{equation}}
\label{eq:A48}
\begin{array}{l}
\frac{1}{{{\rho _1}}}\left\langle {{{\boldsymbol{v}}_{1,l}^{t + 1}},{\lambda _l^{t+1}} \!-\! {\lambda _l^t}} \right\rangle 
 = \frac{1}{{2{\rho _1}}}||{\lambda _l^{t+1}} \!-\! {\lambda _l^t}|{|^2} \! + \! \frac{1}{{2{\rho _1}}}||{{\boldsymbol{v}}_{1,l}^{t + 1}}|{|^2} - \frac{1}{{2{\rho _1}}}||{\lambda _l^t} - {\lambda _l^{t-1}}|{|^2},
\end{array}
\end{equation}}

it follows that,
{\begin{equation}
\renewcommand{\theequation}{A.\arabic{equation}}
\label{eq:A49}
\begin{array}{l}
\sum\limits_{l = 1}^{|{{\bf{A}}^t}|}(\frac{1}{{2{\rho _1}}}||{\lambda _l^{t+1}} \! - \! {\lambda _l^t}|{|^2} \! + \! \frac{1}{{2{\rho _1}}}||{{\boldsymbol{v}}_{1,l}^{t + 1}}|{|^2} \! - \! \frac{1}{{2{\rho _1}}}||{\lambda _l^t} \! - \! {\lambda _l^{t-1}}|{|^2}) \vspace{0.5ex}\\

\! \le\!\sum\limits_{l = 1}^{|{{\bf{A}}^t}|}( \frac{{{L^2}}}{{2{b_1^t}}}(\sum\limits_{j = 1}^N {||{{\boldsymbol{w}}_j^{t+1}} \! - \! {{\boldsymbol{w}}_j^t}|{|^2}}  \! + \! ||{\boldsymbol{z}}^{t+1} \! - \! {\boldsymbol{z}^t}|{|^2} \! + \! ||h^{t+1} \! - \! h^t|{|^2}) \! + \! \frac{{{b_1^t}}}{2}||{\lambda _l^{t+1}} \! - \! {\lambda _l^t}|{|^2} \vspace{0.5ex}\\

\qquad \; +  \frac{{{c_1^{t-1}}  -  {c_1^t}}}{2}(||{\lambda _l^{t+1}}|{|^2} \! - \! ||{\lambda _l^t}|{|^2}) \! - \! \frac{{{c_1^{t-1}}  -  {c_1^t}}}{2}||{\lambda _l^{t+1}} \! - \! {\lambda _l^t}|{|^2} \vspace{1ex}\\

\qquad \; +  \frac{{{\rho _1}}}{2}||{\nabla _{{\lambda _l}}}{\widetilde{L}_p}{\rm{(\{ }}{{\boldsymbol{w}}_j^t}{\rm{\},}}{\boldsymbol{z}^t},h^t,\!\{ {\lambda _l^t}\},\!\{ {{\boldsymbol{\phi}}_j^t}\} {\rm{)}} \! - \! {\nabla _{{\lambda _l}}}{\widetilde{L}_p}{\rm{(\{ }}{{\boldsymbol{w}}_j^t}{\rm{\},}}{\boldsymbol{z}^t},h^t,\!\{ {\lambda _l^{t-1}}\},\!\{ {{\boldsymbol{\phi}}_j^{t-1}}\} {\rm{)}}|{|^2} \! + \! \frac{1}{{2{\rho _1}}}\!||{{\boldsymbol{v}}_{1,l}^{t + 1}}|{|^2} \vspace{1ex}\\

\qquad \; - \frac{1}{{{L_1}'  +  {c_1^{t-1}}}}||{\nabla _{{\lambda _l}}}{\widetilde{L}_p}{\rm{(\{ }}{{\boldsymbol{w}}_j^t}{\rm{\},}}{\boldsymbol{z}^t},h^t, \! \{ {\lambda _l^t}\} , \! \{ {{\boldsymbol{\phi}}_j^t}\} {\rm{)}} \! - \! {\nabla _{{\lambda _l}}}{\widetilde{L}_p}{\rm{(\{ }}{{\boldsymbol{w}}_j^t}{\rm{\},}}{\boldsymbol{z}^t},h^t, \! \{ {\lambda _l^{t-1}}\}  , \!\{ {{\boldsymbol{\phi}}_j^{t-1}}\} {\rm{)}}|{|^2} \vspace{1ex}\\

\qquad \;  - \frac{{{c_1^{t-1}}{L_1}'}}{{{L_1}'  +  {c_1^{t-1}}}}||{\lambda _l^t} \! - \! {\lambda _l^{t-1}}|{|^2}),
\end{array}
\end{equation}}

\noindent where ${b_1^t} > 0$. According to the setting that ${c_1^0} \le {L_1}'$, we have $- \frac{{{c_1^{t-1}}{L_1}'}}{{{L_1}' + {c_1^{t-1}}}} \le  - \frac{{{c_1^{t-1}}{L_1}'}}{{2{L_1}'}} =  - \frac{{{c_1^{t-1}}}}{2} \le  - \frac{{{c_1^t}}}{2}$. Multiplying both sides of the inequality Eq. (A.\ref{eq:A49}) by $\frac{8}{{{\rho _1}{c_1^t}}}$, we have,
{\begin{equation}
\renewcommand{\theequation}{A.\arabic{equation}}
\label{eq:A51}
\begin{array}{l}
\sum\limits_{l = 1}^{|{{\bf{A}}^t}|}(\frac{4}{{{\rho _1}^2{c_1^t}}}||{\lambda _l^{t+1}} - {\lambda _l^t}|{|^2} - \frac{4}{{{\rho _1}}}(\frac{{{c_1^{t-1}} - {c_1^t}}}{{{c_1^t}}})||{\lambda _l^{t+1}}|{|^2}) \\

 \!\le \!\sum\limits_{l = 1}^{|{{\bf{A}}^t}|}\!( \frac{4}{{{\rho _1}^2{c_1^t}}}||{\lambda _l^t} - {\lambda _l^{t-1}}|{|^2} - \frac{4}{{{\rho _1}}}(\frac{{{c_1^{t-1}} - {c_1^t}}}{{{c_1^t}}})||{\lambda _l^t}|{|^2}  + \frac{{4{b_1^t}}}{{{\rho _1}{c_1^t}}}||{\lambda _l^{t+1}} - {\lambda _l^t}|{|^2} - \frac{4}{{{\rho _1}}}||{\lambda _l^t} - {\lambda _l^{t-1}}|{|^2} \\
 
 \qquad \; + \frac{{4{L^2}}}{{{\rho _1}{c_1^t}{b_1^t}}}(\sum\limits_{j = 1}^N {||{{\boldsymbol{w}}_j^{t+1}} - {{\boldsymbol{w}}_j^t}|{|^2}}  \! +\! ||{\boldsymbol{z}}^{t + 1} - {\boldsymbol{z}^t}|{|^2} \! +\! ||h^{t+1} - h^t|{|^2})).
\end{array}
\end{equation}}

\vspace{-1mm}

Setting ${b_1^t} = \frac{{{c_1^t}}}{2}$ in Eq. (A.\ref{eq:A51}) and using the definition of ${S_1^t}$, we have,
{\begin{equation}
\renewcommand{\theequation}{A.\arabic{equation}}
\label{eq:A52}
\begin{array}{l}
{S_1^{t+1}} \! - \! {S_1^t} \vspace{1ex}\\
\! \le\! \sum\limits_{l = 1}^{|{{\bf{A}}^t}|} {(\frac{4}{{{\rho _1}}}(\frac{{{c_1^{t-2}}}}{{{c_1^{t-1}}}} \! - \! \frac{{{c_1^{t-1}}}}{{{c_1^t}}})||{\lambda _l^t}|{|^2}}
 \! +\! \frac{{8{L^2}}}{{{\rho _1}({c_1^t}){^2}}}(\sum\limits_{j = 1}^N {||{{\boldsymbol{w}}_j^{t+1}} \! - \! {{\boldsymbol{w}}_j^t}|{|^2}}  \! +\! ||{\boldsymbol{z}}^{t + 1} \! - \! {\boldsymbol{z}^t}|{|^2} \! +\! ||h^{t+1} \! - \! h^t|{|^2}) \vspace{1ex}\\
 \! + (\frac{2}{{{\rho _1}}} \! +\! \frac{4}{{{\rho_1}^2}}(\frac{1}{{{c_1^{t+1}}}} \! - \! \frac{1}{{{c_1^t}}}))||{\lambda _l^{t+1}} \! - \! {\lambda _l^t}|{|^2} \! - \! \frac{4}{{{\rho _1}}}||{\lambda _l^t} \! - \! {\lambda _l^{t-1}}|{|^2}) \vspace{1ex}\\
\! = \!\sum\limits_{l = 1}^{|{{\bf{A}}^t}|} {\frac{4}{{{\rho _1}}}(\frac{{{c_1^{t-2}}}}{{{c_1^{t-1}}}} \! - \! \frac{{{c_1^{t-1}}}}{{{c_1^t}}})||{\lambda _l^t}|{|^2}}  \! +\! \sum\limits_{l = 1}^{|{{\bf{A}}^t}|} {(\frac{2}{{{\rho _1}}} \! +\! \frac{4}{{{\rho_1}^2}}(\frac{1}{{{c_1^{t+1}}}} \! - \! \frac{1}{{{c_1^t}}}))||{\lambda _l^{t+1}} \! - \! {\lambda _l^t}|{|^2}} \vspace{1ex}\\
\! - \sum\limits_{l = 1}^{|{{\bf{A}}^t}|} {\frac{4}{{{\rho _1}}}||{\lambda _l^t} \! - \! {\lambda _l^{t-1}}|{|^2} \vspace{1ex}}
\! +\! \frac{{8|{{\bf{A}}^t}|{L^2}}}{{{\rho _1}({c_1^t}){^2}}}(\sum\limits_{j = 1}^N {||{{\boldsymbol{w}}_j^{t+1}} \! - \! {{\boldsymbol{w}}_j^t}|{|^2}}  \! +\! ||{\boldsymbol{z}}^{t + 1} \! - \! {\boldsymbol{z}^t}|{|^2} \! +\! ||h^{t+1} \! - \! h^t|{|^2}).
\end{array}
\end{equation}}

Likewise, according to Eq. (15), we have that,
{\begin{equation}
\renewcommand{\theequation}{A.\arabic{equation}}
\label{eq:A53}
\begin{array}{l}
\frac{1}{{{\rho _2}}}\left\langle {{{\boldsymbol{v}}_{2,l}^{t + 1}},{{\boldsymbol{\phi}}_j^{t+1}} \! - \! {{\boldsymbol{\phi}}_j^t}} \right\rangle \vspace{1ex}\\
\! \le \! \left\langle\! {{\nabla _{{{\boldsymbol{\phi}}_j}}}{{\widetilde{L}}_p}{\rm{(\{ }}{{\boldsymbol{w}}_j^{t+1}}{\rm{\},}}{\boldsymbol{z}}^{t+1},h^{t+1}, \! \{ {\lambda _l^{t+1}}\} , \!\{ {{\boldsymbol{\phi}}_j^t}\} {\rm{)  -  }}{\nabla _{{{\boldsymbol{\phi}}_j}}}{{\widetilde{L}}_p}{\rm{(\{ }}{{\boldsymbol{w}}_j^t}{\rm{\},}}{\boldsymbol{z}^t},h^t, \!\{ {\lambda _l^t}\},\{ {{\boldsymbol{\phi}}_j^{t-1}}\} {\rm{)}},{{\boldsymbol{\phi}}_j^{t+1}} \! - \! {{\boldsymbol{\phi}}_j^t}} \! \right\rangle \vspace{1ex}\\
\! = \! \left\langle\! {{\nabla _{{{\boldsymbol{\phi}}_j}}}{{\widetilde{L}}_p}{\rm{(\{ }}{{\boldsymbol{w}}_j^{t+1}}{\rm{\},}}{\boldsymbol{z}}^{t+1},h^{t+1}, \! \{ {\lambda _l^{t+1}}\} , \!\{ {{\boldsymbol{\phi}}_j^t}\} {\rm{)}} \! - \! {\nabla _{{{\boldsymbol{\phi}}_j}}}{{\widetilde{L}}_p}{\rm{(\{ }}{{\boldsymbol{w}}_j^t}{\rm{\} ,}}{\boldsymbol{z}^t},h^t, \! \{ {\lambda _l^t}\} , \!\{ {{\boldsymbol{\phi}}_j^t}\} {\rm{)}},{{\boldsymbol{\phi}}_j^{t+1}} \! - \! {{\boldsymbol{\phi}}_j^t}} \!\right\rangle \vspace{1ex}\\
 \! + \! \left\langle\! {{\nabla _{{{\boldsymbol{\phi}}_j}}}{{\widetilde{L}}_p}{\rm{(\{ }}{{\boldsymbol{w}}_j^t}{\rm{\},}}{\boldsymbol{z}^t},h^t,\{ {\lambda _l^t}\} ,\{ {{\boldsymbol{\phi}}_j^t}\} {\rm{)}} \! - \! {\nabla _{{{\boldsymbol{\phi}}_j}}}{{\widetilde{L}}_p}{\rm{(\{ }}{{\boldsymbol{w}}_j^t}{\rm{\} ,}}{\boldsymbol{z}^t},h^t,\{ {\lambda _l^t}\},\{ {{\boldsymbol{\phi}}_j^{t-1}}\} {\rm{)}},{{\boldsymbol{v}}_{2,l}^{t + 1}}} \!\right\rangle \vspace{1ex}\\
 \! + \! \left\langle \! {{\nabla _{{{\boldsymbol{\phi}}_j}}}{{\widetilde{L}}_p}{\rm{(\{ }}{{\boldsymbol{w}}_j^t}{\rm{\},}}{\boldsymbol{z}^t},h^t, \!\{ {\lambda _l^t}\},\{ {{\boldsymbol{\phi}}_j^t}\} {\rm{)}} \! - \! {\nabla _{{{\boldsymbol{\phi}}_j}}}{{\widetilde{L}}_p}{\rm{(\{ }}{{\boldsymbol{w}}_j^t}{\rm{\},}}{\boldsymbol{z}^t},h^t,\{ {\lambda _l^t}\},\{ {{\boldsymbol{\phi}}_j^{t-1}}\} {\rm{)}},{{\boldsymbol{\phi}}_j^t} \! - \! {{\boldsymbol{\phi}}_j^{t-1}}} \! \right\rangle .
\end{array}
\end{equation}}

In addition, since
{\begin{equation}
\renewcommand{\theequation}{A.\arabic{equation}}
\label{eq:A54}
\begin{array}{l}
\frac{1}{{{\rho _2}}}\left\langle {{{\boldsymbol{v}}_{2,l}^{t+1}},{{\boldsymbol{\phi}}_j^{t+1}} - {{\boldsymbol{\phi}}_j^t}} \right\rangle 
 = \frac{1}{{2{\rho _2}}}||{{\boldsymbol{\phi}}_j^{t+1}} - {{\boldsymbol{\phi}}_j^t}|{|^2} \! +\! \frac{1}{{2{\rho _2}}}||{{\boldsymbol{v}}_{2,l}^{t+1}}|{|^2} - \frac{1}{{2{\rho _2}}}||{{\boldsymbol{\phi}}_j^t} - {{\boldsymbol{\phi}}_j^{t-1}}|{|^2},
\end{array}
\end{equation}}
it follows that,
{\begin{equation}
\renewcommand{\theequation}{A.\arabic{equation}}
\label{eq:A55}
\begin{array}{l}
\frac{1}{{2{\rho _2}}}||{{\boldsymbol{\phi}}_j^{t+1}} \! - \! {{\boldsymbol{\phi}}_j^t}|{|^2} \! +\! \frac{1}{{2{\rho _2}}}||{{\boldsymbol{v}}_{2,l}^{t+1}}|{|^2} \! - \! \frac{1}{{2{\rho _2}}}||{{\boldsymbol{\phi}}_j^t} \! - \! {{\boldsymbol{\phi}}_j^{t-1}}|{|^2} \vspace{0.5ex}\\

 \le \frac{{{L^2}}}{{2{b_2^t}}}(\sum\limits_{j = 1}^N {||{{\boldsymbol{w}}_j^{t+1}} \! - \! {{\boldsymbol{w}}_j^t}|{|^2}}  \! +\! ||{\boldsymbol{z}}^{t + 1} \! - \! {\boldsymbol{z}^t}|{|^2} \! +\! ||h^{t+1} \! - \! h^t|{|^2}) \! +\! \frac{{{b_2^t}}}{2}||{{\boldsymbol{\phi}}_j^{t+1}} \! - \! {{\boldsymbol{\phi}}_j^t}|{|^2} \vspace{0.5ex}\\
 
 \! + \frac{{{c_2^{t-1}}  -  {c_2^t}}}{2}(||{{\boldsymbol{\phi}}_j^{t+1}}|{|^2} \! - \! ||{{\boldsymbol{\phi}}_j^t}|{|^2}) \! - \! \frac{{{c_2^{t-1}}  -  {c_2^t}}}{2}||{{\boldsymbol{\phi}}_j^{t+1}} \! - \! {{\boldsymbol{\phi}}_j^t}|{|^2}  -  \frac{{{c_2^{t-1}}L_2'}}{{L_2'  + {c_2^{t-1}}}}||{{\boldsymbol{\phi}}_j^t} \! - \! {{\boldsymbol{\phi}}_j^{t-1}}|{|^2} \vspace{1ex}\\
 
 \! + \frac{{{\rho _2}}}{2}||{\nabla _{{{\boldsymbol{\phi}}_j}}}{\widetilde{L}_p}{\rm{(\{ }}{{\boldsymbol{w}}_j^t}{\rm{\},}}{\boldsymbol{z}^t},h^t,\{ {\lambda _l^t}\},\{ {{\boldsymbol{\phi}}_j^t}\} {\rm{)}} \! - \! {\nabla _{{{\boldsymbol{\phi}}_j}}}{\widetilde{L}_p}{\rm{(\{ }}{{\boldsymbol{w}}_j^t}{\rm{\},}}{\boldsymbol{z}^t},h^t,\{ {\lambda _l^t}\},\{ {{\boldsymbol{\phi}}_j^{t-1}}\} {\rm{)}}|{|^2} \! +\! \frac{1}{{2{\rho _2}}}||{{\boldsymbol{v}}_{2,l}^{t+1}}|{|^2} \vspace{1ex}\\
 
 \! -  \frac{1}{{L_2'  + {c_2^{t-1}}}}||{\nabla _{{{\boldsymbol{\phi}}_j}}}{\widetilde{L}_p}{\rm{(\{ }}{{\boldsymbol{w}}_j^t}{\rm{\},}}{\boldsymbol{z}^t},h^t,\{ {\lambda _l^t}\},\{ {{\boldsymbol{\phi}}_j^t}\} {\rm{)}} \! - \! {\nabla _{{{\boldsymbol{\phi}}_j}}}{\widetilde{L}_p}{\rm{(\{ }}{{\boldsymbol{w}}_j^t}{\rm{\},}}{\boldsymbol{z}^t},h^t,\{ {\lambda _l^t}\},\{ {{\boldsymbol{\phi}}_j^{t-1}}\} )|{|^2}.
\end{array}
\end{equation}}

According to the setting ${c_2^0} \le {L_2}'$, we have $- \frac{{{c_2^{t-1}}{L_2}'}}{{{L_2}' + {c_2^{t-1}}}} \le  - \frac{{{c_2^{t-1}}{L_2}'}}{{2{L_2}'}} =  - \frac{{{c_2^{t-1}}}}{2} \le  - \frac{{{c_2^t}}}{2}$.  Multiplying both sides of the inequality Eq. (A.\ref{eq:A55})  by $\frac{8}{{{\rho _2}{c_2^t}}}$, we have,
{\begin{equation}
\renewcommand{\theequation}{A.\arabic{equation}}
\label{eq:A57}
\begin{array}{l}
\frac{4}{{{\rho _2}^2{c_2^t}}}||{{\boldsymbol{\phi}}_j^{t+1}} \! - \! {{\boldsymbol{\phi}}_j^t}|{|^2} \! - \! \frac{4}{{{\rho _2}}}(\frac{{{c_2^{t-1}}  -  {c_2^t}}}{{{c_2^t}}})||{{\boldsymbol{\phi}}_j^{t+1}}|{|^2} \vspace{1ex}\\
\! \le \! \frac{4}{{{\rho _2}^2{c_2^t}}}||{{\boldsymbol{\phi}}_j^t} \! - \! {{\boldsymbol{\phi}}_j^{t-1}}|{|^2} \! - \! \frac{4}{{{\rho _2}}}(\frac{{{c_2^{t-1}}  -  {c_2^t}}}{{{c_2^t}}})||{{\boldsymbol{\phi}}_j^t}|{|^2} \! +\! \frac{{4{b_2^t}}}{{{\rho _2}{c_2^t}}}||{{\boldsymbol{\phi}}_j^{t+1}} \! - \! {{\boldsymbol{\phi}}_j^t}|{|^2} \! - \! \frac{4}{{{\rho _2}}}||{{\boldsymbol{\phi}}_j^t} \! - \! {{\boldsymbol{\phi}}_j^{t-1}}|{|^2} \vspace{1ex}\\
 \! + \frac{{4{L^2}}}{{{\rho _2}{c_2^t}{b_2^t}}}(\sum\limits_{j = 1}^N {||{{\boldsymbol{w}}_j^{t+1}} \! - \! {{\boldsymbol{w}}_j^t}|{|^2}}  \! +\! ||{\boldsymbol{z}}^{t + 1} \! - \! {\boldsymbol{z}^t}|{|^2} \! +\! ||h^{t+1} \! - \! h^t|{|^2}).
\end{array}
\end{equation}}

Setting ${b_2^t} = \frac{{{c_2^t}}}{2}$ in Eq. (A.\ref{eq:A57}) and using the definition of ${S_2^t}$, we can obtain,
{\begin{equation}
\renewcommand{\theequation}{A.\arabic{equation}}
\label{eq:A58}
\begin{array}{l}
{S_2^{t+1}} \! - \! {S_2^t} \vspace{1ex}\\
\! \le \! \sum\limits_{j = 1}^N \! {(\frac{4}{{{\rho _2}}}(\frac{{{c_2^{t-2}}}}{{{c_2^{t-1}}}} \! - \! \frac{{{c_2^{t-1}}}}{{{c_2^t}}})||{{\boldsymbol{\phi}}_j^t}|{|^2}}  \vspace{1ex}
 \! +\! \frac{{8{L^2}}}{{{\rho _2}({c_2^t})^2}}(\!\sum\limits_{j = 1}^N\! {||{{\boldsymbol{w}}_j^{t+1}} \! - \! {{\boldsymbol{w}}_j^t}|{|^2}}  \! +\! ||{\boldsymbol{z}}^{t + 1} \! - \! {\boldsymbol{z}^t}|{|^2} \! +\! \sum\limits_{l = 1}^{|{{\bf{A}}^t}|}\! {||{\lambda _l^{t+1}} \! - \! {\lambda _l^t}|{|^2}} ) \vspace{1ex}\\
 \! + (\frac{2}{{{\rho _2}}} \! +\! \frac{2}{{{\rho _2}^2}}(\frac{1}{{{c_2^{t+1}}}} \! - \! \frac{1}{{{c_2^t}}}))||{{\boldsymbol{\phi}}_j^{t+1}} \! - \! {{\boldsymbol{\phi}}_j^t}|{|^2} \! - \! \frac{4}{{{\rho _2}}}||{{\boldsymbol{\phi}}_j^t} \! - \! {{\boldsymbol{\phi}}_j^{t-1}}|{|^2}) \vspace{1ex}\\
\! = \! \sum\limits_{j = 1}^N \! {\frac{4}{{{\rho _2}}}(\frac{{{c_2^{t-2}}}}{{{c_2^{t-1}}}} \! - \! \frac{{{c_2^{t-1}}}}{{{c_2^t}}})||{{\boldsymbol{\phi}}_j^t}|{|^2}}
 \! +\! \sum\limits_{j = 1}^N \!{(\frac{2}{{{\rho _2}}} \! +\! \frac{4}{{{\rho _2}^2}}(\frac{1}{{{c_2^{t+1}}}} \! - \! \frac{1}{{{c_2^t}}}))||{{\boldsymbol{\phi}}_j^{t+1}} \! - \! {{\boldsymbol{\phi}}_j^t}|{|^2}} \vspace{1ex}\\
 \! - \! \sum\limits_{j = 1}^N \! {\frac{4}{{{\rho _2}}}||{{\boldsymbol{\phi}}_j^t} \! - \! {{\boldsymbol{\phi}}_j^{t-1}}|{|^2}}  \! +\! \frac{{8N{L^2}}}{{{\rho _2}({c_2^t})^2}}(\!\sum\limits_{j = 1}^N\! {||{{\boldsymbol{w}}_j^{t+1}} \! - \! {{\boldsymbol{w}}_j^t}|{|^2}}  \! +\! ||{\boldsymbol{z}}^{t + 1} \! - \! {\boldsymbol{z}^t}|{|^2} \! +\! ||h^{t+1} \! - \! h^t|{|^2}).
\end{array}
\end{equation}}

According to the setting about ${c_1^t}$ and ${c_2^t}$, we have $\frac{{{\rho _1}}}{{10}} \ge \frac{1}{{{c_1^{t+1}}}} - \frac{1}{{{c_1^t}}},
\frac{{{\rho _2}}}{{10}} \ge \frac{1}{{{c_2^{t+1}}}} - \frac{1}{{{c_2^t}}},{\rm{    }}\forall t \ge T_1$.
Using the definition of $F^{t+1}$ and combining it with Eq. (A.\ref{eq:A52}) and Eq. (A.\ref{eq:A58}), $\forall t \ge T_1 + \tau$, we have,
{\begin{equation}
\renewcommand{\theequation}{A.\arabic{equation}}
\label{eq:A59}
\begin{array}{l}
F^{t+1} \! - \! F^{t}\\
 \!\le\! (\frac{{L  + 1}}{2} \! - \! \frac{1}{{{\eta _{\boldsymbol{w}}^t}}} \! +\! \frac{{{\rho _1}|{{\bf{A}}^t}|{L^2}}}{2} \! +\! \frac{{{\rho _2}|{{\bf{Q}}^{t + 1}}|{L^2}}}{2} \! +\! \frac{{8|{{\bf{A}}^t}|{L^2}}}{{{\rho _1}({c_1^t})^2}} \! +\! \frac{{8N{L^2}}}{{{\rho _2}({c_2^t})^2}})\sum\limits_{j = 1}^N {||{{\boldsymbol{w}}_j^{t+1}} \! - \! {{\boldsymbol{w}}_j^t}|{|^2}} \\
 
 \! + (\frac{{L  + 3\tau{ k_1 }N{L^2}}}{2} \! - \! \frac{1}{{{\eta _{\boldsymbol{z}}^t}}} \! +\! \frac{{{\rho _1}|{{\bf{A}}^t}|{L^2}}}{2} \! +\! \frac{{{\rho _2}|{{\bf{Q}}^{t + 1}}|{L^2}}}{2} \! +\! \frac{{8|{{\bf{A}}^t}|{L^2}}}{{{\rho _1}({c_1^t})^2}} \! +\! \frac{{8N{L^2}}}{{{\rho _2}({c_2^t})^2}})||\boldsymbol{z}^{t+1} \! - \! \boldsymbol{z}^t|{|^2}\\
 
 \! + (\frac{{L  + 3\tau{ k_1 }N{L^2}}}{2} \! - \! \frac{1}{{{\eta _h^t}}} \! +\! \frac{{{\rho _1}|{{\bf{A}}^t}|{L^2}}}{2} \! +\! \frac{{{\rho _2}|{{\bf{Q}}^{t + 1}}|{L^2}}}{2} \! +\! \frac{{8|{{\bf{A}}^t}|{L^2}}}{{{\rho _1}({c_1^t})^2}} \! +\! \frac{{8N{L^2}}}{{{\rho _2}({c_2^t})^2}})||h^{t+1} \! - \! h^t|{|^2}\\
 
 \! -  (\frac{1}{{10{\rho _1}}} \! - \! \frac{{3\tau{ k_1 }N{L^2}}}{2})\sum\limits_{l = 1}^{|{{\bf{A}}^t}|} {||{\lambda _l^{t+1}} \! - \! {\lambda _l^t}|{|^2}}  \! - \! \frac{1}{{10{\rho _2}}}\sum\limits_{j = 1}^N {||{{\boldsymbol{\phi}}_j^{t+1}} \! - \! {{\boldsymbol{\phi}}_j^t}|{|^2}} 
 \! +\! \frac{{{c_1^{t-1}}  -  {c_1^t}}}{2}\sum\limits_{l = 1}^{|{{\bf{A}}^t}|} {||{\lambda _l^{t+1}}|{|^2}} \\
 
 \! + \frac{{{c_2^{t-1}}  -  {c_2^t}}}{2}\sum\limits_{j = 1}^N {||{{\boldsymbol{\phi}}_j^{t+1}}|{|^2}} 
 \! +\! \frac{4}{{{\rho _1}}}(\frac{{{c_1^{t-2}}}}{{{c_1^{t-1}}}} \! - \! \frac{{{c_1^{t-1}}}}{{{c_1^t}}})\sum\limits_{l = 1}^{|{{\bf{A}}^t}|} {||{\lambda _l^t}|{|^2}}  \! +\! \frac{4}{{{\rho _2}}}(\frac{{{c_2^{t-2}}}}{{{c_2^{t-1}}}} \! - \! \frac{{{c_2^{t-1}}}}{{{c_2^t}}})\sum\limits_{j = 1}^N {||{{\boldsymbol{\phi}}_j^t}|{|^2}}.
\end{array}
\end{equation}}


Next, we will combine Lemma \ref{lemma 1}, Lemma \ref{lemma 2} with Lemma \ref{lemma3} to derive Theorem 1. Firstly, we make some definitions about our problem.

\renewcommand{\thedefinition}{A.\arabic{definition}}
\begin{definition}
\label{definition:A1}
The \textit{stationarity} \textit{gap} at $t^{{th}}$ iteration is defined as:
{\begin{equation}
\label{eq:A61}
\renewcommand{\theequation}{A.\arabic{equation}}
\nabla G^t = \left[ \begin{array}{l}
\{ \frac{1}{{{\alpha _{\boldsymbol{w}}^t}}}({{\boldsymbol{w}}_j^t} - {\mathcal{P}_{{\boldsymbol{\mathcal{W}}}}}({{\boldsymbol{w}}_j^t} - {\alpha _{\boldsymbol{w}}^t}{\nabla _{{{\boldsymbol{w}}_j}}}{L_p}{\rm{(\{ }}{{\boldsymbol{w}}_j^t}{\rm{\} ,}}{\boldsymbol{z}^t},h^t,\{ {\lambda _l^t}\} ,\{ {{\boldsymbol{\phi}}_j^t}\} {\rm{)))\}}} \vspace{1ex}\\
\frac{1}{{{\eta _{\boldsymbol{z}}^t}}}({\boldsymbol{z}^t} - {\mathcal{P}_{{\boldsymbol{\mathcal{Z}}}}}({\boldsymbol{z}^t} - {\eta _{\boldsymbol{z}}^t}{\nabla _{\boldsymbol{z}}}{L_p}{\rm{(\{ }}{{\boldsymbol{w}}_j^t}{\rm{\} ,}}{\boldsymbol{z}^t},h^t,\{ {\lambda _l^t}\} ,\{ {{\boldsymbol{\phi}}_j^t}\} {\rm{))}}) \vspace{1ex}\\
\frac{1}{{{\eta _h^t}}}(h^t - {\mathcal{P}_{{\boldsymbol{\mathcal{H}}}}}(h^t - {\eta _h^t}{\nabla _h}{L_p}{\rm{(\{ }}{{\boldsymbol{w}}_j^t}{\rm{\} ,}}{\boldsymbol{z}^t},h^t,\{ {\lambda _l^t}\} ,\{ {{\boldsymbol{\phi}}_j^t}\} {\rm{)}})) \vspace{1ex}\\
 \{ \frac{1}{{{\rho _1}}}({\lambda _l^t} - {\mathcal{P}_{\bf{\Lambda}} }({\lambda _l^t} \! +\! {\rho _1}{\nabla _{{\lambda _l}}}{L_p}{\rm{(\{ }}{{\boldsymbol{w}}_j^t}{\rm{\} ,}}{\boldsymbol{z}^t},h^t,\{ {\lambda _l^t}\} ,\{ {{\boldsymbol{\phi}}_j^t}\} {\rm{)))\}}} \vspace{1ex}\\
\{ \frac{1}{{{\rho _2}}}({{\boldsymbol{\phi}}_j^t} - {\mathcal{P}_{{\boldsymbol{\Phi}}}}({{\boldsymbol{\phi}}_j^t} \! +\! {\rho _2}{\nabla _{{{\boldsymbol{\phi}}_j}}}{L_p}{\rm{(\{ }}{{\boldsymbol{w}}_j^t}{\rm{\} ,}}{\boldsymbol{z}^t},h^t,\{ {\lambda _l^t}\} ,\{ {{\boldsymbol{\phi}}_j^t}\} {\rm{)))\} }}
\end{array} \right].
\end{equation}}

And we also define:
{\begin{equation}
\label{eq:A62}
\renewcommand{\theequation}{A.\arabic{equation}}
\begin{array}{l}
{(\nabla G^t)_{{{\boldsymbol{w}}_j}}} = \frac{1}{{{\alpha _{\boldsymbol{w}}^t}}}({{\boldsymbol{w}}_j^t} - {\mathcal{P}_{{\boldsymbol{\mathcal{W}}}}}({{\boldsymbol{w}}_j^t} - {\alpha _{\boldsymbol{w}}^t}{\nabla _{{{\boldsymbol{w}}_j}}}{L_p}{\rm{(\{ }}{{\boldsymbol{w}}_j^t}{\rm{\} ,}}{\boldsymbol{z}^t},h^t,\{ {\lambda _l^t}\} ,\{ {{\boldsymbol{\phi}}_j^t}\} {\rm{)))}}, \vspace{1ex}\\
{(\nabla G^t)_{\boldsymbol{z}}} = \frac{1}{{{\eta _{\boldsymbol{z}}^t}}}({\boldsymbol{z}^t} - {\mathcal{P}_{{\boldsymbol{\mathcal{Z}}}}}({\boldsymbol{z}^t} - {\eta _{\boldsymbol{z}}^t}{\nabla _{\boldsymbol{z}}}{L_p}{\rm{(\{ }}{{\boldsymbol{w}}_j^t}{\rm{\} ,}}{\boldsymbol{z}^t},h^t,\{ {\lambda _l^t}\} ,\{ {{\boldsymbol{\phi}}_j^t}\} {\rm{)}})), \vspace{1ex}\\
{(\nabla G^t)_h} = \frac{1}{{{\eta _h^t}}}(h^t - {\mathcal{P}_{{\boldsymbol{\mathcal{H}}}}}(h^t - {\eta _h^t}{\nabla _h}{L_p}{\rm{(\{ }}{{\boldsymbol{w}}_j^t}{\rm{\} ,}}{\boldsymbol{z}^t},h^t,\{ {\lambda _l^t}\} ,\{ {{\boldsymbol{\phi}}_j^t}\} {\rm{)}})), \vspace{1ex}\\
{(\nabla G^t)_{{\lambda _l}}} = \frac{1}{{{\rho _1}}}({\lambda _l^t} - {\mathcal{P}_{\bf{\Lambda}} }({\lambda _l^t} \! +\! {\rho _1}{\nabla _{{\lambda _l}}}{L_p}{\rm{(\{ }}{{\boldsymbol{w}}_j^t}{\rm{\} ,}}{\boldsymbol{z}^t},h^t,\{ {\lambda _l^t}\} ,\{ {{\boldsymbol{\phi}}_j^t}\} {\rm{)))}}, \vspace{1ex}\\
{(\nabla G^t)_{{{\boldsymbol{\phi}}_j}}} = \frac{1}{{{\rho _2}}}({{\boldsymbol{\phi}}_j^t} - {\mathcal{P}_{{\boldsymbol{\Phi}}}}({{\boldsymbol{\phi}}_j^t} \! +\! {\rho _2}{\nabla _{{{\boldsymbol{\phi}}_j}}}{L_p}{\rm{(\{ }}{{\boldsymbol{w}}_j^t}{\rm{\} ,}}{\boldsymbol{z}^t},h^t,\{ {\lambda _l^t}\} ,\{ {{\boldsymbol{\phi}}_j^t}\} {\rm{)))}}.
\end{array}
\end{equation}}

It follows that,
{\begin{equation}
\label{eq:A63}
\renewcommand{\theequation}{A.\arabic{equation}}
||\nabla G^t|{|^2} = \sum\limits_{j = 1}^N {||{{(\nabla G^t)}_{{{\boldsymbol{w}}_j}}}|{|^2}}  \! +\! ||{(\nabla G^t)_{\boldsymbol{z}}}|{|^2} \! +\! ||{(\nabla G^t)_h}|{|^2} \! +\! \sum\limits_{l = 1}^{|{{\bf{A}}^t}|} {||{{(\nabla G^t)}_{{\lambda _l}}}|{|^2}}  \! +\! \sum\limits_{j = 1}^N {||{{(\nabla G^t)}_{{{\boldsymbol{\phi}}_j}}}|{|^2}}. 
\end{equation}}
\end{definition}

\begin{definition}
\label{definition:A2}
At $t^{{th}}$ iteration, the \textit{stationarity} \textit{gap} \textit{w.r.t} ${\widetilde{L}_p}{\rm{(\{ }}{{\boldsymbol{w}}_j}{\rm{\} ,}}{\boldsymbol{z}},h,\{ {\lambda _l}\} ,\{ {{\boldsymbol{\phi}}_j}\} {\rm{)}}$ is defined as:

{\begin{equation}
\label{eq:A64}
\renewcommand{\theequation}{A.\arabic{equation}}
\nabla {\widetilde{G}}^t = \left[ \begin{array}{l}\{ \frac{1}{{{\alpha _{\boldsymbol{w}}^t}}}({{\boldsymbol{w}}_j^t} - {\mathcal{P}_{{\boldsymbol{\mathcal{W}}}}}({{\boldsymbol{w}}_j^t} - {\alpha _{\boldsymbol{w}}^t}{\nabla _{{{\boldsymbol{w}}_j}}}{\widetilde{L}_p}{\rm{(\{ }}{{\boldsymbol{w}}_j^t}{\rm{\} ,}}{\boldsymbol{z}^t},h^t,\{ {\lambda _l^t}\} ,\{ {{\boldsymbol{\phi}}_j^t}\} {\rm{)))\}}} \vspace{1ex}\\
\frac{1}{{{\eta _{\boldsymbol{z}}^t}}}({\boldsymbol{z}^t} - {\mathcal{P}_{{\boldsymbol{\mathcal{Z}}}}}({\boldsymbol{z}^t} - {\eta _{\boldsymbol{z}}^t}{\nabla _{\boldsymbol{z}}}{\widetilde{L}_p}{\rm{(\{ }}{{\boldsymbol{w}}_j^t}{\rm{\} ,}}{\boldsymbol{z}^t},h^t,\{ {\lambda _l^t}\} ,\{ {{\boldsymbol{\phi}}_j^t}\} {\rm{))}}) \vspace{1ex}\\
\frac{1}{{{\eta _h^t}}}(h^t - {\mathcal{P}_{{\boldsymbol{\mathcal{H}}}}}(h^t - {\eta _h^t}{\nabla _h}{\widetilde{L}_p}{\rm{(\{ }}{{\boldsymbol{w}}_j^t}{\rm{\} ,}}{\boldsymbol{z}^t},h^t,\{ {\lambda _l^t}\} ,\{ {{\boldsymbol{\phi}}_j^t}\} {\rm{)}})) \vspace{1ex}\\
\{ \frac{1}{{{\rho _1}}}({\lambda _l^t} - {\mathcal{P}_{\bf{\Lambda}} }({\lambda _l^t} \! +\! {\rho _1}{\nabla _{{\lambda _l}}}{\widetilde{L}_p}{\rm{(\{ }}{{\boldsymbol{w}}_j^t}{\rm{\} ,}}{\boldsymbol{z}^t},h^t,\{ {\lambda _l^t}\} ,\{ {{\boldsymbol{\phi}}_j^t}\} {\rm{)))\} }} \vspace{1ex}\\
\{ \frac{1}{{{\rho _2}}}({{\boldsymbol{\phi}}_j^t} - {\mathcal{P}_{{\boldsymbol{\Phi}}}}({{\boldsymbol{\phi}}_j^t} \! +\! {\rho _2}{\nabla _{{{\boldsymbol{\phi}}_j}}}{\widetilde{L}_p}{\rm{(\{ }}{{\boldsymbol{w}}_j^t}{\rm{\} ,}}{\boldsymbol{z}^t},h^t,\{ {\lambda _l^t}\} ,\{ {{\boldsymbol{\phi}}_j^t}\} {\rm{)))\} }}
\end{array} \right].
\end{equation}}

We further define:
{\begin{equation}
\label{eq:A65}
\renewcommand{\theequation}{A.\arabic{equation}}
\begin{array}{l}
{(\nabla{\widetilde{G}}^t)_{{{\boldsymbol{w}}_j}}} = \frac{1}{{{\alpha _{\boldsymbol{w}}^t}}}({{\boldsymbol{w}}_j^t} - {\mathcal{P}_{{\boldsymbol{\mathcal{W}}}}}({{\boldsymbol{w}}_j^t} - {\alpha _{\boldsymbol{w}}^t}{\nabla _{{{\boldsymbol{w}}_j}}}{\widetilde{L}_p}{\rm{(\{ }}{{\boldsymbol{w}}_j^t}{\rm{\} ,}}{\boldsymbol{z}^t},h^t,\{ {\lambda _l^t}\} ,\{ {{\boldsymbol{\phi}}_j^t}\} {\rm{)))}}, \vspace{1ex}\\
{(\nabla{\widetilde{G}}^t)_{\boldsymbol{z}}} = \frac{1}{{{\eta _{\boldsymbol{z}}^t}}}({\boldsymbol{z}^t} - {\mathcal{P}_{{\boldsymbol{\mathcal{Z}}}}}({\boldsymbol{z}^t} - {\eta _{\boldsymbol{z}}^t}{\nabla _{\boldsymbol{z}}}{\widetilde{L}_p}{\rm{(\{ }}{{\boldsymbol{w}}_j^t}{\rm{\} ,}}{\boldsymbol{z}^t},h^t,\{ {\lambda _l^t}\} ,\{ {{\boldsymbol{\phi}}_j^t}\} {\rm{)}})), \vspace{1ex}\\
{(\nabla{\widetilde{G}}^t)_h} = \frac{1}{{{\eta _h^t}}}(h^t - {\mathcal{P}_{{\boldsymbol{\mathcal{H}}}}}(h^t - {\eta _h^t}{\nabla _h}{\widetilde{L}_p}{\rm{(\{ }}{{\boldsymbol{w}}_j^t}{\rm{\} ,}}{\boldsymbol{z}^t},h^t,\{ {\lambda _l^t}\} ,\{ {{\boldsymbol{\phi}}_j^t}\} {\rm{)}})), \vspace{1ex}\\
{(\nabla{\widetilde{G}}^t)_{{\lambda _l}}} = \frac{1}{{{\rho _1}}}({\lambda _l^t} - {\mathcal{P}_{\bf{\Lambda}} }({\lambda _l^t} \! +\! {\rho _1}{\nabla _{{\lambda _l}}}{\widetilde{L}_p}{\rm{(\{ }}{{\boldsymbol{w}}_j^t}{\rm{\} ,}}{\boldsymbol{z}^t},h^t,\{ {\lambda _l^t}\} ,\{ {{\boldsymbol{\phi}}_j^t}\} {\rm{)))}}, \vspace{1ex}\\
{(\nabla{\widetilde{G}}^t)_{{{\boldsymbol{\phi}}_j}}} = \frac{1}{{{\rho _2}}}({{\boldsymbol{\phi}}_j^t} - {\mathcal{P}_{{\boldsymbol{\Phi}}}}({{\boldsymbol{\phi}}_j^t} \! +\! {\rho _2}{\nabla _{{{\boldsymbol{\phi}}_j}}}{\widetilde{L}_p}{\rm{(\{ }}{{\boldsymbol{w}}_j^t}{\rm{\} ,}}{\boldsymbol{z}^t},h^t,\{ {\lambda _l^t}\} ,\{ {{\boldsymbol{\phi}}_j^t}\} {\rm{)))}}.
\end{array}
\end{equation}}

It follows that,
{\begin{equation}
\label{eq:A66}
\renewcommand{\theequation}{A.\arabic{equation}}
||\nabla{\widetilde{G}}^t|{|^2} = \sum\limits_{j = 1}^N {||{{(\nabla{\widetilde{G}}^t)}_{{{\boldsymbol{w}}_j}}}|{|^2}}  \! +\! ||{(\nabla{\widetilde{G}}^t)_{\boldsymbol{z}}}|{|^2} \! +\! ||{(\nabla{\widetilde{G}}^t)_h}|{|^2} \! +\!\sum\limits_{l = 1}^{|{{\bf{A}}^t}|} {||{{(\nabla{\widetilde{G}}^t)}_{{\lambda _l}}}|{|^2}}  \! +\! \sum\limits_{j = 1}^N {||{{(\nabla{\widetilde{G}}^t)}_{{{\boldsymbol{\phi}}_j}}}|{|^2}}. 
\end{equation}}
\end{definition}

\begin{definition}
\label{definition:A3}
In our asynchronous algorithm, for the worker $j$ in $t^{th}$ iteration, we define the last iteration where worker $j$ was active as $\widetilde{{t}}_j $. And we define the next iteration that  worker $j$ will be active as $\overline{{t}_j}$. For the iteration index set that worker $j$ is active from $T_1^{th}$ to $(T_1 + T+\tau)^{th}$ iteration, we define it as $\mathcal{V}_j(T)$. And the $i^{{th}}$ element in $\mathcal{V}_j(T)$ is defined as $\hat{v}_j(i)$.
\end{definition}

\vspace{3ex}

\emph{\textbf{Proof of Theorem 1:}}

Firstly, setting:
\begin{equation}
\label{eq:A67}
\renewcommand{\theequation}{A.\arabic{equation}}
\begin{array}{l}
{a_5^t} = \frac{{4|{{\bf{A}}^t}|(\gamma  - 2){L^2}}}{{{\rho _1}({c_1^t})^2}} + \frac{{4N(\gamma  - 2){L^2}}}{{{\rho _2}({c_2^t})^2}} + \frac{{{\rho _2}(N - |{{\bf{Q}}^{t + 1}}|){L^2}}}{2} - \frac{1}{2},
\end{array}
\end{equation}
\begin{equation}
\label{eq:A67-1}
\renewcommand{\theequation}{A.\arabic{equation}}
\begin{array}{l}
{a_6^t} = \frac{{4|{{\bf{A}}^t}|(\gamma  - 2){L^2}}}{{{\rho _1}({c_1^t})^2}} + \frac{{4N(\gamma  - 2){L^2}}}{{{\rho _2}({c_2^t})^2}} + \frac{{{\rho _2}(N - |{{\bf{Q}}^{t + 1}}|){L^2}}}{2} - \frac{{3\tau{ k_1 }N{L^2}}}{2},
\end{array}
\end{equation}

\vspace{1ex}

\noindent where $\gamma $ is a constant which satisfies $\gamma > 2$ and $\frac{{4(\gamma  - 2){L^2}}}{{{\rho _1}({c_1^0}){^2}}} + \frac{{4N(\gamma  - 2){L^2}}}{{{\rho _2}({c_2^0}){^2}}} + \frac{{{\rho _2}(N - S){L^2}}}{2} \ge \max \{ \frac{1}{2},\frac{{3\tau{ k_1 }N{L^2}}}{2}\}$. It is seen that the ${a_5^t},{a_6^t}$ are nonnegative sequences. Since $\forall t \ge 0$, $|{{\bf{A}}^0}| \le |{{\bf{A}}^t}|$, $({c_1^0}){^2} \ge ({c_1^t})^2$, $({c_2^0}){^2} \ge ({c_2^t})^2$, and we assume that $|{{\bf{Q}}^{t + 1}}|=S, \forall t$, thus we have ${a_5^0} \le {a_5^t}, {a_6^0} \le {a_6^t},  \forall t$. According to the setting of ${\eta _{\boldsymbol{w}}^t}$, ${\eta _{\boldsymbol{z}}^t}$, ${\eta _h^t}$ and $c_1^t$, $c_2^t$, we have,
\begin{equation}
\renewcommand{\theequation}{A.\arabic{equation}}
\label{eq:A68}
\frac{L\! +\!1}{2} - \frac{1}{{{\eta _{\boldsymbol{w}}^t}}} \! +\! \frac{{{\rho _1}|{{\bf{A}}^t}|{L^2}}}{2} \! +\! \frac{{{\rho _2}|{{\bf{Q}}^{t + 1}}|{L^2}}}{2} \! +\! \frac{{8|{{\bf{A}}^t}|{L^2}}}{{{\rho _1}({c_1^t})^2}} \! +\! \frac{{8N{L^2}}}{{{\rho _2}({c_2^t})^2}} =  - {a_5^t},
\end{equation}
\begin{equation}
\renewcommand{\theequation}{A.\arabic{equation}}
\label{eq:A69}
\frac{L \! +\! 3\tau{ k_1 }N{L^2}}{2} - \frac{1}{{{\eta _{\boldsymbol{z}}^t}}} \! +\! \frac{{{\rho _1}|{{\bf{A}}^t}|{L^2}}}{2} \! +\! \frac{{{\rho _2}|{{\bf{Q}}^{t + 1}}|{L^2}}}{2} \! +\! \frac{{8|{{\bf{A}}^t}|{L^2}}}{{{\rho _1}({c_1^t})^2}} \! +\! \frac{{8N{L^2}}}{{{\rho _2}({c_2^t})^2}} =  - {a_6^t},
\end{equation}
\begin{equation}
\renewcommand{\theequation}{A.\arabic{equation}}
\label{eq:A70}
\frac{L\! +\! 3\tau{ k_1 }N{L^2}}{2} - \frac{1}{{{\eta _h^t}}} \! +\! \frac{{{\rho _1}|{{\bf{A}}^t}|{L^2}}}{2} \! +\! \frac{{{\rho _2}|{{\bf{Q}}^{t + 1}}|{L^2}}}{2} \! +\! \frac{{8|{{\bf{A}}^t}|{L^2}}}{{{\rho _1}({c_1^t})^2}} \! +\! \frac{{8N{L^2}}}{{{\rho _2}({c_2^t})^2}} =  - {a_6^t}.
\end{equation}

Combining Eq. (A.\ref{eq:A68}), (A.\ref{eq:A69}), (A.\ref{eq:A70})  with Lemma \ref{lemma3}, $\forall t \ge T_1 + \tau$, it follows that,
{\begin{equation}
\renewcommand{\theequation}{A.\arabic{equation}}
\label{eq:A71}
\begin{array}{l}
{a_5^t}\sum\limits_{j = 1}^N {||{{\boldsymbol{w}}_j^{t+1}} - {{\boldsymbol{w}}_j^t}|{|^2}}  \! +\! {a_6^t}||\boldsymbol{z}^{t+1} - \boldsymbol{z}^t|{|^2}  + {a_6^t}||h^{t+1} - h^t|{|^2}\\
 \! + {(\frac{1}{{10{\rho _1}}}\! -\! \frac{{3\tau{ k_1 }N{L^2}}}{2})  \sum\limits_{l = 1}^{|{{\bf{A}}^t}|}||{\lambda _l^{t+1}} - {\lambda _l^t}|{|^2}}   \! +\! \frac{1}{{10{\rho _2}}}\sum\limits_{j = 1}^N {||{{\boldsymbol{\phi}}_j^{t+1}} - {{\boldsymbol{\phi}}_j^t}|{|^2}} \\
 
 \le F^{t} - F^{t+1} \! +\! \frac{{{c_1^{t-1}} - {c_1^t}}}{2}\sum\limits_{l = 1}^{|{{\bf{A}}^t}|} {||{\lambda _l^{t+1}}|{|^2}}  \! +\! \frac{{{c_2^{t-1}} - {c_2^t}}}{2}\sum\limits_{j = 1}^N {||{{\boldsymbol{\phi}}_j^{t+1}}|{|^2}} \\
  +  {\frac{4}{{{\rho _1}}}(\frac{{{c_1^{t-2}}}}{{{c_1^{t-1}}}} - \frac{{{c_1^{t-1}}}}{{{c_1^t}}})\sum\limits_{l = 1}^{|{{\bf{A}}^t}|}||{\lambda _l^t}|{|^2}}  \! +\!  {\frac{4}{{{\rho _2}}}(\frac{{{c_2^{t-2}}}}{{{c_2^{t-1}}}} - \frac{{{c_2^{t-1}}}}{{{c_2^t}}})\sum\limits_{j = 1}^N||{{\boldsymbol{\phi}}_j^t}|{|^2}}.
\end{array}
\end{equation}}

Combining the definition of ${(\nabla{\widetilde{G}}^t)_{{{\boldsymbol{w}}_j}}}$ with trigonometric inequality, Cauchy-Schwarz inequality and Assumption 1 and 2, $\forall t \ge T_1 + \tau$, we have,
{\begin{equation}
\label{eq:A72}
\renewcommand{\theequation}{A.\arabic{equation}}
\begin{array}{l}
||{(\nabla {\widetilde{G}}^t)_{{{\boldsymbol{w}}_j}}}|{|^2}
 \!\le\! \frac{2}{\underline{\eta _{\boldsymbol{w}}}^2}||{{\boldsymbol{w}}_j^{\overline{{t}_j}}} \!-\! {{\boldsymbol{w}}_j^t}|{|^2} \! +\! 6\tau{k_1}{L^2}(||\boldsymbol{z}^{t+1} \!-\! \boldsymbol{z}^t|{|^2} \! +\! ||h^{t+1} \!-\! h^t|{|^2} \! +\! \sum\limits_{l = 1}^{|{{\bf{A}}^t}|} \! {||{\lambda _l^{t+1}} \!-\! {\lambda _l^t}|{|^2}}).
\end{array}
\end{equation}}

Combining the definition of ${(\nabla {\widetilde{G}}^t)_{\boldsymbol{z}}}$ with trigonometric inequality and Cauchy-Schwarz inequality, we can obtain the following inequality,
{\begin{equation}
\label{eq:A73}
\renewcommand{\theequation}{A.\arabic{equation}}
\begin{array}{l}
||{(\nabla {\widetilde{G}}^t)_{\boldsymbol{z}}}|{|^2}

\! \le \! 2{L^2}\sum\limits_{j = 1}^N {||{{\boldsymbol{w}}_j^{t+1}} - {{\boldsymbol{w}}_j^t}|{|^2}}  \! +\! \frac{2}{({\eta _{\boldsymbol{z}}^t})^2}||\boldsymbol{z}^{t+1} - \boldsymbol{z}^t|{|^2}.
\end{array}
\end{equation}}

Likewise, combining the definition of ${(\nabla {\widetilde{G}}^t)_h}$ with trigonometric inequality and Cauchy-Schwarz inequality, we have that,
{\begin{equation}
\label{eq:A75}
\renewcommand{\theequation}{A.\arabic{equation}}
\begin{array}{l}
||{(\nabla {\widetilde{G}}^t)_h}|{|^2}
 \le 2{L^2}(\sum\limits_{j = 1}^N {||{{\boldsymbol{w}}_j^{t+1}} \!-\! {{\boldsymbol{w}}_j^t}|{|^2}}  \! +\! ||{\boldsymbol{z}}^{t + 1} \!-\! {\boldsymbol{z}^t}|{|^2}) \! +\! \frac{2}{{({\eta _h^t})^2}}||h^{t+1} \!-\! h^t|{|^2}.
\end{array}
\end{equation}}

Combining the definition of ${(\nabla {\widetilde{G}}^t)_{{\lambda _l}}}$ with trigonometric inequality and Cauchy-Schwarz inequality,
{\begin{equation}
\renewcommand{\theequation}{A.\arabic{equation}}
\label{eq:A76}
\begin{array}{l}
||{(\nabla {\widetilde{G}}^t)_{{\lambda _l}}}|{|^2}\\
\! \le \! \frac{3}{{{\rho_1}^2}}||{\lambda _l^{t+1}} \! -\! {\lambda _l^t}|{|^2} \! +\! 3{L^2}(\!\sum\limits_{j = 1}^N\! {||{{\boldsymbol{w}}_j^{t+1}} \! -\! {{\boldsymbol{w}}_j^t}|{|^2}}  \! +\! ||{\boldsymbol{z}}^{t + 1} \! -\! {\boldsymbol{z}^t}|{|^2} \! +\! ||h^{t+1} \! -\! h^t|{|^2}) 
 \! +\! 3({{c_1^{t-1}} \! -\! {c_1^t})^2}||{\lambda _l^t}|{|^2}\\
\! \le \! \frac{3}{{{\rho_1}^2}}||{\lambda _l^{t+1}} \! -\! {\lambda _l^t}|{|^2} \! +\! 3{L^2}(\!\sum\limits_{j = 1}^N\! {||{{\boldsymbol{w}}_j^{t+1}} \! -\! {{\boldsymbol{w}}_j^t}|{|^2}}  \! +\! ||{\boldsymbol{z}}^{t + 1} \! -\! {\boldsymbol{z}^t}|{|^2} \! +\! ||h^{t+1} \! -\! h^t|{|^2}) 
 \! +\! 3{(({c_1^{t-1}})^2 \! -\! ({c_1^t})^2)}||{\lambda _l^t}|{|^2}.
\end{array}
\end{equation}}

\vspace{-2mm}

Combining the definition of ${(\nabla {\widetilde{G}}^t)_{{{\boldsymbol{\phi}}_j}}}$ with Cauchy-Schwarz inequality and Assumption 2, we have,
{\begin{equation}
\renewcommand{\theequation}{A.\arabic{equation}}
\label{eq:A77}
\begin{array}{l}
||{(\nabla {\widetilde{G}}^t)_{{{\boldsymbol{\phi}}_j}}}|{|^2}\\
\! \le \! \frac{3}{{{\rho_2}^2}}||{{\boldsymbol{\phi}}_j^{\overline{{t}_j}}} \! -\! {{\boldsymbol{\phi}}_j^t}|{|^2} \! +\! 3{L^2}(\!\sum\limits_{j = 1}^N\! {||{{\boldsymbol{w}}_j^{\overline{{t}_j}}} \! -\! {{\boldsymbol{w}}_j^t}|{|^2}}  \! +\! ||{\boldsymbol{z}}^{\overline{{t}_j}} \! -\! {\boldsymbol{z}^t}|{|^2})
 \! +\! 3{({c_2^{\widetilde{{t}}_j  - 1}} \! -\! {c_2^{\overline{{t}_j} - 1}})^2}||{{\boldsymbol{\phi}}_j^t}|{|^2}\\
\! \le\! \frac{3}{{{\rho_2}^2}}||{{\boldsymbol{\phi}}_j^{\overline{{t}_j}}} \! -\! {{\boldsymbol{\phi}}_j^t}|{|^2} \! +\! 3{L^2}(\!\sum\limits_{j = 1}^N\! {||{{\boldsymbol{w}}_j^{\overline{{t}_j}}} \! -\! {{\boldsymbol{w}}_j^t}|{|^2}}  \! +\!\tau{ k_1 }(||\boldsymbol{z}^{t+1} \!-\! \boldsymbol{z}^t|{|^2} \! +\! ||h^{t+1} \!-\! h^t|{|^2} \! +\! \sum\limits_{l = 1}^{|{{\bf{A}}^t}|} \! {||{\lambda _l^{t+1}} \!-\! {\lambda _l^t}|{|^2}}))\\
  + 3{(({c_2^{\widetilde{{t}}_j  - 1}})^2 \! -\! ({c_2^{\overline{{t}_j} -1}})^2)}||{{\boldsymbol{\phi}}_j^t}|{|^2}.
\end{array}
\end{equation}}

\vspace{-2mm}

According to the Definition \ref{definition:A2} as well as Eq. (A.\ref{eq:A72}), (A.\ref{eq:A73}), (A.\ref{eq:A75}), (A.\ref{eq:A76}) and Eq. (A.\ref{eq:A77}), $\forall t \ge T_1 + \tau$, we have that,
{\begin{equation}
\label{eq:A78}
\renewcommand{\theequation}{A.\arabic{equation}}
\begin{array}{l}
||\nabla {\widetilde{G}}^t|{|^2}
 = \sum\limits_{j = 1}^N {||{{(\nabla {\widetilde{G}}^t)}_{{{\boldsymbol{w}}_j}}}|{|^2}}  \! +\! ||{(\nabla {\widetilde{G}}^t)_{\boldsymbol{z}}}|{|^2} \! +\! ||{(\nabla {\widetilde{G}}^t)_h}|{|^2} \! +\! \sum\limits_{l = 1}^{|{{\bf{A}}^t}|} {||{{(\nabla {\widetilde{G}}^t)}_{{\lambda _l}}}|{|^2}}  \! +\! \sum\limits_{j = 1}^N {||{{(\nabla {\widetilde{G}}^t)}_{{{\boldsymbol{\phi}}_j}}}|{|^2}} \\
 
\quad \quad \quad \; \; \; \le (\frac{2}{\underline{\eta _{\boldsymbol{w}}}^2} \! +\! 3NL^2)\sum\limits_{j = 1}^N {||{{\boldsymbol{w}}_j^{\overline{{t}_j}}} - {{\boldsymbol{w}}_j^t}|{|^2}}\! +\! (4 \! +\! 3|{{\bf{A}}^t}|){L^2}\sum\limits_{j = 1}^N {||{{\boldsymbol{w}}_j^{t+1}} - {{\boldsymbol{w}}_j^t}|{|^2}} \\

\quad \quad \quad \; \; \; + (\frac{2}{({\eta _{\boldsymbol{z}}^t})^2} \! +\! (2 \! +\! 9\tau{ k_1 }N \! +\! 3|{{\bf{A}}^t}|){L^2})||\boldsymbol{z}^{t+1} \! -\! \boldsymbol{z}^t|{|^2}
 \! +\! (\frac{2}{({\eta _h^t})^2} \! +\! (9\tau{ k_1 }N \! +\! 3|{{\bf{A}}^t}|){L^2})||h^{t+1}\! -\! h^t|{|^2} \\ 
 
\quad \quad \quad \; \; \; + \sum\limits_{l = 1}^{|{{\bf{A}}^t}|} {(\frac{3}{{{\rho_1}^2}} \! +\! 9\tau{ k_1 }N{L^2})||{\lambda _l^{t+1}} - {\lambda _l^t}|{|^2}}  \! +\! \sum\limits_{l = 1}^{|{{\bf{A}}^t}|} {3{{(({c_1^{t-1}})^2 - ({c_1^t})^2)}}||{\lambda _l^t}|{|^2}} \\

\quad \quad \quad \; \; \; + \sum\limits_{j = 1}^N {\frac{3}{{{\rho_2}^2}}||{{\boldsymbol{\phi}}_j^{\overline{{t}_j}}} - {{\boldsymbol{\phi}}_j^t}|{|^2}}  \! +\! \sum\limits_{j = 1}^N {3{(({c_2^{\widetilde{{t}}_j - 1}})^2 - ({c_2^{\overline{{t}_j}-1}})^2)}||{{\boldsymbol{\phi}}_j^t}|{|^2}} .
\end{array}
\end{equation}}

\vspace{-2mm}

We set constants $d_1$, $d_2$, $d_3$ as,
\begin{equation}
\renewcommand{\theequation}{A.\arabic{equation}}
\label{eq:A79}{d_1} = \frac{{2 k_{\tau} \tau \! +\! (4 \! +\! 3M \! +\! 3 k_{\tau}  \tau N){L^2}\underline{\eta _{\boldsymbol{w}}}^2}}{{\underline{\eta _{\boldsymbol{w}}}^2({a_5^0}){^2}}} \ge \frac{{2 k_{\tau}  \tau \! +\! (4 \! +\! 3|{{\bf{A}}^t}| \! +\! 3k_{\tau}   \tau N){L^2}\underline{\eta _{\boldsymbol{w}}}^2}}{{\underline{\eta _{\boldsymbol{w}}}^2({a_5^t}){^2}}},
\end{equation}
\begin{equation}
\renewcommand{\theequation}{A.\arabic{equation}}
\label{eq:A80}
{d_2} = \frac{{2 \! +\! (2 \! +\!{9\tau{ k_1 }N} \! +\! 3M){L^2}\underline{\eta _{\boldsymbol{z}}}^2}}{{\underline{\eta _{\boldsymbol{z}}}^2({a_6^0}){^2}}} \ge \frac{{2 \! +\! (2 \! +\!{9\tau{ k_1 }N} \! +\! 3|{{\bf{A}}^t}|){L^2}({\eta _{\boldsymbol{z}}^t})^2}}{{({\eta _{\boldsymbol{z}}^t})^2({a_6^t}){^2}}},
\end{equation}
\begin{equation}
\renewcommand{\theequation}{A.\arabic{equation}}
\label{eq:A81}
{d_3} = \frac{{2 \! +\! ({9\tau{ k_1 }N} \! +\! 3M){L^2}\underline{\eta _h}^2}}{{\underline{\eta _h}^2({a_6^0}){^2}}} \ge \frac{{2 \! +\! ({9\tau{ k_1 }N} +3|{{\bf{A}}^t}|){L^2}({\eta _h^t})^2}}{{({\eta _h^t})^2({a_6^t}){^2}}},
\end{equation}

\vspace{1ex}

\noindent where  $k_{\tau}$,  $\underline{\eta _{\boldsymbol{z}}}$ and $\underline{\eta _{h}}$ are positive constants. $ \underline{\eta _{\boldsymbol{z}}} = \frac{2}{{L + {\rho _1}M{L^2} + {\rho _2}N{L^2} + 8(\frac{{M\gamma {L^2}}}{{\rho _1}{\underline{c}_1}^2} + \frac{{N\gamma {L^2}}}{{\rho _2}{\underline{c}_2}^2})}} \le \eta _{\boldsymbol{z}}^t $ and  $\underline{\eta _h} = \frac{2}{{L + {\rho _1}M{L^2} + {\rho _2}N{L^2} + 8(\frac{{M\gamma {L^2}}}{{\rho _1}{\underline{c}_1}^2} + \frac{{N\gamma {L^2}}}{{\rho _2}{\underline{c}_2}^2})}} \le \eta _h^t, \forall t$. Thus, combining Eq. (A.\ref{eq:A78}) with Eq. (A.\ref{eq:A79}), (A.\ref{eq:A80}), (A.\ref{eq:A81}), $\forall t \ge T_1 + \tau$, we have,
{\begin{equation}
\label{eq:A82}
\renewcommand{\theequation}{A.\arabic{equation}}
\begin{array}{l}
||\nabla {\widetilde{G}}^t|{|^2} 
\! \le \! \sum\limits_{j = 1}^N {{d_1}({a_5^t}){^2}||{{\boldsymbol{w}}_j^{t+1}} - {{\boldsymbol{w}}_j^t}|{|^2}}  \! +\! {d_2}({a_6^t}){^2}||{\boldsymbol{z}}^{t + 1} - {\boldsymbol{z}^t}|{|^2} \! +\! {d_3}({a_6^t}){^2}||h^{t+1} - h^t|{|^2}\\

\quad \quad \quad \; \; \; \! +\! \sum\limits_{l = 1}^{|{{\bf{A}}^t}|}\! {(\frac{3}{{{\rho_1}^2}} \! +\! 9\tau{ k_1 }N{L^2})||{\lambda _l^{t+1}}\! -\! {\lambda _l^t}|{|^2}}  \! +\! \sum\limits_{l = 1}^{|{{\bf{A}}^t}|} {3(({c_1^{t - 1}}){^2} \!-\! ({c_1^t})^2)||{\lambda _l^t}|{|^2}} \! +\! \sum\limits_{j = 1}^N  \! {\frac{3}{{{\rho_2}^2}}||{{\boldsymbol{\phi}}_j^{\overline{{t}_j}}} - {{\boldsymbol{\phi}}_j^t}|{|^2}}   \\ 
 
\quad \quad \quad \; \; \;  \! +\! \sum\limits_{j = 1}^N {3{(({c_2^{\widetilde{{t}}_j - 1}})^2 - ({c_2^{\overline{{t}_j}-1}})^2)}||{{\boldsymbol{\phi}}_j^t}|{|^2}} \! +\! (\frac{2}{\underline{\eta _{\boldsymbol{w}}}^2} \! +\! 3NL^2)\sum\limits_{j = 1}^N {||{{\boldsymbol{w}}_j^{\overline{{t}_j}}} - {{\boldsymbol{w}}_j^t}|{|^2}} \\
 
\quad \quad \quad \; \; \; \! - (\frac{2 k_{\tau} \tau}{\underline{\eta _{\boldsymbol{w}}}^2} \! +\! 3 k_{\tau} \tau NL^2)\sum\limits_{j = 1}^N {||{{\boldsymbol{w}}_j^{t+1}} - {{\boldsymbol{w}}_j^t}|{|^2}}.
\end{array}
\end{equation}}

Let $d_4^t$ denote a nonnegative sequence:
{\begin{equation}
\renewcommand{\theequation}{A.\arabic{equation}}
\label{eq:A83}
{d_4^t} = \frac{1}{{\max \{ {d_1}{a_5^t},{d_2}{a_6^t},{d_3}{a_6^t},\frac{{\frac{{30}}{{{\rho _1}}} + 90{\rho _1}\tau{ k_1 }N{L^2}}}{{1 - 15{\rho _1}\tau{ k_1 }N{L^2}}},\frac{{30 \tau }}{{{\rho _2}}}\} }}.
\end{equation}}

It is seen that $d_4^0 \ge d_4^t, \forall t \ge 0$. And we denote the lower bound of $d_4^t$ as $\underline{d_4}$, it appears that $d_4^t\ge \underline{d_4} \ge0, \forall t \ge 0$. And we set the constant $k_{\tau}$ satisfies $k_{\tau} \ge \frac{{{d_4^0}}(\frac{2}{\underline{\eta _{\boldsymbol{w}}}^2} + 3NL^2) }{{\underline{d_4}}(\frac{2}{\overline{\eta _{\boldsymbol{w}}}^2} + 3  NL^2)}$, where ${\overline{\eta _{\boldsymbol{w}}}}$ is the step-size in terms of $\boldsymbol{w}_j$ in the first iteration (it is seen that ${\overline{\eta _{\boldsymbol{w}}}} \ge {\eta _{\boldsymbol{w}}^t}, \forall t $). Then, $\forall t \ge T_1 + \tau$, we can obtain the following inequality from Eq. (A.\ref{eq:A82}) and Eq. (A.\ref{eq:A83}):
{\begin{equation}
\label{eq:A84}
\renewcommand{\theequation}{A.\arabic{equation}}
\begin{array}{l}
{d_4^t}||\nabla {\widetilde{G}}^t|{|^2}
 \!\le\! {a_5^t} \! \sum\limits_{j = 1}^N \! {||{{\boldsymbol{w}}_j^{t+1}}\! -\! {{\boldsymbol{w}}_j^t}|{|^2}}  \! +\! {a_6^t}||{\boldsymbol{z}}^{t + 1}\! -\! {\boldsymbol{z}^t}|{|^2} \! +\! {a_6^t}||h^{t+1} \!-\! h^t|{|^2}\\
 
\quad \quad \quad \quad \; \; \;  \! +  {(\frac{1}{{10{\rho _1}}}\! -\! \frac{{3\tau{ k_1 }N{L^2}}}{2})\sum\limits_{l = 1}^{|{{\bf{A}}^t}|}||{\lambda _l^{t+1}} - {\lambda _l^t}|{|^2}}  \! +\! \frac{1}{{10\tau {\rho _2}}}\sum\limits_{j = 1}^N {||{{\boldsymbol{\phi}}_j^{\overline{{t}_j}}} - {{\boldsymbol{\phi}}_j^t}|{|^2}} \\

\quad \quad \quad \quad \; \; \; \! + 3{d_4^t}(({c_1^{t-1}}){^2} - ({c_1^t}){^2})\sum\limits_{l = 1}^{|{{\bf{A}}^t}|} {||{\lambda _l^t}|{|^2}}  \! +\! 3{d_4^t}\sum\limits_{j = 1}^N(({c_2^{\widetilde{{t}}_j - 1}}){^2} - ({c_2^{\overline{{t}_j}-1}}){^2}) {||{{\boldsymbol{\phi}}_j^t}|{|^2}} \\

\quad \quad \quad \quad  \; \; \; \! +{d_4^t} (\frac{2}{\underline{\eta _{\boldsymbol{w}}}^2} \! +\! 3NL^2)\sum\limits_{j = 1}^N {||{{\boldsymbol{w}}_j^{\overline{{t}_j}}} - {{\boldsymbol{w}}_j^t}|{|^2}} - {d_4^t}(\frac{2 k_{\tau}  \tau}{\underline{\eta _{\boldsymbol{w}}}^2} \! +\! 3 k_{\tau}   \tau NL^2)\sum\limits_{j = 1}^N {||{{\boldsymbol{w}}_j^{t+1}} - {{\boldsymbol{w}}_j^t}|{|^2}}.

\end{array}
\end{equation}}

Combining  Eq. (A.\ref{eq:A84}) with Eq. (A.\ref{eq:A71}) and  according to the setting $||{\lambda _l^t}|{|^2} \le {\sigma _1}^2$, $||{{\boldsymbol{\phi}}_j^t}|{|^2} \le {\sigma _2}^2$ (where ${\sigma _1}^2={\alpha_3}^2$, ${\sigma _2}^2=p{\alpha_4}^2$) and ${d_4^0} \ge {d_4^t} \ge \underline{d_4}$, thus, $\forall t \ge T_1 + \tau$, we have,
{\begin{equation}
\label{eq:A85}
\renewcommand{\theequation}{A.\arabic{equation}}
\begin{array}{l}
{d_4^t}||\nabla {\widetilde{G}}^t|{|^2} \\
  \le F^{t} - F^{t+1} + \frac{{{c_1^{t-1}} - {c_1^t}}}{2}M{\sigma _1}^2  + \frac{{{c_2^{t-1}} - {c_2^t}}}{2}N{\sigma _2}^2 
 + \frac{4}{{{\rho _1}}}(\frac{{{c_1^{t-2}}}}{{{c_1^{t-1}}}} - \frac{{{c_1^{t-1}}}}{{{c_1^t}}})M{\sigma _1}^2 \vspace{0.5ex}\\
 + \frac{4}{{{\rho _2}}}(\frac{{{c_2^{t-2}}}}{{{c_2^{t-1}}}} - \frac{{{c_2^{t-1}}}}{{{c_2^t}}})N{\sigma _2}^2 
 + 3{d_4^0}(({c_1^{t-1}}){^2} - ({c_1^t}){^2})M{\sigma _1}^2 + 3{d_4^0}\sum\limits_{j = 1}^N(({c_2^{\widetilde{{t}}_j - 1}}){^2} - ({c_2^{\overline{{t}_j}-1}}){^2}){\sigma _2}^2 \vspace{0.5ex}\\
 
  + \frac{1}{{10\tau {\rho _2}}}\sum\limits_{j = 1}^N {||{{\boldsymbol{\phi}}_j^{\overline{{t}_j}}} - {{\boldsymbol{\phi}}_j^t}|{|^2}} - \frac{1}{{10{\rho _2}}}\sum\limits_{j = 1}^N {||{{\boldsymbol{\phi}}_j^{t+1}} - {{\boldsymbol{\phi}}_j^t}|{|^2}}\\
 
  +{d_4^0} (\frac{2}{\underline{\eta _{\boldsymbol{w}}}^2} + 3NL^2)\sum\limits_{j = 1}^N {||{{\boldsymbol{w}}_j^{\overline{{t}_j}}} - {{\boldsymbol{w}}_j^t}|{|^2}} - \underline{d_4}(\frac{2 k_{\tau}  \tau}{\underline{\eta _{\boldsymbol{w}}}^2} + 3 k_{\tau}   \tau NL^2)\sum\limits_{j = 1}^N {||{{\boldsymbol{w}}_j^{t+1}} - {{\boldsymbol{w}}_j^t}|{|^2}}.

\end{array}
\end{equation}}

Denoting $\widetilde{T}(\varepsilon )$ as $\widetilde{T}(\varepsilon ) = \min \{ t \ | \; ||\nabla \widetilde{G}^{T_1+t}|| \le \frac{\varepsilon}{2}, t\ge \tau\}$. Summing up Eq. (A.\ref{eq:A85}) from $t=T_1+\tau$ to $t =T_1+{{\widetilde{T}} (\varepsilon )} $, we have,
{\begin{equation}
\label{eq:A86}
\renewcommand{\theequation}{A.\arabic{equation}}
\begin{array}{l}
\sum\limits_{t =T_1\!+\! \tau}^{T_1\!+\!{\widetilde{T}} (\varepsilon )} {{d_4^t}||\nabla {\widetilde{G}}^t|{|^2}} \vspace{1ex}\\
 
 \le F^{T_1 + \tau} - \mathop L\limits_ - 
 + \frac{4}{{{\rho _1}}}(\frac{{{c_1^{T_1+\tau-2}}}}{{{c_1^{T_1 + \tau -1}}}} + \frac{{{c_1^{T_1 + \tau -1}}}}{{{c_1^{T_1 + \tau}}}})M {{\sigma _1}^2}  + \frac{{{c_1^{T_1+ \tau -1}}}}{2}M {{\sigma _1}^2}  + \frac{7}{{2{\rho _1}}}M {{\sigma _3}^2}  + 3{d_4^0} ({{c_1^0}){^2}M {{\sigma _1}^2} } \vspace{0.5ex}\\
 
 + \frac{4}{{{\rho _2}}}(\frac{{{c_1^{T_1+ \tau -2}}}}{{{c_1^{T_1 + \tau -1}}}} + \frac{{{c_1^{T_1 + \tau -1}}}}{{{c_1^{T_1 + \tau}}}})N {{\sigma _2}^2}  + \frac{{{c_2^{T_1+ \tau -1}}}}{2}N {{\sigma _2}^2}  + \frac{7}{{2{\rho _2}}}N {{\sigma _4}^2}  + \sum\limits_{j = 1}^N\sum\limits_{t =T_1\!+\! \tau}^{T_1\!+\!{\widetilde{T}} (\varepsilon )}3{d_4^0}(({c_2^{\widetilde{{t}}_j - 1}}){^2} - ({c_2^{\overline{{t}_j}-1}}){^2}){\sigma _2}^2 \vspace{0.5ex}\\
 
+  \frac{c_1^{T_1+ \tau}}{{2}}M {{\sigma _1}^2} +  \frac{c_2^{T_1+ \tau}}{{2}}N {{\sigma _2}^2} + \frac{1}{{10\tau {\rho _2}}}\sum\limits_{j = 1}^N \sum\limits_{t =T_1\!+\! \tau}^{T_1\!+\!{\widetilde{T}} (\varepsilon )} {||{{\boldsymbol{\phi}}_j^{\overline{{t}_j}}} - {{\boldsymbol{\phi}}_j^t}|{|^2}} - \frac{1}{{10{\rho _2}}}\sum\limits_{j = 1}^N \sum\limits_{t =T_1\!+\! \tau}^{T_1\!+\!{\widetilde{T}} (\varepsilon )} {||{{\boldsymbol{\phi}}_j^{t+1}} - {{\boldsymbol{\phi}}_j^t}|{|^2}}\\
 
  +{d_4^0} (\frac{2}{\underline{\eta _{\boldsymbol{w}}}^2} + 3NL^2)\sum\limits_{j = 1}^N \sum\limits_{t =T_1\!+\! \tau}^{T_1\!+\!{\widetilde{T}} (\varepsilon )} {||{{\boldsymbol{w}}_j^{\overline{{t}_j}}} - {{\boldsymbol{w}}_j^t}|{|^2}} - \underline{d_4}(\frac{2 k_{\tau} \tau}{\overline{\eta _{\boldsymbol{w}}}^2} + 3 k_{\tau} \tau NL^2)\sum\limits_{j = 1}^N \sum\limits_{t =T_1\!+\! \tau}^{T_1\!+\!{\widetilde{T}} (\varepsilon )} {||{{\boldsymbol{w}}_j^{t+1}} - {{\boldsymbol{w}}_j^t}|{|^2}},  \vspace{1.5ex}
\end{array}
\end{equation}}

\vspace{1ex}

\noindent where ${\sigma _3} \!=\! \max \{  ||{\lambda _1} - {\lambda _2}|| \, {\rm{ }}|{\lambda _1},{\lambda _2} \! \in \! {\bf{\Lambda}}  \} $, ${\sigma _4} = \max \{ ||{\boldsymbol{\phi}_1} - {\boldsymbol{\phi}_2}|| \, {\rm{ }}|{\boldsymbol{\phi}_1},{\boldsymbol{\phi}_2} \! \in \! {{\boldsymbol{\Phi}}} \} $ and $\mathop L\limits_ -  \! =  \! \mathop {\min }\limits_{{\rm{\{ }}{{\boldsymbol{w}}_j}  \in  {{\boldsymbol{\mathcal{W}}}}{\rm{\} , }}{\boldsymbol{z}} \in  {{\boldsymbol{\mathcal{Z}}}} , h  \in   {{\boldsymbol{\mathcal{H}}}},\{  {\lambda _l} \in {\bf{\Lambda}}   \} ,\{  {{\boldsymbol{\phi}}_j} \in {{\boldsymbol{\Phi}}}  \} }  {L_p}{\rm{(  \{  }}{{\boldsymbol{w}}_j}{\rm{  \},}}{\boldsymbol{z}},h,  \{  {\lambda _l}  \} ,\{  {{\boldsymbol{\phi}}_j}  \}  {\rm{)}}$, which satisfy that, $\forall t \ge T_1+\tau$,
{\begin{equation}
\label{eq:A87}
\renewcommand{\theequation}{A.\arabic{equation}}
F^{t+1} \!\ge\! \mathop L\limits_ -   -  \frac{4}{{{\rho _1}}}\frac{{{c_1^{T_1 +\tau - 1}}}}{{{c_1^{T_1+\tau}}}}M {{\sigma _1}^2}  - \frac{4}{{{\rho _2}}}\frac{{{c_2^{T_1+\tau - 1}}}}{{{c_2^{T_1+\tau}}}}N {{\sigma _2}^2}  - \frac{7}{{2{\rho _1}}}M {{\sigma _3}^2}  - \frac{7}{{2{\rho _2}}}N {{\sigma _4}^2} -  \frac{c_1^{T_1+\tau}}{{2}}M {{\sigma _1}^2} -  \frac{c_2^{T_1+\tau}}{{2}}N {{\sigma _2}^2}.
\end{equation}}

For each worker $j$, the iterations between the last iteration and the next iteration where it is active is no more than $\tau$, \textit{i.e.}, $\overline{{t}_j} -\widetilde{{t}}_j \le \tau $, we have,
{\begin{equation}
\label{eq:A87-0}
\renewcommand{\theequation}{A.\arabic{equation}}
\begin{array}{l}
\sum\limits_{t =T_1\!+\! \tau}^{T_1\!+\!{\widetilde{T}} (\varepsilon )}3{d_4^0}(({c_2^{\widetilde{{t}}_j - 1}}){^2} - ({c_2^{\overline{{t}_j}-1}}){^2}){\sigma _2}^2 \vspace{1ex}\\

\le \tau \sum\limits_{\scriptstyle{{\hat v}_j}(i) \in \mathcal{V}_j({\widetilde{T}}(\varepsilon )),\hfill\atop 
\scriptstyle{T_1 + \tau} \le {{\hat v}_j}(i) \le T_1+{\widetilde{T}} (\varepsilon ) \hfill} 3{d_4^0}(({c_2^{{\hat v}_j(i) - 1}}){^2} - ({c_2^{{\hat v}_j(i+1)-1}}){^2}){\sigma _2}^2
\vspace{1ex}\\

\le 3\tau {d_4^0}({c_2^0}){^2}{\sigma _2}^2.
\end{array}
\end{equation}}

Since the idle workers do not update their variables in each iteration, for any $t$ that satisfies ${\hat v_j}(i - 1) \le t < {\hat v_j}(i)$, we have ${\boldsymbol{\phi}_j^t} = {\boldsymbol{\phi}_j^{{\hat v_j}(i) - 1}}$. And for $t \notin \mathcal{V}_j(T)$, we have ${||{{\boldsymbol{\phi}}_j^t} - {{\boldsymbol{\phi}}_j^{t-1}}|{|^2}}=0$. Combing with ${\hat v_j}(i) - {\hat v_j}(i - 1) \le \tau $, we can obtain that, 
{\begin{equation}
\label{eq:A87-1}
\renewcommand{\theequation}{A.\arabic{equation}}
\begin{array}{l}
\sum\limits_{j = 1}^N\sum\limits_{t =T_1\!+\! \tau}^{T_1\!+\!{\widetilde{T}} (\varepsilon )}\! { {||{{\boldsymbol{\phi}}_j^{\overline {{t_j}}}} \!-\! {{\boldsymbol{\phi}}_j^t}|{|^2}} } 

\!\le\! \tau \sum\limits_{j = 1}^N \sum\limits_{\scriptstyle{{\hat v}_j}(i) \in \mathcal{V}_j({\widetilde{T}}(\varepsilon )),\hfill\atop
\scriptstyle T_1 + \tau + 1 \le {{\hat v}_j}(i)\hfill} { {||{\boldsymbol{\phi}_j^{{{\hat v}_j}(i)}} - {\boldsymbol{\phi}_j^{{{\hat v}_j}(i) - 1}}|{|^2}} } 
\vspace{0.5ex}\\

\quad \, \quad \quad \quad \quad \quad \quad \quad \quad \;  = \tau \! \sum\limits_{j = 1}^N \sum\limits_{t =T_1\!+\! \tau}^{T_1\!+\!{\widetilde{T}} (\varepsilon )}\! { {||{{\boldsymbol{\phi}}_j^{t+1}}\! -\! {{\boldsymbol{\phi}}_j^t}|{|^2}} }  + \tau \!\sum\limits_{j = 1}^N \sum\limits_{t =T_1\!+\! {\widetilde{T}}(\varepsilon ) + 1}^{T_1\!+\!{\widetilde{T}}(\varepsilon ) + \tau  - 1} { \!{||{{\boldsymbol{\phi}}_j^{t+1}} \!-\! {{\boldsymbol{\phi}}_j^t}|{|^2}} } \vspace{0.5ex}\\

\quad \, \quad \quad \quad \quad \quad \quad \quad \quad \;  \le \tau\!\sum\limits_{j = 1}^N \sum\limits_{t =T_1\!+\! \tau}^{T_1\!+\!{\widetilde{T}} (\varepsilon )}\! { {||{{\boldsymbol{\phi}}_j^{t+1}} \!-\! {{\boldsymbol{\phi}}_j^t}|{|^2}} }  + 4\tau (\tau \! -\! 1)N{\sigma _2}^2.
\end{array}
\end{equation}}

Similarly, for any $t$ that satisfies ${\hat v_j}(i - 1) \le t < {\hat v_j}(i)$, we have ${\boldsymbol{w}_j^t} = {\boldsymbol{w}_j^{{\hat v_j}(i) - 1}}$. And for $t \notin \mathcal{V}_j(T)$, we have ${||{{\boldsymbol{w}}_j^t} - {{\boldsymbol{w}}_j^{t-1}}|{|^2}}=0$. Combing with ${\hat v_j}(i) - {\hat v_j}(i - 1) \le \tau $, we can obtain,
{\begin{equation}
\label{eq:A87-2}
\renewcommand{\theequation}{A.\arabic{equation}}
\begin{array}{l}
\sum\limits_{j = 1}^N \sum\limits_{t =T_1\!+\! \tau}^{T_1\!+\!{\widetilde{T}} (\varepsilon )}\! { {||{{\boldsymbol{w}}_j^{\overline {{t_j}}}} \!-\! {{\boldsymbol{w}}_j^t}|{|^2}} } 

\!\le\! \tau\! \sum\limits_{j = 1}^N \sum\limits_{\scriptstyle{{\hat v}_j}(i) \in \mathcal{V}_j({\widetilde{T}}(\varepsilon )),\hfill\atop
\scriptstyle T_1+ \tau + 1 \le {{\hat v}_j}(i)\hfill}\! { {||{\boldsymbol{w}_j^{{{\hat v}_j}(i)}} - {\boldsymbol{w}_j^{{{\hat v}_j}(i) - 1}}|{|^2}} } 
\vspace{0.5ex}\\

\quad \,  \; \quad \quad \quad \quad \quad \quad  \quad \quad \;  = \tau \! \sum\limits_{j = 1}^N \sum\limits_{t =T_1\!+\! \tau}^{T_1\!+\!{\widetilde{T}} (\varepsilon )}\! { {||{{\boldsymbol{w}}_j^{t+1}}\! -\! {{\boldsymbol{w}}_j^t}|{|^2}} } \! +\! \tau\! \sum\limits_{j = 1}^N \sum\limits_{t = {T_1\!+\!\widetilde{T}}(\varepsilon ) + 1}^{T_1\!+\!{\widetilde{T}}(\varepsilon ) + \tau  - 1} { {||{{\boldsymbol{w}}_j^{t+1}} \!-\! {{\boldsymbol{w}}_j^t}|{|^2}} } \vspace{0.5ex}\\

\quad \,  \; \quad \quad \quad \quad \quad \quad  \quad \quad \;  \le \tau \! \sum\limits_{j = 1}^N \sum\limits_{t =T_1\!+\! \tau}^{T_1\!+\!{\widetilde{T}} (\varepsilon )}\! { {||{{\boldsymbol{w}}_j^{t+1}} \!-\! {{\boldsymbol{w}}_j^t}|{|^2}} }  + 4\tau (\tau  \!- \!1)pN{\alpha _1}^2.
\end{array}
\end{equation}}

It follows from Eq. (A.\ref{eq:A86}), (A.\ref{eq:A87-0}), (A.\ref{eq:A87-1}), (A.\ref{eq:A87-2}) that,
{\begin{equation}
\label{eq:A87-3}
\renewcommand{\theequation}{A.\arabic{equation}}
\begin{array}{l}
\sum\limits_{t =T_1\!+\! \tau}^{T_1\!+\!{\widetilde{T}} (\varepsilon )} {{d_4^t}||\nabla {\widetilde{G}}^t|{|^2}} \vspace{0.5ex}\\
 
 \le F^{T_1 + \tau} - \mathop L\limits_ -  
 + \frac{4}{{{\rho _1}}}(\frac{{{c_1^{T_1 + \tau - 2}}}}{{{c_1^{T_1 + \tau - 1}}}} + \frac{{{c_1^{T_1  + \tau - 1}}}}{{{c_1^{T_1  + \tau}}}})M {{\sigma _1}^2}  + \frac{{{c_1^{T_1+ \tau - 1}}}}{2}M {{\sigma _1}^2}  + \frac{7}{{2{\rho _1}}}M {{\sigma _3}^2}  + 3{d_4^0} {({c_1^0}){^2}M {{\sigma _1}^2} } \vspace{0.5ex}\\
 
 + \frac{4}{{{\rho _2}}}(\frac{{{c_1^{T_1+ \tau - 2}}}}{{{c_1^{T_1 + \tau - 1}}}} + \frac{{{c_1^{T_1 + \tau - 1}}}}{{{c_1^{T_1 + \tau}}}})N {{\sigma _2}^2}  + \frac{{{c_2^{T_1+ \tau - 1}}}}{2}N {{\sigma _2}^2}  + \frac{7}{{2{\rho _2}}}N {{\sigma _4}^2}  + 3\tau{d_4^0} {({c_2^{0}}){^2}N {{\sigma _2}^2} } \vspace{0.5ex}\\
 
+  \frac{c_1^{T_1+ \tau}}{{2}}M {{\sigma _1}^2} +  \frac{c_2^{T_1+ \tau}}{{2}}N {{\sigma _2}^2} + (\frac{{2N{\sigma _2}^2}}{{5{\rho _2}}} + 4{d_4^0}(\frac{2}{\underline{\eta _{\boldsymbol{w}}}^2} + 3N{L^2})pN{\alpha _1}^2\tau) (\tau  - 1) \\
 = \mathop d\limits^ -  + k_d(\tau -1),
\end{array}
\end{equation}}

\noindent where $\mathop d\limits^ -$ and $k_d$ are constants. And constant $d_5$ is given by, 
{\begin{equation}
\renewcommand{\theequation}{A.\arabic{equation}}
\label{eq:A88}
\begin{array}{l}
{d_5} = \max \{ \frac{{{d_1}}}{{{a_6^0}}},\frac{{{d_2}}}{{{a_5^0}}},\frac{{{d_3}}}{{{a_5^0}}},\frac{{\frac{{30}}{{{\rho _1}}} + 90{\rho _1}\tau{ k_1 }N{L^2}}}{{(1 - 15{\rho _1}\tau{ k_1 }N{L^2}){a_5^0}{a_6^0}}},\frac{{30\tau}}{{{\rho _2}{a_5^0}{a_6^0}}}\} \vspace{1ex}\\
 \ge \max \{ \frac{{{d_1}}}{{{a_6^t}}},\frac{{{d_2}}}{{{a_5^t}}},\frac{{{d_3}}}{{{a_5^t}}},\frac{{\frac{{30}}{{{\rho _1}}} + 90{\rho _1}\tau{ k_1 }N{L^2}}}{{(1 - 15{\rho _1}\tau{ k_1 }N{L^2}){a_5^t}{a_6^t}}},\frac{{30\tau}}{{{\rho _2}{a_5^t}{a_6^t}}}\} \vspace{1ex}\\
 = \frac{1}{{{d_4^t}{a_5^t}{a_6^t}}}.
\end{array}
\end{equation}}

\vspace{2ex}

Thus, we can obtain that,
{\begin{equation}
\label{eq:A89}
\renewcommand{\theequation}{A.\arabic{equation}}
\sum\limits_{t =T_1\!+\! \tau}^{T_1\!+\!{\widetilde{T}} (\varepsilon )} {\frac{1}{{{d_5}{a_5^t}{a_6^t}}}||\nabla \widetilde{G}^{T_1+\widetilde{T}(\varepsilon )}|{|^2}}  \le \sum\limits_{t =T_1\!+\! \tau}^{T_1\!+\!{\widetilde{T}} (\varepsilon )} {\frac{1}{{{d_5}{a_5^t}{a_6^t}}}||\nabla \widetilde{G}^t|{|^2}}  \le \sum\limits_{t =T_1\!+\! \tau}^{T_1\!+\!{\widetilde{T}} (\varepsilon )} {{d_4^t}||\nabla \widetilde{G}^t|{|^2}}  \le \mathop d\limits^ -  + k_d(\tau -1) .
\end{equation}}

And it follows from Eq. (A.\ref{eq:A89}) that,
{\begin{equation}
\label{eq:A90}
\renewcommand{\theequation}{A.\arabic{equation}}
||\nabla \widetilde{G}^{T_1+\widetilde{T}(\varepsilon )}|{|^2} \le \frac{{(\mathop d\limits^ -  + k_d(\tau -1) )  {d_5}}}{{\sum\limits_{t =T_1\!+\! \tau}^{T_1\!+\!{\widetilde{T}} (\varepsilon )} {\frac{1}{{{a_5^t}{a_6^t}}}}}}.  
\end{equation}}

According to the setting of ${c_1^t}$, ${c_2^t}$ and Eq. (A.\ref{eq:A67}), (A.\ref{eq:A67-1}), we have,

{\begin{equation}
\renewcommand{\theequation}{A.\arabic{equation}}
\label{eq:A91}
\frac{1}{{{a_5^t}{a_6^t}}} \ge \frac{1}{{{{(4(\gamma  - 2){L^2}(M{\rho _1} + N{\rho _2}){(t+1)^{\frac{1}{3}}} + \frac{{{\rho _2}(N - S){L^2}}}{2})}^2}}}.
\end{equation}}

Summing up $\frac{1}{{{a_5^t}{a_6^t}}}$ from ${t =T_1 \!+\! \tau}$ to ${t =T_1 \!+\! {{\widetilde{T}} (\varepsilon) }}$, it follows that,
{\begin{equation}
\label{eq:A92}
\renewcommand{\theequation}{A.\arabic{equation}}
\begin{array}{l}
\sum\limits_{t =T_1\!+\! \tau}^{T_1\!+\!{\widetilde{T}} (\varepsilon )} {\frac{1}{{{a_5^t}{a_6^t}}}} 
 \ge \sum\limits_{t =T_1\!+\! \tau}^{T_1\!+\!{\widetilde{T}} (\varepsilon )} {\frac{1}{{{{(4(\gamma  - 2){L^2}(M{\rho _1} + N{\rho _2}){(t+1)^{\frac{1}{3}}} + \frac{{{\rho _2}(N - S){L^2}}}{2})}^2}}}} \vspace{1ex}\\
\quad \,  \quad \quad \quad \quad   \ge \sum\limits_{t =T_1\!+\! \tau}^{T_1\!+\!{\widetilde{T}} (\varepsilon )} {\frac{1}{{{{(4(\gamma  - 2){L^2}(M{\rho _1} + N{\rho _2}){(t+1)^{\frac{1}{3}}} + \frac{{{\rho _2}(N - S){L^2}}}{2}{(t+1)^{\frac{1}{3}}})}^2}}}} \vspace{1ex}\\
 
\quad \,  \quad \quad \quad \quad  \ge \frac{{(T_1+{\widetilde{T}} {{(\varepsilon ))}^{\frac{1}{3}}} - (T_1+\tau)^{\frac{1}{3}}}}{{{{(4(\gamma  - 2){L^2}(M{\rho _1} + N{\rho _2}) + \frac{{{\rho _2}(N - S){L^2}}}{2})}^2}}}.
\end{array}
\end{equation}}

The second inequality in Eq. (A.\ref{eq:A92}) is due to that $\forall t \ge T_1 + \tau$, we have,
{\begin{equation}
\renewcommand{\theequation}{A.\arabic{equation}}
\label{eq:A93}
4(\gamma  - 2){L^2}(M{\rho _1} + N{\rho _2}){(t\!+\!1)^{\frac{1}{3}}} + \frac{{{\rho _2}(N \!-\! S){L^2}}}{2} \!\le\! (4(\gamma  - 2){L^2}(M{\rho _1} + N{\rho _2}) + \frac{{{\rho _2}(N \!-\! S){L^2}}}{2}){(t\!+\!1)^{\frac{1}{3}}}.
\end{equation}}

The last inequality in Eq. (A.\ref{eq:A92}) follows from the fact that  $\sum\limits_{t =T_1\!+\! \tau}^{T_1\!+\!{\widetilde{T}} (\varepsilon )} {\frac{1}{{{(t+1)^{\frac{2}{3}}}}}}  \ge (T_1+{\widetilde{T}} {(\varepsilon )})^{\frac{1}{3}} - (T_1+\tau)^{\frac{1}{3}}$.

\vspace{1ex}

Thus, plugging Eq. (A.\ref{eq:A92}) into Eq. (A.\ref{eq:A90}), we can obtain:
{\begin{equation}
\label{eq:A94}
\renewcommand{\theequation}{A.\arabic{equation}}
||\nabla {\widetilde{G}}^{T_1 + {\widetilde{T}} (\varepsilon )} |{|^2} \! \le\! \frac{{\mathop (\mathop d\limits^ -  + k_d(\tau\! -\!1))   {d_5}}}{{\sum\limits_{t =T_1\!+\! \tau}^{T_1\!+\!{\widetilde{T}} (\varepsilon )} {\frac{1}{{{a_5^t}{a_6^t}}}} }} \! \le \! \frac{{{{(4(\gamma  \!-\! 2){L^2}(M{\rho _1} \!+\! N{\rho _2}) + \frac{{{\rho _2}(N - S){L^2}}}{2})}^2}(\mathop d\limits^ -  + k_d(\tau\! -\! 1))   {d_5}}}{{(T_1+{\widetilde{T}} {{(\varepsilon )}){^{\frac{1}{3}}}} - (T_1+\tau)^{\frac{1}{3}}}}.
\end{equation}}

According to the definition of ${\widetilde{T}(\varepsilon )}$, we have:
{\begin{equation}
\label{eq:A95}
\renewcommand{\theequation}{A.\arabic{equation}}
T_1 + {\widetilde{T}} (\varepsilon ) \ge {(\frac{{4{{(4(\gamma  - 2){L^2}(M{\rho _1} + N{\rho _2}) + \frac{{{\rho _2}(N - S){L^2}}}{2})}^2}(\mathop d\limits^ -  + k_d(\tau -1))   {d_5}}}{{{\varepsilon ^2}}} + (T_1+\tau)^{\frac{1}{3}})^3}.
\end{equation}}

Combining the definition of $\nabla G^t$ and $\nabla {\widetilde{G}}^t$ with trigonometric inequality, we then get:
{\begin{equation}
\renewcommand{\theequation}{A.\arabic{equation}}
\label{eq:A96}
||\nabla G^t|| - ||\nabla {\widetilde{G}}^t|| \le ||\nabla G^t - \nabla {\widetilde{G}}^t|| \le \sqrt {\sum\limits_{l = 1}^{|{{\bf{A}}^t}|} {||{c_1^{t-1}}{\lambda _l^t}|{|^2}}  + \sum\limits_{j = 1}^N {||{c_2^{t-1}}{{\boldsymbol{\phi}}_j^t}|{|^2}} }.
\end{equation}}

Denoting constant $d_6$ as $d_6 = 4(\gamma  -  2){{L}^2}( M{\rho _1}  +  N{\rho _2})$. If $t > {(\frac{{4M{\sigma _1}^2}}{{{\rho _1}^2}} + \frac{{4N{\sigma _2}^2}}{{{\rho _2}^2}})^3}\frac{1}{{{\varepsilon ^6}}}$, then we have $\sqrt {\sum\limits_{l = 1}^{|{{\bf{A}}^t}|} {||{c_1^{t-1}}{\lambda _l^t}|{|^2}}  + \sum\limits_{j = 1}^N {||{c_2^{t-1}}{{\boldsymbol{\phi}}_j^t}|{|^2}} }  \le \frac{\varepsilon }{2}$. Combining it with Eq. (A.\ref{eq:A95}), we can conclude that there exists a
\begin{equation}
\label{eq:A97}
\renewcommand{\theequation}{A.\arabic{equation}}
T( \varepsilon ) \! \sim  \! \mathcal{O}(\max  \{ {(\frac{{4M\!{\sigma _1}^2}}{{{\rho _1}^2}}\! +\! \frac{{4N\!{\sigma _2}^2}}{{{\rho _2}^2}}\!)^3}\!\frac{1}{{{\varepsilon ^6}}}, 
{(\!\frac{{4{{{(d_6 \!+\! \frac{{{\rho _2}(N \! - \! S){{L}^2}}}{2}\!)}}^2}\! (\mathop d\limits^ -  +  k_d(\tau \! - \! 1))  {d_5}}}{{{\varepsilon ^2}}}\! +\! (T_1\!+\!\tau)^{\frac{1}{3}})^3}\}),
\end{equation}
such that $||\nabla G^t|| \le ||\nabla {\widetilde{G}}^t|| + \sqrt {\sum\limits_{l = 1}^{|{{\bf{A}}^t}|} {||{c_1^{t-1}}{\lambda _l^t}|{|^2}}  + \sum\limits_{j = 1}^N {||{c_2^{t-1}}{{\boldsymbol{\phi}}_j^t}|{|^2}} }  \le \varepsilon $, which concludes our proof.

\section{Time Efficiency Comparison}\label{appendix:time efficiency}
In a distributed communication network, the communication and computation delays of workers are inevitable. Due to differences in system configuration, communication and computation delays vary across different workers, the existence of lagging workers (\textit{i.e.}, stragglers and stale workers) is inevitable. For synchronous algorithm, it will lead to idling and
 wastage of computing resources since the master only updates the variables after receiving updates from all workers (as illustrated in Figure \ref{fig:delay}). Different from the synchronous algorithm, the asynchronous algorithm allows the master updates the variables whenever
it receives updates from a subset of workers, which is more efficient. In this section, we compare the time for our asynchronous and synchronous algorithms to return an $\varepsilon$-stationary point.

\begin{fact}
 \label{theorem 2}
Let ${\mathcal{T}_1}$ and $T_1(\varepsilon )$ denote the convergence time and iterations for the proposed asynchronous algorithm. Let ${\mathcal{T}_2}$ and $T_2(\varepsilon )$ denote the convergence time and iterations for the synchronous algorithm, we have,
\begin{equation}
\renewcommand{\theequation}{B.\arabic{equation}}
\label{eq:A102} \frac{{{\mathcal{T}_1}}}{{{\mathcal{T}_2}}} = \frac{{{T_1}(\varepsilon ) \times S }}{{{T_{2}}(\varepsilon )  \times (\frac{{\widetilde{d}}}{{\hat{d}_1}}+\frac{{\widetilde{d}}}{{\hat{d}_2}}   \cdots  + \frac{{\widetilde{d}}}{{\hat{d}_N}} )}},
\end{equation}
where ${\widetilde{d}} $ is the maximum (computation + communication) delay of all workers. 
\end{fact}

\emph{\textbf{Proof of Fact \ref{theorem 2}:}}

\setcounter{figure}{0}
\begin{figure}[t]  
\renewcommand{\thefigure}{B\arabic{figure}}
\begin{center}
\includegraphics[height=0.37\textwidth,scale=1]{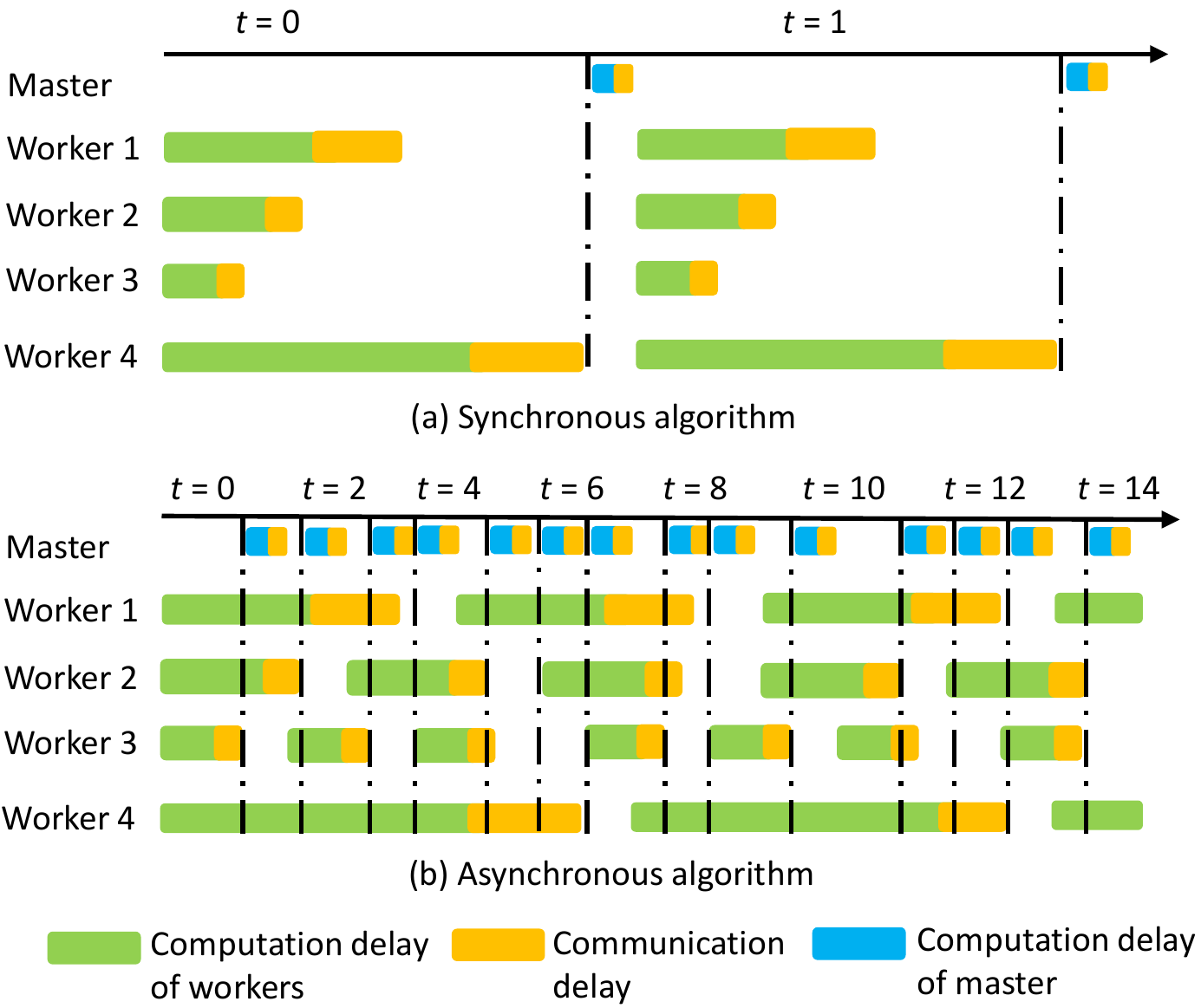} 
\caption{The illustration of synchronous and asynchronous algorithms. $t$ represents the number of iterations. In the asynchronous algorithm (at the bottom), the master begins to update its parameters after receiving the update from one worker.} 
\label{fig:delay}
\end{center}
\end{figure}

In this part, we do not consider the delays of master. Suppose that there are $N$ workers in a distributed system and the number of active workers is $S$. For brevity, we assume the delay for each work remains the same during the iteration. Let $[{\hat{d}_1},{\hat{d}_2}, \cdots {\hat{d}_N}] \in {\mathbb{R}^N}$  denote the (computation + communication) delay for $N$ workers. And we define the maximum delay of all workers as ${\widetilde{d}}$.  For the $j^{\rm{th}}$ worker, it has communicated with the master $\frac{{{\mathcal{T}_1}}}{{{d_j}}}$ times during time ${{\mathcal{T}_1}}$. Thus, for time ${{\mathcal{T}_1}}$, it satisfies that,
{\begin{equation}
\renewcommand{\theequation}{B.\arabic{equation}}
\label{eq:A104}
\frac{{{\mathcal{T}_1}}}{{{\hat{d}_1}}} + \frac{{{\mathcal{T}_1}}}{{{\hat{d}_2}}} +  \cdots  + \frac{{{\mathcal{T}_1}}}{{{\hat{d}_N}}} = {T_1}(\varepsilon ) \times S.
\end{equation}}


For the synchronous algorithm, the time  ${\mathcal{T}_2}$ needs to satisfy that:
\begin{equation}
\renewcommand{\theequation}{B.\arabic{equation}}
\label{eq:A106}
{\mathcal{T}_2} = {T_{2}}(\varepsilon ) \times  {\widetilde{d}}.
\end{equation}
Thus, we have that,
{\begin{equation}
\renewcommand{\theequation}{B.\arabic{equation}}
\label{eq:A107} \frac{{{\mathcal{T}_1}}}{{{\mathcal{T}_2}}} = \frac{{{T_1}(\varepsilon ) \times S }}{{{T_{2}}(\varepsilon ) \times (\frac{{\widetilde{d}}}{{\hat{d}_1}}+\frac{{\widetilde{d}}}{{\hat{d}_2}}   \cdots  + \frac{{\widetilde{d}}}{{\hat{d}_N}} )}}.
\end{equation}}

For the special case that the asynchronous algorithm degrades to the synchronous algorithm, \textit{i.e.}, the master is required to update its parameters only after receiving the updates from all workers, we have $S = N$, ${T_1}(\varepsilon ) = {T_2}(\varepsilon )$ and ${\hat{d}_j} = {\widetilde{d}}  ,\forall j = 1, \cdots ,N$ since all workers are required to wait the slowest worker. Back to Eq. (B.\ref{eq:A107}), we can obtain that $\frac{{\mathcal{T}_1}}{{\mathcal{T}_2}} = 1$.

\section{Experiments}
\label{appendix:experiment}
\setcounter{figure}{0}
\begin{figure}[tbp]
\renewcommand{\thefigure}{C\arabic{figure}}
\centering    
 \subfigure[Clean images whose labels are T-shirt.] 
{\begin{minipage}[t]{0.5\linewidth}
	\centering      
	\includegraphics[scale=0.5]{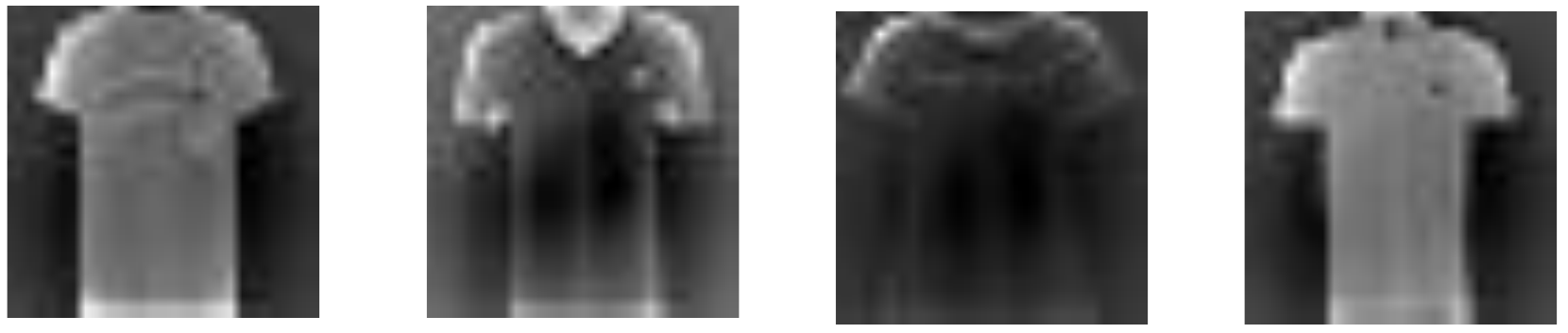}   
	\end{minipage}}
\subfigure[Attacked images whose target labels are Pullover.] 
{\begin{minipage}[t]{0.5\linewidth}
	\centering      
	\includegraphics[scale=0.5]{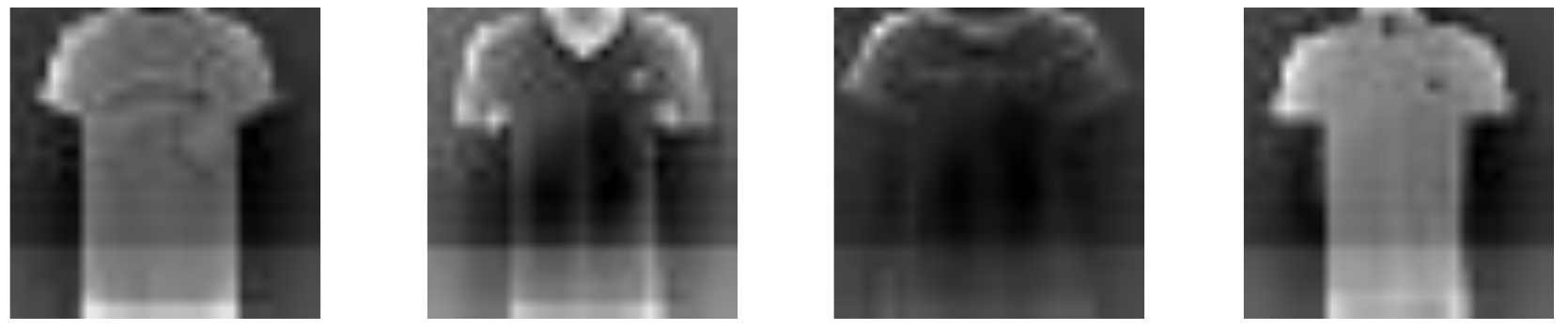}   
	\end{minipage}}
\caption{Backdoor attacks on Fashion MNIST dataset. Through adding triggers on local patch of clean images, the attacked images are misclassified as the target labels.} 
\label{fig:backdoor_fashion}  

\end{figure}

In this section, we present the detailed results of our experiments. We first give a detailed description of the datasets and baseline methods used in our experiments. 

\subsection{Datasets and Baseline Methods}

\setcounter{table}{0}

\renewcommand\arraystretch{1.5}
\renewcommand\tabcolsep{15pt}
\begin{table}[t]
\renewcommand{\thetable}{C\arabic{table}}
\caption{The number of workers and categories of datasets}
\centering
\scalebox{0.85}{
\begin{tabular}{lcccc}
\toprule
    & SHL     & Person Activity         & SC-MA  &   Fashion MNIST      \\ \hline
Number of workers & 6 & 5 & 15   &  3 \\ 
Number of categories & 8 & 11 & 7         &  3 \\ 
\bottomrule  
\end{tabular}}
\label{tab:dataset}
\end{table}

\renewcommand\arraystretch{1.5}
\renewcommand\tabcolsep{15pt}
\begin{table}[t]
\renewcommand{\thetable}{C\arabic{table}}
\caption{Model structure that used for SHL dataset.}
\centering
\scalebox{0.85}{
\begin{tabular}{llcc}
\toprule
No.    & Layer type     & Number of neurons          & Activation        \\ \hline
1 & Fully-connected & 96 & ReLU\\ 
2 & Fully-connected & 48 & ReLU\\ 
3 & Fully-connected & 24 & ReLU\\ 
4 & Output & 8 & Softmax\\
\bottomrule  
\end{tabular}}
\label{model:SHL}
\end{table}

In this section, we provide a detailed introduction to datasets and baseline methods. The number of workers and categories of every dataset are summarized in Table \ref{tab:dataset}.

\noindent \textbf{Datasets}:

\begin{enumerate}
    \item \textbf{SHL dataset}: The SHL dataset was collected using four cellphones on four body locations where people usually carry cellphones. The SHL dataset provides multimodal locomotion and transportation data collected in real-world settings using eight various modes of transportation. We separated the data into six workers with varied proportions based on the four body locations of smartphones to imitate the different tendencies of workers (users) in positioning cellphones.
    
    \item \textbf{Person Activity dataset}: Data contains recordings of five participants performing eleven different activities. Each participant wears four sensors in four different body locations (ankle left, ankle right, belt, and chest) while performing the activities. Each participant corresponds to one worker in the experiment.
    
    \item \textbf{Single Chest-Mounted Accelerometer dataset}: Data was collected from fifteen participants engaged in seven distinct activities. Each participant (worker) wears an accelerometer mounted on the chest.
    
    \item \textbf{Fashion MNIST}: Similar to MNIST, Fashion MNIST is a dataset where images are grouped into ten categories of clothing. The subset of the data labeled with Pullover, Shirt and T-shirt are extracted as three workers and each worker consists of one class of clothing.
    
\end{enumerate}

\noindent \textbf{Baseline Methods}:

\begin{enumerate}

\item \textbf{Ind}${}_j$: It learns the model from an individual worker $j$.

\item \textbf{Mix}{$\rm{_{Even}}$}: It learns the model from all workers with even weights using the proposed distributed algorithm. 

\item \textbf{FedAvg}: It learns the model from all workers with even weights. It aggregates the local model parameters from workers through using model averaging.

\item \textbf{AFL}: It aims to address the fairness issues in federated learning. AFL adopts the strategy that alternately update the model parameters and the weight of each worker through alternating projected gradient descent/ascent.

\item \textbf{DRFA-Prox}: It aims to mitigate the data heterogeneity issue in federated learning. Compared with AFL, it is communication-efficient which requires fewer communication rounds. Moreover, it leverages the prior distribution and introduces it as a regularizer in the objective function.

\item \textbf{ASPIRE-EASE(-)}: The proposed ASPIRE-EASE without asynchronous setting.

\item \textbf{ASPIRE-CP}: The proposed ASPIRE with cutting plane method.

\item \textbf{ASPIRE-EASE$_{\rm{per}}$}: The proposed ASPIRE-EASE with periodic communication.

\end{enumerate}

\renewcommand\arraystretch{1.5}
\renewcommand\tabcolsep{15pt}
\begin{table}[t]
\renewcommand{\thetable}{C\arabic{table}}
\caption{Model structure that used for Peson Activity dataset.}
\centering
\scalebox{0.85}{
\begin{tabular}{llcc}
\toprule
No.    & Layer type     & Number of neurons          & Activation        \\ \hline
1 & Fully-connected & 64 & ReLU\\ 
2 & Fully-connected & 32 & ReLU\\ 
3 & Fully-connected & 16 & ReLU\\ 
4 & Output & 11 & Softmax\\
\bottomrule  
\end{tabular}}
\label{model:person activity}
\end{table}

\renewcommand\arraystretch{1.5}
\renewcommand\tabcolsep{15pt}
\begin{table}[t]
\renewcommand{\thetable}{C\arabic{table}}
\caption{Model structure that used for SC-MA dataset.}
\centering
\scalebox{0.85}{
\begin{tabular}{llcc}
\toprule
No.    & Layer type     & Number of neurons          & Activation        \\ \hline
1 & Fully-connected & 32 & ReLU\\ 
2 & Fully-connected & 16 & ReLU\\ 
3 & Output & 7 & Softmax\\
\bottomrule  
\end{tabular}}
\label{model:SCMA}
\end{table}


\subsection{Training Details}
\label{training details}
In our empirical studies, since the downstream tasks are multi-class classification, the cross entropy loss is used on each worker (\textit{i.e.}, ${\mathcal{L}_j}( \cdot ),\forall j$). For SHL, Person Activity and SM-AC datasets, we adopt the deep multilayer perceptron~\citep{wang2017time} as the base model. Specifically, we exhibit the model structures that are used for SHL, Person Activity and SM-AC datasets in Table \ref{model:SHL}, Table \ref{model:person activity} and Table \ref{model:SCMA}. And we use the same logistic regression model as in \citep{mohri2019agnostic,deng2021distributionally} for the Fashion MNIST dataset. In the experiments, we use the SGD optimizer for model training, and we implement our model with PyTorch and conduct all the experiments on a server with two TITAN V GPUs.

\renewcommand{\thefigure}{C\arabic{figure}}
	\begin{figure*}[h]
\centering    
\subfigure[SHL] 
{\begin{minipage}[t]{0.36\linewidth}
	\centering      
	\includegraphics[scale=0.32]{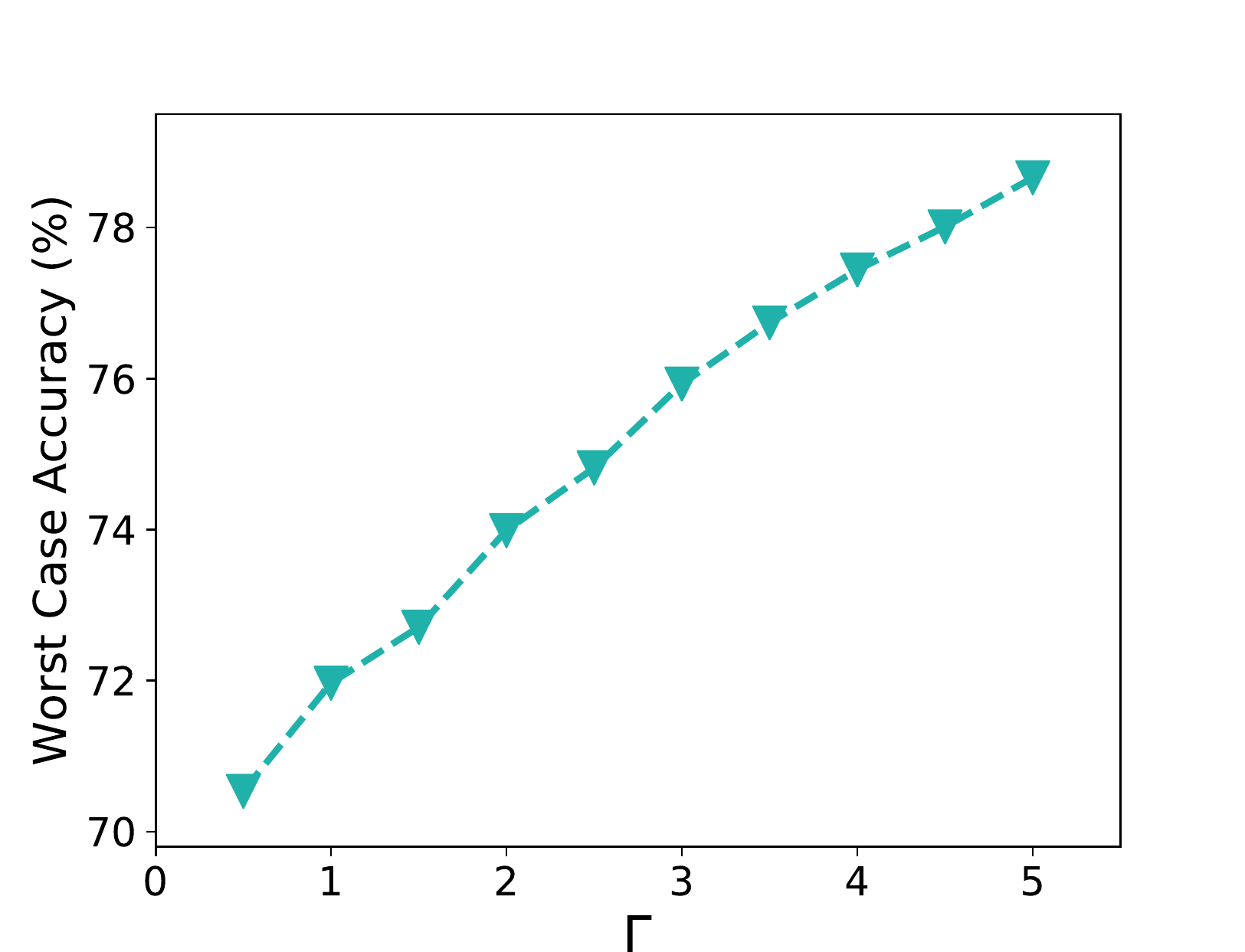}   
	\end{minipage}}
\subfigure[Fashion MNIST] 
{\begin{minipage}[t]{0.36\linewidth}
	\centering      
	\includegraphics[scale=0.32]{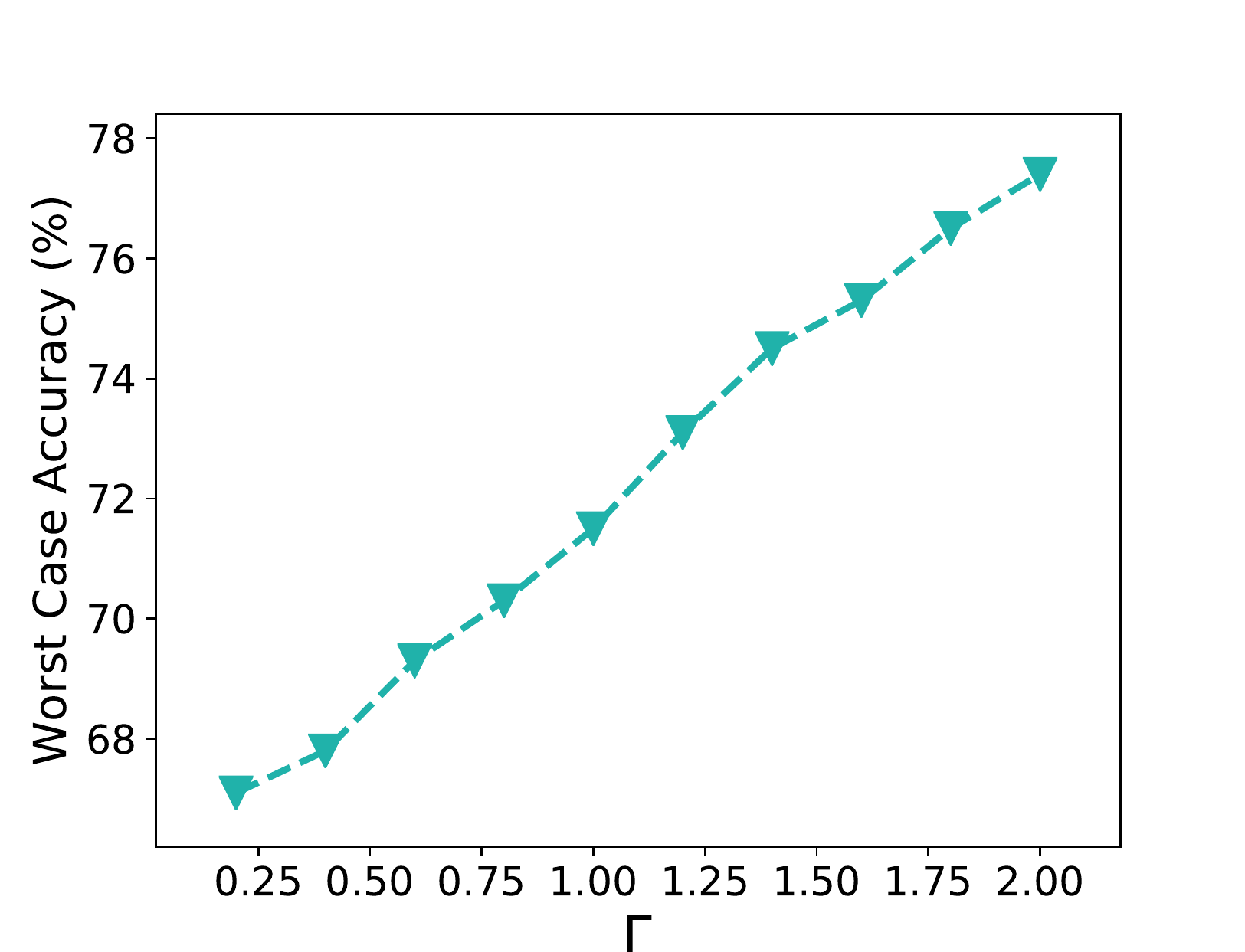}   
	\end{minipage}}
\caption{$\Gamma$ control the degree of robustness (worst case performance in the problem) on (a) SHL, (b) Fashion MNIST datasets.} 
\label{appendix:fig:gamma} 
\end{figure*}

\renewcommand{\thefigure}{C\arabic{figure}}
	\begin{figure*}[h]
\centering    
\setlength{\abovecaptionskip}{-0.1cm} 
\subfigure[SHL] 
{\begin{minipage}[t]{0.36\linewidth}
	\centering      
	\includegraphics[scale=0.32]{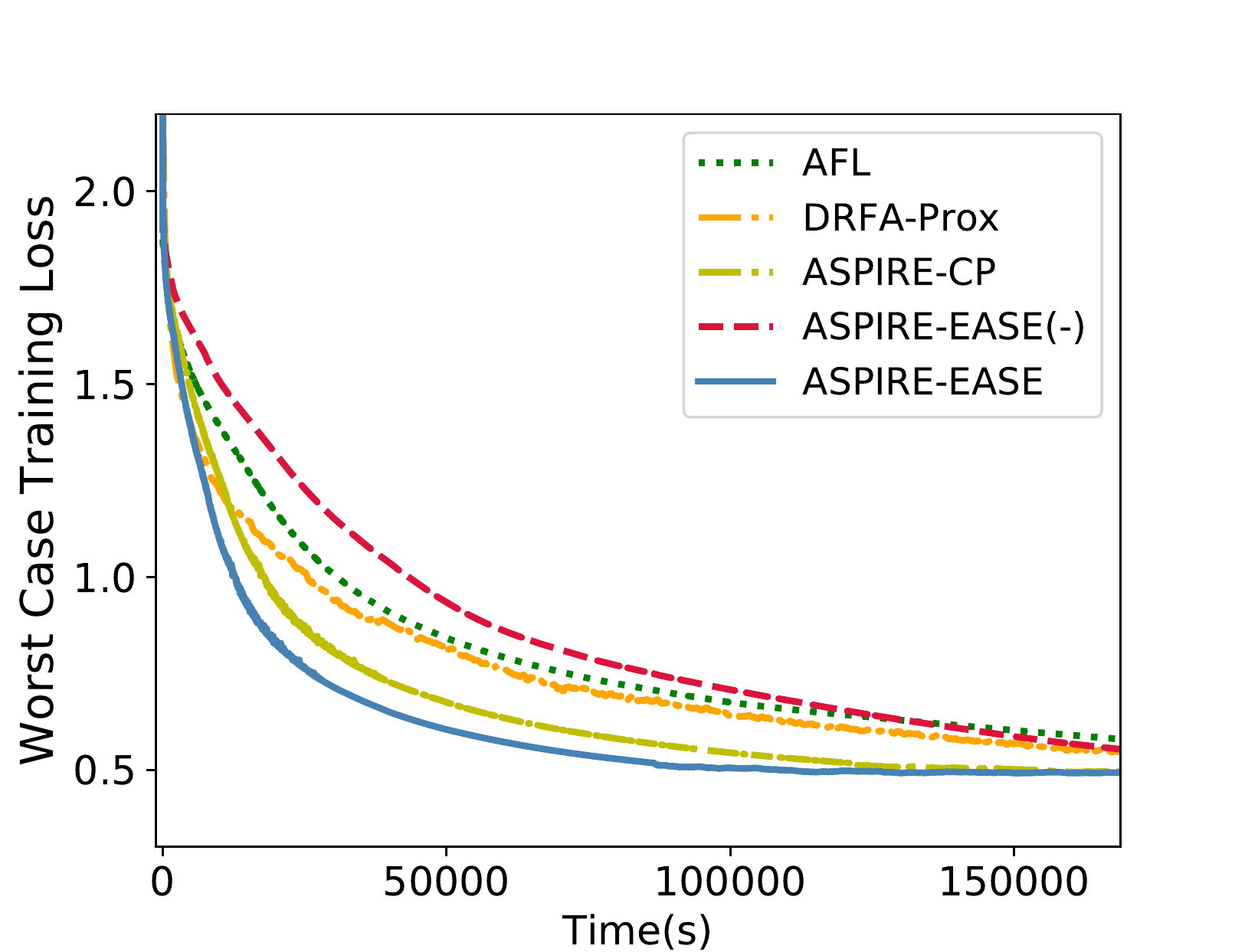}   
	\end{minipage}}
\subfigure[Fashion MNIST] 
{\begin{minipage}[t]{0.36\linewidth}
	\centering      
	\includegraphics[scale=0.32]{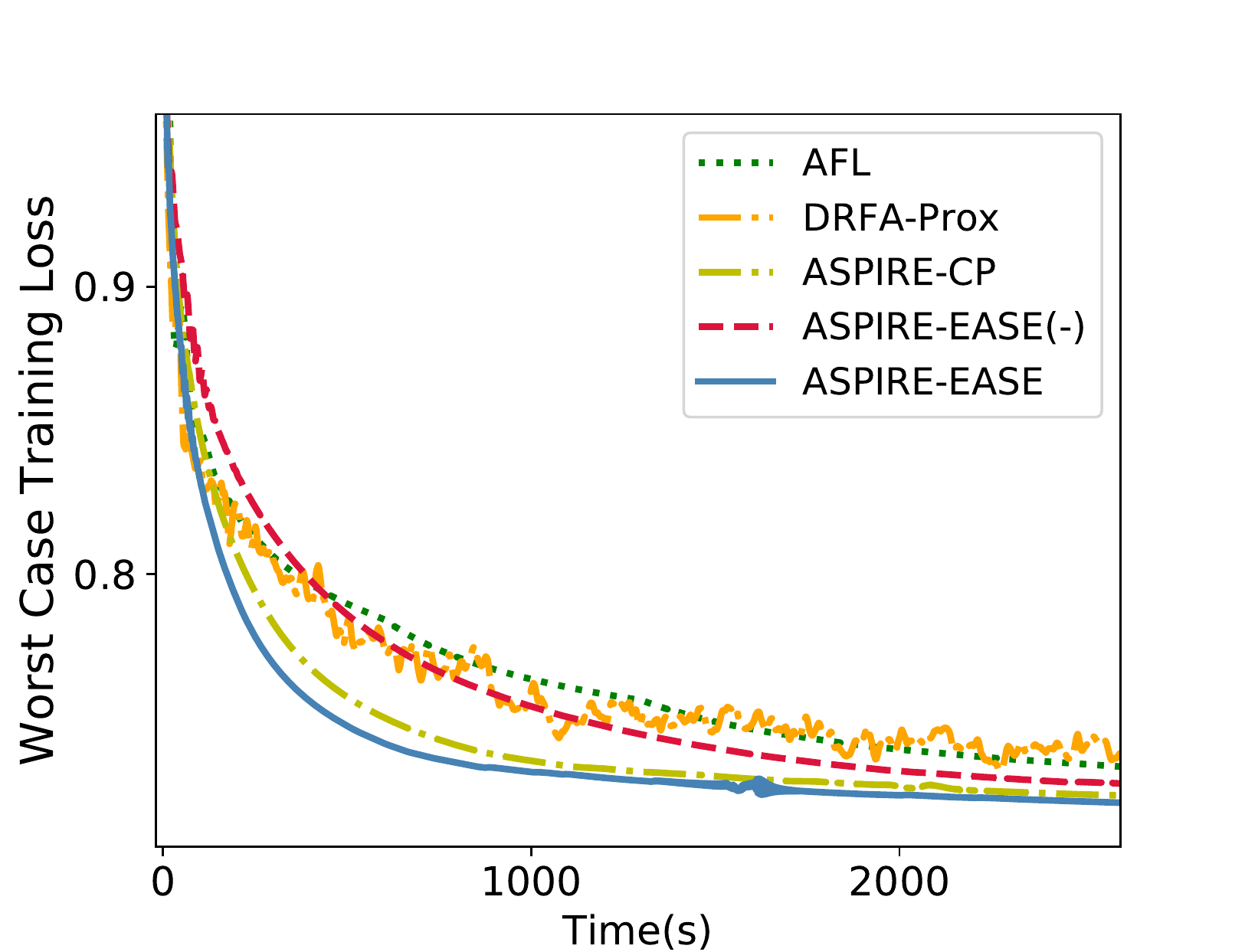}   
	\end{minipage}}
\caption{Comparison of the convergence time on worst case worker on  (a) SHL, (b) Fashion MNIST datasets.} 
\label{appendix:fig:time efficientcy} 
\end{figure*}

\renewcommand{\thefigure}{C\arabic{figure}}
	\begin{figure*}[h]
\centering    
\setlength{\abovecaptionskip}{-0.1cm} 
 \subfigure[SHL] 
{\begin{minipage}[t]{0.36\linewidth}
	\centering          
	\includegraphics[scale=0.32]{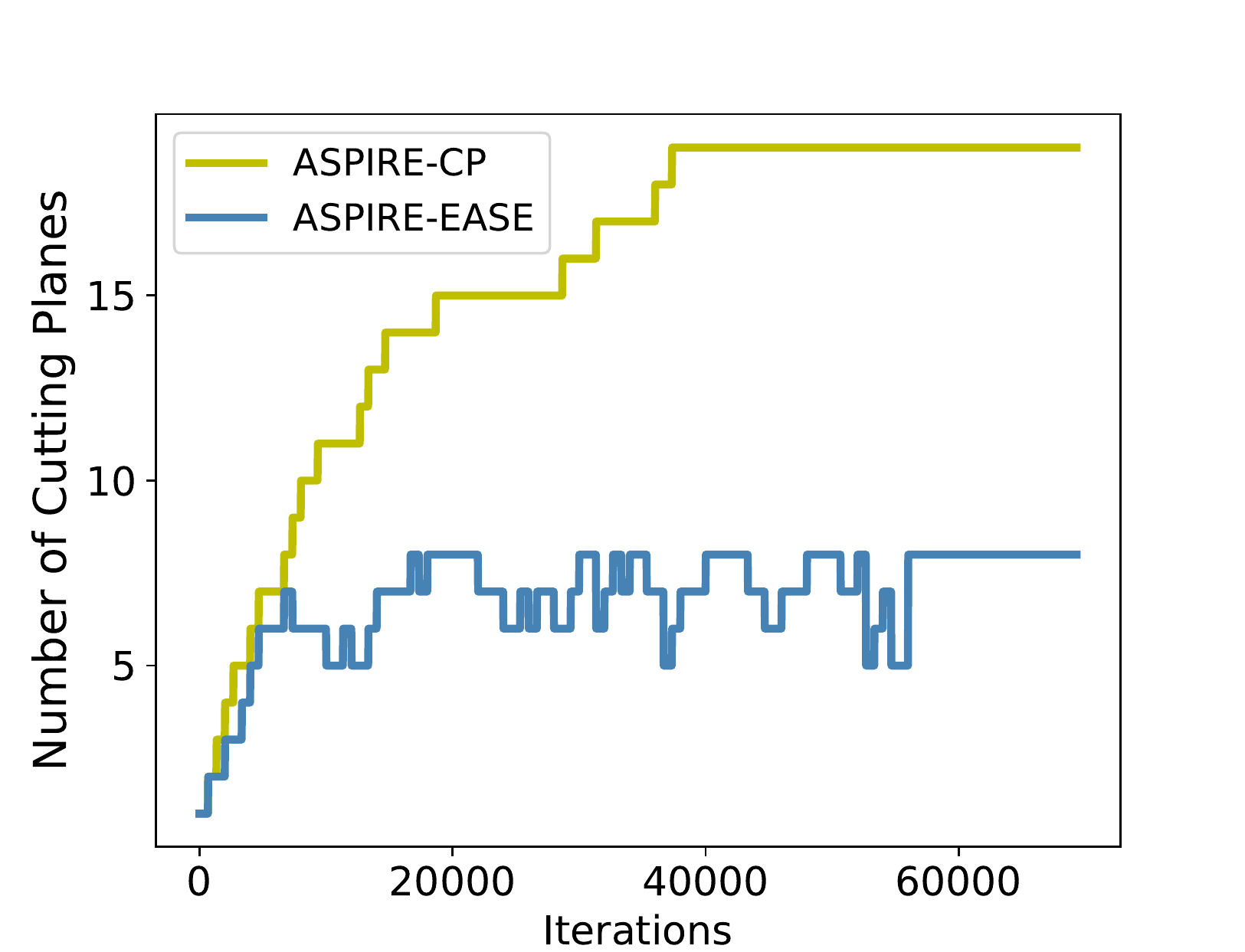}   
	\end{minipage}}
\subfigure[Fashion MNIST] 
{\begin{minipage}[t]{0.36\linewidth}
	\centering      
	\includegraphics[scale=0.32]{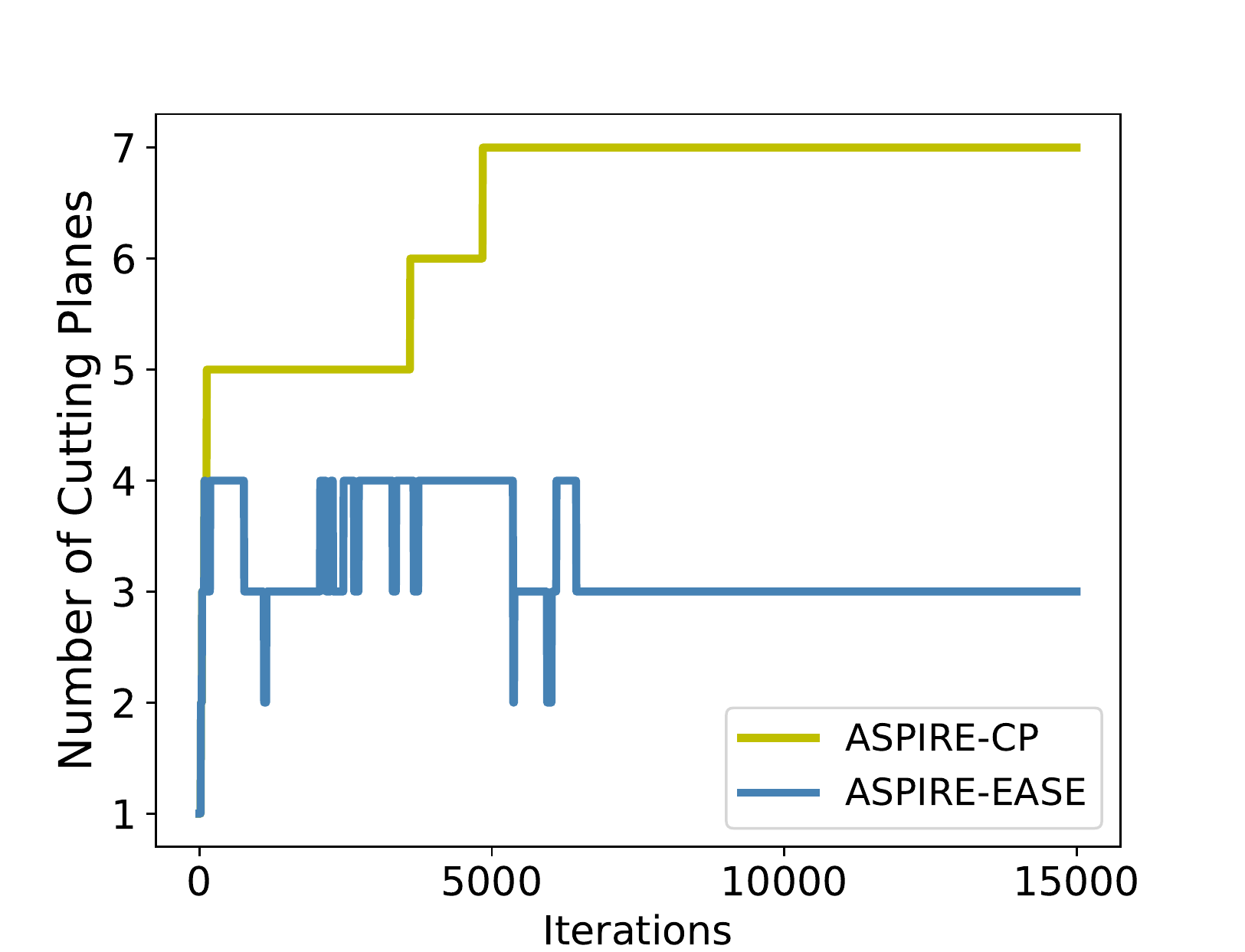}   
	\end{minipage}}
\caption{Comparison of ASPIRE-CP and ASPIRE-EASE regarding the number of cutting planes on  (a) SHL, (b) Fashion MNIST datasets. ASPIRE-CP represents ASPIRE with cutting plane method.} 
\label{appendix:fig:cutting plane}  
\end{figure*}

\subsection{Additional Results}
We first show the detailed experiment settings about robustness against malicious attacks. We conduct experiments in the setting where there are malicious workers which attempt to mislead the model training process. The backdoor attack  \citep{bagdasaryan2020backdoor,wang2019neural} is adopted in the experiment which aims to bury the backdoor during the training phase of the model. The buried backdoor will be activated by the preset trigger. When the backdoor is not activated, the attacked model performs normally to other local models. When the backdoor is activated, the output of the attack model is misled as the target label which is pre-specified by the attacker. In the experiment, one worker is chosen as the malicious worker.  We add triggers to a small part of the data and change their primal labels to target labels (\textit{e.g.}, triggers are added on the local patch of clean images on the Fashion MNIST dataset, which are shown in Figure \ref{fig:backdoor_fashion}). Furthermore, the malicious worker can purposefully raise the training loss to mislead the master. To evaluate the model's robustness against malicious attacks, following \citep{dai2019backdoor}, we calculate the success attack rate of the backdoor attacks. The success attack rate can be calculated by checking how many instances in the backdoor dataset can be misled into the target labels. The lower success attack rate indicates better robustness against backdoor attacks. The success attack rates of different models on three datasets are reported in Table 2. In Table 2, we observe that AFL can be attacked easily since it could assign higher weights to malicious workers. Compared to AFL, FedAvg and ${\rm{Mix}}{\rm{_{Even}}}$ achieve relatively lower success attack rates since they assign equal weights to the malicious workers and other workers. DRFA-Prox can achieve even lower success attack rates since it can leverage the prior distribution to assign lower weights for malicious workers.  The proposed ASPIRE-EASE achieves the lowest success attack rates since it can leverage the prior distribution more effectively. Specifically, it will assign lower weights to malicious workers with tight theoretical guarantees.

We also report additional experiment results on SHL and Fashion MNIST datasets. We first show that the proposed ASPIRE-EASE can flexibly control the level of robustness by adjusting $\Gamma$, which is presented in Figure \ref{appendix:fig:gamma}. It is seen that the robustness of ASPIRE-EASE can be gradually enhanced when $\Gamma$ increases. Next, the comparison of convergence speed by considering different communication and computation delays for each worker is exhibited in Figure \ref{appendix:fig:time efficientcy}. We can observe that the proposed ASPIRE-EASE is generally the most efficient since the ASPIRE is an asynchronous algorithm and the proposed EASE is effective. Finally, to further demonstrate the efficiency of EASE, we compare  ASPIRE-EASE with ASPIRE-CP concerning the number of cutting planes used during the training. As shown in Theorem 1, a smaller number of cutting planes (which corresponds to a smaller $M$) will need fewer iterations to achieve convergence.  In Figure \ref{appendix:fig:cutting plane}, we can see that ASPIRE-EASE uses fewer cutting planes and thus is more efficient.

\section{Solve PD-DRO in Centralized Manner}
Considering to solve the PD-DRO problem in Eq. (\ref{eq:3}) in centralized manner, we can rewrite the problem in Eq. (\ref{eq:3}) as:
\begin{equation}
\renewcommand{\theequation}{D.\arabic{equation}}
\label{eq:b102}
\mathop {{\rm{min}}}\limits_{\boldsymbol{w}\in{{\boldsymbol{\mathcal{W}}}}} \mathop {{\rm{max}}} \limits_{{\bf{p}}  \in \boldsymbol{\mathcal{P}}} \sum\limits_{j = 1}^N {{p_j}{f_j}({\boldsymbol{w}})} 
\end{equation}
where ${\boldsymbol{w}}\! \in \! {\mathbb{R}^p}$ is the model parameter. Utilizing the cutting plane method, we can obtain the approximate problem of Eq. (D.\ref{eq:b102}),
{\renewcommand{\theequation}{D.\arabic{equation}}
\begin{align}
\label{eq:b103}
\mathop {{\rm{min}}}\limits_{\boldsymbol{w}\in{{\boldsymbol{\mathcal{W}}}}, h\in{{\boldsymbol{\mathcal{H}}}}} \quad \quad h &\\
{\rm{s.t.}} \; \sum\limits_{j = 1}^N {(\mathop p\limits^ -   + {a_{l,j}}){{f}_j}({{\boldsymbol{w}}})} & - h \le 0,{\rm{     }}\forall {\boldsymbol{a}_l} \in {{\bf{A}}^t},\nonumber \\
{\rm{var.}}\quad \quad  \boldsymbol{w}, h&. \nonumber
\end{align}}

Thus, the Lagrangian function of Eq. (\ref{eq:b103}) can be written as:
\begin{equation}
\renewcommand{\theequation}{D.\arabic{equation}}
\label{eq:B104}
 {L_p}({\boldsymbol{w}},h,\{ {\lambda _l}\}) = h \! + \! \sum\limits_{l = 1}^{|{{\bf{A}}^t}|} {{\lambda _l}(\sum\limits_{j = 1}^N {(\mathop p\limits^ -   + {a_{l,j}}){f_j}({\boldsymbol{w}})}  \!-\! h)}.
\end{equation}
Following \citep{xu2020unified}, the regularized version of (D.\ref{eq:B104}) is employed to update all variables as follows,
\begin{equation}
\renewcommand{\theequation}{D.\arabic{equation}}
\label{eq:B105}
 {\widetilde{L}_p}({\boldsymbol{w}},h,\{ {\lambda _l}\}) \!  = \! h +  \sum\limits_{l = 1}^{|{{\bf{A}}^t}|} {{\lambda _l}(\sum\limits_{j = 1}^N {(\mathop p\limits^ -   + {a_{l,j}}){f_j}({\boldsymbol{w}})}  \!-\! h)} - \sum\limits_{l = 1}^{|{{\bf{A}}^t}|} {\frac{{{c_1^t}}}{2}||{\lambda _l}|{|^2}},
\end{equation}
where ${c_1^t}$ denotes the regularization term in $(t+1)^{\rm{th}}$ iteration. To avoid enumerating the whole dataset, the mini-batch loss  ${\hat{f}_j}({\boldsymbol{w}})  =  \sum\limits_{i = 1}^m {\frac{1}{m}{{\cal L}_j}({\bf{x}}_j^i,y_j^i;{\boldsymbol{w}})}$ can be used, where $m$ is the mini-batch size. It is evident that $\mathbb{E}[{\hat{f}_j}({\boldsymbol{w}})]={f_j}({\boldsymbol{w}})$ and $\mathbb{E}[\nabla {\hat{f}_j}({\boldsymbol{w}})]=\nabla{f_j}({\boldsymbol{w}})$. The centralized algorithm, which aims to solve problem in Eq. (\ref{eq:3}) in centralized manner,  proceeds as follows in $(t+1)^{\rm{th}}$ iteration:
\begin{enumerate}
\item Updating the model parameter ${\boldsymbol{w}}$ as follows,
\begin{equation}
\renewcommand{\theequation}{D.\arabic{equation}}
\label{eq:B106}
{\boldsymbol{w}}^{t+1} = {\mathcal{P}_{{\boldsymbol{\mathcal{W}}}}}({\boldsymbol{w}}^{t} - {\eta _{\boldsymbol{w}}^t}{\nabla _{{\boldsymbol{w}}}}{ \widetilde{L}_p} ({\boldsymbol{w}}^{t},h^{t},\{ {\lambda _l^{t}}\})),
\end{equation}
where ${\eta _{\boldsymbol{w}}^t}$ represents the step-size and $\mathcal{P}_{{\boldsymbol{\mathcal{W}}}}$ represents the projection onto the convex set ${{\boldsymbol{\mathcal{W}}}}$.

\item Updating the additional variable $h$ as follows,
\begin{equation}
\renewcommand{\theequation}{D.\arabic{equation}}
\label{eq:B107}
h^{t+1} = {\mathcal{P}_{{\boldsymbol{\mathcal{H}}}}}(h^{t} - {\eta _h^t}{\nabla _{h}}{ \widetilde{L}_p} ({\boldsymbol{w}}^{t+1},h^{t},\{ {\lambda _l^{t}}\})),
\end{equation}
where ${\eta _h^t}$ represents the step-size and  ${\mathcal{P}_{{\boldsymbol{\mathcal{H}}}}}$ represents the projection onto the convex set ${{\boldsymbol{\mathcal{H}}}}$.

\item Updating the dual variable ${\lambda _l}$ as follows,
\begin{equation}
\renewcommand{\theequation}{D.\arabic{equation}}
\label{eq:B108}
{\lambda _l^{t+1}} = {\mathcal{P}_{\bf{\Lambda}} }({\lambda _l^{t}} + {\rho _1}{\nabla _{\lambda _l}}{ \widetilde{L}_p} ({\boldsymbol{w}}^{t+1},h^{t+1},\{ {\lambda _l^{t}}\})), \; l=1,\!\cdots\!, |{{\bf{A}}^t}|,
\end{equation}
where $\rho_1$ represents the step-size and  ${\mathcal{P}_{\bf{\Lambda}}}$ represents the projection onto the convex set ${\bf{\Lambda}}$.
\end{enumerate}

Then, during $T_1$ iterations, EASE is utilized to update the set ${{\rm \bf{A}}^{t + 1}}$ every $k$ iterations.  

\setcounter{definition}{0}
\renewcommand{\thedefinition}{D.\arabic{definition}}
\begin{definition}
\label{definition:B1}
Following \citep{xu2020unified,lu2020hybrid,xu2021zeroth}, the \textit{stationarity} \textit{gap} at $t^{{th}}$ iteration is defined as,
{\begin{equation}
\renewcommand{\theequation}{D.\arabic{equation}}
\label{eq:B109}
\nabla G^t = \left[ \begin{array}{l}
\frac{1}{{{\eta _{\boldsymbol{w}}^t}}}({\boldsymbol{w}}^{t} - {\mathcal{P}_{{\boldsymbol{\mathcal{W}}}}}({\boldsymbol{w}}^{t} - {\eta _{\boldsymbol{w}}^t}{\nabla _{\boldsymbol{w}}}{L_p}({\boldsymbol{w}}^{t},h^{t},\{ {\lambda _l^{t}}\} {\rm{))}}) \vspace{0.5ex}\\
\frac{1}{{{\eta _h^t}}}(h^{t} - {\mathcal{P}_{{\boldsymbol{\mathcal{H}}}}}(h^{t} - {\eta _h^t}{\nabla _h}{L_p}({\boldsymbol{w}}^{t},h^{t},\{ {\lambda _l^{t}}\} {\rm{))}}) \vspace{0.5ex}\\
 \{ \frac{1}{{{\rho _1}}}({\lambda _l^{t}} - {\mathcal{P}_{\bf{\Lambda}} }({\lambda _l^{t}} \! +\! {\rho _1}{\nabla _{{\lambda _l}}}{L_p}({\boldsymbol{w}}^{t},h^{t},\{ {\lambda _l^{t}}\} {\rm{))}})\}
\end{array} \right].
\end{equation}}

And we also define:
{\begin{equation}
\renewcommand{\theequation}{D.\arabic{equation}}
\label{eq:B110}
\begin{array}{l}
{(\nabla G^t)_{\boldsymbol{w}}} = \frac{1}{{{\eta _{\boldsymbol{w}}^t}}}({\boldsymbol{w}}^{t} - {\mathcal{P}_{{\boldsymbol{\mathcal{W}}}}}({\boldsymbol{w}}^{t} - {\eta _{\boldsymbol{w}}^t}{\nabla _{\boldsymbol{w}}}{L_p}({\boldsymbol{w}}^{t},h^{t},\{ {\lambda _l^{t}}\} {\rm{))}}), \vspace{0.5ex}\\
{(\nabla G^t)_h} = \frac{1}{{{\eta _h^t}}}(h^{t} - {\mathcal{P}_{{\boldsymbol{\mathcal{H}}}}}(h^{t} - {\eta _h^t}{\nabla _h}{L_p}({\boldsymbol{w}}^{t},h^{t},\{ {\lambda _l^{t}}\} {\rm{))}}), \vspace{0.5ex}\\
{(\nabla G^t)_{{\lambda _l}}} = \frac{1}{{{\rho _1}}}({\lambda _l^{t}} - {\mathcal{P}_{\bf{\Lambda}} }({\lambda _l^{t}} \! +\! {\rho _1}{\nabla _{{\lambda _l}}}{L_p}({\boldsymbol{w}}^{t},h^{t},\{ {\lambda _l^{t}}\} {\rm{))}}).
\end{array}
\end{equation}}

It follows that:
{\begin{equation}
\renewcommand{\theequation}{D.\arabic{equation}}
\label{eq:B112}
||\nabla G^t|{|^2} = {||{{(\nabla G^t)}_{{{\boldsymbol{w}}}}}|{|^2}}  \! +\! ||{(\nabla G^t)_h}|{|^2} \! +\! \sum\limits_{l = 1}^{|{{\bf{A}}^t}|} {||{{(\nabla G^t)}_{{\lambda _l}}}|{|^2}}. 
\end{equation}}
\end{definition}

\begin{definition}
\label{definition:B2}
At $t^{{th}}$ iteration, the \textit{stationarity} \textit{gap} \textit{w.r.t} ${\widetilde{L}_p}( {\boldsymbol{w}},h,\{ {\lambda _l}\} {\rm{)}}$ is defined as:
{\begin{equation}
\renewcommand{\theequation}{D.\arabic{equation}}
\label{eq:B112}
\nabla \widetilde{G}^t = \left[ \begin{array}{l}
\frac{1}{{{\eta _{\boldsymbol{w}}^t}}}({\boldsymbol{w}}^{t} - {\mathcal{P}_{{\boldsymbol{\mathcal{W}}}}}({\boldsymbol{w}}^{t} - {\eta _{\boldsymbol{w}}^t}{\nabla _{\boldsymbol{w}}}{\widetilde{L}_p}({\boldsymbol{w}}^{t},h^{t},\{ {\lambda _l^{t}}\} {\rm{))}}) \vspace{0.5ex}\\
\frac{1}{{{\eta _h^t}}}(h^{t} - {\mathcal{P}_{{\boldsymbol{\mathcal{H}}}}}(h^{t} - {\eta _h^t}{\nabla _h}{\widetilde{L}_p}({\boldsymbol{w}}^{t},h^{t},\{ {\lambda _l^{t}}\} {\rm{))}}) \vspace{0.5ex}\\
 \{ \frac{1}{{{\rho _1}}}({\lambda _l^{t}} - {\mathcal{P}_{\bf{\Lambda}} }({\lambda _l^{t}} \! +\! {\rho _1}{\nabla _{{\lambda _l}}}{\widetilde{L}_p}({\boldsymbol{w}}^{t},h^{t},\{ {\lambda _l^{t}}\} {\rm{))}})\}
\end{array} \right].
\end{equation}}

And we also define:
{\begin{equation}
\renewcommand{\theequation}{D.\arabic{equation}}
\label{eq:B113}
\begin{array}{l}
{(\nabla \widetilde{G}^t)_{\boldsymbol{w}}} = \frac{1}{{{\eta _{\boldsymbol{w}}^t}}}({\boldsymbol{w}}^{t} - {\mathcal{P}_{{\boldsymbol{\mathcal{W}}}}}({\boldsymbol{w}}^{t} - {\eta _{\boldsymbol{w}}^t}{\nabla _{\boldsymbol{w}}}{\widetilde{L}_p}({\boldsymbol{w}}^{t},h^{t},\{ {\lambda _l^{t}}\} {\rm{))}}), \vspace{0.5ex}\\
{(\nabla \widetilde{G}^t)_h} = \frac{1}{{{\eta _h^t}}}(h^{t} - {\mathcal{P}_{{\boldsymbol{\mathcal{H}}}}}(h^{t} - {\eta _h^t}{\nabla _h}{\widetilde{L}_p}({\boldsymbol{w}}^{t},h^{t},\{ {\lambda _l^{t}}\} {\rm{))}}), \vspace{0.5ex}\\
{(\nabla \widetilde{G}^t)_{{\lambda _l}}} = \frac{1}{{{\rho _1}}}({\lambda _l^{t}} - {\mathcal{P}_{\bf{\Lambda}} }({\lambda _l^{t}} \! +\! {\rho _1}{\nabla _{{\lambda _l}}}{\widetilde{L}_p}({\boldsymbol{w}}^{t},h^{t},\{ {\lambda _l^{t}}\} {\rm{))}}).
\end{array}
\end{equation}}

It follows that:
{\begin{equation}
\renewcommand{\theequation}{D.\arabic{equation}}
\label{eq:B114}
||\nabla \widetilde{G}^t|{|^2} = {||{{(\nabla \widetilde{G}^t)}_{{{\boldsymbol{w}}}}}|{|^2}}  \! +\! ||{(\nabla \widetilde{G}^t)_h}|{|^2} \! +\! \sum\limits_{l = 1}^{|{{\bf{A}}^t}|} {||{{(\nabla \widetilde{G}^t)}_{{\lambda _l}}}|{|^2}}. 
\end{equation}}
\end{definition}

\setcounter{assumption}{0}
\renewcommand{\theassumption}{D.\arabic{assumption}}
\begin{assumption}\label{assumption:B1}
$L_p$ has Lipschitz continuous gradients. We assume that there exists $L>0$ satisfying that,

\centerline{$\begin{array}{l}
||{\nabla \! _\theta }{L_p}({\boldsymbol{w}},h,\{ {\lambda _l}\}) \! - \! {\nabla \! _\theta }{L_p}({\hat{\boldsymbol{w}}},\hat{h},\{ {\hat{\lambda} _l}\})||  \le L||[ {\boldsymbol{w}} \! - \!{\hat{\boldsymbol{w}}}  ;h \! - \! \hat{h} ; {\boldsymbol{\lambda} _{\rm{cat}}} \! - \! {{\hat{\boldsymbol{\lambda}} _{\rm{cat}}}}  ]||,
\end{array}$}

 where $\theta \!  \in \! \{  {\boldsymbol{w}},h,\{ {\lambda _l}\} \} $, $[;]$ represents the concatenation and ${\boldsymbol{\lambda} _{\rm{cat}}} \! - \! {{\hat{\boldsymbol{\lambda}} _{\rm{cat}}}} = [{\lambda_1} \! - \! {\hat{\lambda}_1};\cdots;{\lambda}_{|{{\bf{A}}^t}|} \! - \! {\hat{\lambda}_{|{{\bf{A}}^t}|}}]\in \! {\mathbb{R}^{|{{\bf{A}}^t}|}}$.
\end{assumption}

\setcounter{setting}{0}
\renewcommand{\thesetting}{D.\arabic{setting}}
\begin{setting}\label{assumption:B3}
$|{{\bf{A}}^t}| \le M,{\rm{    }}\forall t$, \textit{i.e.}, an upper bound is set for the number of cutting planes.
\end{setting}

\begin{setting}\label{assumption:B2}
${c_1^t} = \frac{1}{{{\rho _1}{(t+1)^{\frac{1}{4}}}}} \ge \underline{c}_1$ is nonnegative non-increasing sequence, where $\underline{c}_1>0$  meets ${\underline{c}_1}^2 \le \frac{{{\varepsilon ^2}}}{4M}$.
\end{setting}

\setcounter{theorem}{0}
\renewcommand{\thetheorem}{D.\arabic{theorem}}
\begin{theorem}
\label{theorem B1}
Suppose Assumption \ref{assumption:B1} holds. We set 
${\eta _{\boldsymbol{w}}^t} = {\eta _h^t} = \frac{2}{{L + {\rho _1}|{{\bf{A}}^t}|{L^2} + 8\frac{{|{{\bf{A}}^t}|\gamma {L^2}}}{{{\rho _1}({c_1^t})^2}}}}$, and we set constant  $ \rho _1 \! \le\! \frac{2}{{L + 2c_1^0}} $. For a given $\varepsilon $, we have:
{\begin{equation}
\renewcommand{\theequation}{D.\arabic{equation}}
\label{eq:B156}
T(\varepsilon )\sim \mathcal{O}(\max \{{(\frac{{{{16(\gamma  - 2){L^2}M{\rho _1} }}\mathop d\limits^ - {d_5}}}{{{\varepsilon ^2}}} + (T_1+2)^{\frac{1}{2}})^2},{\frac{{16M^2{\sigma _1}^4}}{{{\rho _1}^4}}}\frac{1}{{{\varepsilon ^4}}}\}),
\end{equation}} 

\vspace{-1.5mm}

where ${\sigma _1}$,  $\gamma$, $\mathop d\limits^ -  $, ${d_5}$ and $T_1$ are constants.
\end{theorem}

\setcounter{lemma}{0}
\renewcommand{\thelemma}{D.\arabic{lemma}}
\begin{lemma} \label{lemma B1}
Suppose Assumption \ref{assumption:B1} holds, we have:
\begin{equation}
\label{eq:B109}
\renewcommand{\theequation}{D.\arabic{equation}}
\begin{array}{l}
{L_p}({\boldsymbol{w}}^{t+1},h^{t},\{ {\lambda _l^{t}}\}) \!-\! {L_p}({\boldsymbol{w}}^{t},h^{t},\{ {\lambda _l^{t}}\})
\! \le \!  (\frac{L}{2}-\frac{1}{{{\eta _{\boldsymbol{w}}^t}}})||{{\boldsymbol{w}}}^{t+1} \!-\! {{\boldsymbol{w}}}^t|{|^2},
\end{array}
\end{equation}
\begin{equation}
\label{eq:B110}
\renewcommand{\theequation}{D.\arabic{equation}}
\begin{array}{l}
{L_p}({\boldsymbol{w}}^{t+1},h^{t+1},\{ {\lambda _l^{t}}\}) \!-\! {L_p}({\boldsymbol{w}}^{t+1},h^{t},\{ {\lambda _l^{t}}\})
 \!\le\! (\frac{L}{2} - \frac{1}{{{\eta _h^t}}})||h^{t+1} \!-\! h^{t}|{|^2}.
\end{array}
\end{equation}
\end{lemma}

\emph{\textbf{Proof}:} 

According to Assumption \ref{assumption:B1}, we have,
\begin{equation}
\label{eq:B111}
\renewcommand{\theequation}{D.\arabic{equation}}
\begin{array}{l}
{L_p}({\boldsymbol{w}}^{t+1},h^{t},\{ {\lambda _l^{t}}\}) \!-\! {L_p}({\boldsymbol{w}}^{t},h^{t},\{ {\lambda _l^{t}}\})\vspace{1ex}\\
\! \le \! {\left\langle {{\nabla _{{{\boldsymbol{w}}}}}{L_p}{\rm{( }}{{\boldsymbol{w}}}^t,h^{t},\{ {\lambda _l^{t}}\}  {\rm{),}}{{\boldsymbol{w}}}^{t+1} \!-\! {{\boldsymbol{w}}}^t} \right\rangle  \! + \! \frac{L}{2}||{{\boldsymbol{w}}}^{t+1} \!-\! {{\boldsymbol{w}}}^t|{|^2}}.
\end{array}
\end{equation}

According to the optimal condition for Eq. (D.\ref{eq:B106}) and ${\nabla _{\boldsymbol{w}}}\widetilde{L}_p({\boldsymbol{w}}^{t},h^{t},\{ {\lambda _l^{t}}\})\! =\!{\nabla _{\boldsymbol{w}}} {L_p}({\boldsymbol{w}}^{t},h^{t},\{ {\lambda _l^{t}}\})$, we have,
\begin{equation}
\label{eq:B112}
\renewcommand{\theequation}{D.\arabic{equation}}
\begin{array}{l}
\left\langle {{{\boldsymbol{w}}}^{t+1} \!-\! {{\boldsymbol{w}}}^t,{\nabla _{{{\boldsymbol{w}}}}}{L_p}{\rm{( }}{{\boldsymbol{w}}}^t,h^{t},\{ {\lambda _l^{t}}\}} \right\rangle  \le  -\frac{1}{{{\eta _{\boldsymbol{w}}^t}}}||{{\boldsymbol{w}}}^{t+1} \!-\! {{\boldsymbol{w}}}^t|{|^2}.
\end{array}
\end{equation}

Combining Eq. (D.\ref{eq:B111}) with Eq. (D.\ref{eq:B112}), we have that,
\begin{equation*}
\begin{array}{l}
{L_p}({\boldsymbol{w}}^{t+1},h^{t},\{ {\lambda _l^{t}}\}) \!-\! {L_p}({\boldsymbol{w}}^{t},h^{t},\{ {\lambda _l^{t}}\})
\! \le \!  (\frac{L}{2}-\frac{1}{{{\eta _{\boldsymbol{w}}^t}}})||{{\boldsymbol{w}}}^{t+1} \!-\! {{\boldsymbol{w}}}^t|{|^2}.
\end{array}
\end{equation*}

Similar to Eq. (D.\ref{eq:B109}), we can easily have Eq. (D.\ref{eq:B110}).

\vspace{3ex}

\begin{lemma} \label{lemma B2}
Suppose Assumption \ref{assumption:B1} holds, $\forall t \ge T_1$, we have:
\begin{equation}
\renewcommand{\theequation}{D.\arabic{equation}}
\label{eq:B113}
\begin{array}{l}
{L_p}({\boldsymbol{w}}^{t+1},h^{t+1},\{ {\lambda _l^{t+1}}\})-{L_p}({\boldsymbol{w}}^{t},h^{t},\{ {\lambda _l^{t}}\})\vspace{0.5ex}\\
\!\le\! (\frac{L}{2}-\frac{1}{{{\eta _{\boldsymbol{w}}^t}}}+\frac{{|{{\bf{A}}^t}|{L^2}}}{{2{a_1}}}){||{{\boldsymbol{w}}}^{t+1} \! - \! {{\boldsymbol{w}}}^t|{|^2}}  + (\frac{L}{2}-\frac{1}{{{\eta _h^t}}}+\frac{{|{{\bf{A}}^t}|{L^2}}}{{2{a_1}}})||h^{t+1} \! - \! h^{t}|{|^2} \\
 \!  +  (\frac{{{a_1}}}{2} \! - \! \frac{{{c_1^{t-1}}  -  {c_1^t}}}{2} \!  + \! \frac{1}{{2{\rho _1}}})\!\sum\limits_{l = 1}^{|{{\bf{A}}^t}|}\! {||{\lambda _l^{t+1}} \! - \! {\lambda _l^{t}}|{|^2}} 
  \!+ \! \frac{{{c_1^{t-1}} }}{2}\! \sum\limits_{l = 1}^{|{{\bf{A}}^t}|}\! {(||{\lambda _l^{t+1}}|{|^2} \! - \! ||{\lambda _l^{t}}|{|^2})} \! + \! \frac{1}{{2{\rho _1}}}\! \sum\limits_{l = 1}^{|{{\bf{A}}^t}|} \!{||{\lambda _l^{t}} \! - \! {\lambda _l^{t-1}}|{|^2}}.
\end{array}
\end{equation}

\end{lemma}

\emph{\textbf{Proof}:} 

According to Eq. (D.\ref{eq:B108}), in $(t+1)^{\rm{th}}$ iteration, $\forall \lambda  \in {\bf{\Lambda}} $, it follows that,
\begin{equation}
\renewcommand{\theequation}{D.\arabic{equation}}
\label{eq:B114}
\left\langle {{\lambda _l^{t+1}} \!-\! {\lambda _l^{t}} \!-\! {\rho _1}{\nabla _{{\lambda _l}}}{{\widetilde{L}}_p}({\boldsymbol{w}}^{t+1},h^{t+1},\{ {\lambda _l^{t}}\}),\lambda  \!-\! {\lambda _l^{t+1}}} \right\rangle  \ge 0.
\end{equation}

Let $\lambda  = {\lambda _l^{t}}$, we can obtain,
\begin{equation}
\renewcommand{\theequation}{D.\arabic{equation}}
\label{eq:B115}
\left\langle {{\nabla _{{\lambda _l}}}{{\widetilde{L}}_p}{{({\boldsymbol{w}}^{t+1},h^{t+1},\{ {\lambda _l^{t}}\})}} \!-\! \frac{1}{{{\rho _1}}}({\lambda _l^{t+1}} \!-\! {\lambda _l^{t}}),{\lambda _l^{t}} \!-\! {\lambda _l^{t+1}}} \right\rangle  \le 0.
\end{equation}

Likewise,  in $t^{\rm{th}}$ iteration, we have that,
\begin{equation}
\renewcommand{\theequation}{D.\arabic{equation}}
\label{eq:B116}
\left\langle {{\nabla _{{\lambda _l}}}{{\widetilde{L}}_p}({\boldsymbol{w}}^{t},h^{t},\{ {\lambda _l^{t-1}}\}) \!-\! \frac{1}{{{\rho _1}}}({\lambda _l^{t}} \!-\! {\lambda _l^{t-1}}),{\lambda _l^{t+1}} \!-\! {\lambda _l^{t}}} \right\rangle  \le 0.
\end{equation}

$\forall t \ge T_1$, since ${\widetilde{L}_p}({{\boldsymbol{w}}} ,h,\{ {\lambda _l}\})$ is concave with respect to ${\lambda _l}$, we have,
\begin{equation}
\renewcommand{\theequation}{D.\arabic{equation}}
\label{eq:B117}
\begin{array}{l}
{\widetilde{L}_p}({\boldsymbol{w}}^{t+1},h^{t+1},\{ {\lambda _l^{t+1}}\}) - {\widetilde{L}_p}({\boldsymbol{w}}^{t+1},h^{t+1},\{ {\lambda _l^{t}}\})\vspace{1ex}\\
\! \le \! \sum\limits_{l = 1}^{|{{\bf{A}}^t}|} \! {\left\langle \! {{\nabla _{{\lambda _l}}}{{\widetilde{L} }_p}({\boldsymbol{w}}^{t+1},h^{t+1},\{ {\lambda _l^{t}}\}),{\lambda _l^{t+1}} - {\lambda _l^{t}}} \! \right\rangle } \\

\! \le \! \sum\limits_{l  =  1}^{|{{\bf{A}}^t}|} \! {( \left\langle \! {{\nabla _{{\lambda _l}}}{{\widetilde{L} }_p}({\boldsymbol{w}}^{t+1},h^{t+1},\{ {\lambda _l^{t}}\}) \! -\! {\nabla _{{\lambda _l}}}{{\widetilde{L}}_p}({\boldsymbol{w}}^{t},h^{t},\{ {\lambda _l^{t-1}}\}),{\lambda _l^{t+1}}\! -\! {\lambda _l^{t}}} \! \right\rangle } \vspace{0.5ex}\\
\qquad \quad + \frac{1}{{{\rho _1}}}\left\langle {{\lambda _l^{t}} - {\lambda _l^{t-1}},{\lambda _l^{t+1}} - {\lambda _l^{t}}} \right\rangle ).
\end{array}
\end{equation}

Denoting ${{\boldsymbol{v}}_{1,l}^{t+1}} = {\lambda _l^{t+1}} - {\lambda _l^{t}} - ({\lambda _l^{t}} - {\lambda _l^{t-1}})$, we have,
\begin{equation}
\renewcommand{\theequation}{D.\arabic{equation}}
\label{eq:B118}
\begin{array}{l}
\sum\limits_{l  =  1}^{|{{\bf{A}}^t}|} \! {( \! \left\langle \! {{\nabla _{{\lambda _l}}}{{\widetilde{L} }_p}({\boldsymbol{w}}^{t+1},h^{t+1},\{ {\lambda _l^{t}}\}) \! -\! {\nabla _{{\lambda _l}}}{{\widetilde{L}}_p}({\boldsymbol{w}}^{t},h^{t},\{ {\lambda _l^{t-1}}\}),{\lambda _l^{t+1}}\! -\! {\lambda _l^{t}}} \! \right\rangle } \\
 \!= \! \sum\limits_{l = 1}^{|{{\bf{A}}^t}|} \! {\left\langle \! {{\nabla _{{\lambda _l}}}{{\widetilde{L}}_p}({\boldsymbol{w}}^{t+1},h^{t+1},\{ {\lambda _l^{t}}\})  \! -\! {\nabla _{{\lambda _l}}}{{\widetilde{L}}_p}({\boldsymbol{w}}^{t},h^{t},\{ {\lambda _l^{t}}\}),{\lambda _l^{t+1}}  \! -\! {\lambda _l^{t}}} \! \right\rangle } (1a)\\
  \! +\! \sum\limits_{l = 1}^{|{{\bf{A}}^t}|} \! {\left\langle \! {{\nabla _{{\lambda _l}}}{{\widetilde{L}}_p}({\boldsymbol{w}}^{t},h^{t},\{ {\lambda _l^{t}}\})  \! -\! {\nabla _{{\lambda _l}}}{{\widetilde{L}}_p}({\boldsymbol{w}}^{t},h^{t},\{ {\lambda _l^{t-1}}\}), {{\boldsymbol{v}}_{1,l}^{t+1}}} \right\rangle } (1b)\\
  \! +\! \sum\limits_{l = 1}^{|{{\bf{A}}^t}|} \! {\left\langle {{\nabla _{{\lambda _l}}}{{\widetilde{L}}_p}({\boldsymbol{w}}^{t},h^{t},\{ {\lambda _l^{t}}\})  \! -\! {\nabla _{{\lambda _l}}}{{\widetilde{L}}_p}({\boldsymbol{w}}^{t},h^{t},\{ {\lambda _l^{t-1}}\}),{\lambda _l^{t}}  \! -\! {\lambda _l^{t-1}}} \! \right\rangle } (1c).
\end{array}
\end{equation}

We firstly focus on ($1a$) in Eq. (D.\ref{eq:B118}). According to the Cauchy-Schwarz inequality and Assumption \ref{assumption:B1}, we have,
\begin{equation}
\renewcommand{\theequation}{D.\arabic{equation}}
\label{eq:B119}
\begin{array}{l}
\sum\limits_{l = 1}^{|{{\bf{A}}^t}|} \! {\left\langle \! {{\nabla _{{\lambda _l}}}{{\widetilde{L}}_p}({\boldsymbol{w}}^{t+1},h^{t+1},\{ {\lambda _l^{t}}\})  \! -\! {\nabla _{{\lambda _l}}}{{\widetilde{L}}_p}({\boldsymbol{w}}^{t},h^{t},\{ {\lambda _l^{t}}\}),{\lambda _l^{t+1}}  \! -\! {\lambda _l^{t}}} \! \right\rangle }\\
 \!\le \sum\limits_{l = 1}^{|{{\bf{A}}^t}|} ( \frac{{{L^2}}}{{2{a_1}}}({||{{\boldsymbol{w}}}^{t+1} \! - \! {{\boldsymbol{w}}}^t|{|^2}}  + ||h^{t+1} \! - \! h^{t}|{|^2}) + \frac{{{a_1}}}{2}||{\lambda _l^{t+1}} \! - \! {\lambda _l^{t}}|{|^2}\vspace{1ex}\\
 \qquad \quad + \frac{{{c_1^{t-1}}  -  {c_1^t}}}{2}(||{\lambda _l^{t+1}}|{|^2} \! - \! ||{\lambda _l^{t}}|{|^2}) \! - \! \frac{{{c_1^{t-1}} -  {c_1^t}}}{2}||{\lambda _l^{t+1}} \! - \! {\lambda _l^{t}}|{|^2}),
\end{array}
\end{equation}
where  $a_1 >0$ is a constant. Secondly, according to Cauchy-Schwarz inequality we write ($1b$) in Eq. (D.\ref{eq:B118}) as,
\begin{equation}
\renewcommand{\theequation}{D.\arabic{equation}}
\label{eq:B120}
\begin{array}{l}
\sum\limits_{l = 1}^{|{{\bf{A}}^t}|} \! {\left\langle \! {{\nabla _{{\lambda _l}}}{{\widetilde{L}}_p}({\boldsymbol{w}}^{t},h^{t},\{ {\lambda _l^{t}}\})  \! -\! {\nabla _{{\lambda _l}}}{{\widetilde{L}}_p}({\boldsymbol{w}}^{t},h^{t},\{ {\lambda _l^{t-1}}\}), {{\boldsymbol{v}}_{1,l}^{t+1}}} \right\rangle }\\
\! \le \! \sum\limits_{l = 1}^{|{{\bf{A}}^t}|}\! {( \frac{{{a_2}}}{2}\!||{{\nabla _{{\lambda _l}}}{{\widetilde{L}}_p}({\boldsymbol{w}}^{t},h^{t},\{ {\lambda _l^{t}}\})  \! -\! {\nabla _{{\lambda _l}}}{{\widetilde{L}}_p}({\boldsymbol{w}}^{t},h^{t},\{ {\lambda _l^{t-1}}\})}|{|^2}\! +\! \frac{1}{{2{a_2}}}\!||{{\boldsymbol{v}}_{1,l}^{t+1}}|{|^2})},
\end{array}
\end{equation}
where $a_2 >0$  is a constant. Then, we focus on the ($1c$) in Eq. (D.\ref{eq:B118}).  Denoting ${L_1}' = L + {c_1^0}$, according to Assumption \ref{assumption:B1}, trigonometric inequality and the strong  concavity of ${\widetilde{L}}_p{\rm{( }}{{\boldsymbol{w}}},h,\{ {\lambda _l}\})$ \textit{w.r.t} ${\lambda _l}$ \citep{nesterov2003introductory,xu2020unified}, we have,
{\begin{equation}
\renewcommand{\theequation}{D.\arabic{equation}}
\label{eq:B121}
\begin{array}{l}
\sum\limits_{l = 1}^{|{{\bf{A}}^t}|} \! {\left\langle {{\nabla _{{\lambda _l}}}{{\widetilde{L}}_p}({\boldsymbol{w}}^{t},h^{t},\{ {\lambda _l^{t}}\})  \! -\! {\nabla _{{\lambda _l}}}{{\widetilde{L}}_p}({\boldsymbol{w}}^{t},h^{t},\{ {\lambda _l^{t-1}}\}),{\lambda _l^{t}}  \! -\! {\lambda _l^{t-1}}} \! \right\rangle } \\
\! \le\! \sum\limits_{l = 1}^{|{{\bf{A}}^t}|}\! (  - \frac{1}{{{L_1}' + {c_1^{t-1}}}}||{{\nabla _{{\lambda _l}}}{{\widetilde{L}}_p}({\boldsymbol{w}}^{t},h^{t},\{ {\lambda _l^{t}}\})  \! -\! {\nabla _{{\lambda _l}}}{{\widetilde{L}}_p}({\boldsymbol{w}}^{t},h^{t},\{ {\lambda _l^{t-1}}\}), {{\boldsymbol{v}}_{1,l}^{t+1}}}|{|^2} 
 \!-\! \frac{{{c_1^{t-1}}{L_1}'}}{{{L_1}' + {c_1^{t-1}}}}||{\lambda _l^{t}} \! -\! {\lambda _l^{t-1}}|{|^2}).
\end{array}
\end{equation}}

In addition, we can obtain the following inequality,
{\begin{equation}
\renewcommand{\theequation}{D.\arabic{equation}}
\label{eq:B122}
\begin{array}{l}
\frac{1}{{{\rho _1}}}\left\langle {{\lambda _l^{t}} - {\lambda _l^{t-1}},{\lambda _l^{t+1}} - {\lambda _l^{t}}} \right\rangle 
 \le \frac{1}{{2{\rho _1}}}||{\lambda _l^{t+1}} - {\lambda _l^{t}}|{|^2} - \frac{1}{{2{\rho _1}}}||{{\boldsymbol{v}}_{1,l}^{t+1}}|{|^2} + \frac{1}{{2{\rho _1}}}||{\lambda _l^{t}} - {\lambda _l^{t-1}}|{|^2}.
\end{array}
\end{equation}}

Combining Eq.  (D.\ref{eq:B117}), (D.\ref{eq:B119}), (D.\ref{eq:B120}), (D.\ref{eq:B121}), (D.\ref{eq:B122}) with $  \frac{{{\rho _1}}}{2} \le \frac{1}{{{L_1}' + c_1^0}} $, and setting  ${a_2} = {\rho _1}$, $\forall t \ge T_1$, we have,
\begin{equation}
\renewcommand{\theequation}{D.\arabic{equation}}
\label{eq:B123}
\begin{array}{l}
{L_p}({\boldsymbol{w}}^{t+1},h^{t+1},\{ {\lambda _l^{t+1}}\})-{L_p}({\boldsymbol{w}}^{t+1},h^{t+1},\{ {\lambda _l^{t}}\})\\
\!\le\! \frac{{|{{\bf{A}}^t}|{L^2}}}{{2{a_1}}}({||{{\boldsymbol{w}}}^{t+1} \! - \! {{\boldsymbol{w}}}^t|{|^2}}  + ||h^{t+1} \! - \! h^{t}|{|^2})
 \!  + \! (\frac{{{a_1}}}{2} \! - \! \frac{{{c_1^{t-1}}  -  {c_1^t}}}{2} \!  + \! \frac{1}{{2{\rho _1}}} )\sum\limits_{l = 1}^{|{{\bf{A}}^t}|} {||{\lambda _l^{t+1}} \! - \! {\lambda _l^{t}}|{|^2}} \\
  +  \frac{{{c_1^{t-1}}}}{2}\sum\limits_{l = 1}^{|{{\bf{A}}^t}|} {(||{\lambda _l^{t+1}}|{|^2} \! - \! ||{\lambda _l^{t}}|{|^2})}  + \frac{1}{{2{\rho _1}}}\sum\limits_{l = 1}^{|{{\bf{A}}^t}|} {||{\lambda _l^{t}} \! - \! {\lambda _l^{t-1}}|{|^2}}.
\end{array}
\end{equation}

By combining Lemma \ref{lemma B1} with Eq. (D.\ref{eq:B123}), we conclude the proof of Lemma \ref{lemma B2}.

\vspace{3ex}

\begin{lemma} \label{lemma B3}
Denote:
\begin{equation}
\renewcommand{\theequation}{D.\arabic{equation}}
\label{eq:B124}
{S_1^{t+1}} = \frac{4}{{{\rho _1}^2{c_1^{t+1}}}}\sum\limits_{l = 1}^{|{{\bf{A}}^t}|} {||{\lambda _l^{t+1}} \! - \! {\lambda _l^{t}}|{|^2}}  \! - \! \frac{4}{{{\rho _1}}}(\frac{{{c_1^{t-1}}}}{{{c_1^t}}} \! - \! 1)\sum\limits_{l = 1}^{|{{\bf{A}}^t}|} {||{\lambda _l^{t+1}}|{|^2}},
\end{equation}
\begin{equation}
\renewcommand{\theequation}{D.\arabic{equation}}
\label{eq:B125}
\begin{array}{l}
F^{t+1} = {L_p}{\rm{( }}{{\boldsymbol{w}}}^{t+1},h^{t+1},\{ {\lambda _l^{t+1}}\}) \!  + \! {S_1^{t+1}}  \! - \! \frac{7}{{2{\rho _1}}}\!\sum\limits_{l = 1}^{|{{\bf{A}}^t}|}\! {||{\lambda _l^{t+1}} \! - \! {\lambda _l^{t}}|{|^2}} \! - \! \frac{c_1^t}{{2}}\!\sum\limits_{l = 1}^{|{{\bf{A}}^t}|}\! {||{\lambda _l^{t+1}}|{|^2}},
\end{array}
\end{equation}
then $\forall t \ge T_1$, we have:
{\begin{equation}
\renewcommand{\theequation}{D.\arabic{equation}}
\label{eq:B126}
\begin{array}{l}
F^{t+1} \! - \! F^t\\
 \!\le\! (\frac{{L}}{2} \! - \! \frac{1}{{{\eta _{\boldsymbol{w}}^t}}} \! +\! \frac{{{\rho _1}|{{\bf{A}}^t}|{L^2}}}{2}  \! +\! \frac{{8|{{\bf{A}}^t}|{L^2}}}{{{\rho _1}({c_1^t})^2}}) {||{{\boldsymbol{w}}}^{t+1} \! - \! {{\boldsymbol{w}}}^t|{|^2}} \! + (\frac{{L}}{2} \! - \! \frac{1}{{{\eta _h^t}}} \! +\! \frac{{{\rho _1}|{{\bf{A}}^t}|{L^2}}}{2}  \! +\! \frac{{8|{{\bf{A}}^t}|{L^2}}}{{{\rho _1}({c_1^t})^2}})||h^{t+1} \! - \! h^{t}|{|^2}\\
 \! -  \frac{1}{{10{\rho _1}}}  \sum\limits_{l = 1}^{|{{\bf{A}}^t}|} {||{\lambda _l^{t+1}} \! - \! {\lambda _l^{t}}|{|^2}} 
 \! +\! \frac{{{c_1^{t-1}}  -  {c_1^t}}}{2}\sum\limits_{l = 1}^{|{{\bf{A}}^t}|} {||{\lambda _l^{t+1}}|{|^2}}  \! + \frac{4}{{{\rho _1}}}(\frac{{{c_1^{t-2}}}}{{{c_1^{t-1}}}} \! - \! \frac{{{c_1^{t-1}}}}{{{c_1^t}}})\sum\limits_{l = 1}^{|{{\bf{A}}^t}|} {||{\lambda _l^{t}}|{|^2}}.
\end{array}
\end{equation}}

\end{lemma}

\emph{\textbf{Proof}:} 

Let ${a_1} = \frac{1}{{{\rho _1}}}$ and substitute it into Lemma \ref{lemma B2}, $\forall t \ge T_1$, we have,
\begin{equation}
\renewcommand{\theequation}{D.\arabic{equation}}
\label{eq:B127}
\begin{array}{l}
{L_p}({\boldsymbol{w}}^{t+1},h^{t+1},\{ {\lambda _l^{t+1}}\})-{L_p}({\boldsymbol{w}}^{t},h^{t},\{ {\lambda _l^{t}}\})\vspace{0.5ex}\\

\!\le\! (\frac{L}{2}-\frac{1}{{{\eta _{\boldsymbol{w}}^t}}}+\frac{{\rho_1|{{\bf{A}}^t}|{L^2}}}{{2}}){||{{\boldsymbol{w}}}^{t+1} \! - \! {{\boldsymbol{w}}}^t|{|^2}}  + (\frac{L}{2}-\frac{1}{{{\eta _h^t}}}+\frac{{\rho_1|{{\bf{A}}^t}|{L^2}}}{{2}})||h^{t+1} \! - \! h^{t}|{|^2} \\

 \!  +  ( -  \frac{{{c_1^{t-1}}  -  {c_1^t}}}{2} \!  + \! \frac{1}{{{\rho _1}}} )\sum\limits_{l = 1}^{|{{\bf{A}}^t}|} {||{\lambda _l^{t+1}} \! - \! {\lambda _l^{t}}|{|^2}} 
  +  \frac{{{c_1^{t-1}} }}{2}\sum\limits_{l = 1}^{|{{\bf{A}}^t}|} {(||{\lambda _l^{t+1}}|{|^2} \! - \! ||{\lambda _l^{t}}|{|^2})}  + \frac{1}{{2{\rho _1}}}\sum\limits_{l = 1}^{|{{\bf{A}}^t}|} {||{\lambda _l^{t}} \! - \! {\lambda _l^{t-1}}|{|^2}}.
\end{array}
\end{equation}

Firstly, $\forall t \ge T_1$, we can obtain the following inequality, 
{\begin{equation}
\renewcommand{\theequation}{D.\arabic{equation}}
\label{eq:103_1}
\begin{array}{l}
\! \sum\limits_{l = 1}^{|{{\bf{A}}^t}|}\! \frac{1}{{{\rho _1}}} \left\langle {{{\boldsymbol{v}}_{1,l}^{t + 1}},{\lambda _l^{t+1}} \! - \! {\lambda _l^t}} \right\rangle  \\
\! \le \! \sum\limits_{l = 1}^{|{{\bf{A}}^t}|} ( {\left\langle \! {{\nabla _{{\lambda _l}}}{{\widetilde{L}}_p}({\boldsymbol{w}}^{t+1},h^{t+1},\{ {\lambda _l^{t}}\})  \! -\! {\nabla _{{\lambda _l}}}{{\widetilde{L}}_p}({\boldsymbol{w}}^{t},h^{t},\{ {\lambda _l^{t}}\}),{\lambda _l^{t+1}}  \! -\! {\lambda _l^{t}}} \! \right\rangle }  \vspace{0.5ex}\\

 \qquad \;  +   {\left\langle \! {{\nabla _{{\lambda _l}}}{{\widetilde{L}}_p}({\boldsymbol{w}}^{t},h^{t},\{ {\lambda _l^{t}}\})  \! -\! {\nabla _{{\lambda _l}}}{{\widetilde{L}}_p}({\boldsymbol{w}}^{t},h^{t},\{ {\lambda _l^{t-1}}\}), {{\boldsymbol{v}}_{1,l}^{t+1}}} \right\rangle }
  \vspace{0.5ex}\\
 
 \qquad \; +   {\left\langle {{\nabla _{{\lambda _l}}}{{\widetilde{L}}_p}({\boldsymbol{w}}^{t},h^{t},\{ {\lambda _l^{t}}\})  \! -\! {\nabla _{{\lambda _l}}}{{\widetilde{L}}_p}({\boldsymbol{w}}^{t},h^{t},\{ {\lambda _l^{t-1}}\}),{\lambda _l^{t}}  \! -\! {\lambda _l^{t-1}}} \! \right\rangle } ). 
\end{array}
\end{equation}}

Since
{\begin{equation}
\renewcommand{\theequation}{D.\arabic{equation}}
\label{eq:B128}
\begin{array}{l}
\frac{1}{{{\rho _1}}}\left\langle {{{\boldsymbol{v}}_{1,l}^{t+1}},{\lambda _l^{t+1}} \!-\! {\lambda _l^{t}}} \right\rangle 
 = \frac{1}{{2{\rho _1}}}||{\lambda _l^{t+1}} \!-\! {\lambda _l^{t}}|{|^2} \! + \! \frac{1}{{2{\rho _1}}}||{{\boldsymbol{v}}_{1,l}^{t+1}}|{|^2} - \frac{1}{{2{\rho _1}}}||{\lambda _l^{t}} - {\lambda _l^{t-1}}|{|^2},
\end{array}
\end{equation}}
it follows that,
{\begin{equation}
\renewcommand{\theequation}{D.\arabic{equation}}
\label{eq:B129}
\begin{array}{l}
\sum\limits_{l = 1}^{|{{\bf{A}}^t}|}\!(\frac{1}{{2{\rho _1}}}||{\lambda _l^{t+1}} \! - \! {\lambda _l^{t}}|{|^2} \! + \! \frac{1}{{2{\rho _1}}}||{{\boldsymbol{v}}_{1,l}^{t+1}}|{|^2} \! - \! \frac{1}{{2{\rho _1}}}||{\lambda _l^{t}} \! - \! {\lambda _l^{t-1}}|{|^2})\\

\!\le\!\sum\limits_{l = 1}^{|{{\bf{A}}^t}|}\!( \frac{{{L^2}}}{{2{b_1^t}}}({||{{\boldsymbol{w}}}^{t+1} \! - \! {{\boldsymbol{w}}}^t|{|^2}}  + ||h^{t+1} \! - \! h^{t}|{|^2})
 \!  + \! \frac{{{b_1^t}}}{2} {||{\lambda _l^{t+1}} \! - \! {\lambda _l^{t}}|{|^2}} \\
 
 \qquad \; +  \frac{{{c_1^{t-1}} \! - \! {c_1^t}}}{2} {(||{\lambda _l^{t+1}}|{|^2} \! - \! ||{\lambda _l^{t}}|{|^2})}  \! - \! \frac{{{c_1^{t-1}} \! - \! {c_1^t}}}{2}||{\lambda _l^{t+1}} \! - \! {\lambda _l^{t}}|{|^2}\vspace{0.5ex}\\
  
 \qquad \; +\frac{{{\rho_1}}}{2}\!||{{\nabla _{{\lambda _l}}}{{\widetilde{L}}_p}({\boldsymbol{w}}^{t},h^{t},\{ {\lambda _l^{t}}\})  \! -\! {\nabla _{{\lambda _l}}}{{\widetilde{L}}_p}({\boldsymbol{w}}^{t},h^{t},\{ {\lambda _l^{t-1}}\}), {{\boldsymbol{v}}_{1,l}^{t+1}}}|{|^2} \! + \! \frac{1}{{2{\rho _1}}}\!||{{\boldsymbol{v}}_{1,l}^{t+1}}|{|^2}\vspace{0.5ex} \\
  
 \qquad \;  - \frac{1}{{{L_1}' + {c_1^{t-1}}}}||{{\nabla _{{\lambda _l}}}{{\widetilde{L}}_p}({\boldsymbol{w}}^{t},h^{t},\!\{ {\lambda _l^{t}}\})  \! -\! {\nabla _{{\lambda _l}}}{{\widetilde{L}}_p}({\boldsymbol{w}}^{t},h^{t},\!\{ {\lambda _l^{t-1}}\}), {{\boldsymbol{v}}_{1,l}^{t+1}}}|{|^2} 
 \!- \frac{{{c_1^{t-1}}{L_1}'}}{{{L_1}' + {c_1^{t-1}}}}||{\lambda _l^{t}} \! -\! {\lambda _l^{t-1}}|{|^2}),
\end{array}
\end{equation}}

\noindent where ${b_1^t} > 0$. According to the setting that ${c_1^0} \le {L_1}'$, we have $- \frac{{{c_1^{t-1}}{L_1}'}}{{{L_1}' + {c_1^{t-1}}}} \le  - \frac{{{c_1^{t-1}}{L_1}'}}{{2{L_1}'}} =  - \frac{{{c_1^{t-1}}}}{2} \le  - \frac{{{c_1^t}}}{2}$. Multiplying both sides of the inequality Eq. (D.\ref{eq:B129}) by $\frac{8}{{{\rho _1}{c_1^t}}}$ and setting ${b_1^t} = \frac{{{c_1^t}}}{2}$, $\forall t \ge T_1$, we have,
{\begin{equation}
\renewcommand{\theequation}{D.\arabic{equation}}
\label{eq:B130}
\begin{array}{l}
{S_1^{t+1}} \! - \! {S_1^{t}}
\! \le \!\sum\limits_{l = 1}^{|{{\bf{A}}^t}|} {\frac{4}{{{\rho _1}}}(\frac{{{c_1^{t-2}}}}{{{c_1^{t-1}}}} \! - \! \frac{{{c_1^{t-1}}}}{{{c_1^t}}})||{\lambda _l^{t}}|{|^2}}  \! +\! \sum\limits_{l = 1}^{|{{\bf{A}}^t}|} {(\frac{2}{{{\rho _1}}} \! +\! \frac{4}{{\rho_1}^2}(\frac{1}{{{c_1^{t+1}}}} \! - \! \frac{1}{{{c_1^t}}}))||{\lambda _l^{t+1}} \! - \! {\lambda _l^{t}}|{|^2}} \vspace{1ex}\\
\quad \quad \quad \quad  \quad  \! - \sum\limits_{l = 1}^{|{{\bf{A}}^t}|} {\frac{4}{{{\rho _1}}}||{\lambda _l^{t}} \! - \! {\lambda _l^{t-1}}|{|^2} \vspace{1ex}}
\! +\! \frac{{8|{{\bf{A}}^t}|{L^2}}}{{{\rho _1}({c_1^t})^2}}({||{{\boldsymbol{w}}}^{t+1} \! - \! {{\boldsymbol{w}}}^t|{|^2}}  + ||h^{t+1} \! - \! h^{t}|{|^2}).
\end{array}
\end{equation}}

According to the setting about ${c_1^t}$, we have $\frac{{{\rho _1}}}{{10}} \ge \frac{1}{{{c_1^{t+1}}}} - \frac{1}{{{c_1^t}}}, \forall t \ge T_1$.
Using the definition of $F^{t+1}$ and combining it with Eq. (D.\ref{eq:B130}) and Eq. (D.\ref{eq:B127}), $\forall t \ge T_1$, we have,
{\begin{equation*}
\renewcommand{\theequation}{A.\arabic{equation}}
\label{eq:A59}
\begin{array}{l}
F^{t+1} \! - \! F^t\\

 \!\le\! (\frac{{L}}{2} \! - \! \frac{1}{{{\eta _{\boldsymbol{w}}^t}}} \! +\! \frac{{{\rho _1}|{{\bf{A}}^t}|{L^2}}}{2}  \! +\! \frac{{8|{{\bf{A}}^t}|{L^2}}}{{{\rho _1}({c_1^t})^2}}) {||{{\boldsymbol{w}}}^{t+1} \! - \! {{\boldsymbol{w}}}^t|{|^2}} \! + (\frac{{L}}{2} \! - \! \frac{1}{{{\eta _h^t}}} \! +\! \frac{{{\rho _1}|{{\bf{A}}^t}|{L^2}}}{2}  \! +\! \frac{{8|{{\bf{A}}^t}|{L^2}}}{{{\rho _1}({c_1^t})^2}})||h^{t+1} \! - \! h^{t}|{|^2}\\
 \! -  \frac{1}{{10{\rho _1}}} \sum\limits_{l = 1}^{|{{\bf{A}}^t}|} {||{\lambda _l^{t+1}} \! - \! {\lambda _l^{t}}|{|^2}} 
 \! +\! \frac{{{c_1^{t-1}}  -  {c_1^t}}}{2}\sum\limits_{l = 1}^{|{{\bf{A}}^t}|} {||{\lambda _l^{t+1}}|{|^2}}  \! + \frac{4}{{{\rho _1}}}(\frac{{{c_1^{t-2}}}}{{{c_1^{t-1}}}} \! - \! \frac{{{c_1^{t-1}}}}{{{c_1^t}}})\sum\limits_{l = 1}^{|{{\bf{A}}^t}|} {||{\lambda _l^{t}}|{|^2}}.
\end{array}
\end{equation*}}

\emph{\textbf{Proof of Theorem \ref{theorem B1}:}}

Firstly, we set that ${a_5^t} = \frac{{4|{{\bf{A}}^t}|(\gamma  - 2){L^2}}}{{{\rho _1}({c_1^t})^2}}$, where $\gamma>2$ is a constant. According to the setting of ${\eta _{\boldsymbol{w}}^t}$, ${\eta _h^t}$ and $c_1^t$, we have,
{\begin{equation}
\renewcommand{\theequation}{D.\arabic{equation}}
\label{eq:B131}
\frac{{L}}{2} \! - \! \frac{1}{{{\eta _{\boldsymbol{w}}^t}}} \! +\! \frac{{{\rho _1}|{{\bf{A}}^t}|{L^2}}}{2}  \! +\! \frac{{8|{{\bf{A}}^t}|{L^2}}}{{{\rho _1}({c_1^t})^2}} =  - {a_5^t}, \quad \frac{{L}}{2} \! - \! \frac{1}{{{\eta _h^t}}} \! +\! \frac{{{\rho _1}|{{\bf{A}}^t}|{L^2}}}{2}  \! +\! \frac{{8|{{\bf{A}}^t}|{L^2}}}{{{\rho _1}({c_1^t})^2}} =  - {a_5^t}.
\end{equation}}

Combining with Lemma \ref{lemma B3}, $\forall t \ge T_1$, it follows that,
{\begin{equation}
\renewcommand{\theequation}{D.\arabic{equation}}
\label{eq:B139}
\begin{array}{l}
{a_5^t} {||{{\boldsymbol{w}}}^{t+1} - {{\boldsymbol{w}}}^t|{|^2}}  \! +\! {a_5^t}||h^{t+1} - h^{t}|{|^2}
 \! +\! {\frac{1}{{10{\rho _1}}} \sum\limits_{l = 1}^{|{{\bf{A}}^t}|}||{\lambda _l^{t+1}} - {\lambda _l^{t}}|{|^2}}  \\
 
 \le F^t - F^{t+1} \! +\! \frac{{{c_1^{t-1}} - {c_1^t}}}{2}\sum\limits_{l = 1}^{|{{\bf{A}}^t}|} {||{\lambda _l^{t+1}}|{|^2}}  \! +\!  {\frac{4}{{{\rho _1}}}(\frac{{{c_1^{t-2}}}}{{{c_1^{t-1}}}} - \frac{{{c_1^{t-1}}}}{{{c_1^t}}})\sum\limits_{l = 1}^{|{{\bf{A}}^t}|}||{\lambda _l^{t}}|{|^2}}.
\end{array}
\end{equation}}

\vspace{-2mm}

According to the definition of ${(\nabla{\widetilde{G}} (t))_{{{\boldsymbol{w}}}}}$, we have,
{\begin{equation}
\renewcommand{\theequation}{D.\arabic{equation}}
\label{eq:B140}
\begin{array}{l}
||{(\nabla {\widetilde{G}}^t)_{{{\boldsymbol{w}}}}}|{|^2}
 \le  \frac{1}{{({\eta _{\boldsymbol{w}}^t})^2}}||{{\boldsymbol{w}}}^{t+1} \!-\! \boldsymbol{w}^t|{|^2}.
\end{array}
\end{equation}}

\vspace{-2mm}

Combining the definition of ${(\nabla{\widetilde{G}} (t))_{{h}}}$ with trigonometric, Cauchy-Schwarz inequality and Assumption \ref{assumption:B1}, we have,
{\begin{equation}
\renewcommand{\theequation}{D.\arabic{equation}}
\label{eq:B141}
\begin{array}{l}
||{(\nabla {\widetilde{G}}^t)_h}|{|^2}
 \le 2{L^2}{||{{\boldsymbol{w}}}^{t+1} \!-\! {{\boldsymbol{w}}}^t|{|^2}} \! +\! \frac{2}{{{({\eta _h^t})^2}}}||h^{t+1} \!-\! h^{t}|{|^2}.
\end{array}
\end{equation}}

\vspace{-2mm}

Combining the definition of ${(\nabla {\widetilde{G}}^t)_{{\lambda _l}}}$ with trigonometric inequality and Cauchy-Schwarz inequality,
{\begin{equation}
\renewcommand{\theequation}{D.\arabic{equation}}
\label{eq:B142}
\begin{array}{l}
||{(\nabla {\widetilde{G}}^t)_{{\lambda _l}}}|{|^2}
 \!\le \!\frac{3}{{\rho_1}^2}||{\lambda _l^{t+1}} \! -\! {\lambda _l^{t}}|{|^2} \! +\! 3{L^2}({||{{\boldsymbol{w}}}^{t+1} \! -\! {{\boldsymbol{w}}}^t|{|^2}} \! +\! ||h^{t+1} \! -\! h^{t}|{|^2})  \! +\! 3{(({c_1^{t-1}})^2 \! -\! ({c_1^t})^2)}||{\lambda _l^{t}}|{|^2}.
\end{array}
\end{equation}}

According to the Definition \ref{definition:B2} as well as Eq. (D.\ref{eq:B140}), (D.\ref{eq:B141}) and Eq. (D.\ref{eq:B142}), we can obtain,
{\begin{equation}
\renewcommand{\theequation}{D.\arabic{equation}}
\label{eq:B143}
\begin{array}{l}
||\nabla {\widetilde{G}}^t|{|^2}

 \le (\frac{1}{{({\eta _{\boldsymbol{w}}^t})^2}} \! +\! 2L^2 \! +\! 3|{{\bf{A}}^t}|{L^2}) {||{{\boldsymbol{w}}}^{t+1} \! - \! {{\boldsymbol{w}}}^t|{|^2}} \! +\! (\frac{2}{{({\eta _h^t})^2}} \! +\! 3|{{\bf{A}}^t}|{L^2})||h^{t+1} - h^{t}|{|^2} \\ 
 
\quad \quad \quad \quad \quad \quad \! +\! \sum\limits_{l = 1}^{|{{\bf{A}}^t}|} {\frac{3}{{\rho_1}^2}||{\lambda _l^{t+1}} \! - \! {\lambda _l^{t}}|{|^2}}  \! +\! \sum\limits_{l = 1}^{|{{\bf{A}}^t}|} {3{{(({c_1^{t-1}})^2 \! - \! ({c_1^t})^2)}}||{\lambda _l^{t}}|{|^2}}.
\end{array}
\end{equation}}

\vspace{-2mm}

We set constants $d_1$, $d_2$ as
{\begin{equation}
\renewcommand{\theequation}{D.\arabic{equation}}
\label{eq:B144}
{d_1} = \frac{{1 \! +\! (2 \! +\! 3M){L^2}\underline{{\eta _{\boldsymbol{w}}}}^2}}{{\underline{{\eta _{\boldsymbol{w}}}}^2({a_5^0})}^2} \ge \frac{{1  \! +\! (2 \! +\! 3|{{\bf{A}}^t}|){L^2}({\eta _{\boldsymbol{w}}^t})^2}}{{({\eta _{\boldsymbol{w}}^t})^2({a_5^t})^2}},
\end{equation}}
{\begin{equation}
\renewcommand{\theequation}{D.\arabic{equation}}
\label{eq:B145}
{d_2} = \frac{{2 \! +\!  3M{L^2}\underline{{\eta _h}}^2}}{{\underline{{\eta _h}}^2({a_5^0})^2}} \ge \frac{{2 \! +\! 3|{{\bf{A}}^t}|{L^2}({\eta _h^t})^2}}{{({\eta _h^t})^2({a_5^t})^2}},
\end{equation}}

\noindent where  $\underline{\eta _{\boldsymbol{w}}} = \frac{2}{{L + {\rho _1}M{L^2} + 8\frac{{M\gamma {L^2}}}{{\rho _1}\underline{c}_1^2}}} \le \eta _{\boldsymbol{w}}^t$ and  $\underline{\eta _h} = \frac{2}{{L + {\rho _1}M{L^2} + 8\frac{{M\gamma {L^2}}}{{\rho _1}\underline{c}_1^2}}} \le \eta _h^t, \forall t$ are positive constants. Thus, combining Eq. (D.\ref{eq:B143}) with Eq. (D.\ref{eq:B144}) and Eq. (D.\ref{eq:B145}), we can obtain,
{\begin{equation}
\renewcommand{\theequation}{D.\arabic{equation}}
\label{eq:B146}
\begin{array}{l}
||\nabla {\widetilde{G}}^t|{|^2}

\! \le \! {{d_1}({a_5^t})^2||{{\boldsymbol{w}}}^{t+1} - {{\boldsymbol{w}}}^t|{|^2}} \! +\! {d_2}({a_5^t})^2||h^{t+1} - h^{t}|{|^2}\\
 
\quad \quad \quad \quad \quad \! +\! \sum\limits_{l = 1}^{|{{\bf{A}}^t}|} {\frac{3}{{\rho_1}^2} ||{\lambda _l^{t+1}} - {\lambda _l^{t}}|{|^2}}  \! +\! \sum\limits_{l = 1}^{|{{\bf{A}}^t}|} {3(({c_1^{t-1}})^2- ({c_1^t})^2)||{\lambda _l^{t}}|{|^2}}.
\end{array}
\end{equation}}

Let $d_3^t$ denote a nonnegative sequence, 
${d_3^t} = \frac{1}{{\max \{ {d_1}{a_5^t},{d_2}{a_5^t},\frac{{30  }}{{{\rho _1}}}\} }}$, and we have,
{\begin{equation}
\renewcommand{\theequation}{D.\arabic{equation}}
\label{eq:B147}
\begin{array}{l}
{d_3^t}||\nabla {\widetilde{G}}^t|{|^2}
 \le {a_5^t} {||{{\boldsymbol{w}}}^{t+1} - {{\boldsymbol{w}}}^t|{|^2}}  \! +\! {a_5^t}||h^{t+1} - h^{t}|{|^2}\\
 \quad \quad  \quad  \quad \quad  \quad \! +  {\frac{1}{{10{\rho _1}}}\!\sum\limits_{l = 1}^{|{{\bf{A}}^t}|}\!||{\lambda _l^{t+1}}\! -\! {\lambda _l^{t}}|{|^2}} 
\! +\! 3{d_3^t}(({c_1^{t-1}})^2 \!-\! ({c_1^t})^2)\!\sum\limits_{l = 1}^{|{{\bf{A}}^t}|}\! {||{\lambda _l^{t}}|{|^2}}.

\end{array}
\end{equation}}

\vspace{-2mm}

Combining Eq. (D.\ref{eq:B147}) with Eq. (D.\ref{eq:B139}) and  according to the setting $||{\lambda _l^{t}}|{|^2} \! \le \! {\sigma _1}^2$ (where ${\sigma _1}^2={\alpha_3}^2$) and $d_3^0\ge d_3^t$, $\forall t \ge T_1$, we have that,
{\begin{equation}
\renewcommand{\theequation}{D.\arabic{equation}}
\label{eq:B148}
\begin{array}{l}
{d_3^t}||\nabla {\widetilde{G}}^t|{|^2} \! \le\!  F^t \! - \! F^{t+1} \! +\! \frac{{{c_1^{t-1}}  -  {c_1^t}}}{2}M{\sigma _1}^2  \! +\!  {\frac{4}{{{\rho _1}}}(\frac{{{c_1^{t-2}}}}{{{c_1^{t-1}}}}  \! - \! \frac{{{c_1^{t-1}}}}{{{c_1^t}}})M{\sigma _1}^2} \! +\! 3{d_3^0}(({c_1^{t-1}})^2 \!-\! ({c_1^t})^2)M{\sigma _1}^2.
\end{array}
\end{equation}}

Denoting $\widetilde{T}(\varepsilon )$ as $\widetilde{T}(\varepsilon ) = \min \{ t \ | \quad ||\nabla \widetilde{G}^{T_1 + t}|| \le \frac{\varepsilon}{2}, t\ge 2 \}$. Summing up Eq. (D.\ref{eq:B148}) from $t=T_1+2$ to $t =T_1+{{\widetilde{T}} (\varepsilon )} $, we have,
{\begin{equation}
\renewcommand{\theequation}{D.\arabic{equation}}
\label{eq:B149}
\begin{array}{l}
\sum\limits_{t =T_1\!+\! 2}^{T_1\!+\!{\widetilde{T}} (\varepsilon )}\! {{d_3^t}||\nabla {\widetilde{G}}^t|{|^2}} 
 
 \le F^{T_1+2} \!-\! \mathop L\limits_ -
 \! +  \frac{4}{{{\rho _1}}}(\frac{{{c_1^{T_1}}}}{{{c_1^{T_1 + 1}}}} + \frac{{{c_1^{T_1 + 1}}}}{{{c_1^{T_1 + 2}}}})M {{\sigma _1}^2}  \! + \! \frac{{{c_1^{T_1+1}}}}{2}M {{\sigma _1}^2} \\
 
  \qquad \qquad \qquad \qquad \;\; +  \frac{7}{{2{\rho _1}}}M {{\sigma _3}^2} + \frac{c_1^{T_1+2}}{{2}}M {{\sigma _1}^2}  \! + \! 3{d_3^0} {({c_1^0})^2M {{\sigma _1}^2} }\\
  
 \qquad \qquad \qquad \quad \;\;  = \mathop d\limits^ - ,
\end{array}
\end{equation}}

where ${\sigma _3} \!=\! \max \{  ||{\lambda _1} - {\lambda _2}|| \, {\rm{ }}|{\lambda _1},{\lambda _2} \! \in \! {\bf{\Lambda}}  \} $ and $\mathop L\limits_ -  \! =  \! \mathop {\min }\limits_{\!{{\boldsymbol{w}}}  \in  {{\boldsymbol{\mathcal{W}}}}\! , h  \in   {{\boldsymbol{\mathcal{H}}}},\{ \! {\lambda _l} \in {\bf{\Lambda}}  \! \} } \! {L_p}{\rm{(  }}{{\boldsymbol{w}}},h, \! \{  {\lambda _l}  \}  {\rm{)}}$, which satisfy that,
{\begin{equation}
\renewcommand{\theequation}{D.\arabic{equation}}
\label{eq:B150}
F^{t+1} \ge \mathop L\limits_ -   - \frac{4}{{{\rho _1}}}\frac{{{c_1^{T_1 +1}}}}{{{c_1^{T_1 + 2}}}}M {{\sigma _1}^2}   - \frac{7}{{2{\rho _1}}}M {{\sigma _3}^2} -  \frac{c_1^{T_1+2}}{{2}}M {{\sigma _1}^2},\; \forall t \ge T_1 + 2,
\end{equation}}

and $\mathop d\limits^ -$ is a constant. Constant $d_5$ is given by,
{\begin{equation}
\renewcommand{\theequation}{D.\arabic{equation}}
\label{eq:B151}
\begin{array}{l}
{d_5} = \max \{ {d_1},{d_2},\frac{{30}}{{{{{\rho _1}}}{a_5^0}}}\} \ge 
\max \{{d_1},{d_2},\frac{{30}}{{{{{\rho _1}}}{a_5^t}}}\}   = \frac{1}{{{d_3^t}{a_5^t}}}
\end{array}.
\end{equation}} 

Thus, we can obtain that,
{\begin{equation}
\renewcommand{\theequation}{D.\arabic{equation}}
\label{eq:B152}
\sum\limits_{t =T_1\!+\! 2}^{T_1\!+\!{\widetilde{T}} (\varepsilon )} {\frac{1}{{{d_5}{a_5^t}}}||\nabla \widetilde{G}^{T_1+\widetilde{T}(\varepsilon )}|{|^2}}   \le \sum\limits_{t =T_1\!+\! 2}^{T_1\!+\!{\widetilde{T}} (\varepsilon )} {{d_3^t}||\nabla \widetilde{G}^t|{|^2}}  \le \mathop d\limits^ - .
\end{equation}} 

Summing up $\frac{1}{{{a_5^t}}}$ from ${t =T_1+ 2}$ to ${t =T_1+ {{\widetilde{T}} (\varepsilon) }}$, it follows that, 
{\begin{equation}
\renewcommand{\theequation}{D.\arabic{equation}}
\label{eq:B153}
\begin{array}{l}
\sum\limits_{t =T_1\!+\! 2}^{T_1\!+\!{\widetilde{T}} (\varepsilon )} {\frac{1}{{{a_5^t}}}} 
 \ge \sum\limits_{t =T_1\!+\! 2}^{T_1\!+\!{\widetilde{T}} (\varepsilon )} {\frac{1}{{{{4(\gamma  - 2){L^2}M{\rho _1} {(t+1)^{\frac{1}{2}}} }}}}}
 
 \ge \frac{(T_1+{\widetilde{T}} {(\varepsilon ))}^{\frac{1}{2}} - (T_1+2)^{\frac{1}{2}}}{{{{4(\gamma  - 2){L^2}M{\rho _1} }}}}.
\end{array}
\end{equation}}  

Combining Eq. (D.\ref{eq:B152}), (D.\ref{eq:B153}) with the definition of ${\widetilde{T}(\varepsilon )}$, we have that,  
{\begin{equation}
\renewcommand{\theequation}{D.\arabic{equation}}
\label{eq:B154}
T_1 + {\widetilde{T}} (\varepsilon ) \ge {(\frac{{{{16(\gamma  - 2){L^2}M{\rho _1} }}\mathop d\limits^ - {d_5}}}{{{\varepsilon ^2}}} + (T_1+2)^{\frac{1}{2}})^2}.
\end{equation}}  

According to trigonometric inequality, we then get $||\nabla G^t|| - ||\nabla {\widetilde{G}}^t|| \le ||\nabla G^t - \nabla {\widetilde{G}}^t|| \le \sqrt {\sum\limits_{l = 1}^{|{{\bf{A}}^t}|} {||{c_1^{t-1}}{\lambda _l^{t}}|{|^2}}}$. If $t > {\frac{{16M^2{\sigma _1}^4}}{{{\rho _1}^4}}}\frac{1}{{{\varepsilon ^4}}}$, we have $\sqrt {\sum\limits_{l = 1}^{|{{\bf{A}}^t}|} {||{c_1^{t-1}}{\lambda _l^{t}}|{|^2}} }  \le \frac{\varepsilon }{2}$. Combining it with Eq. (D.\ref{eq:B154}), we can conclude that there exists a  
{\begin{equation}
\renewcommand{\theequation}{D.\arabic{equation}}
\label{eq:B156}
T(\varepsilon )\sim \mathcal{O}(\max \{{(\frac{{{{16(\gamma  - 2){L^2}M{\rho _1} }}\mathop d\limits^ - {d_5}}}{{{\varepsilon ^2}}} + (T_1+2)^{\frac{1}{2}})^2},{\frac{{16M^2{\sigma _1}^4}}{{{\rho _1}^4}}}\frac{1}{{{\varepsilon ^4}}}\}),
\end{equation}} 

\noindent such that $||\nabla G^t||  \le \varepsilon $, which concludes our proof.

\section{Convergence Rate Analysis}
In this section, we compare the convergence results of the proposed method against the existing methods in the literature (with centralized and distributed setting). GDmax \citep{jin2020local} is proposed recently, which can be utilized to solve the nonconvex-concave minimax
problems (related to the setting of our problem) with iteration complexity $\mathcal{O}(\frac{1}{{{\varepsilon ^6}}})$ to obtain the $\varepsilon$-stationary point (\textit{i.e.}, $||\Phi ( \cdot )|{|^2} \le {\varepsilon ^2}$, where $\Phi ( \cdot ) = {\max _y}f( \cdot ,y)$). However, GDmax is nested-loop which has to solve the inner subproblem every iteration \citep{xu2020unified}. Gradient descent-ascent (GDA) method \citep{lin2020gradient} is proposed, which performs alternating gradient descent-ascent every iteration. The iteration complexity of GDA to obtain the $\varepsilon$-stationary point (\textit{i.e.}, $||\Phi ( \cdot )|{|^2} \le {\varepsilon ^2}$) for nonconvex-concave minimax
problems is upper bounded by $\mathcal{O}(\frac{1}{{{\varepsilon ^6}}})$.  COVER \citep{qi2021online} is proposed to solve the distributionally robust optimization with nonconvex objectives, which can obtain the $\varepsilon$-stationary point (\textit{i.e.},  $||\mathcal{G}_{\eta} ( \cdot )|{|^2} \le {\varepsilon ^2}$, where $\mathcal{G}_{\eta}$ is a proximal gradient measure) with the complexity $\mathcal{O}(\frac{1}{{{\varepsilon ^3}}})$. Nevertheless, all the algorithms mentioned above do not discuss about the distributed algorithms. Recently, GCIVR \citep{haddadpour2022learning} is proposed to solve the distributionally robust optimization problem in centralized and distributed manners. GCIVR is effective, which can respectively obtain the $\varepsilon$-stationary point (\textit{i.e.},  $||\mathcal{G}_{\eta} ( \cdot )|{|^2} \le {\varepsilon ^2}$) with the complexity  $\mathcal{O}(\min\{\frac{\sqrt N }{{{\varepsilon ^2}}}, \frac{1}{{{\varepsilon ^3}}}\})$ and $\mathcal{O}(\min\{\frac{\sqrt N }{{{p\varepsilon ^2}}}+\frac{\sqrt N }{{{\varepsilon ^2}}}, \frac{1}{{{p\varepsilon ^3}}}+\frac{1}{{{\varepsilon ^3}}}\})$ ($p$ is the number of workers, in this problem $p=N$) in centralized and distributed manners when the objective is nonconvex. 

The proposed algorithm differs significantly from the aforementioned methods because it is designed for solving the PD-DRO problem in Eq. (4) in an \emph{asynchronous} \emph{distributed} \emph{manner}. The asynchronous distributed algorithm does not suffer from the straggler problem \citep{jiang2021asynchronous} and therefore is critical for large scale distributed optimization in practice. On the contrary, synchronous distributed algorithm suffers from the straggler problem, \textit{i.e.}, its speed is limited by the worker with maximum delay \citep{chang2016asynchronous} and may not scale well with the size of a distributed system. For instance, we assume that the delays of workers follow a heavy-tailed distribution as given in \citep{cohen2021asynchronous}. With the increase of the number of workers in the distributed system, the maximum delay may increase dramatically as shown in Figure E1. Hence, the synchronous algorithm may incur huge delays and become practically infeasible for a large-scale distributed systems with tens of thousands of workers. Moreover, if a few workers fail to respond, which is very common in real-world large-scale data centers, the synchronous algorithm will come to an immediate halt \citep{zhang2014asynchronous}. Therefore, the asynchronous algorithm is strongly preferred in practice. 

The asynchronous setting is considered when we design the distributed algorithm. Compared with centralized algorithm, the asynchronous distributed algorithm is more complicated, which pose the major challenge against the theoretical analysis. In the future work, how to improve the iteration complexity will be taken into consideration. And we summarize the convergence results of different methods in Table \ref{tab:convergence rate compare}.

\setcounter{table}{0}
\renewcommand\arraystretch{1.5}
\renewcommand\tabcolsep{10pt}
\begin{table}[t]
\renewcommand{\thetable}{E\arabic{table}}
\caption{Convergence rate of algorithms related to our work (with centralized and distributed setting).}
{\centering
\scalebox{0.85}{
\begin{tabular}{l|c|c|c}
\toprule
Method    & Centralized     & Synchronous (Distributed) & Asynchronous (Distributed)      \\ \hline
GDmax \citep{jin2020local} & $\mathcal{O}(\frac{1}{{{\varepsilon ^6}}})^{1,3}$  & NA$^5$ & NA$^5$ \\ 
GDA  \citep{lin2020gradient}  & $\mathcal{O}(\frac{1}{{{\varepsilon ^6}}})^{1}$ & NA$^5$ & NA$^5$ \\ 
COVER \citep{qi2021online} & $\mathcal{O}(\frac{1}{{{\varepsilon ^3}}})^{2}$ & NA$^5$ & NA$^5$ \\ 
GCIVR \citep{haddadpour2022learning} & $\mathcal{O}(\min\{\frac{\sqrt N }{{{\varepsilon ^2}}}, \frac{1}{{{\varepsilon ^3}}}\})^{2}$ & $\mathcal{O}(\min\{\frac{\sqrt N }{{{p\varepsilon ^2}}}\!+\!\frac{\sqrt N }{{{\varepsilon ^2}}}, \frac{1}{{{p\varepsilon ^3}}}\!+\!\frac{1}{{{\varepsilon ^3}}}\})^{2, 4}$ & NA$^5$ \\
\textbf{ASPIRE-EASE} & $\mathcal{O}(\frac{1}{{{\varepsilon ^4}}})$ & NA$^5$ & $\mathcal{O}(\frac{1}{{{\varepsilon ^6}}})$\\
\bottomrule  
\end{tabular}}
\label{tab:convergence rate compare}}
\\\footnotesize{$^1$ This complexity is to find an $\varepsilon$-stationary point  of $\Phi ( \cdot ) = {\max _y}f( \cdot ,y)$, that is $||\Phi ( \cdot )|{|^2} \le {\varepsilon ^2}$.}
\\\footnotesize{$^2$ This complexity is to find an $\varepsilon$-stationary point  such that  $||\mathcal{G}_{\eta} ( \cdot )|{|^2} \le {\varepsilon ^2}$, where $\mathcal{G}_{\eta}$ is a proximal gradient measure.}
\\\footnotesize{$^3$ This complexity corresponds to the number of iterations to solve the inner subproblem.  It does not consider the complexity of solving the inner subproblem.}
\\\footnotesize{$^4$ $p$ is the number of workers, in this problem $p=N$.}
\\\footnotesize{$^5$ NA represents not applicable.}
\end{table}

\setcounter{figure}{0}
\begin{figure}[t]  
\renewcommand{\thefigure}{E\arabic{figure}}
\begin{center}
\includegraphics[height=0.42\textwidth,scale=1]{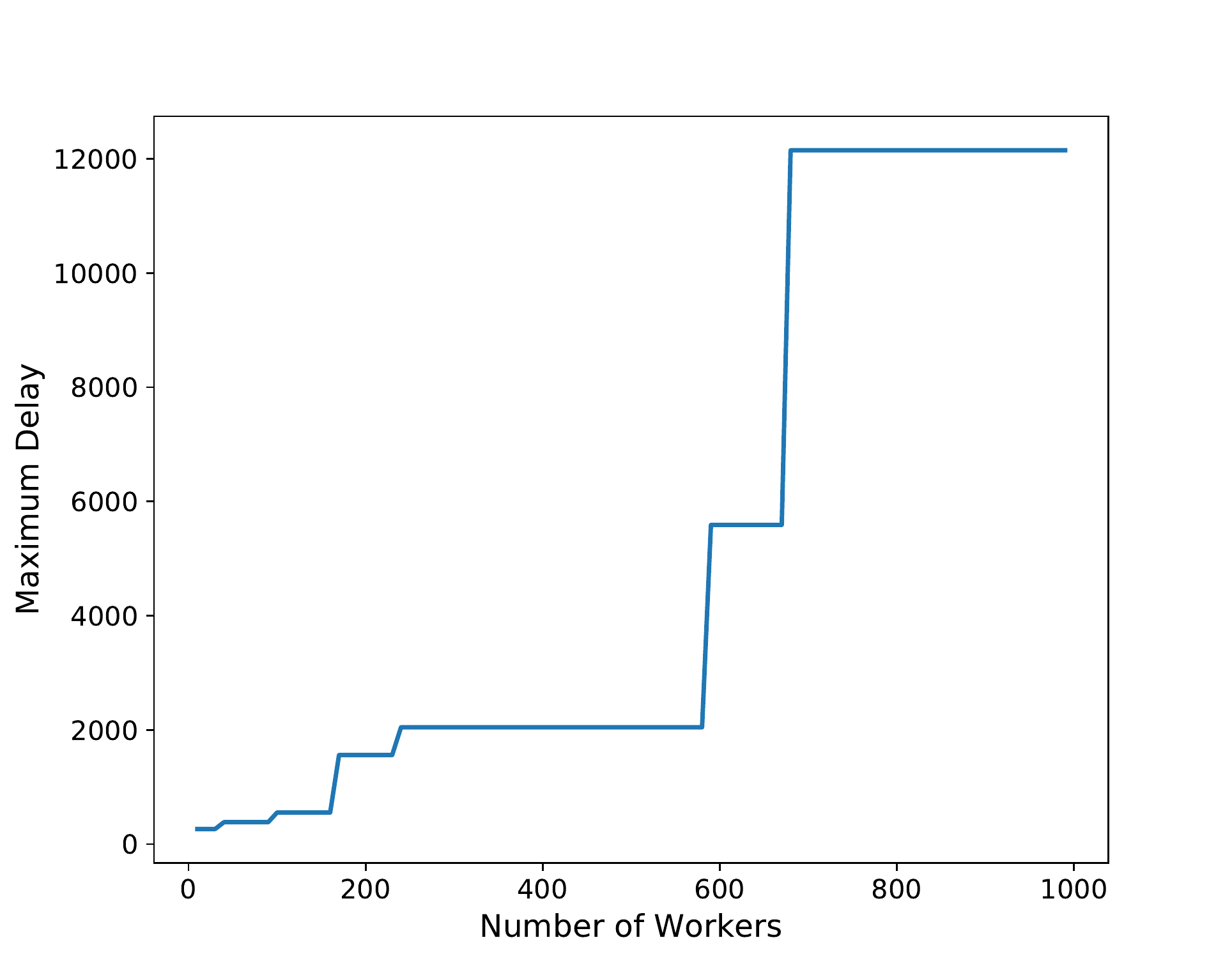} 
\caption{With the increase of the number of workers in the distributed system, the maximum delay would increase dramatically. The delay follows log-normal distribution LN(1, 0.4) in the experiment.} 
\label{fig:delay_ratio}
\end{center}
\end{figure}

\section{Explanation about Assumption}
The gradient Lipschitz (or smoothness) is a common assumption that has been widely used \citep{deng2021distributionally,qi2021online,lin2020gradient}. In some other works \citep{thekumparampil2019efficient,zhou2018fenchel}, if the function $L_p$ is $\widetilde{L}$-smooth, it has to satisfy,

\begin{equation}
\label{eq:qppendix 108}
\renewcommand{\theequation}{F.\arabic{equation}}
    \begin{array}{l}
||{\nabla  _\theta }{L_p}( \{ {{\boldsymbol{w}}_j}\},{\boldsymbol{z}},h,  \{ {\lambda _l}\}, \{ {{\boldsymbol{\phi}}_j}\} ) -  {\nabla  _\theta }{L_p}( \{ {\hat{\boldsymbol{w}}_j}\},\hat{\boldsymbol{z}},\hat{h},  \{ {\hat{\lambda }_l}\}, \{ {\hat{\boldsymbol{\phi}}_j}\}  )|| \\
 \le \widetilde{L}(\sum\limits_{j = 1}^N {||{{\boldsymbol{w}}_j} - {\hat{\boldsymbol{w}}_j}||}  + ||{\boldsymbol{z}} - \hat{\boldsymbol{z}}|| + ||h - \hat{h}|| + \sum\limits_{l = 1}^M {||{\lambda _l} - {\hat{\lambda}_l}||}  + \sum\limits_{j = 1}^N {||{{\boldsymbol{\phi}}_j} - {\hat{\boldsymbol{\phi}}_j}||)} ,
\end{array}
\end{equation}

where $\theta  \in  \{ \{ {{\boldsymbol{w}}_j}\} ,{\boldsymbol{z}},h,\{ {\lambda _l}\} ,\{ {{\boldsymbol{\phi}}_j}\} \} $ and we demonstrate $L_p$ that satisfies Eq. (F. \ref{eq:qppendix 108}) is also satisfied with our Assumption 1.

From Eq. (F. \ref{eq:qppendix 108}) and according to Cauchy-Schwarz inequality, we can obtain,
\begin{equation}
\label{eq:qppendix 109}
\renewcommand{\theequation}{F.\arabic{equation}}
    \begin{array}{l}
||{\nabla  _\theta }{L_p}( \{ {{\boldsymbol{w}}_j}\},{\boldsymbol{z}},h,  \{ {\lambda _l}\}, \{ {{\boldsymbol{\phi}}_j}\} ) \! - \! {\nabla  _\theta }{L_p}( \{ {\hat{\boldsymbol{w}}_j}\},\hat{\boldsymbol{z}},\hat{h},  \{ {\hat{\lambda }_l}\}, \{ {\hat{\boldsymbol{\phi}}_j}\}  )||^2\\

\! \le\! (2N  +  M   +   2)\widetilde{L}{^2}(\sum\limits_{j = 1}^N {||{{\boldsymbol{w}}_j} \!-\! {\hat{\boldsymbol{w}}_j}|{|^2}}  + ||{\boldsymbol{z}}\! -\! \hat{\boldsymbol{z}}|{|^2} + ||h \!-\! \hat{h}|{|^2} + \sum\limits_{l = 1}^M {||{\lambda _l} \!-\! {\hat{\lambda}_l}|{|^2}}  + \sum\limits_{j = 1}^N {||{{\boldsymbol{\phi}}_j} \!-\! {\hat{\boldsymbol{\phi}}_j}|{|^2}} ).
\end{array}
\end{equation}

Let $L = \sqrt {(2N + M + 2)} \widetilde{L}$, we can obtain,
\begin{equation}
\renewcommand{\theequation}{F.\arabic{equation}}
\label{eq:qppendix 110}
\begin{array}{l}
||{\nabla  _\theta }{L_p}( \{ {{\boldsymbol{w}}_j}\},{\boldsymbol{z}},h, \! \{ {\lambda _l}\}, \!\{ {{\boldsymbol{\phi}}_j}\} ) \! - \! {\nabla  _\theta }{L_p}( \{ {\hat{\boldsymbol{w}}_j}\},\hat{\boldsymbol{z}},\hat{h}, \! \{ {\hat{\lambda }_l}\}, \!\{ {\hat{\boldsymbol{\phi}}_j}\}  )|| \vspace{0.5ex}\\
 \le L||[ {{\boldsymbol{w}}_{\rm{cat}}} \! - \!  {{\hat{\boldsymbol{w}}_{\rm{cat}}}}   ;{\boldsymbol{z}} \! - \!{\hat{\boldsymbol{z}}}  ;h \! - \! \hat{h} ; {\boldsymbol{\lambda} _{\rm{cat}}} \! - \! {{\hat{\boldsymbol{\lambda}} _{\rm{cat}}}}  ; {{\boldsymbol{\phi}}_{\rm{cat}}} \! - \! {{\hat{\boldsymbol{\phi}}_{\rm{cat}}}}  ]||.
\end{array}
\end{equation}

\section{Discussion about $CD$-norm Uncertainty Set}
In this paper, we utilize the $CD$-norm uncertainty set in our framework. Compared with ellipsoid and KL-divergence uncertainty sets, whose cutting plane generation subproblems are respectively a second-order cone optimization (SOCP) problem and a relative entropy programming (REP) problem, the cutting plane generation subproblem (Eq. (17)) is an LP-type problem when utilizing $CD$-norm uncertainty set. Please note that the LP-type problem in Eq. (17) can be efficiently solved by merge sort. Therefore, the cutting plane generation subproblem with $CD$-norm uncertainty set is much easier to solve than those with the ellipsoid and KL-divergence uncertainty sets.

\end{document}